\pdfoutput=1

\documentclass[11pt]{article}

\usepackage{acl}

\usepackage{times}
\usepackage{latexsym}

\usepackage[T1]{fontenc}

\usepackage[utf8]{inputenc}

\usepackage{microtype}

\usepackage{inconsolata}

\usepackage{graphicx}
\usepackage{caption}
\usepackage{subcaption}
\usepackage{booktabs}
\usepackage{multirow}
\usepackage{bm}
\usepackage{hyperref}
\usepackage{amsmath}
\usepackage{cleveref}

\usepackage{todonotes}
\makeatletter
\newcommand*\iftodonotes{\if@todonotes@disabled\expandafter\@secondoftwo\else\expandafter\@firstoftwo\fi}  %
\makeatother

\usepackage[normalem]{ulem}

\title{The Hidden Space of Transformer Language Adapters}

\author{Jesujoba O. Alabi\textsuperscript{1} \,
 Marius Mosbach\textsuperscript{2} \,
 Matan Eyal\textsuperscript{3} \, Dietrich Klakow\textsuperscript{1} \, Mor Geva\textsuperscript{4} \ \vspace{3px} \\ \vspace{3px}
 \textsuperscript{1}Saarland University, Saarland Informatic Campus \\
 \textsuperscript{2}Mila, McGill University~~~
\textsuperscript{3}Google Research~~~
\textsuperscript{4}Tel Aviv University\\
\small{{\tt  jalabi@lsv.uni-saarland.de}\quad {\tt  marius.mosbach@mila.quebec}\quad {\tt morgeva@tauex.tau.ac.il}}
}

\begin{document}
\maketitle
\begin{abstract}

We analyze the operation of transformer language adapters, which are small modules trained on top of a frozen language model to adapt its predictions to new target languages. We show that adapted predictions mostly evolve in the source language the model was trained on, while the target language becomes pronounced only in the very last layers of the model. Moreover, the adaptation process is gradual and distributed across layers, where it is possible to skip small groups of adapters without decreasing adaptation performance. Last, we show that adapters operate on top of the model’s frozen representation space while largely preserving its structure, rather than on an ``isolated'' subspace. Our findings provide a deeper view into the adaptation process of language models to new languages, showcasing the constraints imposed on it by the underlying model and introduces practical implications to enhance its efficiency.\footnote{We release our code and models publicly at \url{https://github.com/uds-lsv/hidden-space-adapters}.}

\end{abstract}

\section{Introduction} 
\label{sec:introduction}

The adaptation of pre-trained transformer-based language models (LMs) to new languages has become a widely adopted practice in natural language processing (NLP).
A prominent approach for achieving this is to train small feed-forward network neural modules, called \textit{adapters}, on top of the LM layers, while freezing the LM parameters~\citep{pmlr-v97-houlsby19a}.
To successfully adapt an LM to a new language, adapters should introduce changes that steer the prediction but still can be processed by subsequent layers.

Adapters have demonstrated impressive capabilities in zero-shot cross-lingual settings~\citep{pfeiffer-etal-2020-mad,lee-etal-2022-fad,parovic-etal-2022-bad}, multi-task learning~\citep{pfeiffer-etal-2021-adapterfusion}, and integrating knowledge into pre-trained LMs~\cite{lauscher-etal-2020-common, wang-etal-2021-k}. 
However, despite this vast success, how LM predictions are being adapted internally is largely unknown. 
Nonetheless, a better understanding of the internal operation of LM adapters would be valuable to inform language selection for multilingual pre-training and for designing more effective adaptation approaches. 

In this work, we tackle the question of how language adapters work, while focusing on LM predictions adapted from one or more languages (source) to another language (target). For our study, we train an auto-regressive decoder-only LM from scratch on English and then adapt it separately to four different target languages: German, French, Hebrew and Arabic (\S\ref{sec:experimental-setup}). To demonstrate the generality of our findings, we additionally experiment with bilingual models as well as mBERT \citep{devlin-etal-2019-bert}, a massively multilingual LM pre-trained on 104 languages.\looseness-1

First, we consider the sequence of hidden representations across the layers as an information stream being evolved through additive updates \citep{elhage2021mathematical}, and analyze the adapters' updates to the prediction (\S\ref{sec:think}).
We find that throughout most of the inference pass, the prediction is evolved in the source language and only in the very last layers the target language abruptly becomes pronounced. Then, we explore the contributions of individual adapters (\S\ref{sec:adapt_dist}) and observe that adapters introduce gradual updates, which often can be canceled without any decrease in performance, while the last few adapter layers are critical for adaptation success.\looseness-1 

Based on these findings, we explore two alternative hypotheses for how adapters interact with the underlying frozen LM (\S\ref{sec:hypotheses}). Either adapters operate in an ``isolated'' subspace of the representation space that is unused for predictions in the source language, or they operate on top of the model's representation space, while preserving its structure. We test the first hypothesis by training sparse probing classifiers to detect features that change the most during adaptation and then intervene on these features. We observe that while the identified features indeed deteriorate adaptation performance upon intervention, intervening on random features also leads to substantial drops in adaptation performance, and therefore refute the first hypothesis.\looseness-1 

To test the second hypothesis we analyze principal components in the representation space of the monolingual base model as well as the adapted models. For different properties such as part-of-speech, number, and tense, we observe a strong alignment between the original representation space and its adapted counterpart. 

In summary, we provide novel insights into the prevalent language adaptation process of pre-training LMs, showing that adapted predictions are gradually evolved on top of the representation space of the source language, while being shifted towards the target language only at the end of the computation and that adapters operate on top of the existing structure of the pre-trained model's representation space. Our work not only provides a first step towards a better understanding of language model adaptation but also opens interesting avenues for future work on making language adaptation more efficient.

\section{Experimental setup}
\label{sec:experimental-setup}

We employ a controlled setting of pre-training our own LMs from scratch, and then adapting them to a target language by training language adapters \citep{pfeiffer-etal-2020-mad}. This is important because some of our experiments require language identification based on sub-tokens. Achieving this with existing multilingual LMs is challenging due to the diverse languages on which the models and their tokenizers have been trained. Where possible, we extend our experiments to mBERT \cite{devlin-etal-2019-bert}, a multiligual LM trained on 104 languages.

\paragraph{Models} 

As our base model, we pre-trained an auto-regressive decoder-only LM on English (\texttt{en}) texts. Our model follows the GPT-NeoX architecture~\citep{black-etal-2022-gpt} with $L=24$ decoder layers, 16 attention heads per layer, and a hidden dimensionality of 1024. We used a vocabulary size of 250K\footnote{Resulting in a total of 814M trainable parameters.}, which is large enough to support multiple languages, and trained a BPE tokenizer \citep{wang2019neural} from scratch on a combination of sentences sampled from these languages. For our experiments in \S\ref{sec:think}, we additionally train a bilingual model on English and a small fraction of German data.\looseness-1

\paragraph{Adaptation to a new language}

To adapt the model, we trained a separate set of adapters (via language modeling) for each of the following target languages: Arabic (\texttt{ar}), English (\texttt{en}), French (\texttt{fr}), German (\texttt{de}), and Hebrew (\texttt{he}). We choose German and French because of their high similarity to each other, and English and French having a higher lexical overlap than English and German. The choice of Hebrew and Arabic is motivated by the fact that they are relatively similar to each other while highly dissimilar to English, in part by their different non-Latin scripts.
Further details on our pre-training and adaptation procedure are provided in \Cref{apx:model_training}.\looseness-1

\begin{figure*}[t]
\setlength{\belowcaptionskip}{-4pt}
  \centering
    \begin{subfigure}{0.235\textwidth}
        \includegraphics[width=\textwidth]{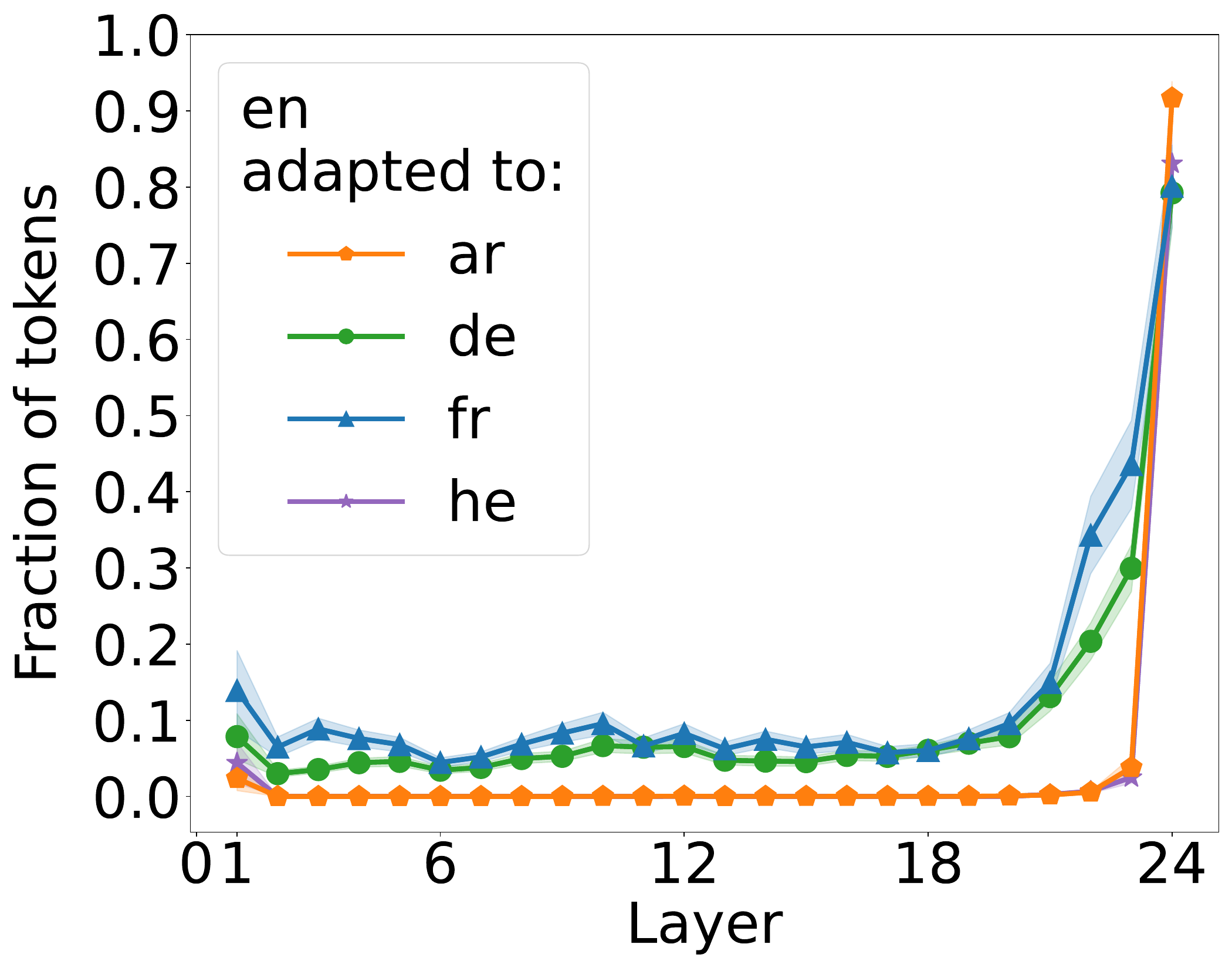}
        \caption{}
        \label{fig:results_proj_toy}
    \end{subfigure}
     ~
    \begin{subfigure}{0.235\textwidth}
        \includegraphics[width=\textwidth]{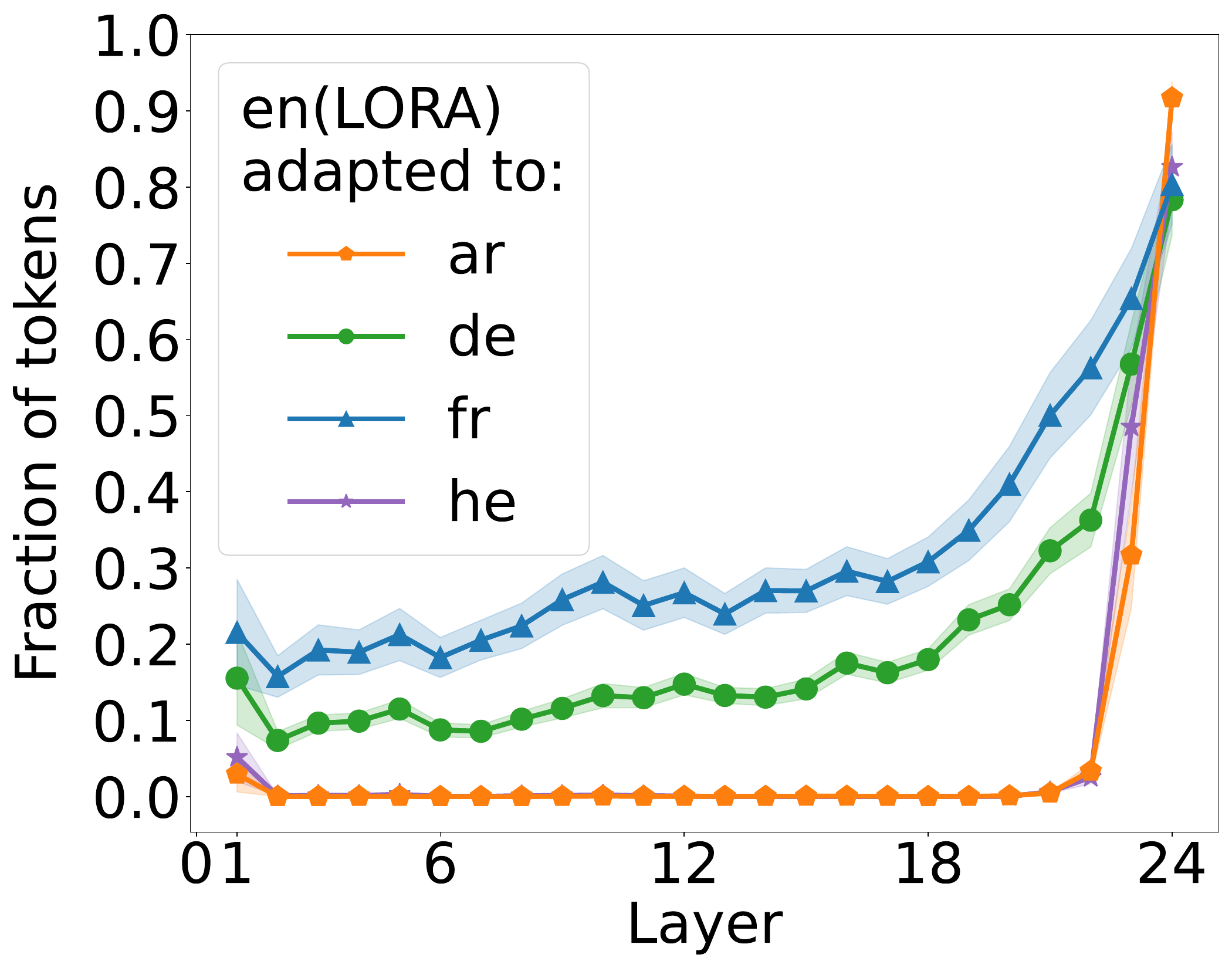}
        \caption{}
        \label{fig:results_proj_lora_toy}
    \end{subfigure}
     ~     
     \begin{subfigure}{0.235\textwidth}
         \includegraphics[width=\textwidth]{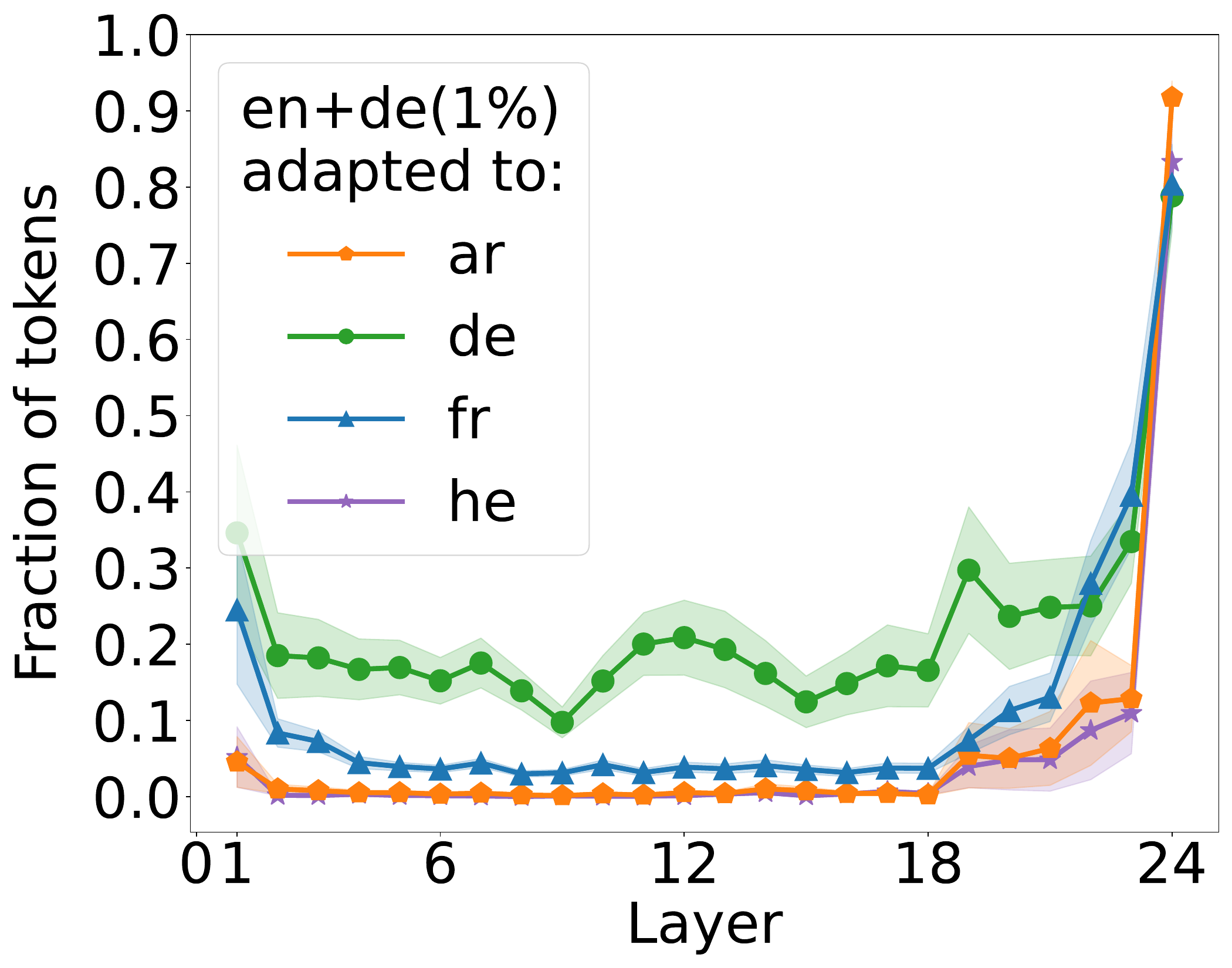}
         \caption{}
         \label{fig:results_proj_toy_ende}
     \end{subfigure}
     ~
    \begin{subfigure} {0.235\textwidth} 
        \includegraphics[width=\textwidth]{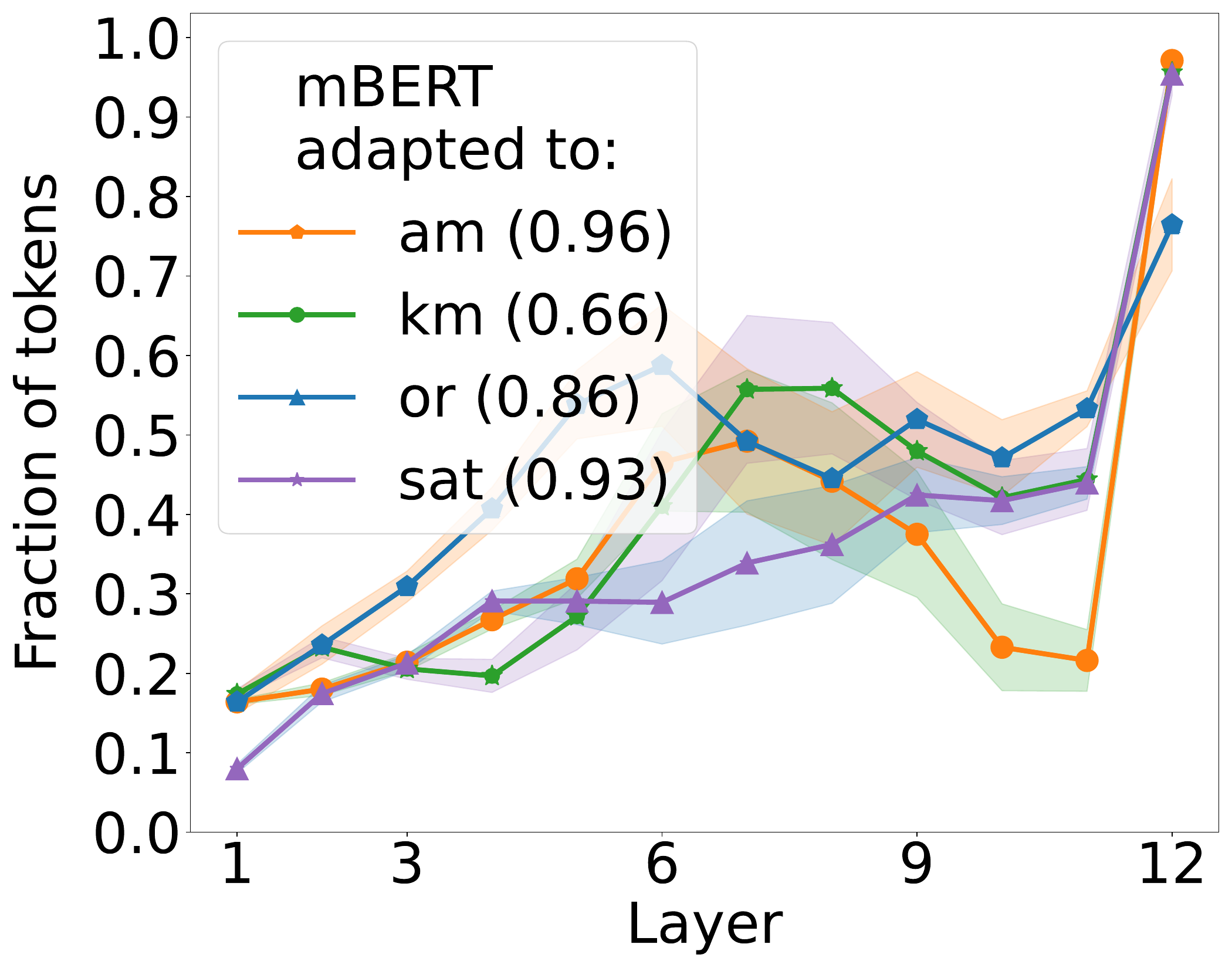}
        \caption{}
    \label{fig:results_proj_mbert}
    \end{subfigure}
\caption{The fraction of target language tokens in the top-10 predicted tokens when projecting hidden layer representations to the vocabulary space. (\subref{fig:results_proj_toy}) shows results for a model pre-trained only on English, (\subref{fig:results_proj_lora_toy}) shows results for a model pre-trained only on English, and adapted with LoRA, (\subref{fig:results_proj_toy_ende}) shows results for a model pre-trained on English and 1\% of the German data, (\subref{fig:results_proj_mbert}) shows results for mBERT.}
\label{fig:fraction_of_tokens}
\end{figure*} 

\paragraph{Adapter setup}

For most of our experiments, we focus on the adapter architecture proposed by \citet{pfeiffer-etal-2020-mad}, which is the most commonly used technique for multilingual LM adaptation and train using its default hyperparameters\footnote{Adapters with a reduction factor of 16, i.e. from 1024 to 64, and the ReLU activation function as implemented in \citet{pfeiffer2020AdapterHub}}. This technique relies on placing an adapter layer in each LM block. Specifically, 
for every transformer decoder block \citep{vaswani2017attention}, an adapter is stacked on top of the feed-forward network (FFN), such that it gets as input the hidden states after the FFN layer and outputs an update to every hidden state that is added via a residual connection. Formally, let $t_1, \dots, t_N$ denote an input sequence consisting of $N$ tokens, the hidden representation $\mathbf{x}_{\texttt{out}}^{l,i} \in \mathcal{R}^d $ at the $i$-th position after the $l$-th block is obtained by:

\vspace{-0.25cm}
\begin{align}
    \mathbf{x}_{\texttt{attn}}^{l,i} &= \mathbf{x}_{\texttt{out}}^{l-1,i} + \textcolor{black}{\texttt{self-attn}}(X_{\texttt{out}}^{l-1})_i \\
    \mathbf{x}_{\texttt{ffn}}^{l,i}\; &= \mathbf{x}_{\texttt{attn}}^{l,i} + \textcolor{black}{\texttt{feed-forward}}(\mathbf{x}_{\texttt{attn}}^{l,i}) \\
    \mathbf{x}_{\texttt{out}}^{l,i}\; &= \mathbf{x}_{\texttt{ffn}}^{l,i} + \textcolor{black}{\texttt{adapter}}(\mathbf{x}_{\texttt{ffn}}^{l,i})
\end{align}
where $X_{\texttt{out}}^{l-1} \in \mathcal{R}^{N\times d}$ are the hidden representations from the preceding layer, and $\textcolor{black}{\texttt{self-attn}}(X_{\texttt{out}}^{l-1})_i \in \mathcal{R}^d$ 
is the output from the self-attention layer being added to the $i$-th hidden representation. The adapter layer is typically a small FFN with a low-dimensional bottleneck:
\begin{align}
    \textcolor{black}{\texttt{adapter}}(\mathbf{x}) &= \mathbf{W}_2 \sigma(\mathbf{W}_1\mathbf{x})~,
\end{align}
where $\sigma$ is a non-linear activation function, $\mathbf{W}_1 \in \mathcal{R}^{b \times d}$, $\mathbf{W}_2 \in \mathcal{R}^{d \times b}$ and $b << d$. We will refer to this setting as Pfeiffer adapters for the remainder of this paper.

In addition to analyzing Pfeiffer adapters, we extend our analysis to LM adaptation with low-rank adapters (LoRA)~\citep{hu2021lora}. To ensure comparison with the Pfeiffer adapter setup, we add low-rank adapters only to the FFN layers within each transformer decoder block. We describe and formalize the LoRA setup in more detail in \Cref{sec:appendix:lora}.

In our analysis, we view each layer (self-attention, feed-forward, and adapter) as reading from and writing to the model's residual stream \cite{elhage2021mathematical}. 
Crucially, during adaptation all pre-trained weights, except for the input and output embeddings, are frozen. Training the embeddings in addition to the adapters is necessary, as our original model was pre-trained exclusively on English, and it also leads to improved adaptation performance \cite{artetxe-etal-2020-cross,conneau-etal-2020-unsupervised,yong2022adapting}.\footnote{An alternative approach is to use invertible adapters \citep{pfeiffer-etal-2020-mad}, but it requires that the input and output embeddings are tied, which does not hold in our case.}

\paragraph{Data} We used data sourced from Wikipedia for both the pre-training and adaptation stages. Details on the data sizes are provided in \Cref{apx:details_token_statistics}. For our analysis, we use data sampled from FLORES-101 \citep{goyal-etal-2022-flores}, a dataset of 3k English sentences carefully translated to various languages. We combined the dev and devtest splits for our analysis. In addition, we used the parallel universal dependencies (PUD) treebanks \citep{zeman-etal-2017-conll} for \texttt{en}, \texttt{de}, and \texttt{fr} for POS analysis, and the  multilingual morphosyntactic probing dataset for \texttt{de}, and \texttt{fr} \citep{Acs_Kornai_2023}.

\section{Adapted predictions evolve in the source language}
\label{sec:think}

Adapters operate on top of the computation of the frozen pre-trained LM, introducing changes that the subsequent layer does not ``expect'' since its parameters were not updated during the adapter training. 
This raises the question of whether the adapted model ``thinks'' in the distribution of the source language(s) it predominantly saw during pre-training
and translates its predictions during adaptation or whether it operates exclusively in the target language it was adapted to.

\paragraph{Experiment}

To address this question, we inspect the adapted LM  predictions across layers via projection to the vocabulary space~\citep{nostalgebraist2020interpreting, geva-etal-2022-transformer}. We feed examples in each target language into the model and extract the hidden representations created during inference at the last position, from which the next-token prediction is obtained. Then, we project each representation to the vocabulary by multiplying it by the LM head, and apply the softmax function to obtain an intermediate output distribution for every hidden representation.
For each example we analyze in which language --- the source language the LM was trained on or the target language --- the LM constructs its adapted prediction. This is done by taking the top-$k$ tokens with the highest probabilities in each layer, and categorizing them to their respective languages. We perform this classification based on the script for \texttt{ar} and \texttt{he} and based on token statistic for \texttt{en}, \texttt{de}, and \texttt{fr} (see \Cref{apx:details_lang_id} for details).\looseness-1

\paragraph{Results} 

\Cref{fig:results_proj_toy} shows the portion of predicted tokens from the target language with $k=10$, across the layers of a monolingual English model adapted to each of the target languages independently.\footnote{We choose $k=10$ following \citet{geva-etal-2022-transformer}. Similar trends were observed for $k=100,1000$.}
For all target languages, we observe that starting from the second layer
and up until the last two layers, the presence of target language tokens remains notably limited, suggesting that the vast majority of the layers promotes tokens in the pre-training language rather in the target language. In contrast, for the last two layers we observe a dramatic increase of target language tokens presence in the top of the output distribution. \Cref{fig:results_proj_lora_toy} shows a similar trend when adapting with LoRA instead.

We additionally pre-trained a model on $1\%$ of the German data together with all of the English data. We adapt this model to all target languages and repeat our previous analysis. \Cref{fig:results_proj_toy_ende} shows the result. As expected, the fraction of German tokens is now higher, even for the middle layers. Overall, however, we still see the same pattern as before, with most of the predictions being in the source language except for the last two layers.

\begin{figure*}[t]
\setlength{\belowcaptionskip}{-2pt}
  \centering
    \begin{subfigure}{0.235\textwidth}
        \includegraphics[width=\textwidth]{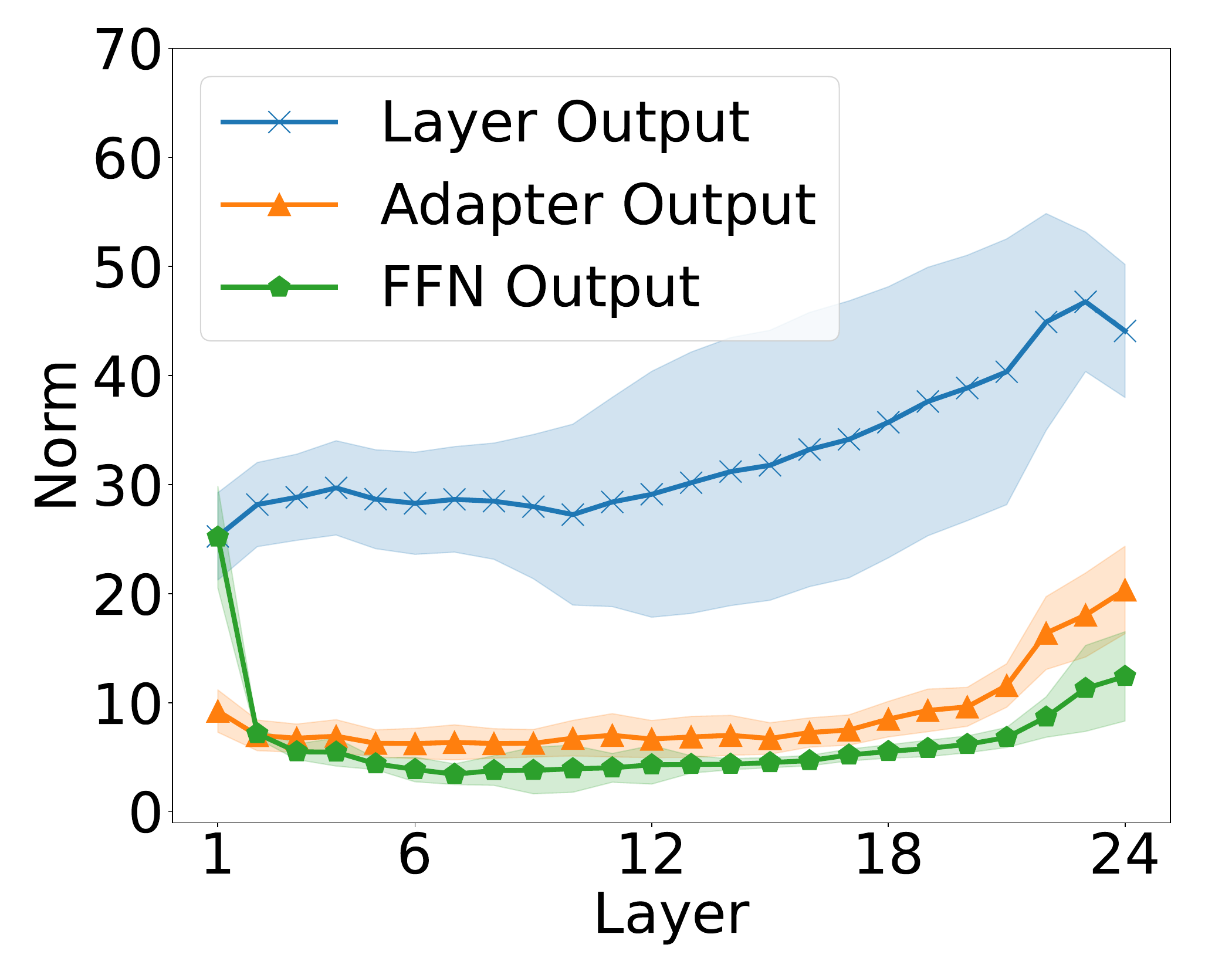}
        \caption{\texttt{en}$\rightarrow$\texttt{he}}
    \end{subfigure}
    ~
    \begin{subfigure}{0.235\textwidth}
        \includegraphics[width=\textwidth]{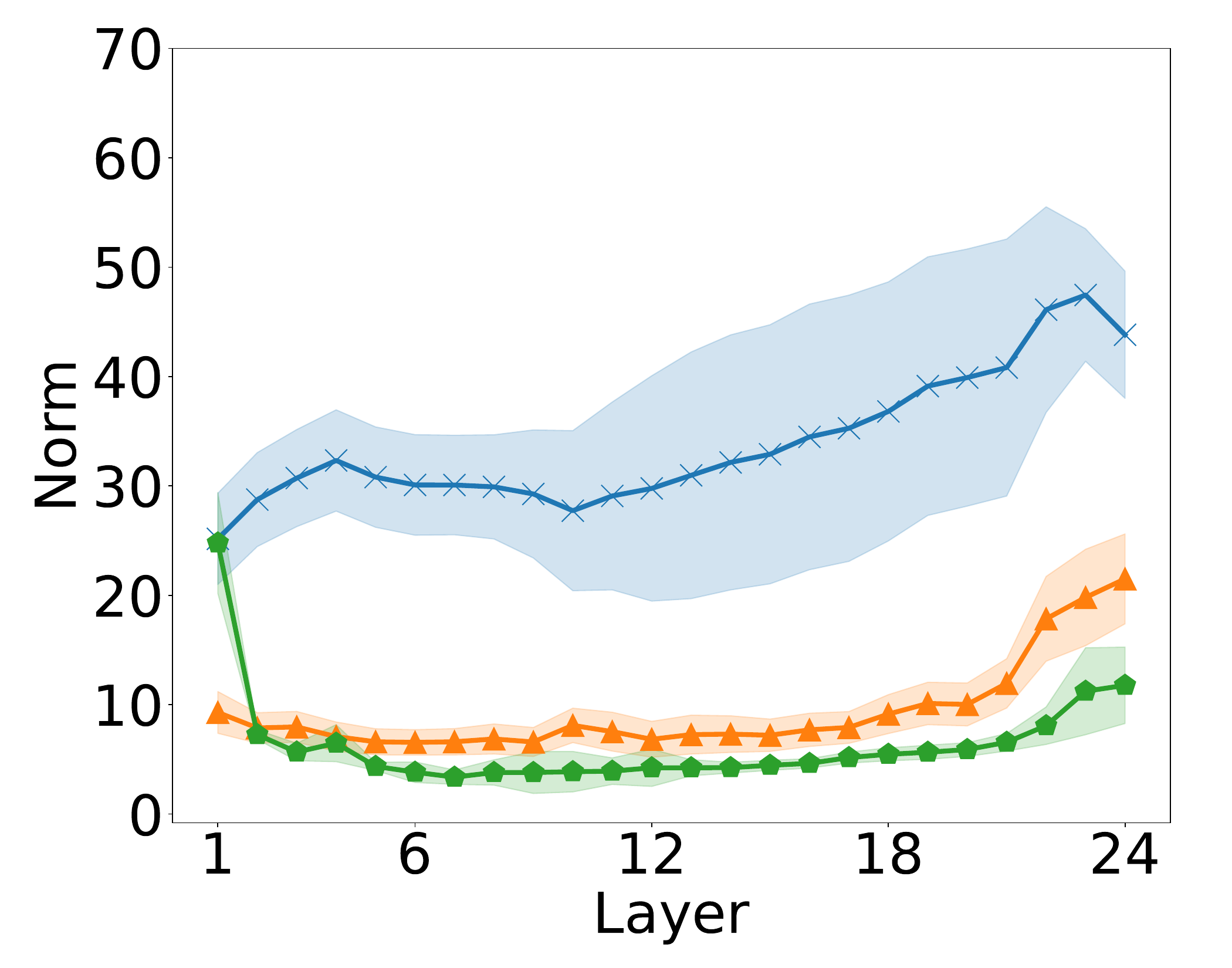}
        \caption{\texttt{en}$\rightarrow$\texttt{ar}}
    \end{subfigure}
    ~
    \begin{subfigure}{0.235\textwidth}
        \includegraphics[width=\textwidth]{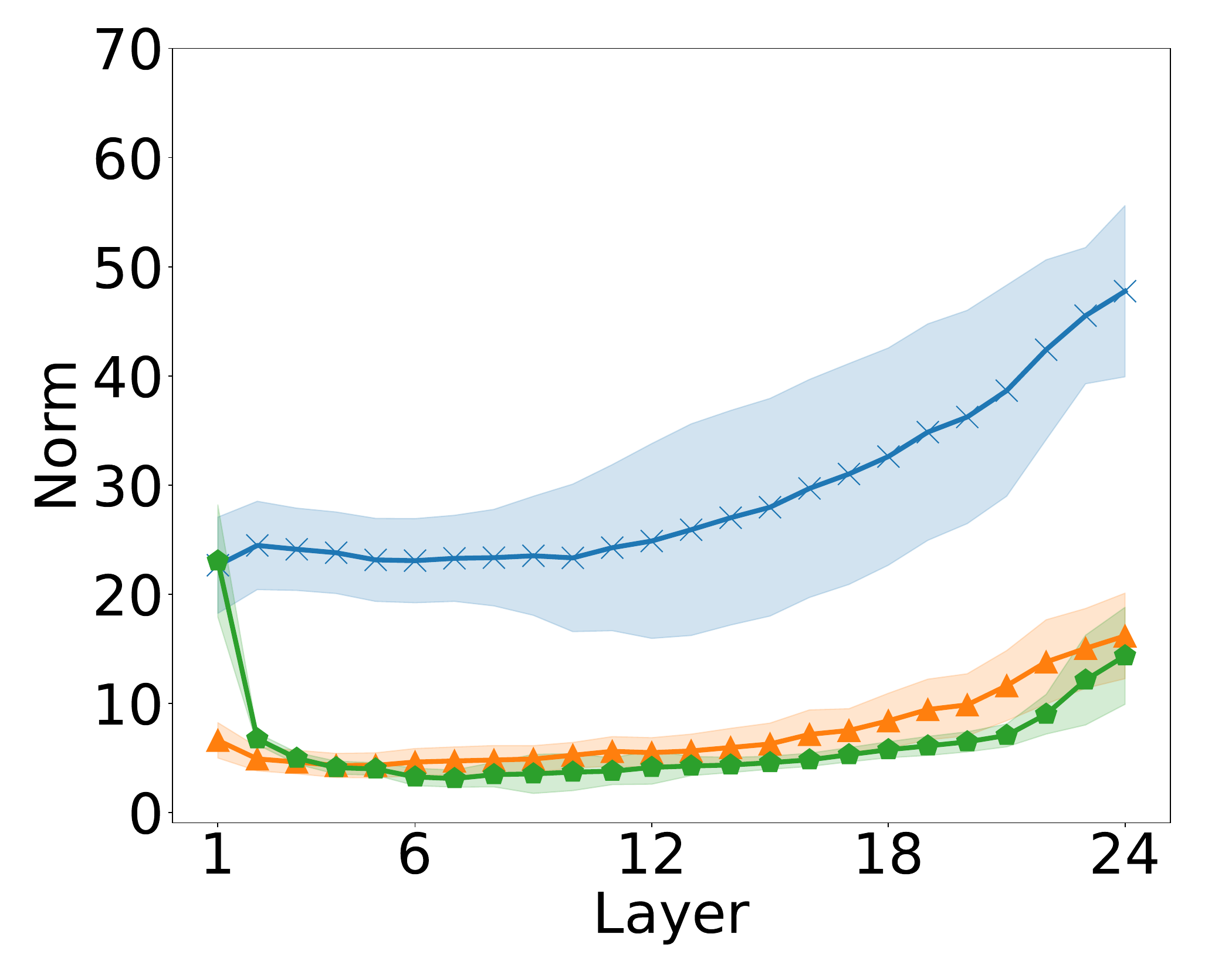}
        \caption{\texttt{en}$\rightarrow$\texttt{de}}
    \end{subfigure}
    ~
    \begin{subfigure}{0.235\textwidth}
        \includegraphics[width=\textwidth]{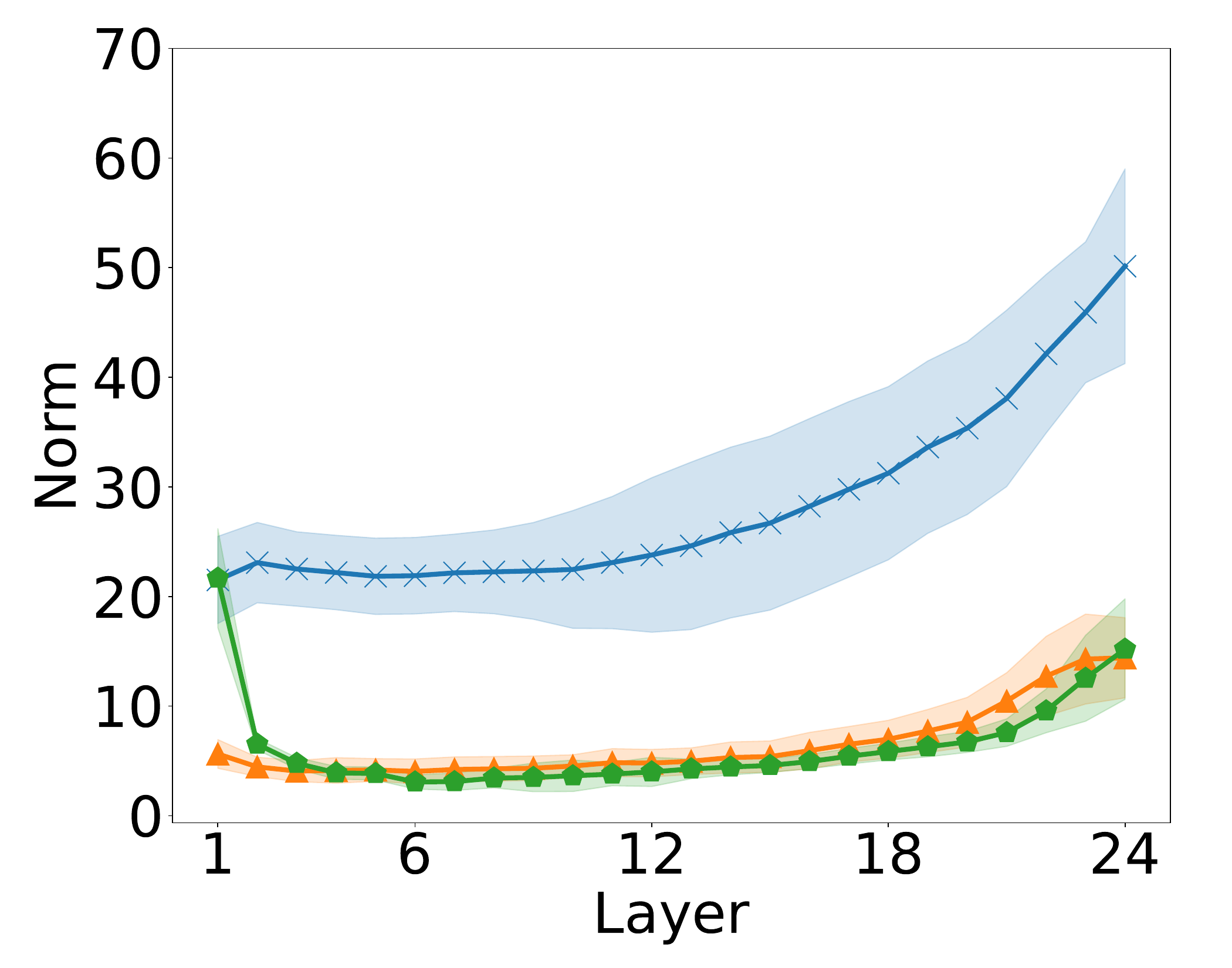}
        \caption{\texttt{en}$\rightarrow$\texttt{fr}}
    \end{subfigure}
\caption{The average L2 norm of the adapter output in comparison to the feed-forward and layer outputs. 
}
\label{fig:norm_results}
\end{figure*} 

\paragraph{Generalization to multilingual models}
\label{sec:todo}

We repeat the same experiment but for an existing multilingual LM, namely mBERT \citep{devlin-etal-2019-bert}. We exploit the fact that mBERT has not been trained on Amharic (\texttt{am}), Khmer (\texttt{km}), Odia (\texttt{or}), and Santali (\texttt{sat}) for which we can obtain sufficient data for adaptation. For language identification, we follow the same approach as described above and rely on the script of each language. To adapt mBERT, we first train new tokenizer jointly on all of these new languages and extend mBERT's vocabulary with the newly created vocabulary. As with our models, we train only the embeddings and adapters and project the hidden representations of every layer to the vocabulary at inference time.

\Cref{fig:results_proj_mbert} shows that in contrast to the mono- and bilingual LMs, for mBERT we observe a substantial increase in target language tokens already in the middle layers, reaching up to 60\% of the top tokens in the projection. We attribute this to the multilingual representation space of mBERT, which potentially helps the model to adapt to new languages. Nonetheless, we again observe that the fraction of target tokens increases dramatically at the final layer.\looseness-1

\paragraph{Conclusion} We take these observations as evidence that adapted predictions are evolved in the distribution of the source languages the model saw during pre-training, while being shifted to the target language only at the end of the inference pass.

\section{The adaptation process is distributed across layers}
\label{sec:adapt_dist}

Our previous findings raise the question of how distributed the adaptation process is, i.e., is it concentrated on specific adapters or distributed across all layers of the model? Asked differently, are only the last adapters important while those at the early layers can be ignored or removed? In the following, we tackle this question by analyzing the magnitude of adapter outputs and the effect of canceling individual and consecutive groups of adapters during inference on the adaptation performance.

\subsection{Adapters gradually steer the prediction} 
\label{sec:layer_norms}

Every adapter at layer $l$ contributes an additive update to the evolving residual stream representation.\footnote{$\texttt{adapter}(\mathbf{x}_{\texttt{ffn}}^{l, N})$ in the Pfeiffer configuration and $\textcolor{black}{\texttt{LoRA}_2}(\mathbf{x}_{\texttt{ffn}_1}^{l,i})$ with LoRA (see \Cref{sec:appendix:lora}).} Here we analyze how pronounced these updates are in terms of their magnitude. To this end, for each target language, we feed the entire FLORES-101 devtest data to our models and we obtain the adapter, feed-forward, and layer outputs for 6500 tokens at randomly chosen positions.
Then, we compare the average L2 norm of these representations at every layer. If not stated otherwise, we focus on the Pfeiffer configuration in the following, i.e. we compare $\lVert\texttt{adapter}(\mathbf{x}_{\texttt{ffn}}^{l,N})\rVert_2$ with $\lVert\mathbf{x}_{\texttt{ffn}}^{l, N}\rVert_2$ and $\lVert\texttt{feed-forward}(\mathbf{x}_{\texttt{attn}}^{l,N})\rVert_2$. 

\Cref{fig:norm_results} shows that adapters introduce updates that are substantially smaller in magnitude compared to the residual stream representations. Moreover, the adapter outputs are similar in magnitude to those of the FFN layers, which previously have been shown to introduce gradual changes to the prediction \cite{geva-etal-2021-transformer}.
While upper-layer adapters introduce updates of larger magnitude compared to early-layer adapters, this increase in norm is also observed for the hidden representations, and thus may not explain the shift towards the target language observed in \S\ref{sec:think}.
Interestingly, the norm of adapter outputs in the last layers is larger for Arabic and Hebrew than for German and French, suggesting that larger updates are needed to steer predictions to more distant languages.\looseness-1

\begin{figure*}[t]
\setlength{\belowcaptionskip}{-4pt}
  \centering
      \begin{subfigure}{0.235\textwidth}
        \includegraphics[width=\textwidth]{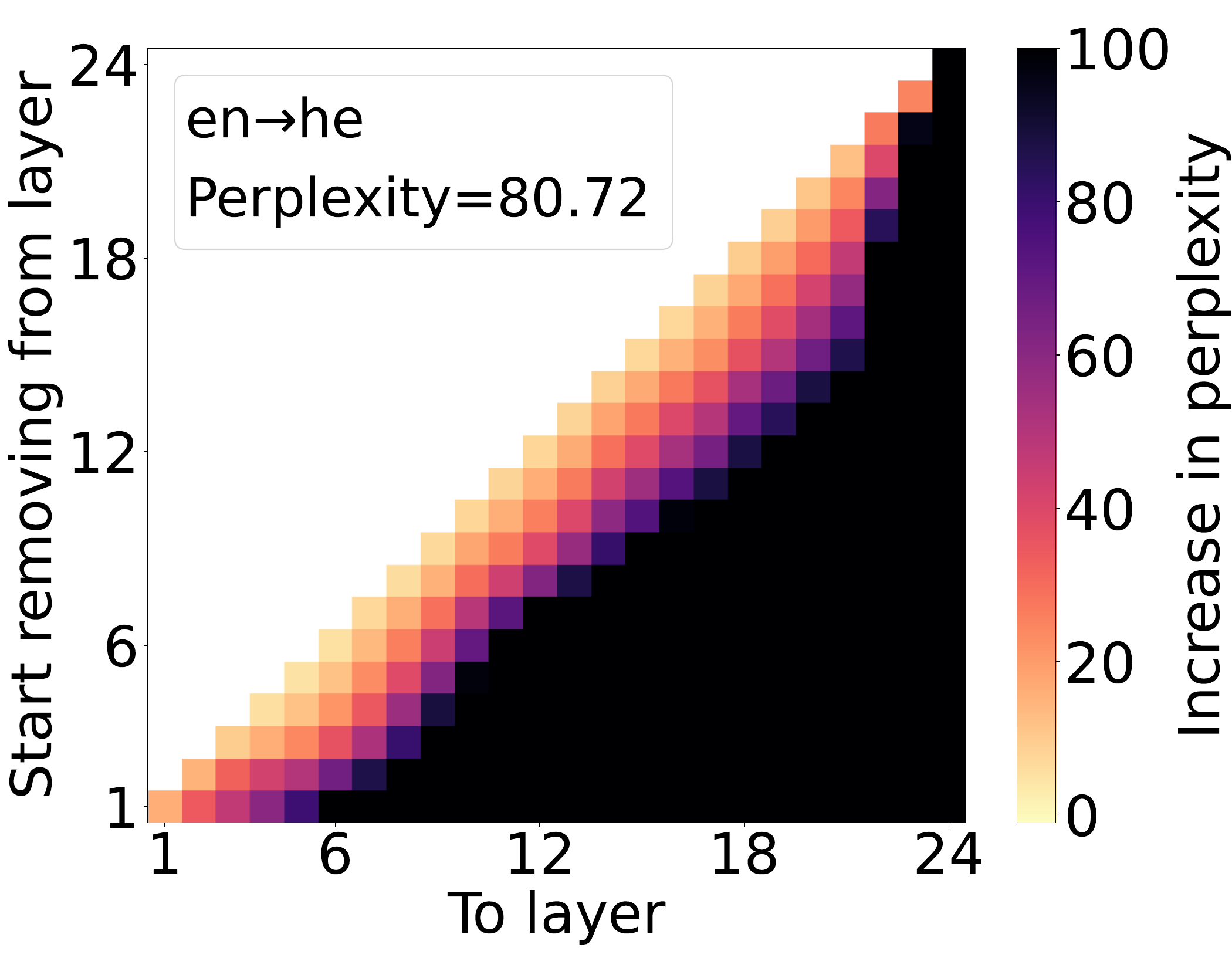}
        \caption{\texttt{en} $\rightarrow$ \texttt{he}}
        \label{fig:ppl_ar}
    \end{subfigure}
    ~
    \begin{subfigure}{0.235\textwidth} 
        \includegraphics[width=\textwidth]{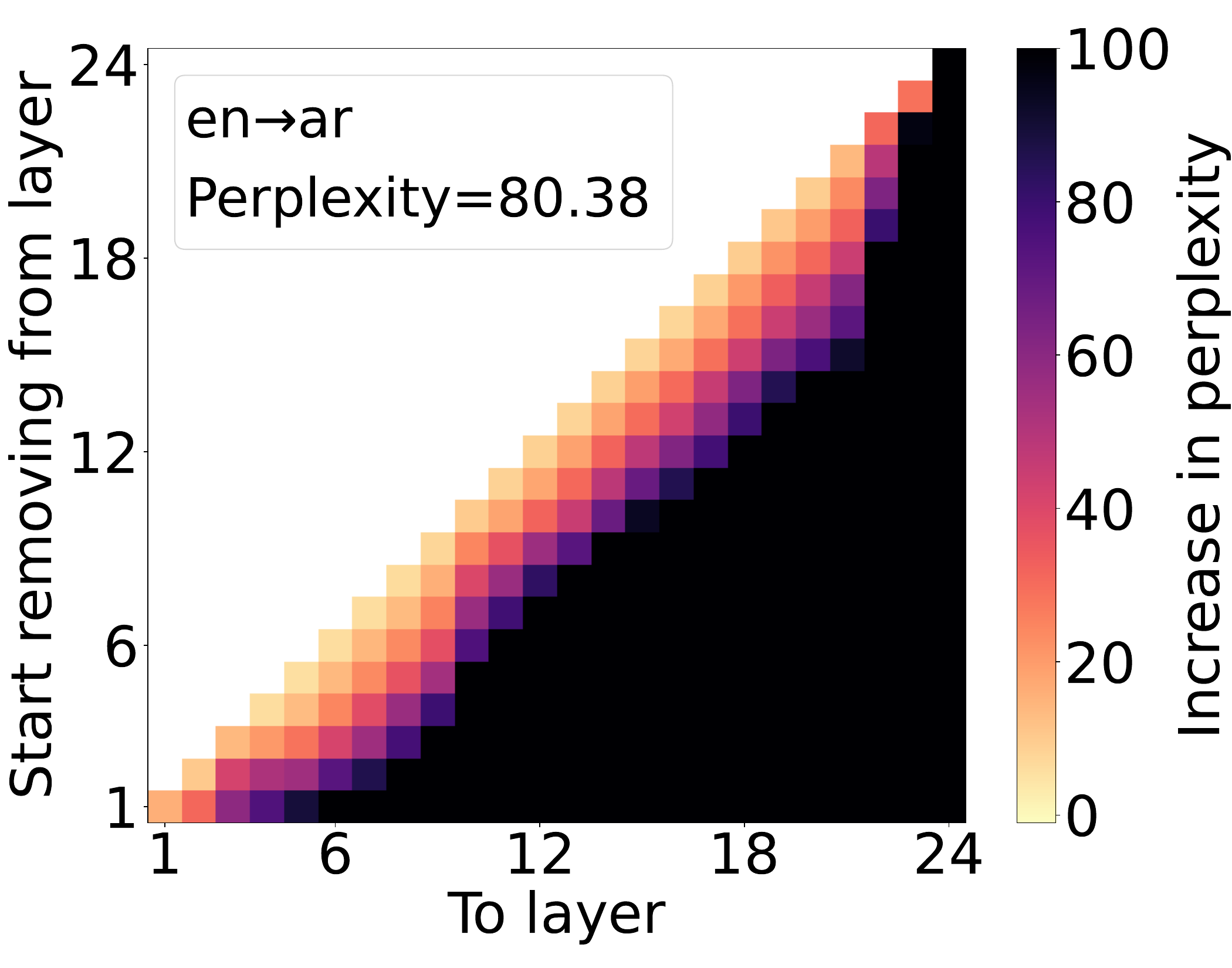}
        \caption{\texttt{en} $\rightarrow$ \texttt{ar}}
        \label{fig:ppl_he}
    \end{subfigure}
    ~
    \begin{subfigure}{0.235\textwidth} 
        \includegraphics[width=\textwidth]{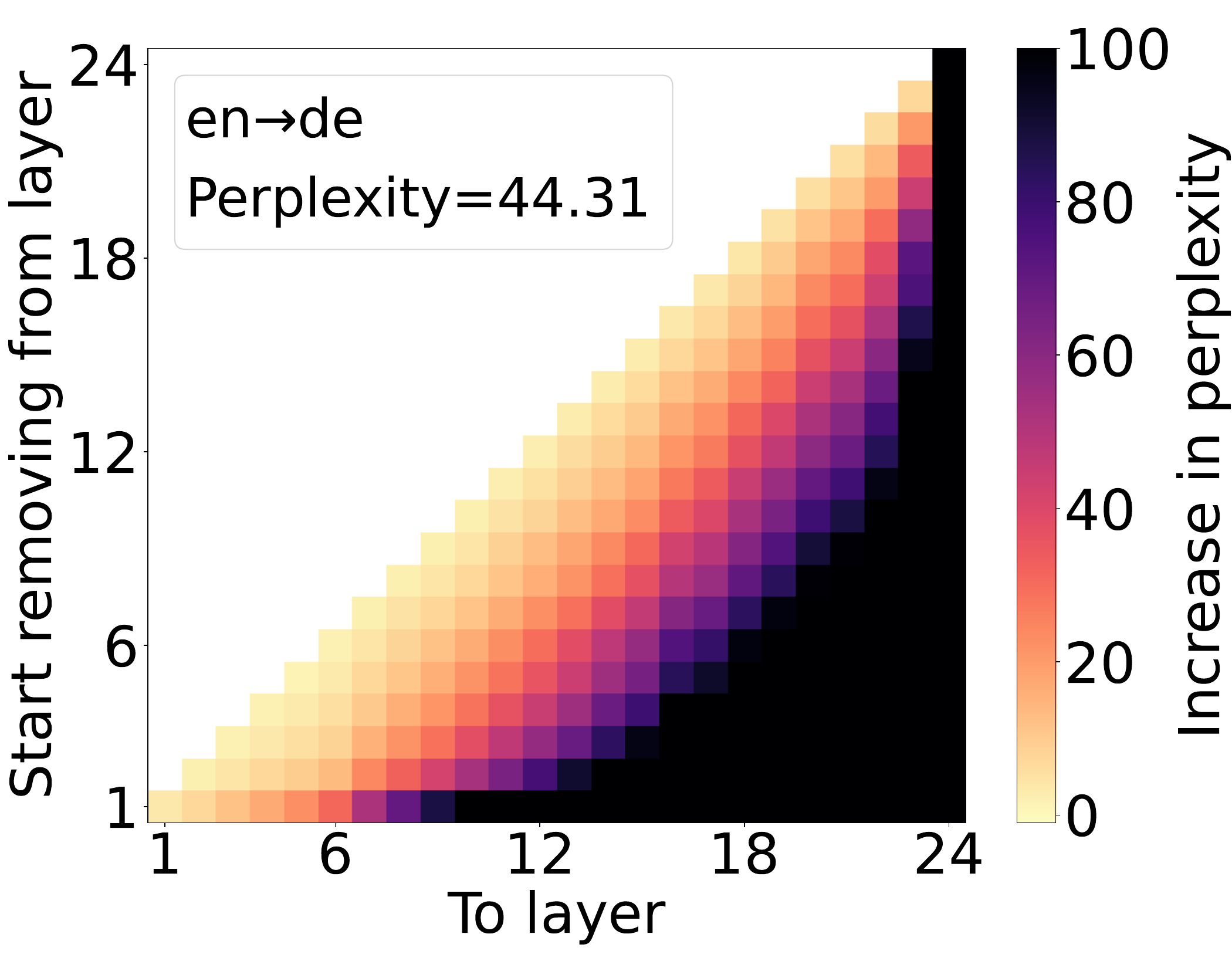}
        \caption{\texttt{en} $\rightarrow$ \texttt{de}}
        \label{fig:ppl_de}
    \end{subfigure}
    ~     
    \begin{subfigure}{0.235\textwidth}
        \includegraphics[width=\textwidth]{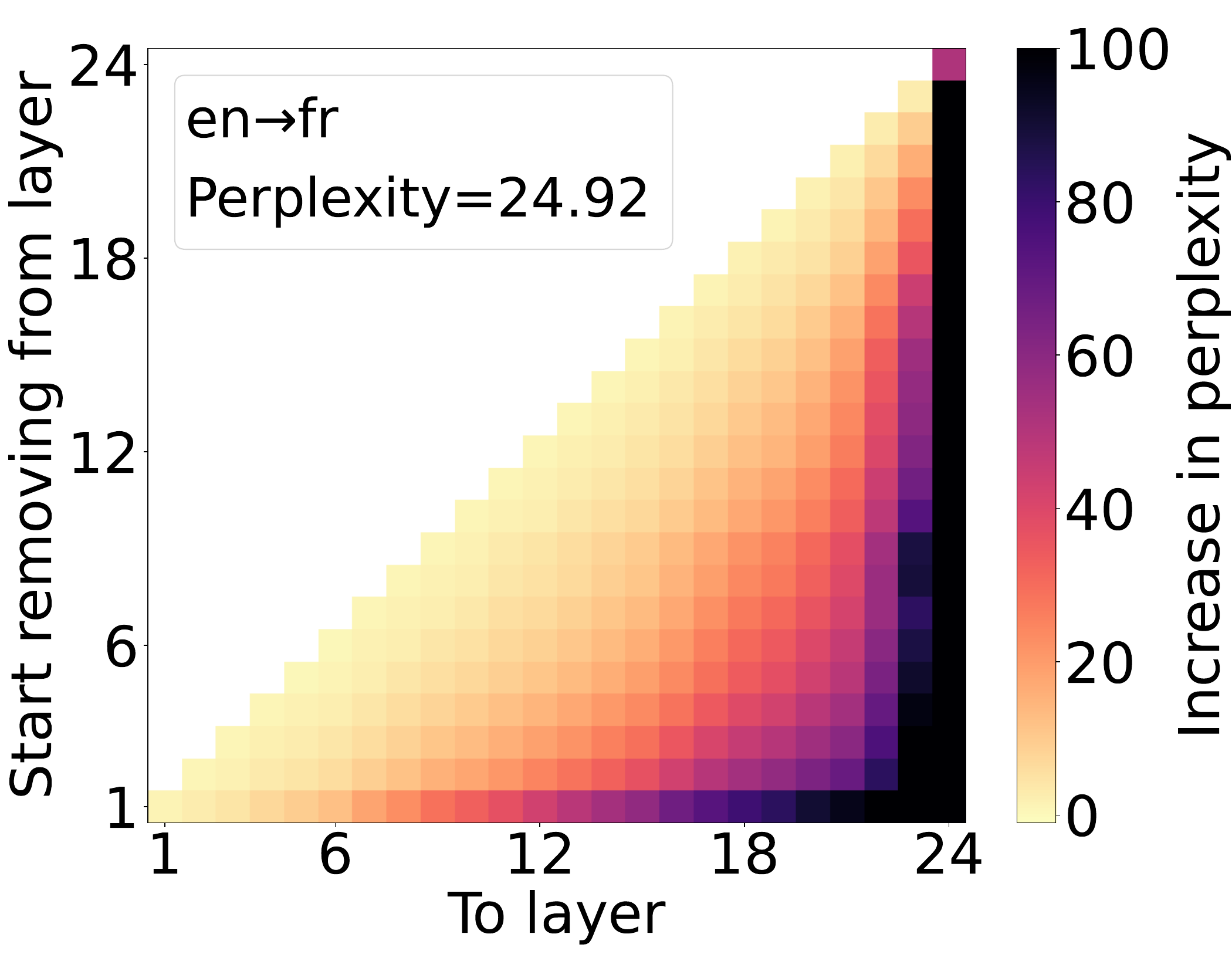}
        \caption{\texttt{en} $\rightarrow$ \texttt{fr}}
        \label{fig:ppl_fr}
    \end{subfigure}
\caption{The increase in perplexity when removing adapter layers during inference for models adapted from \texttt{en} to \texttt{he}, \texttt{ar}, \texttt{de}, \texttt{fr}, respectively. To aid visibility, we cap the increase in perplexity at 100.}
\label{fig:ppl_results}
\end{figure*} 

\subsection{Adapters often can be removed with only a small effect on perplexity} 

To investigate the importance of individual adapter layers, we zero-out the output from single or consecutive adapter layers during inference (similarly to \citet{haviv-etal-2023-understanding, wang2023interpretability}) and measure how this intervention affects perplexity on the held out Wikipedia validation set.
Concretely, for every layer $l'$ and any of its succeeding layers $l''=l'+1,...,L$, we intervene on the inference pass and set the outputs of all the adapters between layers $l''$ and $l'$ to zero, i.e. $\mathbf{x}_{N}^{l, \texttt{out}}\; \xleftarrow{} \mathbf{x}_{N}^{l, \texttt{ffn}} \;\;\forall l\in [l', l'']$.

\Cref{fig:ppl_ar,fig:ppl_he,fig:ppl_de,fig:ppl_fr} show the average increase in validation perplexity of our adapted models for every intervention and each target language. We observe consistent results when adapting with LoRA and also when adapting mBERT, a massively multilingual model, and discuss these results in more detail in \Cref{sec:appendix:adapter_drop_mbert}. We observe that removing individual layers typically has a minor impact on perplexity, except for the last two layers where perplexity dramatically increase to over 100 across all languages. This effect is consistent with our findings in \S\ref{sec:think}, where we observe the largest increase in target tokens at the last layer.

The increase in perplexity becomes even more pronounced as the number of removed adapters increases and when considering target languages that are more distant from the source language. For example, for most layers, removing three consecutive adapters leads to an increase of up to 25 points perplexity for Arabic and Hebrew compared to less than 10 points for German and French.\looseness-1 

This suggests that adapting from English to Hebrew or Arabic is more difficult than adaptation to German or French as individual adapters (and especially small groups of adapters) have a more profound impact on the adaptation performance. Stated differently, we hypothesize that for language more distant from the source language, adapters have to ``do more work'', therefore, removing them leads to larger drops in performance.

\section{Two adaptation hypotheses}
\label{sec:hypotheses}

Our experiments so far have analyzed how LM adapters influence the predictions of the underlying model, making several key observations: 1) adapted LM predictions are evolved in the source language distribution; decoding the output distribution from most hidden layers shows that source language tokens are substantially more pronounced than target language tokens, until reaching the very last layers where the target language abruptly becomes pronounced; 2) the adaptation process is gradual across most layers, as it is possible to skip multiple adapters without decrease in performance, while the last few adapters are crucial for adaptation success; 3) the adaptation of our pretrained model is notably better for French and German than for Hebrew and Arabic. This is reflected in the larger norm of the adapter updates for these two languages and the bigger impact on perplexity when removing chunks of adapters. We attribute this finding to the fact that English shares a linguistic and etymological connection with German and French. In contrast, for Arabic and Hebrew, which utilize non-Latin scripts, adaptation appears to be more difficult due to the difference in linguistic and orthographic characteristics.

From these findings, we conclude that the adaptation process during the forward pass is distributed across all layers, with each individual layer---except the last few layers--- making gradual updates to the overall adaptation. In what follows, we set out to investigate two alternative hypotheses that could explain the interplay between the adapter updates and the underlying prediction space.

\paragraph{Hypothesis 1: Adapters operate in an isolated subspace}

We hypothesize that adapters operate in a specific subspace of the model's residual stream which is not utilized by the underlying model, leaving most of the representation space untouched. This isolated representation space then becomes pronounced in the last few layers, leading to the prediction of tokens in the target language. 

\paragraph{Hypothesis 2: Adapters operate on top of the model's representation space}

An alternative hypothesis is that adapters operate on top of the structure in the underlying model's representation space, preserving its overall structure while gradually pushing representations closer to the embedding space of the target language.

\begin{figure}[t]
\setlength{\belowcaptionskip}{-2pt}
  \centering
    \begin{subfigure}{0.45\columnwidth}
        \includegraphics[width=\textwidth]{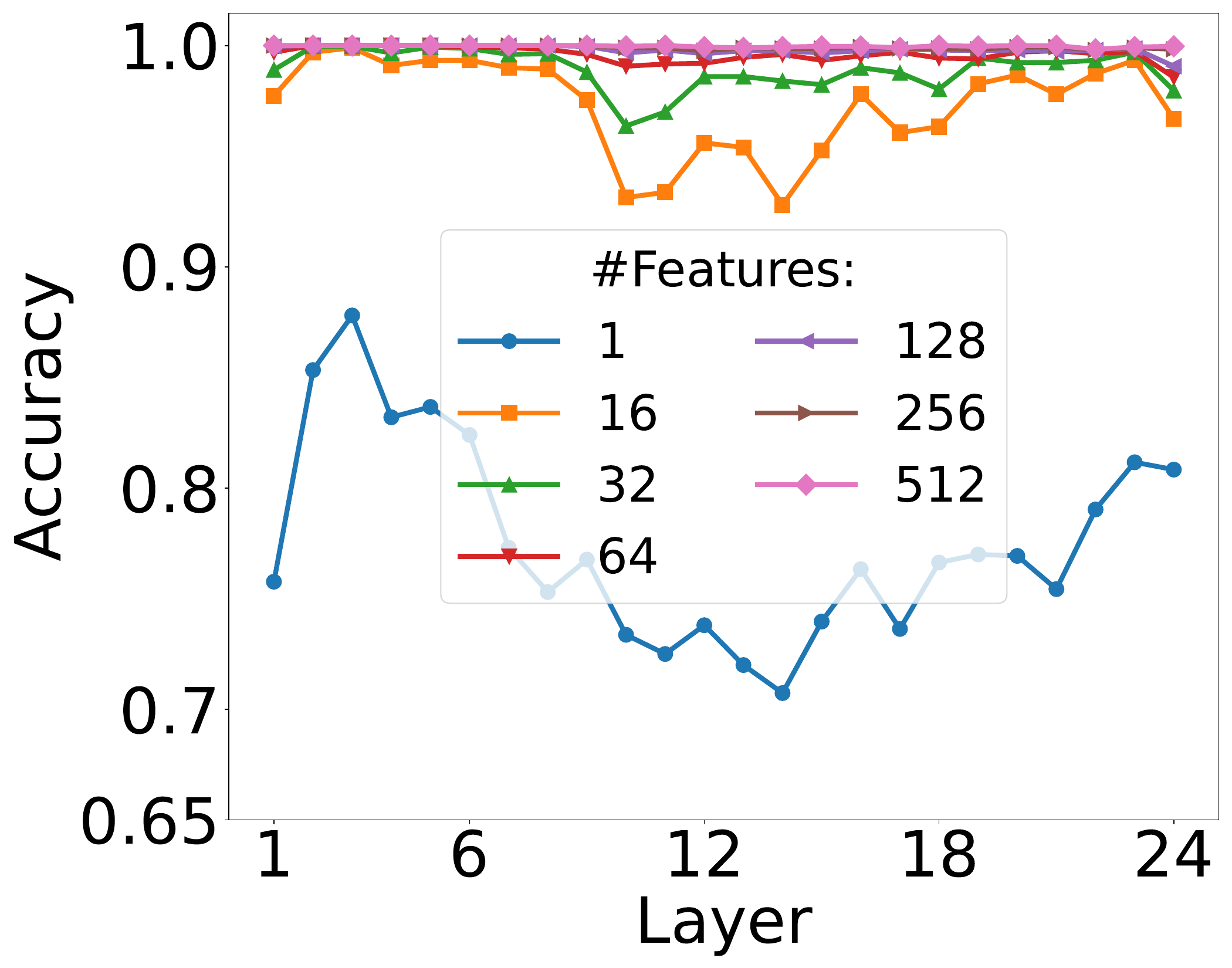}
        \caption{\texttt{en} $\rightarrow$ \texttt{he}}
        \label{fig:probe_acc_de}
    \end{subfigure}
    ~
    \begin{subfigure}{0.45\columnwidth}
        \includegraphics[width=\textwidth]{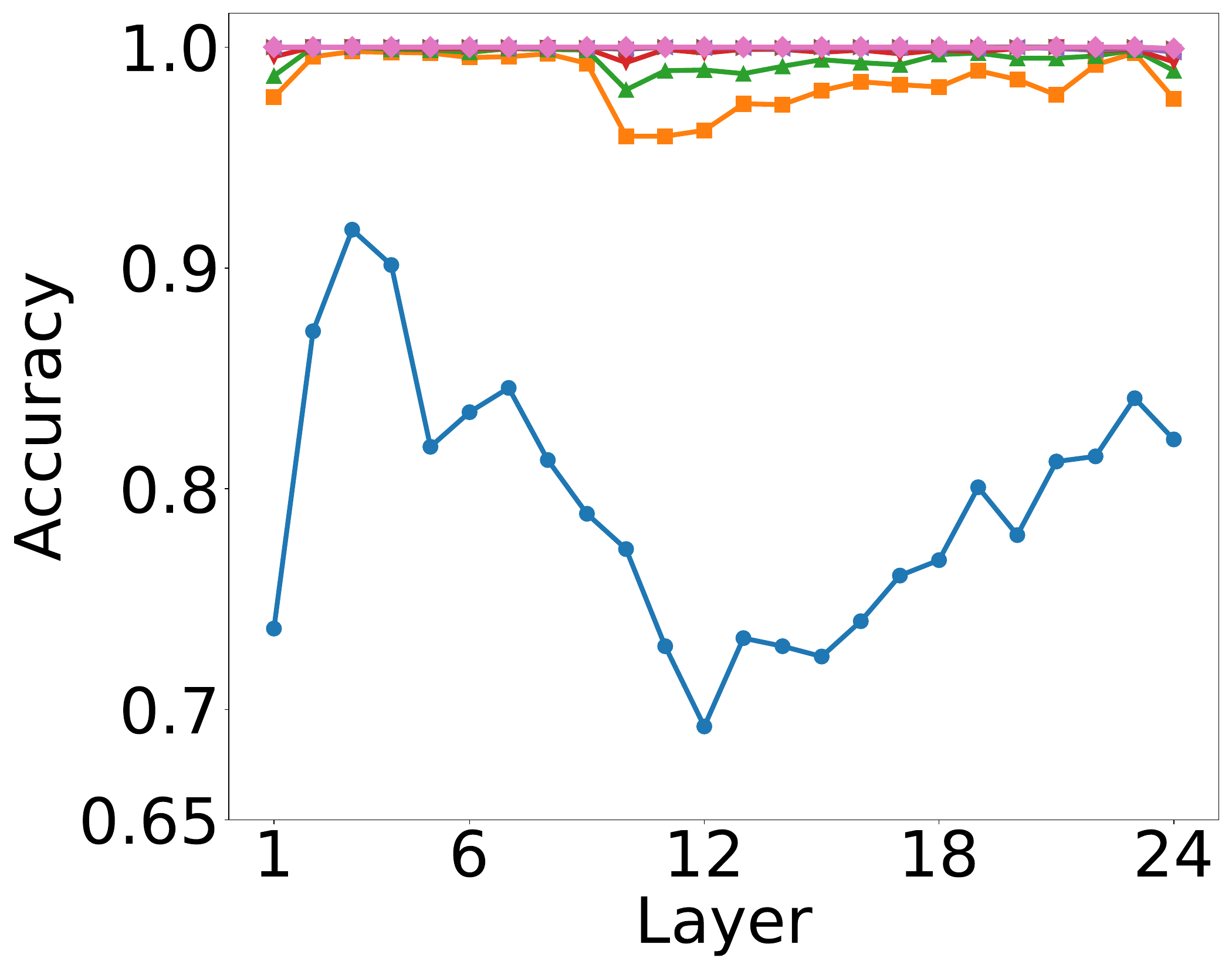}
        \caption{\texttt{en} $\rightarrow$ \texttt{ar}}
        \label{fig:probe_acc_fr}
    \end{subfigure}
    \\
    \begin{subfigure}{0.45\columnwidth}
        \includegraphics[width=\textwidth]{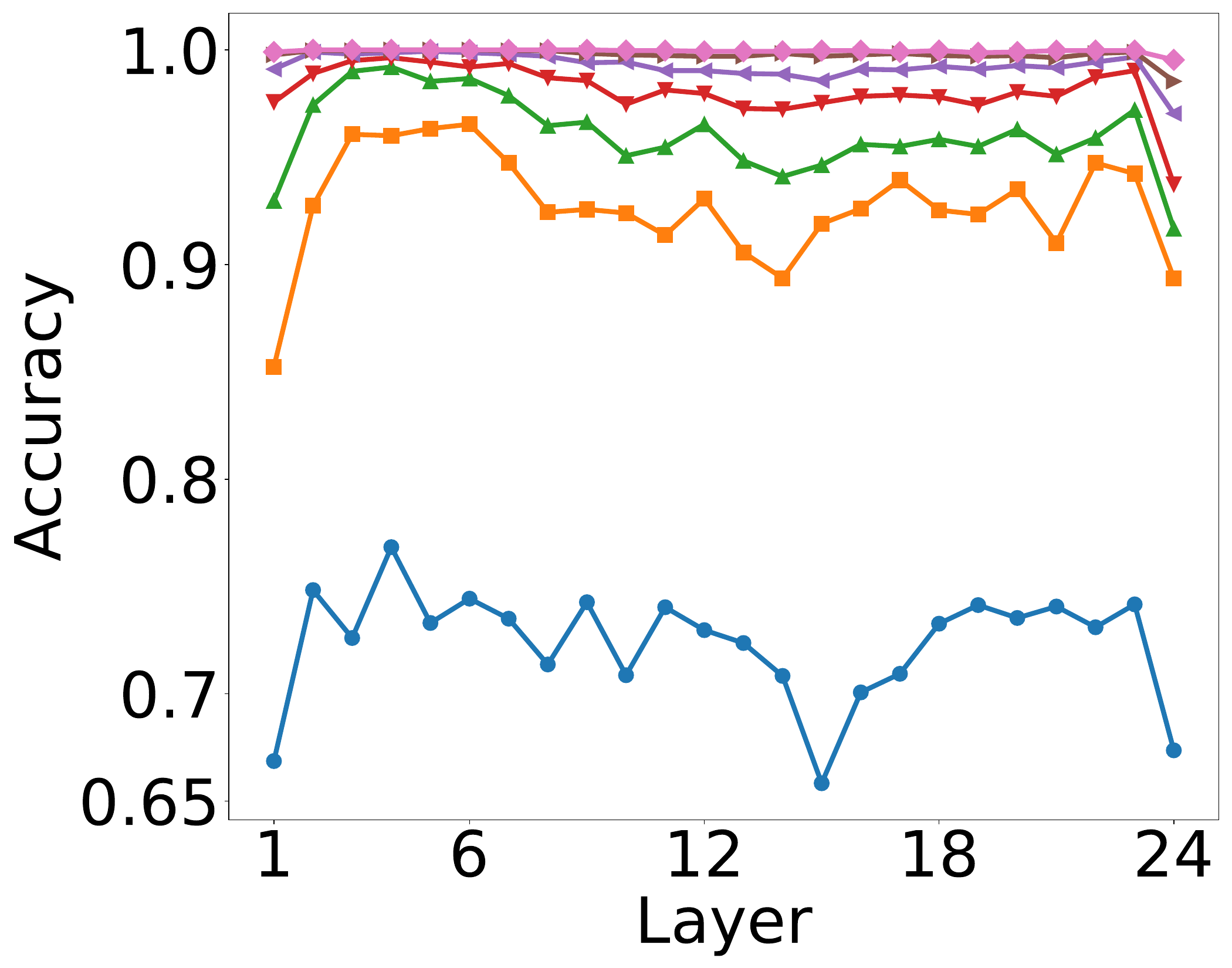}
        \caption{\texttt{en} $\rightarrow$ \texttt{de}}
        \label{fig:probe_acc_he}
    \end{subfigure}
    ~
    \begin{subfigure}{0.45\columnwidth}
        \includegraphics[width=\textwidth]{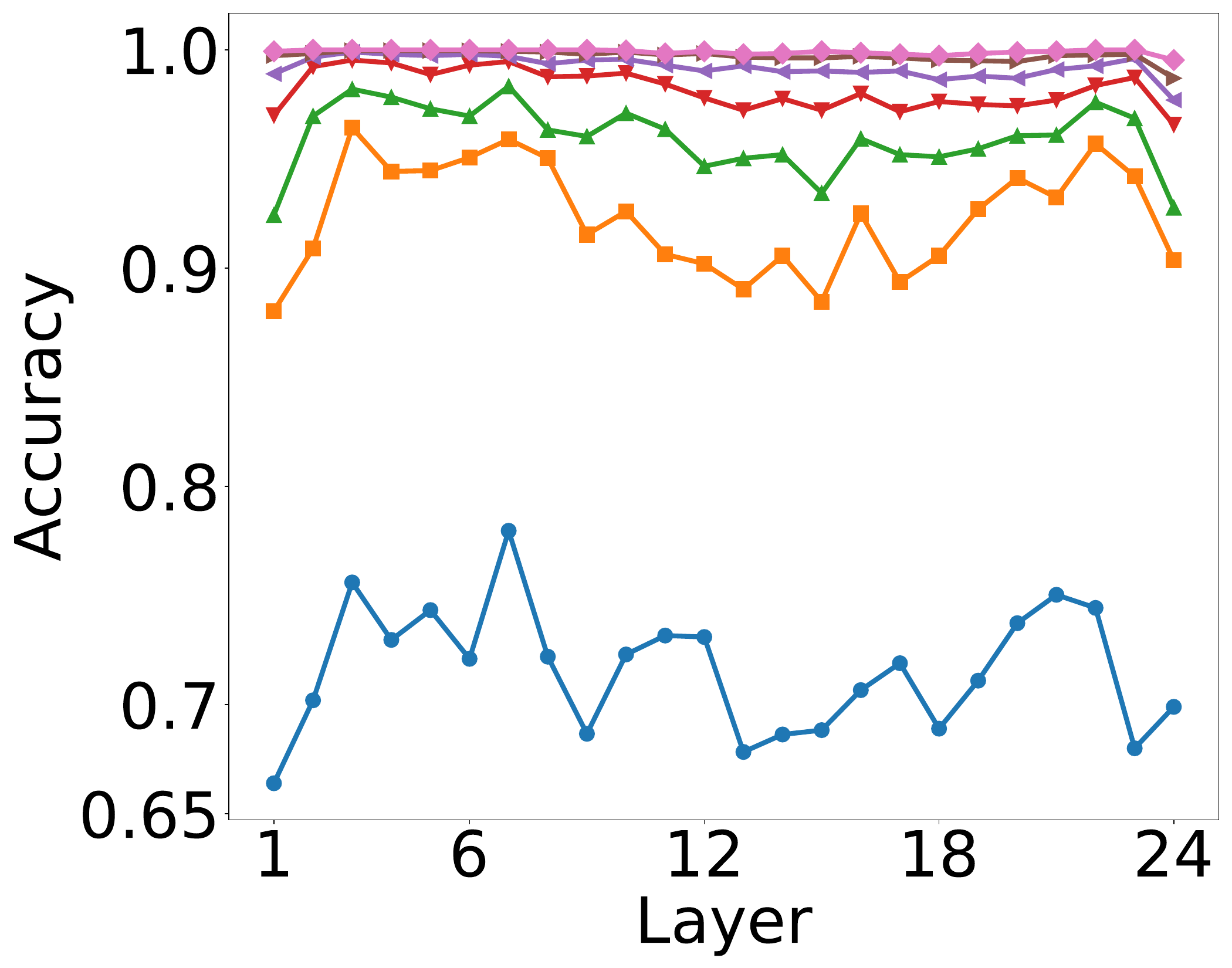}
        \caption{\texttt{en} $\rightarrow$ \texttt{fr}}
        \label{fig:probe_acc_ar}
    \end{subfigure}
\caption{Sparse probing classifiers can detect language adaptation with high accuracy.
}
\label{fig:probe_acc}
\end{figure} 

\subsection{Do adapters operate in an isolated subspace of the residual stream?}

We approach the first hypothesis via a sparse probing experiment. Specifically, we train probing classifiers to predict whether a given layer output has been adapted or not. In a first step, we follow \citet{gurnee2023finding} and identify critical features for adaptation using the maximum mean differences (MMD) algorithm which ranks features based on their importance for distinguishing between adapted and non-adapted features (a detailed description is provided in \Cref{sec:subspace_id}). Next, we take the top-k most important features identified by MMD and use them as input features for a binary logistic regression classifier which we train to predict whether or not a representation has been adapted at that layer. 

\Cref{fig:probe_acc} shows that for various levels of probe sparsity, a linear probing classifier can predict with high accuracy whether a given hidden representation has been updated by a specific adapter or not. Interestingly, across all levels of sparsity, adaptation is easier to predict for Hebrew and Arabic than for German and French. This provides further evidence that adaptation is more pronounced on the underlying model representations for these languages.\looseness-1 

Next, we intervene on the features identified by MMD to investigate the extend to which these features are involved in predicting tokens from the target language. We zero-out individual dimensions of the adapter outputs which correspond to the most important features identified by MMD and compare the target language perplexity on our adaptation validation split before and after intervention.\footnote{Replacing the value of individual dimensions by the average of other feature directions leads to very similar results (see \Cref{sec:appendix:intervention}).} As baselines, we intervene on the least important features identified by MMD as well as on randomly selected features.

\begin{figure}[t]
\setlength{\belowcaptionskip}{-2pt}
  \centering
    \begin{subfigure}{0.45\columnwidth}
        \includegraphics[width=\textwidth]{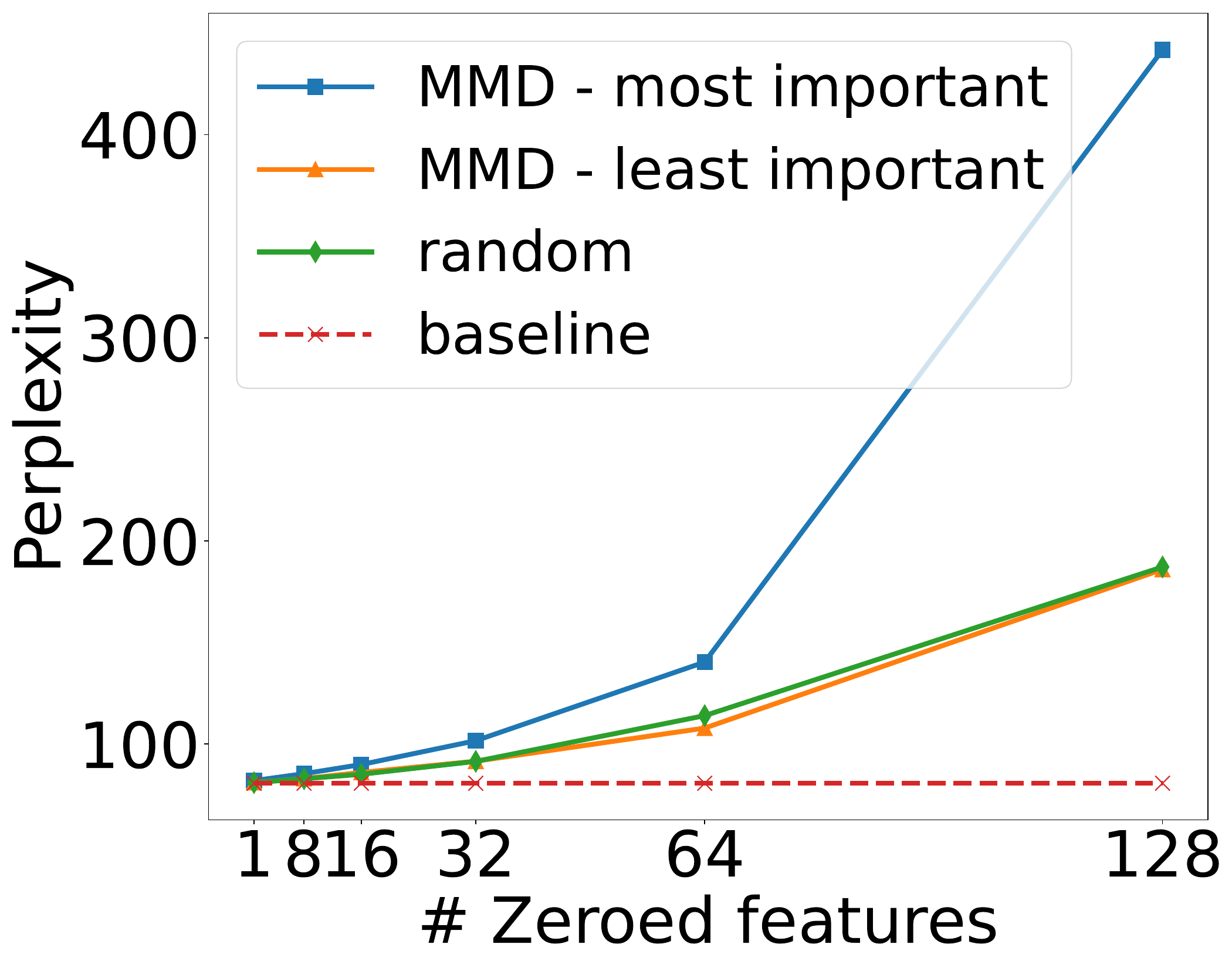}
        \caption{\texttt{en} $\rightarrow$ \texttt{he}}
        \label{fig:zero-intervention-he}
    \end{subfigure}
    ~
    \begin{subfigure}{0.45\columnwidth} 
        \includegraphics[width=\textwidth]{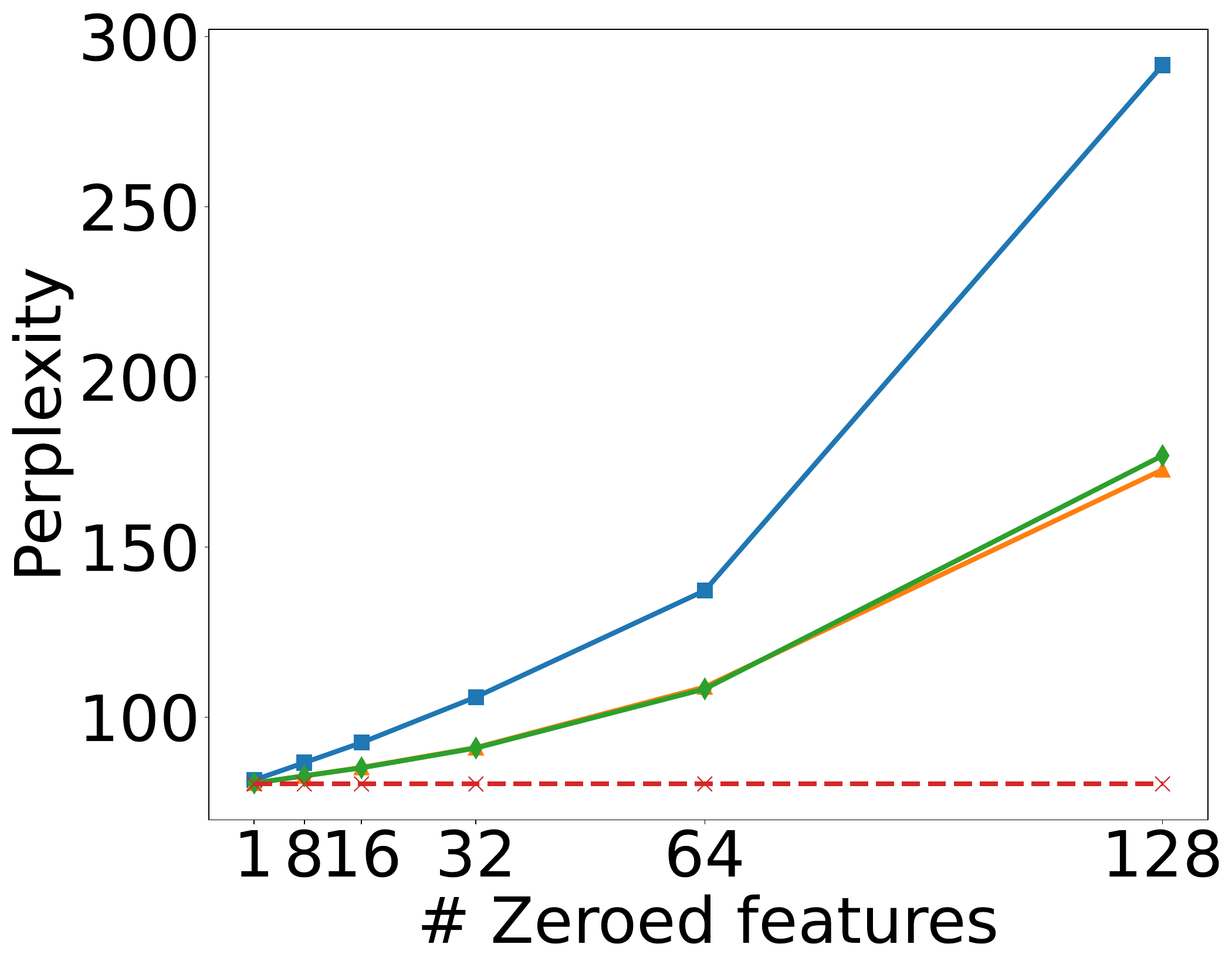}
        \caption{\texttt{en} $\rightarrow$ \texttt{ar}}
        \label{fig:zero-intervention-ar}
    \end{subfigure}
    \\
    \begin{subfigure}{0.45\columnwidth}
        \includegraphics[width=\textwidth]{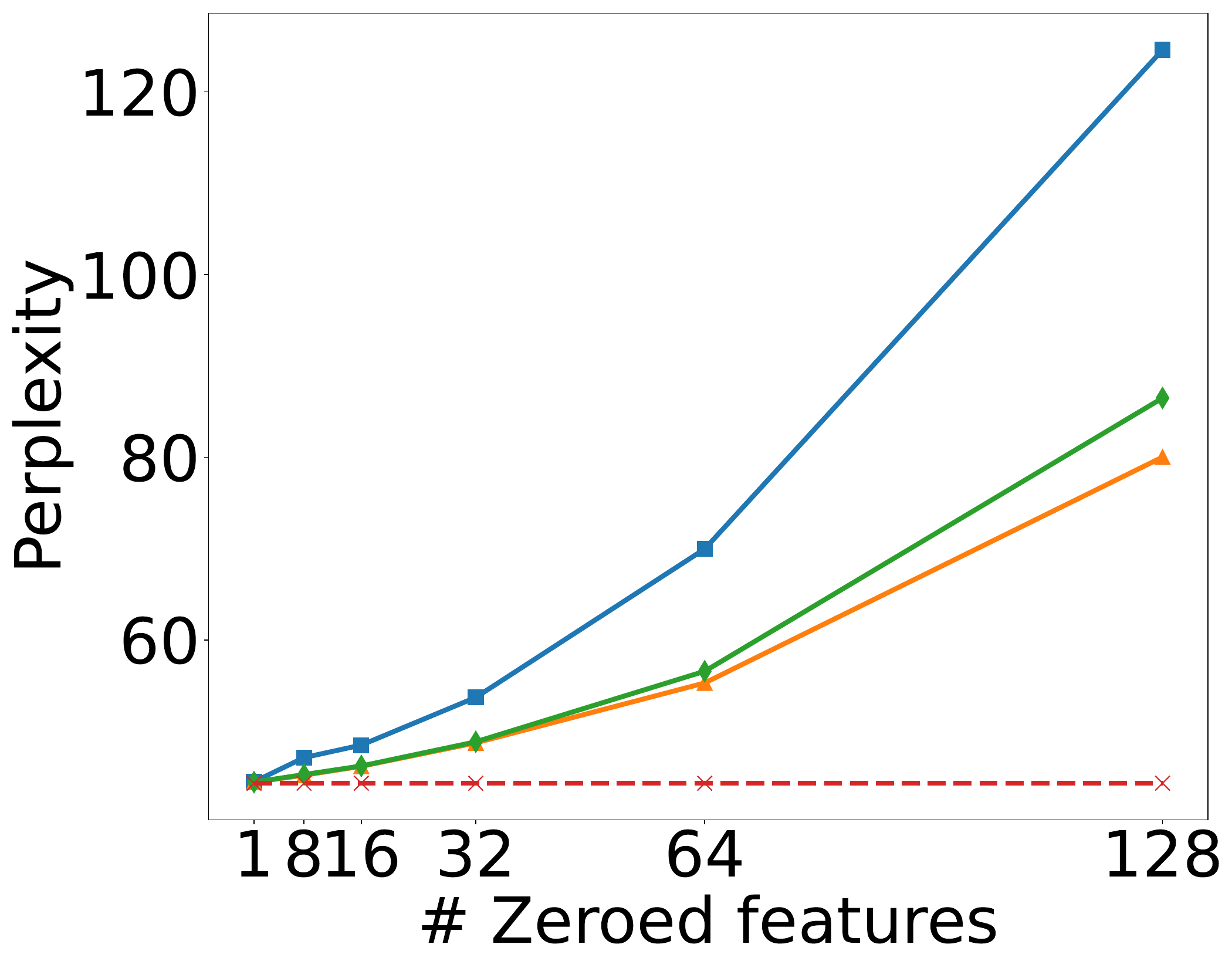}
        \caption{\texttt{en} $\rightarrow$ \texttt{de}}
       \label{fig:zero-intervention-de}
    \end{subfigure}
    ~
    \begin{subfigure}{0.45\columnwidth} 
        \includegraphics[width=\textwidth]{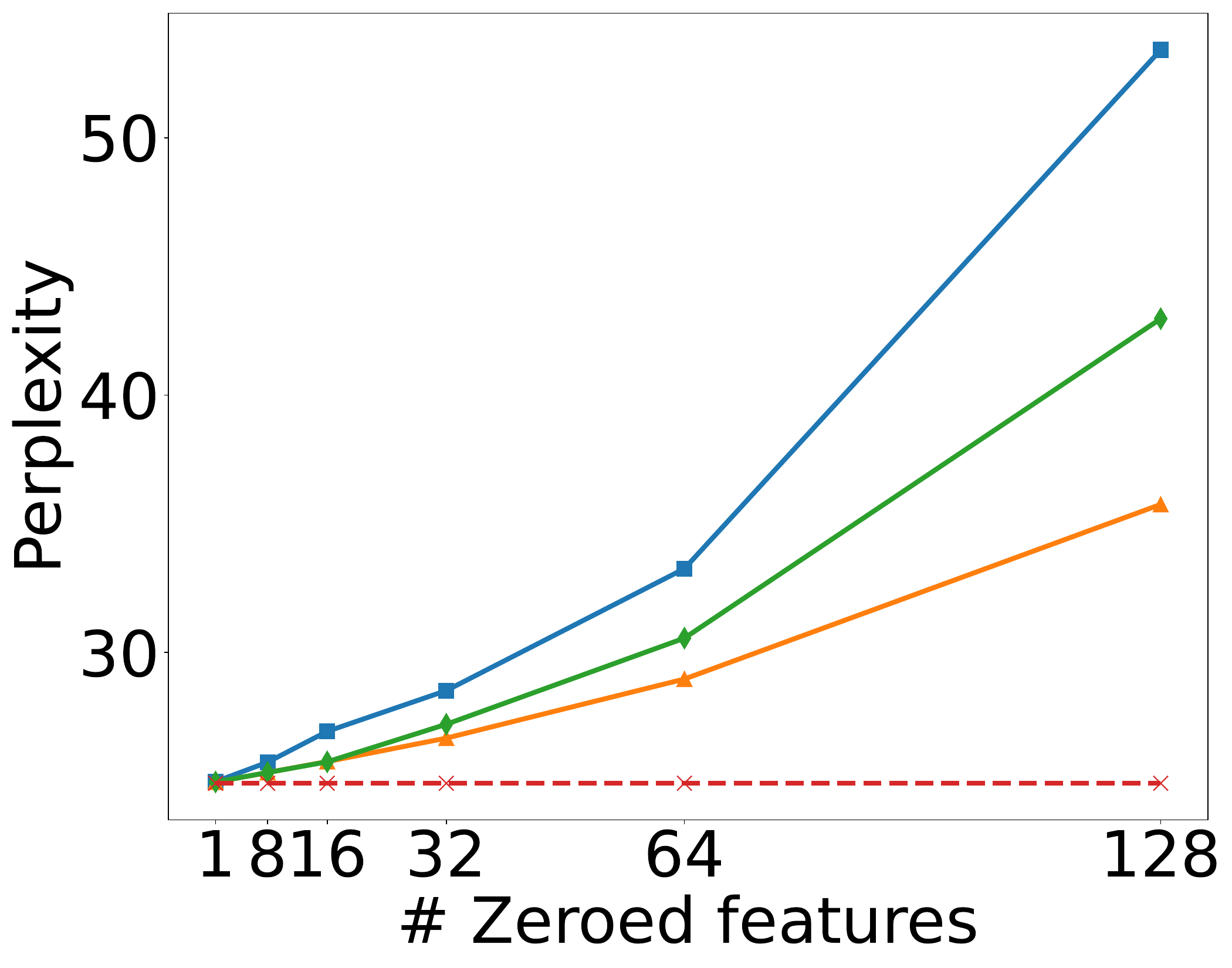}
        \caption{\texttt{en} $\rightarrow$ \texttt{fr}}
        \label{fig:zero-intervention-fr}
    \end{subfigure}
\caption{Adapted model perplexity after intervention.
}
\label{fig:interventions}
\end{figure}

\Cref{fig:interventions} depicts the adapted model perplexity for increasing number of intervened features. While intervening on the most important features indeed leads to a dramatic increase in perplexity on all target languages, intervening on the least important or even on random features also leads to a substantial increase in perplexity.

\paragraph{Refuting Hypothesis 1}

From the findings above we refute the hypothesis that adapters operate in an isolated subspace of the residual stream of the underlying model, as even intervening on  random features leads to a substantial increase in perplexity. We note that adapters might still operate in a subspace, however, our findings show that they do not operate in an isolated subspace which is uniquly used by the adapters.

\begin{figure*}[t]
\setlength{\belowcaptionskip}{-1px}
    \centering    
    \begin{subfigure}{0.25\textwidth}
        \includegraphics[width=\textwidth]{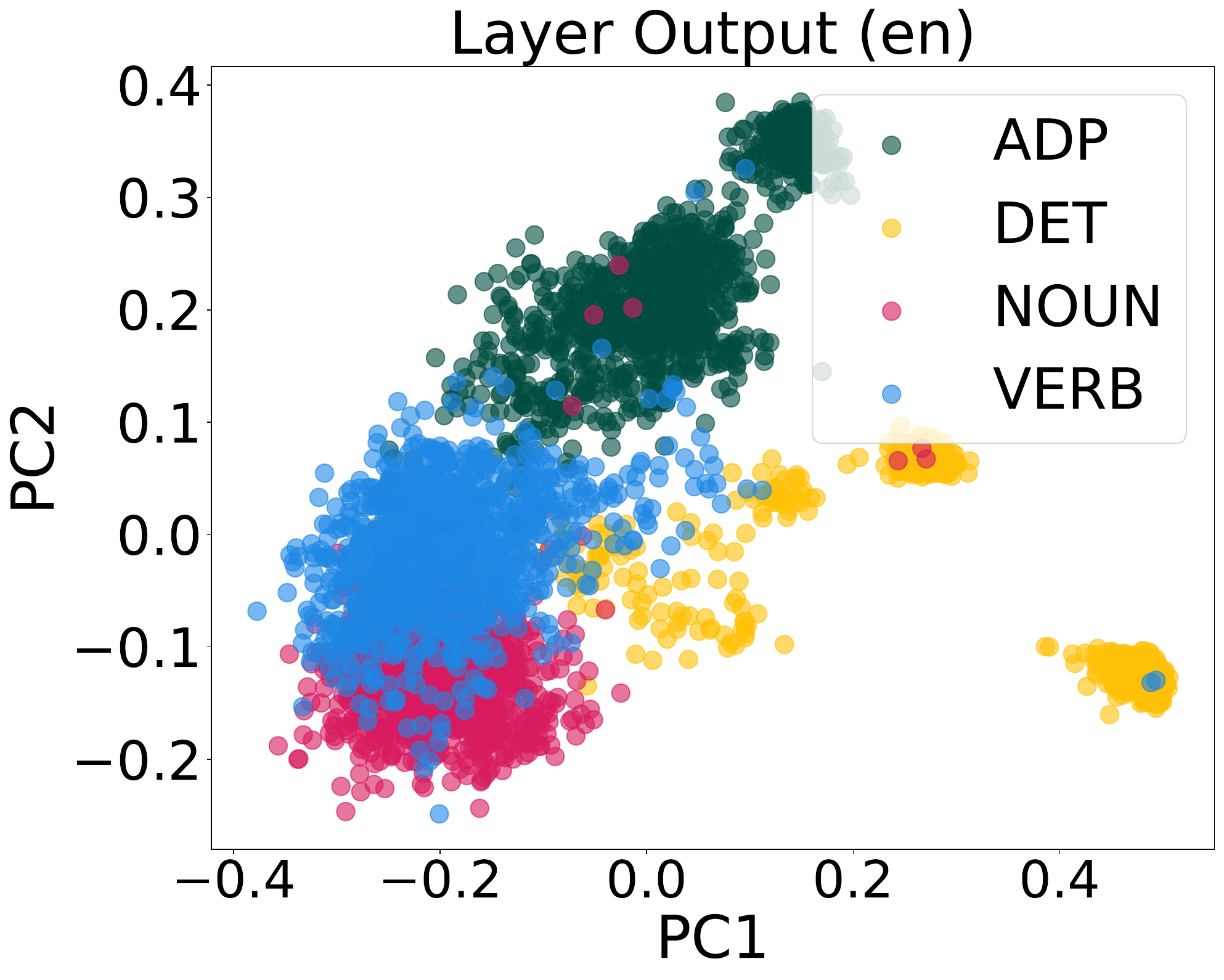}
        \caption{Layer 6}
    \end{subfigure}
    ~
    \begin{subfigure}{0.25\textwidth}
        \includegraphics[width=\textwidth]{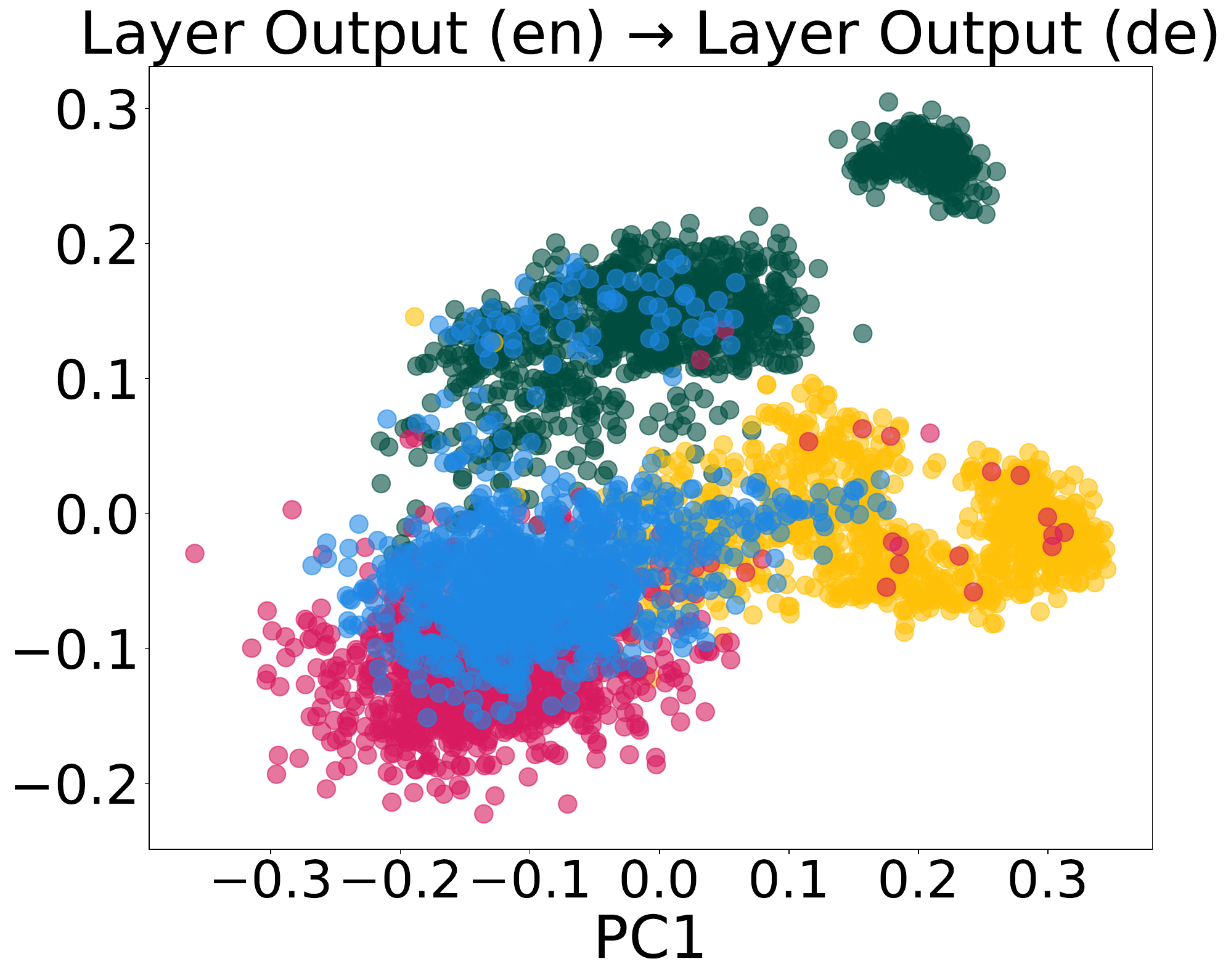}
        \caption{Layer 6}
    \end{subfigure}
    ~
    \begin{subfigure}{0.25\textwidth}
        \includegraphics[width=\textwidth]{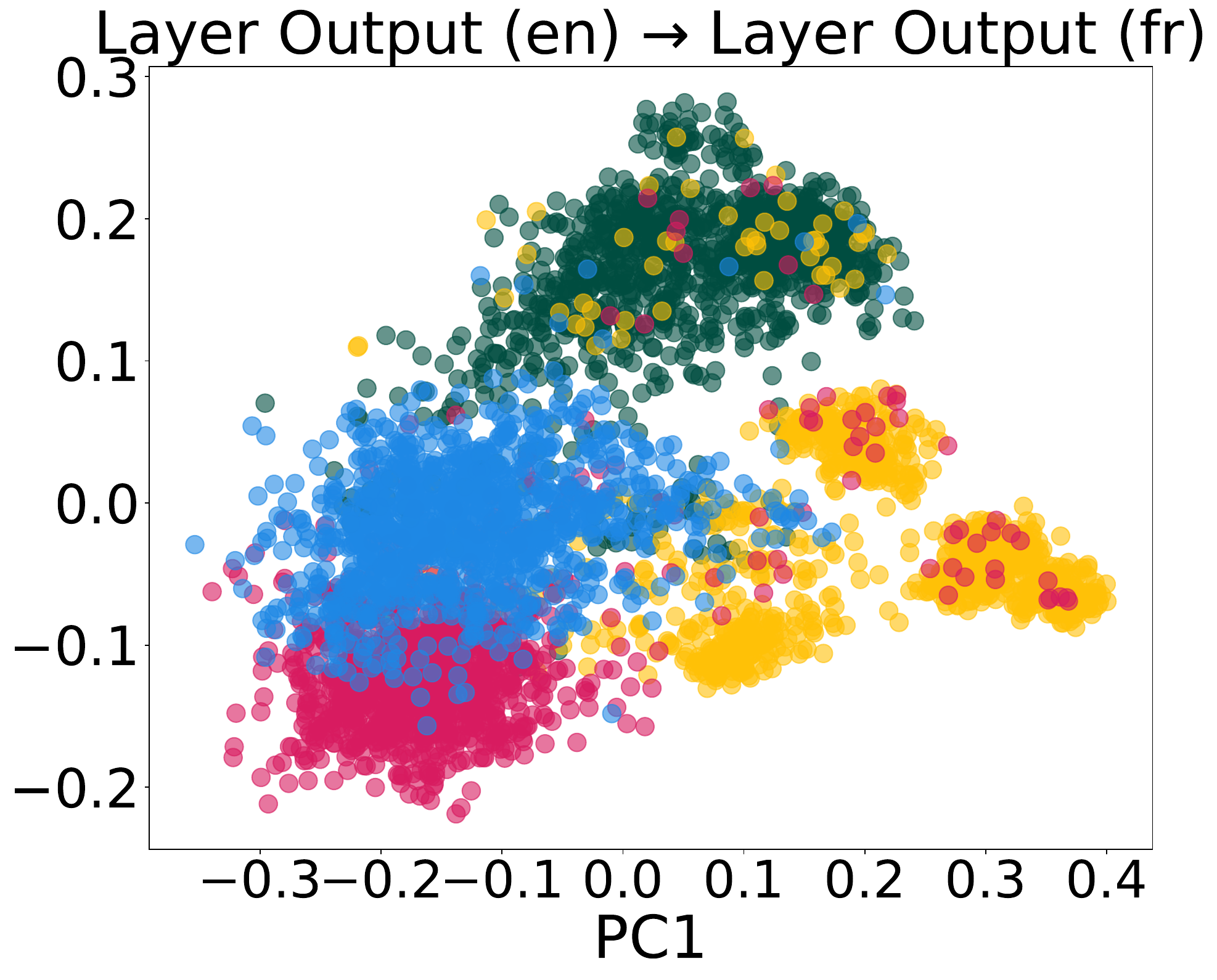}
        \caption{Layer 6}
    \end{subfigure}
    \\
    \begin{subfigure}{0.25\textwidth}
        \includegraphics[width=\textwidth]{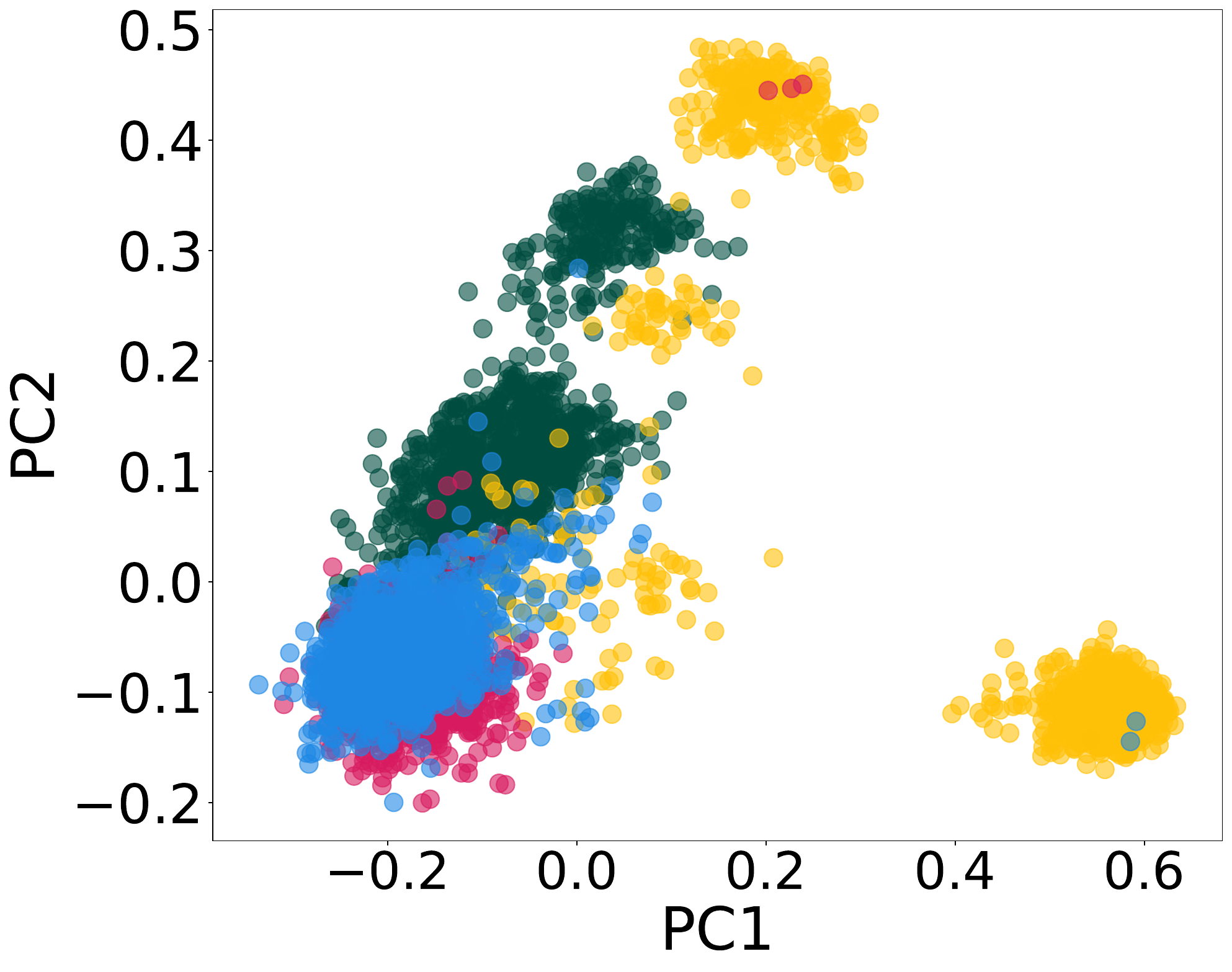}
        \caption{Layer 12}
    \end{subfigure}
    ~
    \begin{subfigure}{0.25\textwidth}
        \includegraphics[width=\textwidth]{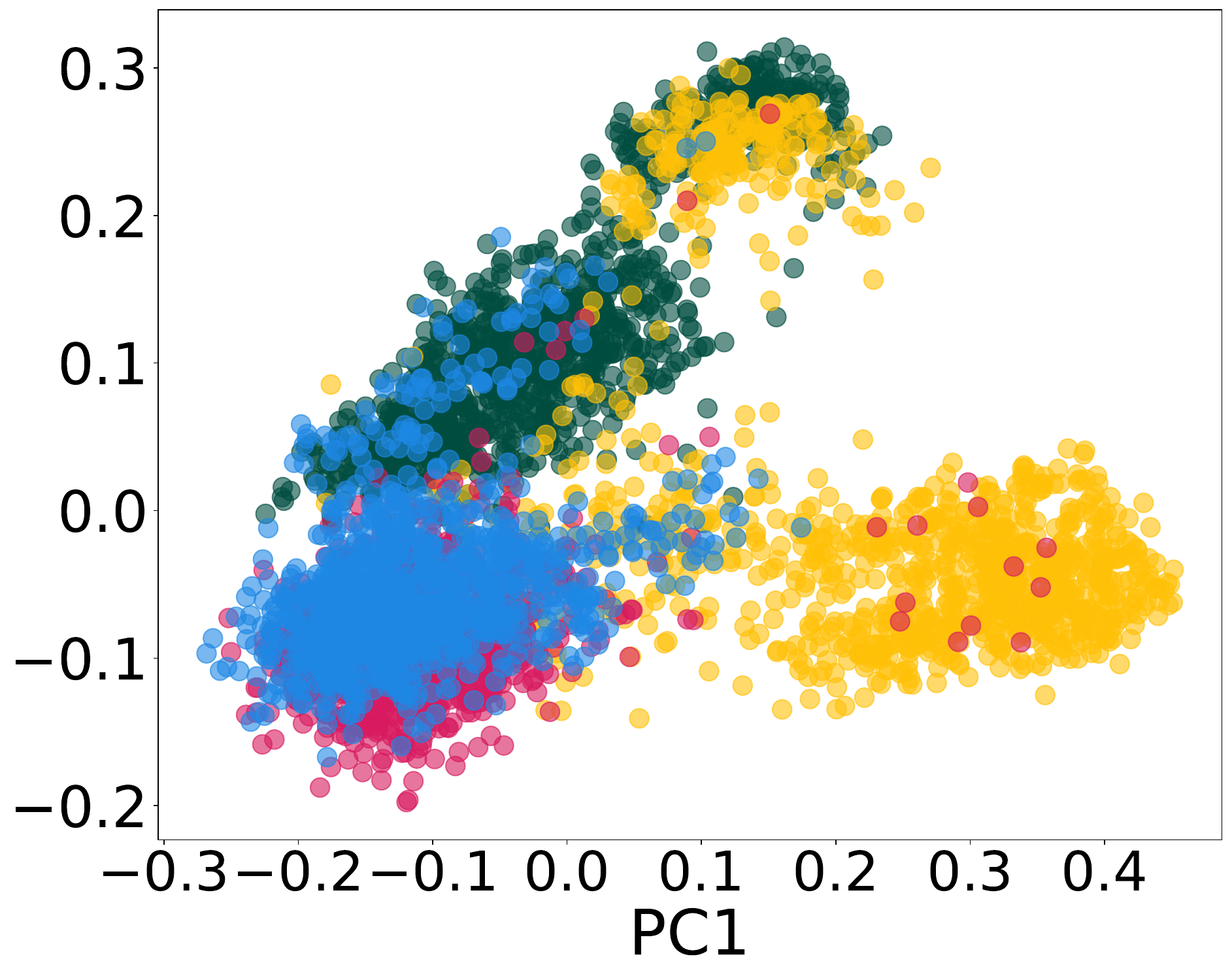}
        \caption{Layer 12}
    \end{subfigure}
    ~
    \begin{subfigure}{0.25\textwidth}
        \includegraphics[width=\textwidth]{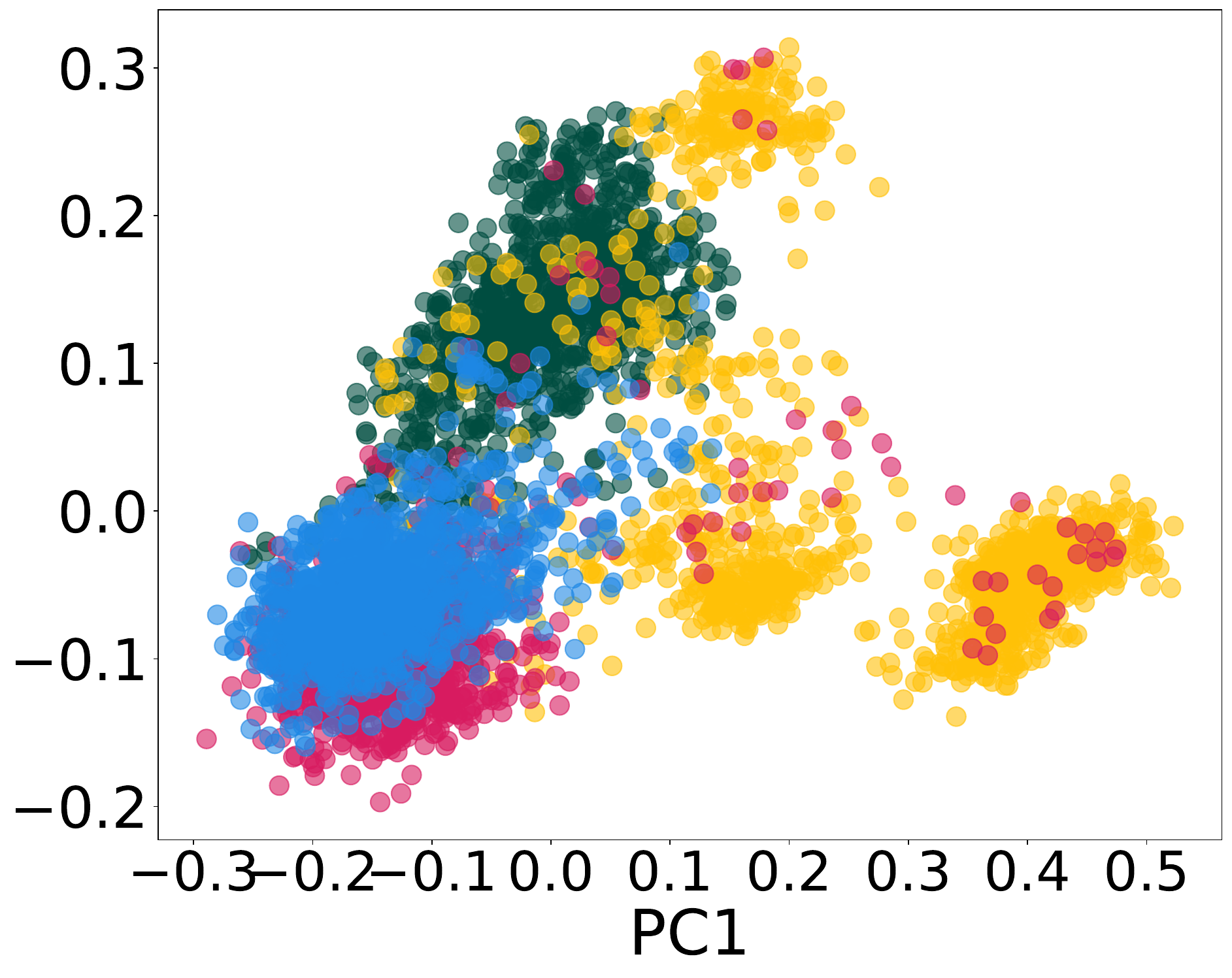}
        \caption{Layer 12}
    \end{subfigure}
    \\
    \begin{subfigure}{0.25\textwidth}
        \includegraphics[width=\textwidth]{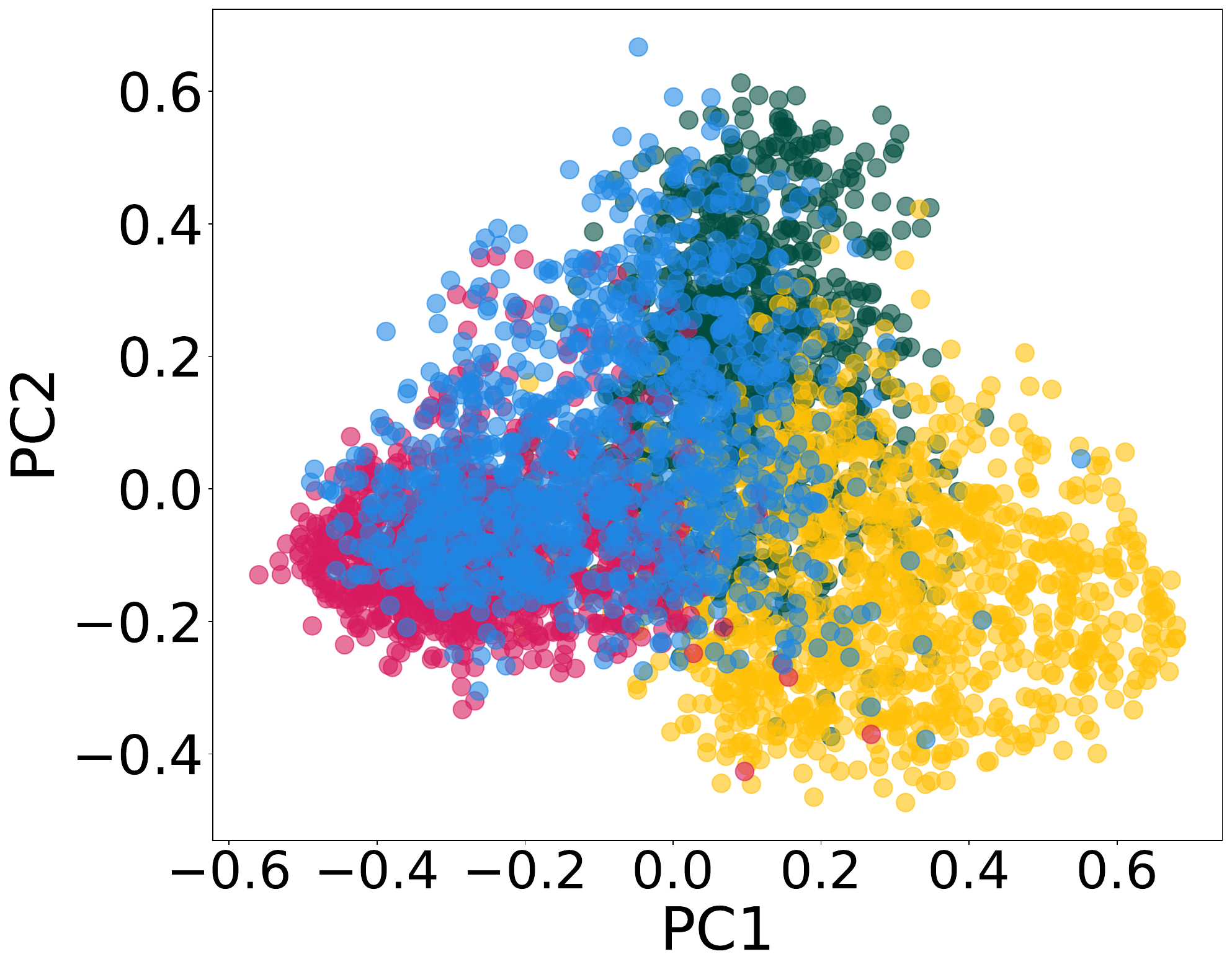}
        \caption{Layer 24}
    \end{subfigure}
    ~
    \begin{subfigure}{0.25\textwidth}
        \includegraphics[width=\textwidth]{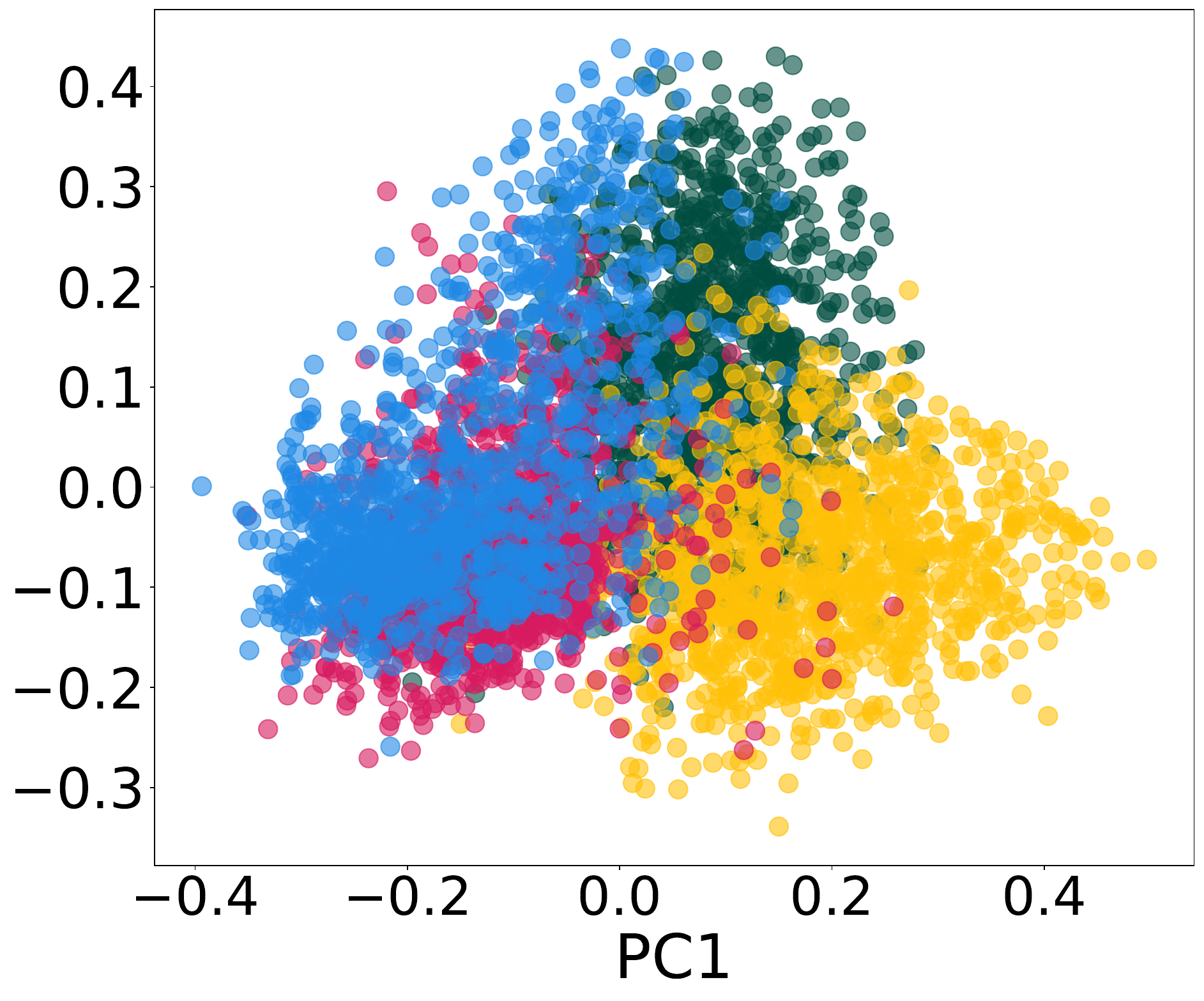}
        \caption{Layer 24}
    \end{subfigure}
    ~
    \begin{subfigure}{0.25\textwidth}
        \includegraphics[width=\textwidth]{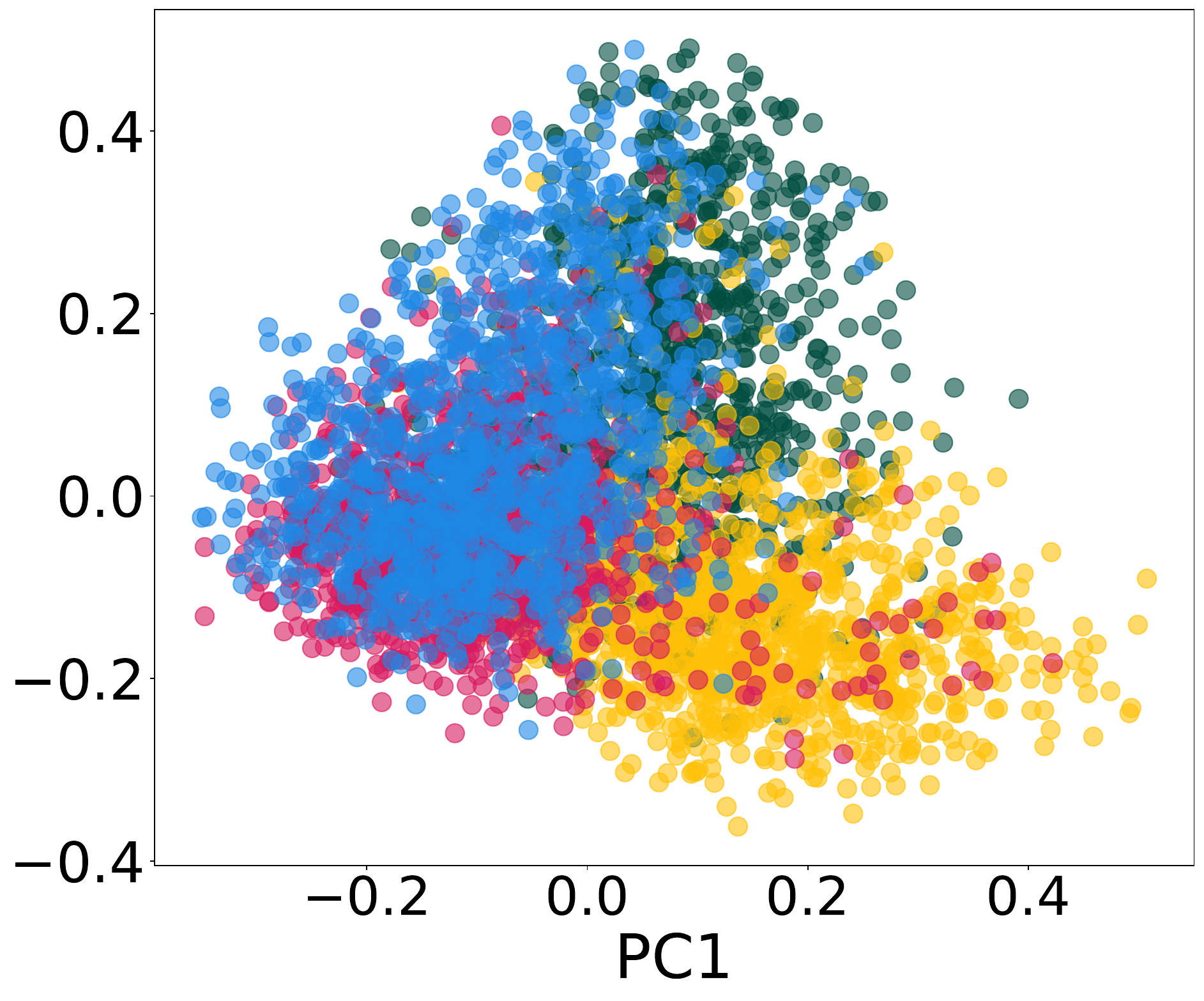}
        \caption{Layer 24}
    \end{subfigure}
    \caption{2D projections for tokens with different POS of the model pre-trained on English (first column) and the adapted models trained on German (second column) and French (third column) at various layers. In all three cases, the projection matrix is computed via PCA on the English representations only.}
    \label{fig:pos_pca_de}
\end{figure*}

\subsection{Adapters largely preserve the structure of the underlying model}
\label{sec:discussion}

To approach the second hypothesis, we compare the structure of the residual stream in adapted versus non-adapted predictions with respect to different linguistic properties. Specifically, we analyze the structure corresponding to part of speech (adposition/determiner/noun/verb), verb tense (past/present), and grammatical number (singular/plural). Part of speech (POS) labels were obtained from PUD \citep{zeman-etal-2017-conll} and data for verb tense and grammatical number where obtained from the data released by \citet{Acs_Kornai_2023}.

Given a property, we first obtain token representations from the monolingual model for 1100 inputs in English which correspond to different values of this property. 
For every layer $l$, we run PCA on the hidden representations and create a projection matrix $P^{l}\in \mathcal{R}^{2\times d}$ consisting of the first two principal components. Next, we apply the projection matrix obtained from the English representations to layer output representations from the adapted models for German and French inputs,
again focusing on representations of tokens with different values of the property.\looseness-1

\Cref{fig:pos_pca_de} visualizes the projection results for different layers of the models for POS. Results for additional layers, with LoRA, as well as tense and number all show very consistent trends and are discussed in \Cref{sec:appendix:overlap}. Focusing on the English representation space first (first column), the POS structure is clearly visible across all layers. Interestingly, using the same projection matrix derived from the English layer outputs, reveals a highly similar structure in the adapted representation space for German and French (second and third columns).

\begin{figure}[t]
\setlength{\belowcaptionskip}{-4pt}
  \centering
    \begin{subfigure}{0.48\columnwidth}
        \includegraphics[width=\textwidth]{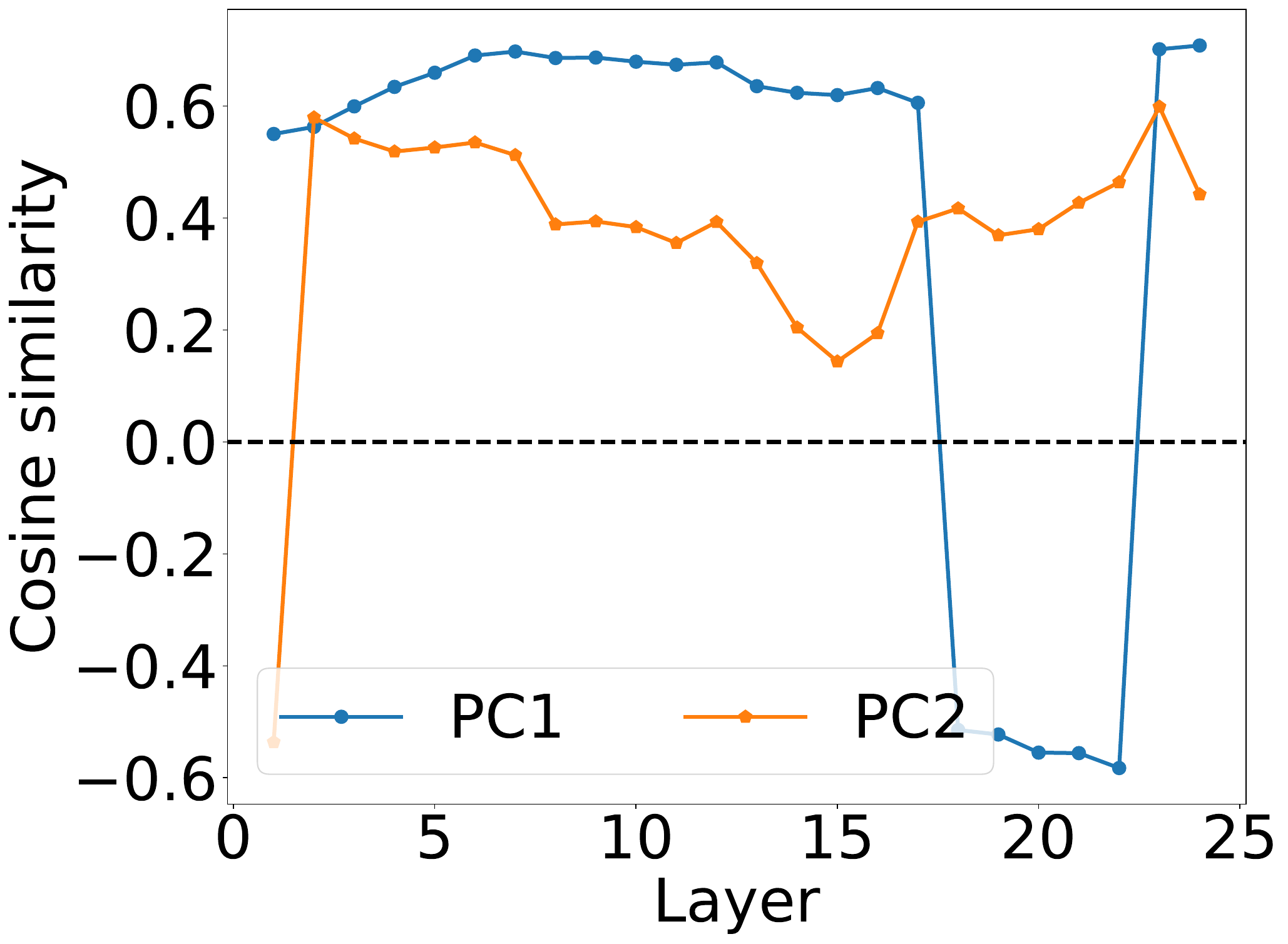}
        \caption{\texttt{en} \& \texttt{de} (POS)}
        \label{fig:cosine_en_de}
    \end{subfigure}
    ~
    \begin{subfigure}{0.48\columnwidth}
        \includegraphics[width=\textwidth]{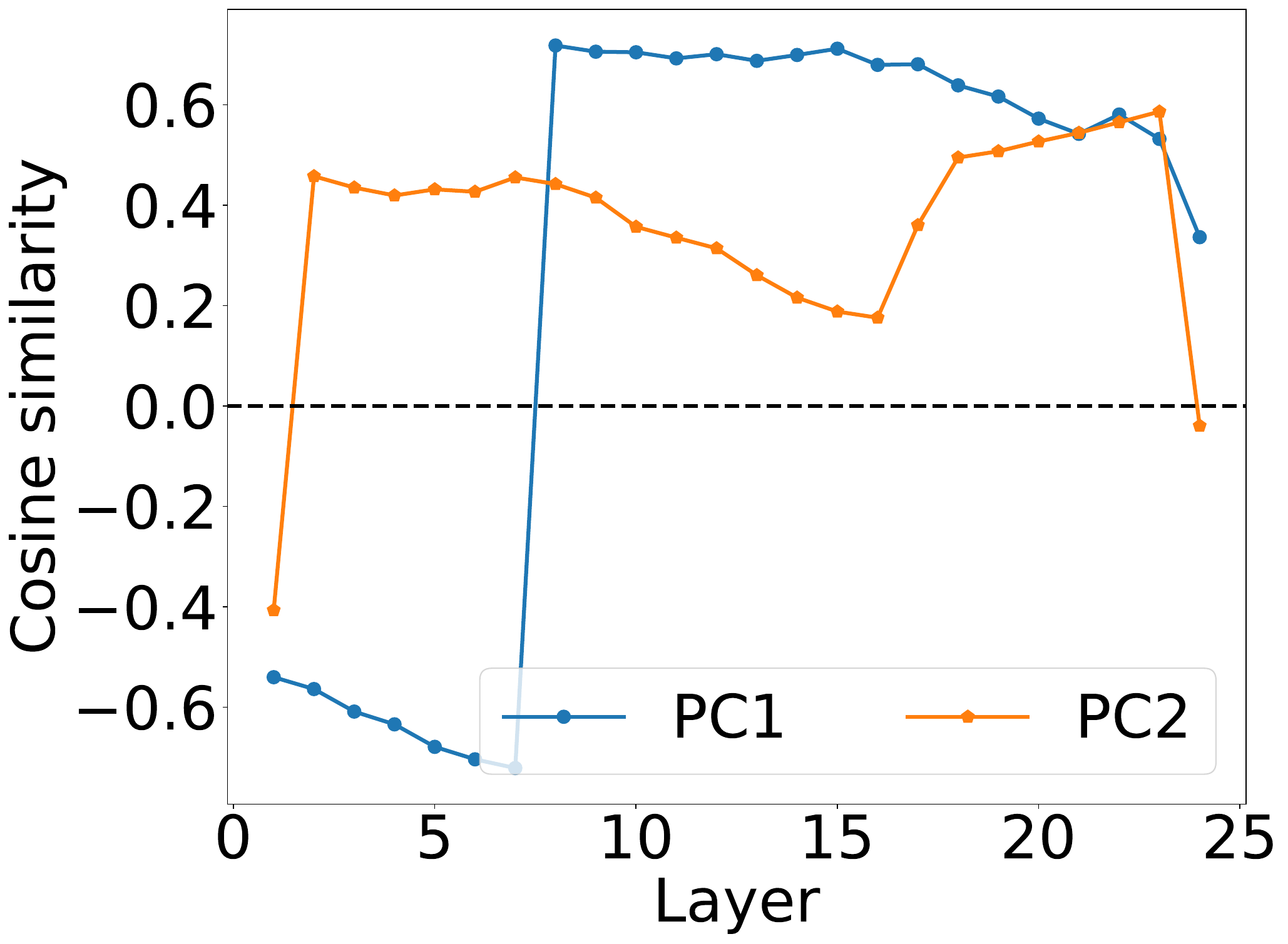}
        \caption{\texttt{en} \& \texttt{fr} (POS)}
        \label{fig:cosine_en_fr}
    \end{subfigure}
\caption{Cosine similarity between the first two principal components of residual stream. PCA is computed on token representations with different POS on English and the target languages separately.
 }
\label{fig:cosine}
\end{figure}

\Cref{fig:cosine} 
provides an alternative way to view the results, showing for every layer the cosine similarity between the first two principal components obtained from applying PCA to the English layer outputs and each of the German and French layer outputs, separately. For almost all layers, we observe a very high alignment (absolute cosine similarity $\approx 0.6$) between the principal components (similar plots for number and tense are provided in \Cref{sec:appendix:overlap}).\looseness-1 

\paragraph{Conclusion}

Overall, we take these observations as evidence that the modifications of the adapter layers operate on top of the already existing structure in the representation space of the pre-trained model. This also means that the adapters are constrained by this space and making more drastic changes will require more adaptation (more adapters/larger-in-magnitude updates), which is consistent with our findings in previous sections (\S\ref{sec:think}, \S\ref{sec:adapt_dist}).

\section{Related work}
\label{sec:related_work}

\paragraph{Language adaptation}
Various efforts have been aiming to extend the capabilities of pre-trained LMs to previously unseen languages \cite[\textit{inter alia}]{chau-etal-2020-parsing, pfeiffer-etal-2021-unks, yong2022adapting, alabi-etal-2022-adapting, imanigooghari-etal-2023-glot500}. While prior approaches have largely relied on continued or adaptive pre-training which updates all the model's parameters~\citep{chi-etal-2021-infoxlm,alabi-etal-2022-adapting,imanigooghari-etal-2023-glot500}, more recent approaches rely on parameter efficient methods~\citep{pmlr-v97-houlsby19a, pfeiffer-etal-2020-mad,marchisio-etal-2023-mini} and have been shown to be effective across many languages.
Our work seeks to provide a better understanding of the adaptation process via adapters, which could potentially lead to concrete ways to improve it.\looseness-1

\paragraph{How transformer-based LMs build predictions}
Several prior works have studied the evolution of predictions in transformer-based LMs \cite{voita-etal-2019-bottom, tenney-etal-2019-bert, nostalgebraist2020interpreting, geva-etal-2022-lm, din-etal-2023-jump, belrose-etal-2023-eliciting, ferrando-etal-2023-explaining}.
Specifically, \citet{geva-etal-2021-transformer, geva-etal-2022-transformer} showed that FFN layers in transformers gradually build predictions by promoting concepts that are interpretable in the vocabulary space. Unlike these works, here we analyze the evolution of \textit{adapted predictions}, providing a unique perspective on how adapters steer the frozen LM prediction process.

\paragraph{Analyzing multilingual language models}
Several works have analyzed multilingual LMs with special focus on the representations of these models. Some of these works have shown that these models learn language-agnostic representations which are necessary for cross-lingual transfer \citep{pires-etal-2019-multilingual, libovicky-etal-2020-language, muller-etal-2021-first,  zhao-etal-2021-inducing, xie-etal-2022-discovering, chang-etal-2022-geometry, abdullah-etal-2023-nature}. %

\paragraph{Analyzing adapters}
Only few works have analyzed the specific role of adapters during adaptation. \citet{ruckle-etal-2021-adapterdrop} proposed AdapterDrop, which involves either dynamically dropping adapters during training to enhance robustness or to speed up inference. 
\citet{he-etal-2021-effectiveness} showed that adapter-based fine-tuning results in representations with less deviation from the original model at each layer, leading to better training stability and generalization without forgetting compared to full fine-tuning. Recently, \citet{kunz2024impact} investigate the impact of target language adapter in zero-shot cross-lingual transfer, revealing that language adapters have only a minor impact on downstream performance.

\section{Conclusion and Discussion}
\label{sec:conclusion}

We experiment with language adaptation applied to mono-, bi-, and multilingual pre-trained LMs and study how language adapters interact with the underlying model.
We show that adapted LM predictions are mostly evolved in the distribution of the source language(s) the model saw during pre-training and that the adaptation is gradually happening on top of the existing representation structure of the underlying models. We not only provide a unique perspective on the inner-working of language adaptation but our findings also open up several interesting avenues for future work on language adaptation.\looseness-1

Future research on designing more efficient adaptation approaches could build on our findings on the relationship between the ``ease of adaptation'' and source-target language similarity, to automatically reduce the number of adapters during inference and to study the transfer of adapters across languages.
Our insights on the alignment between adapters and the underlying representation space could be extended to investigate the extend to which this alignment restricts the amount of adaptation possible informing studies on alternative ways of adaptation which are less constrained by the existing structure in the pre-trained model, which might lead to better performance on languages that are more distant from the source language.

\section*{Limitations}
\label{sec:limitation}

Most of our experiments are conducted on models we trained from scratch using a relatively small corpus. These models are easier to analyze compared to existing multilingual models, which allows us to perform analyses in controlled settings. Extending these experiments to existing multilingual models, such as mGPT, is a valuable non-trivial effort to pursue, which we leave for future work.

Our experiments that show how adapted predictions are evolved in the distribution of source language(s) (\Cref{sec:think}) rely on the recent method of projecting hidden representations to the LM vocabulary space, which only provides an approximation to the information encoded in intermediate representations \cite{din-etal-2023-jump}. Nonetheless, the low rate of target language tokens is unlikely to be explained only by this, as it is also low in the middle-upper layers where approximation is better \cite{geva-etal-2022-transformer, geva2023dissecting, merullo2023mechanism, hendel-etal-2023-context}.

Lastly, we focus on 8 specific target languages for our experiments. We leave it to future work to extend our analysis beyond these languages.

\section*{Acknowledgements}
\label{sec:ack}

We thank Badr Abdullah, Vagrant Gautam, Jonathan Hertzig, Shauli Ravfogel, and Julius Steuer for their valuable feedback and discussions. Jesujoba O. Alabi was supported by the BMBF's (German Federal Ministry of Education and Research) SLIK project under the grant 01IS22015C.

\bibliography{anthology,custom}

\clearpage
\appendix

\section{Details on model training}
\label{apx:model_training}

\paragraph{Tokenization and training data}

We pretrain and adapt our models using data sourced from Wikipedia using WikiExtractor \citep{Wikiextractor2015}. For tokenization, we use the BPE (Byte Pair Encoding) tokenizer developed by~\cite{wang2019neural}. Given the noisy nature of webcrawled data such as Wikipedia, we filtered the collected data using the OpenLID~\citep{burchell-etal-2023-open} language identification model. To train the tokenizer, we sampled 4M sentences from each language's monolingual data. Thus, in total, we obtained 20M sentences and trained a single tokenizer on data of all languages with using a vocabulary of size 250K. Given this pretrained tokenizer, we uniformly sampled sentences based on the count of tokens from each language's monolingual data and used the resulting data for pre-training and adaptation.

\paragraph{Adapter training}

The adapters were trained for 150K steps. We evaluated the model's on the validation monolingual data every $5000$ steps and selected the model (adapter) that yielded the lowest validation perplexity.

\section{Details on pre-training and adaptation token statistics}
\label{apx:details_token_statistics}
\Cref{tab:train_token} shows the number of sentences and tokens used for both training and adaptation. In our controlled experiments, Hebrew had the least amount of available monolingual data on Wikipedia. Therefore, to ensure equitable representation across all languages, we opted to sample an equal amount of tokens from Wikipedia for the other languages. Specifically, given the approximately 145 million tokens in Hebrew, we sampled an equal number of tokens from the other languages' monolingual data. Additionally, $2\%$ of the data was reserved as a development set, with the remainder used for training or adaptation purposes. 

We followed a similar approach for the mBERT experiment, where we sampled 4 millions tokens from all languages, and reserved $5\%$ as the development set for each language.

\begin{table}[ht]
  \centering
  \scalebox{0.78}{ %
  \begin{tabular}{llccc}
    \toprule
    \textbf{Language} & \textbf{Code} & \textbf{\# Sentences} & \multicolumn{2}{c}{\textbf{\# Tokens}} \\
    {} & {} & {} & Train & Dev \\
    \midrule
    Arabic & \texttt{ar} & 4,001,963 & 142,492,237 & 2,908,012 \\
    German & \texttt{de} & 2,148,944 & 142,492,054 & 2,908,064 \\
    French & \texttt{fr} & 2,187,922 & 142,491,947 & 2,908,244 \\ 
    English & \texttt{en} & 1,742,844 & 142,492,202 & 2,908,014 \\ 
    Hebrew & \texttt{he} & 3,078,001 & 142,492,172 & 2,908,062 \\
    \midrule
    Amharic & \texttt{am} & 80,060 & 3,869,262 & 203,101 \\
    Khmer & \texttt{km} & 152,375 & 3,869,233 & 203,008 \\
    Odia & \texttt{or} & 124,900 & 3,869,209 & 203,112 \\
    Santali & \texttt{sat} & 78,887 & 3,869,077 & 203,161 \\
    \bottomrule
  \end{tabular}
  }
  \caption{Pre-training and adaptation data sizes.}
\label{tab:train_token}
\end{table}

\section{Details of language token identification}
\label{apx:details_lang_id}

In \Cref{sec:think}, we mention that for French and German, language identification is performed by looking at language statistics. Considering the token counts in three languages (including English) based on the Latin script. We determine the predominant language for any randomly selected token based on its frequency. Once we identify the language with the highest token count for that particular token, we calculate the ratio of its occurrences in the English corpus to the occurrences in the identified language. If this ratio is equal to or less than a specified threshold, which we have set at $0.7$ through manual inspection of the resulting classification, we classify the token as belonging to that language.

For mBERT, we used basic script identification just as for Arabic and Hebrew, Ge’ez, Khmer, Oriya, and Ol Chiki have distinct scripts that can be easily identified. 

\section{Details on mBERT's language converge}
\label{apx:mbert_coverage}
In order to determine the languages to include in the mBERT experiment, we sought languages with scripts not adequately represented in mBERT. We opted for several languages from FLORES, particularly those using scripts distinct from those covered by mBERT. \Cref{tab:mbert_token} shows four languages that we finally selected, their scripts, and the unknown token rates (UNK rates) according to mBERT's tokenizer when evaluated on each language's FLORES dev split. The UNK rate confirms that these languages were not adequately covered by mBERT, with Amharic and Santali having the highest values.
\begin{table}[ht]
  \centering
  \begin{tabular}{llll}
    \toprule
    \textbf{Language} & \textbf{Code} & \textbf{Script} & \textbf{UNK rate} \\
    \midrule
    Amharic & \texttt{am} & Ge'ez & 0.956 \\
    Khmer & \texttt{km} & Khmer & 0.664 \\
    Odia & \texttt{or} & Oriya & 0.858 \\
    Santali & \texttt{sat} & Ol Chiki & 0.933 \\
    \bottomrule
  \end{tabular}
  \caption{The langauges not covered in mBERT and their UNK rate when FLORES dev splits for these languages were tokenized.}
\label{tab:mbert_token}
\end{table}

\subsection{LoRA Adapters}
\label{sec:appendix:lora}

We add low-rank adapters (LoRA) to the FFN sublayers within each decoder block. With LoRA, the forward-pass of a single layer can be formalized as follows:
\vspace{-0.25cm}
\begin{align}
    \mathbf{x}_{\texttt{attn}}^{l,i} &= \mathbf{x}_{\texttt{out}}^{l-1,i} + \textcolor{black}{\texttt{self-attn}}(X_{\texttt{out}}^{l-1})_i \\
    \mathbf{x}_{\texttt{ffn}_1}^{l,i}\; &= \textcolor{black}{\texttt{FFN}}_1(\mathbf{x}_{\texttt{attn}}^{l,i}) + \textcolor{black}{\texttt{LoRA}_1}(\mathbf{x}_{\texttt{attn}}^{l,i}) \\
    \mathbf{x}_{\texttt{ffn}_2}^{l,i}\; &= \textcolor{black}{\texttt{FFN}}_2(\mathbf{x}_{\texttt{ffn}_1}^{l,i}) + \textcolor{black}{\texttt{LoRA}_2}(\mathbf{x}_{\texttt{ffn}_1}^{l,i}) \\
    \mathbf{x}_{\texttt{out}}^{l,i}\; &= \mathbf{x}_{\texttt{attn}}^{l,i} + \mathbf{x}_{\texttt{ffn}_2}^{l,i}
\end{align}

Where $\textcolor{black}{\texttt{FFN}_1}$ and $\textcolor{black}{\texttt{FFN}_2}$ are the feed forward sublayers within the decoder block. Unlike the Pfeiffer adapters, LoRA uses rank decomposition matrices without non-linearity between them: 
\begin{align}
    \textcolor{black}{\texttt{LoRA}}(\mathbf{x}) &= \mathbf{W}_2 (\mathbf{W}_1\mathbf{x})~,
\end{align}

where $\mathbf{W}_1$ and $\mathbf{W}_2$ are  trainable parameters, and $\textcolor{black}{\texttt{LoRA}_1}$ and $\textcolor{black}{\texttt{LoRA}}_2$ have distinct parameters. During adaptation, we only update the parameters of the LoRA and the embedding layer while keeping all other parameters frozen.

\begin{figure}[t]
\setlength{\belowcaptionskip}{-2pt}
  \centering
    \begin{subfigure}{0.45\columnwidth}
        \includegraphics[width=\textwidth]{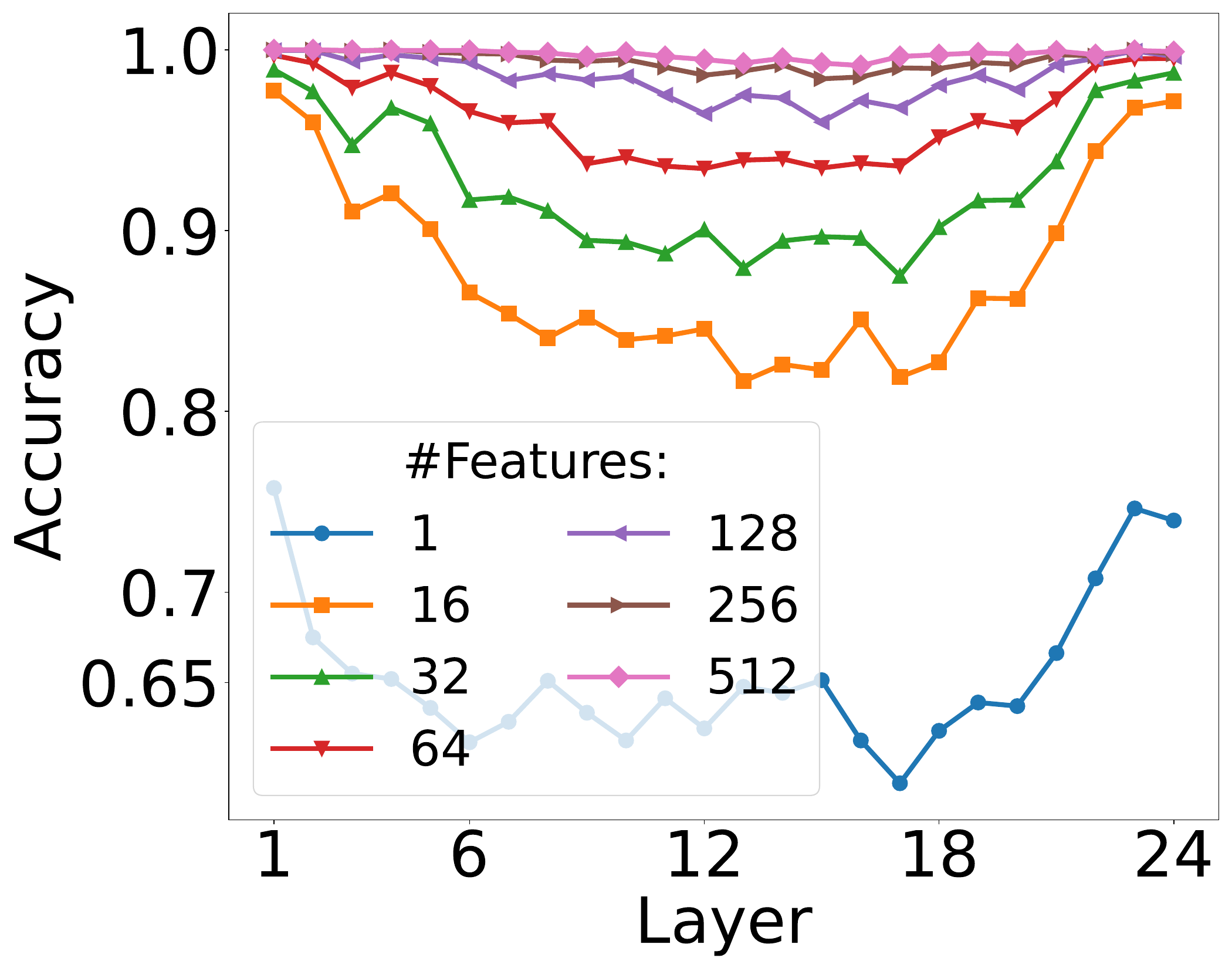}
        \caption{\texttt{en} $\rightarrow$ \texttt{he}}
        \label{fig:probe_acc_de2}
    \end{subfigure}
    ~
    \begin{subfigure}{0.45\columnwidth}
        \includegraphics[width=\textwidth]{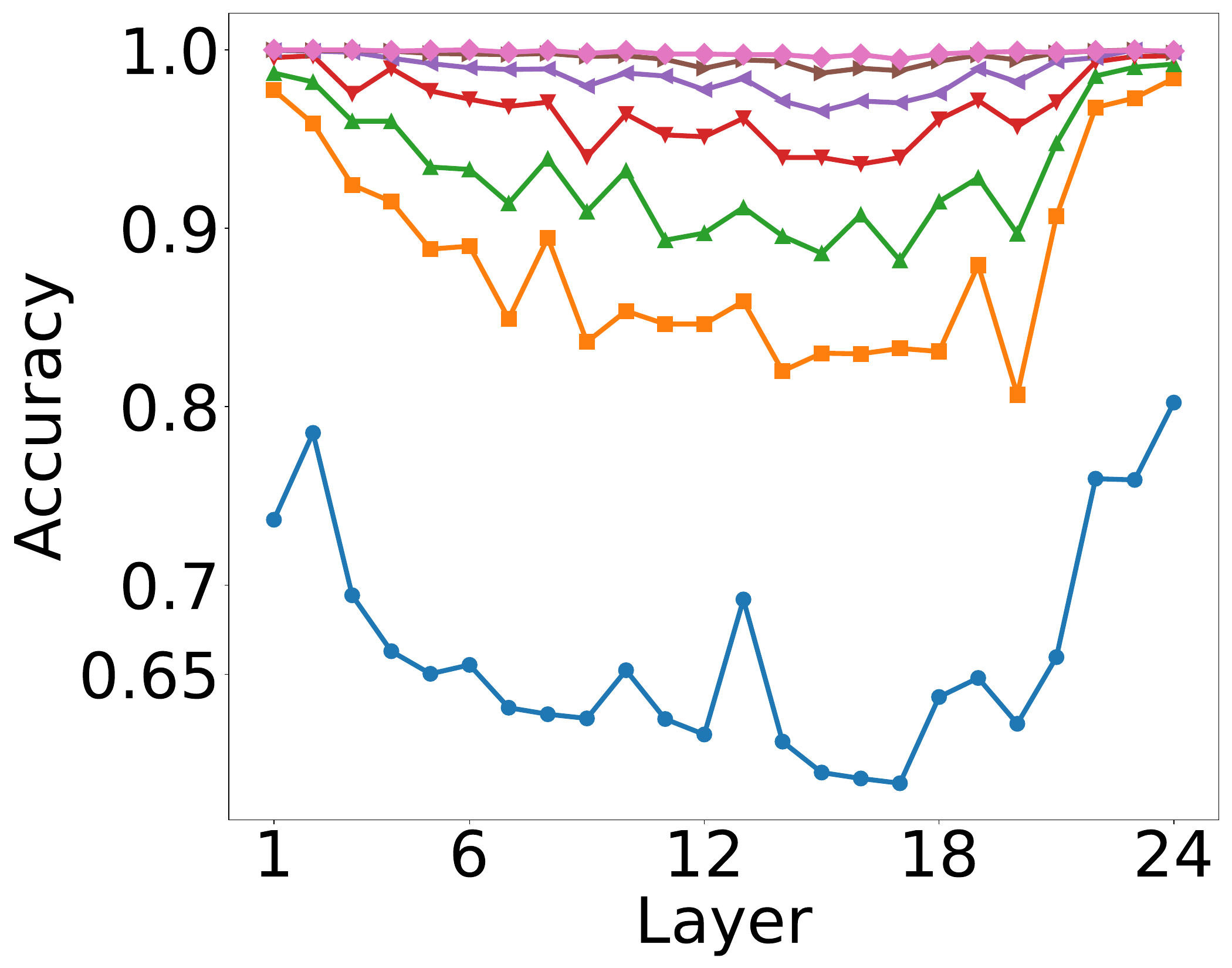}
        \caption{\texttt{en} $\rightarrow$ \texttt{ar}}
        \label{fig:probe_acc_fr2}
    \end{subfigure}
    \\
    \begin{subfigure}{0.45\columnwidth}
        \includegraphics[width=\textwidth]{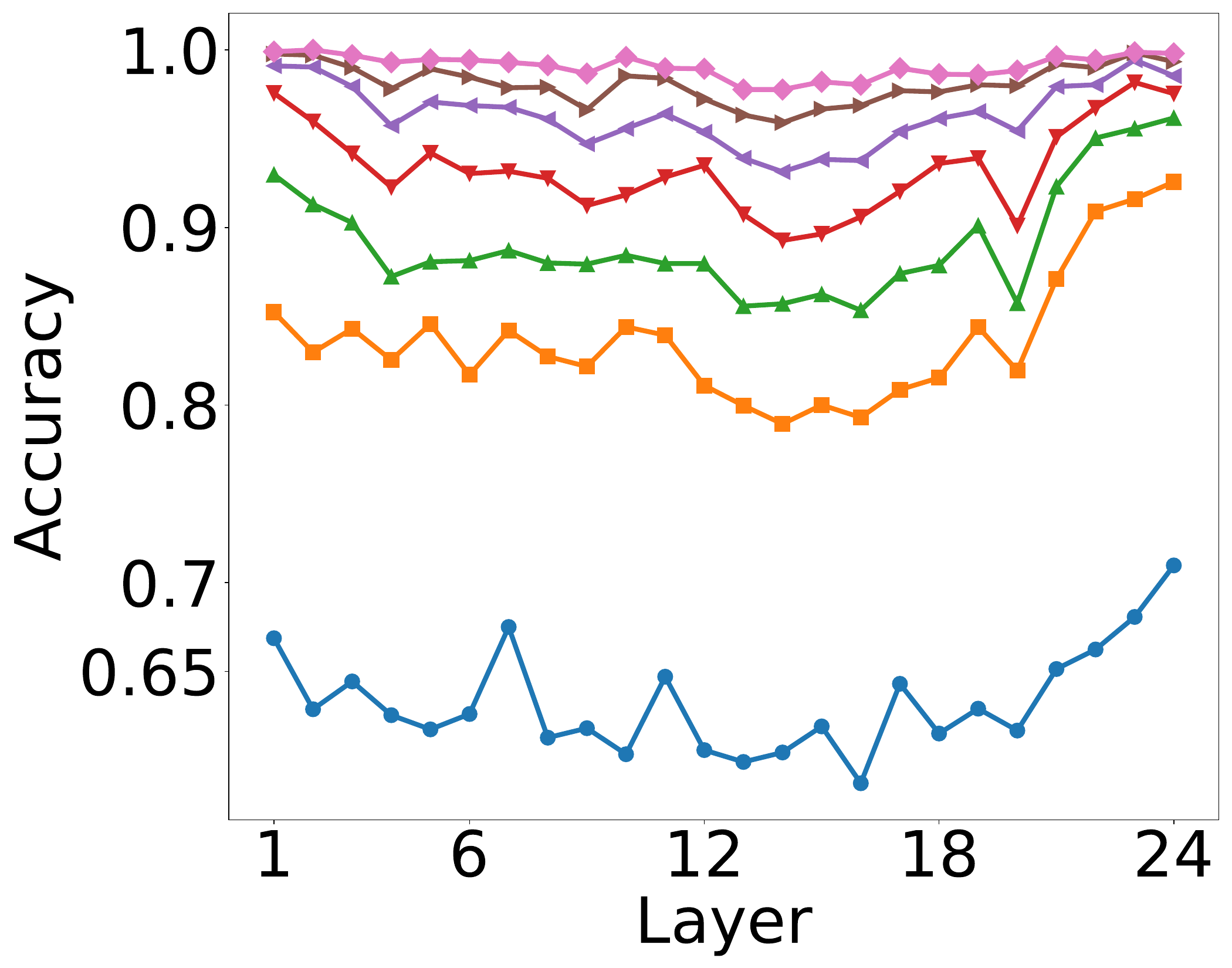}
        \caption{\texttt{en} $\rightarrow$ \texttt{de}}
        \label{fig:probe_acc_he2}
    \end{subfigure}
    ~
    \begin{subfigure}{0.45\columnwidth}
        \includegraphics[width=\textwidth]{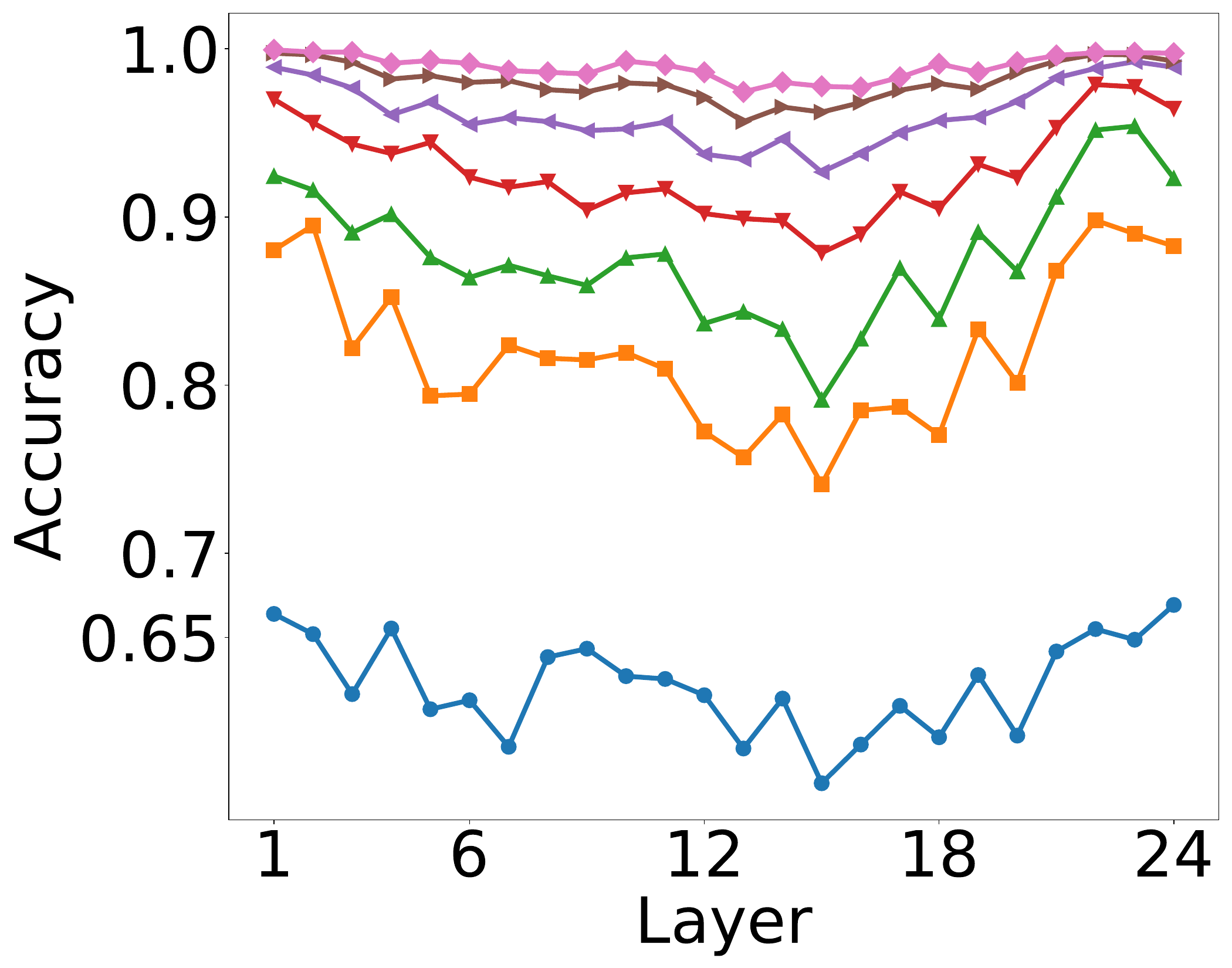}
        \caption{\texttt{en} $\rightarrow$ \texttt{fr}}
        \label{fig:probe_acc_ar2}
    \end{subfigure}
\caption{Sparse probing classifiers can detect language adaptation with high accuracy (\textbf{case 2}).
}
\label{fig:probe_acc2}
\end{figure} 

\section{Additional results}
\label{apx:additional_results}

\begin{figure*}[t]
  \centering
    \begin{subfigure}{0.235\textwidth}
        \includegraphics[width=\textwidth]{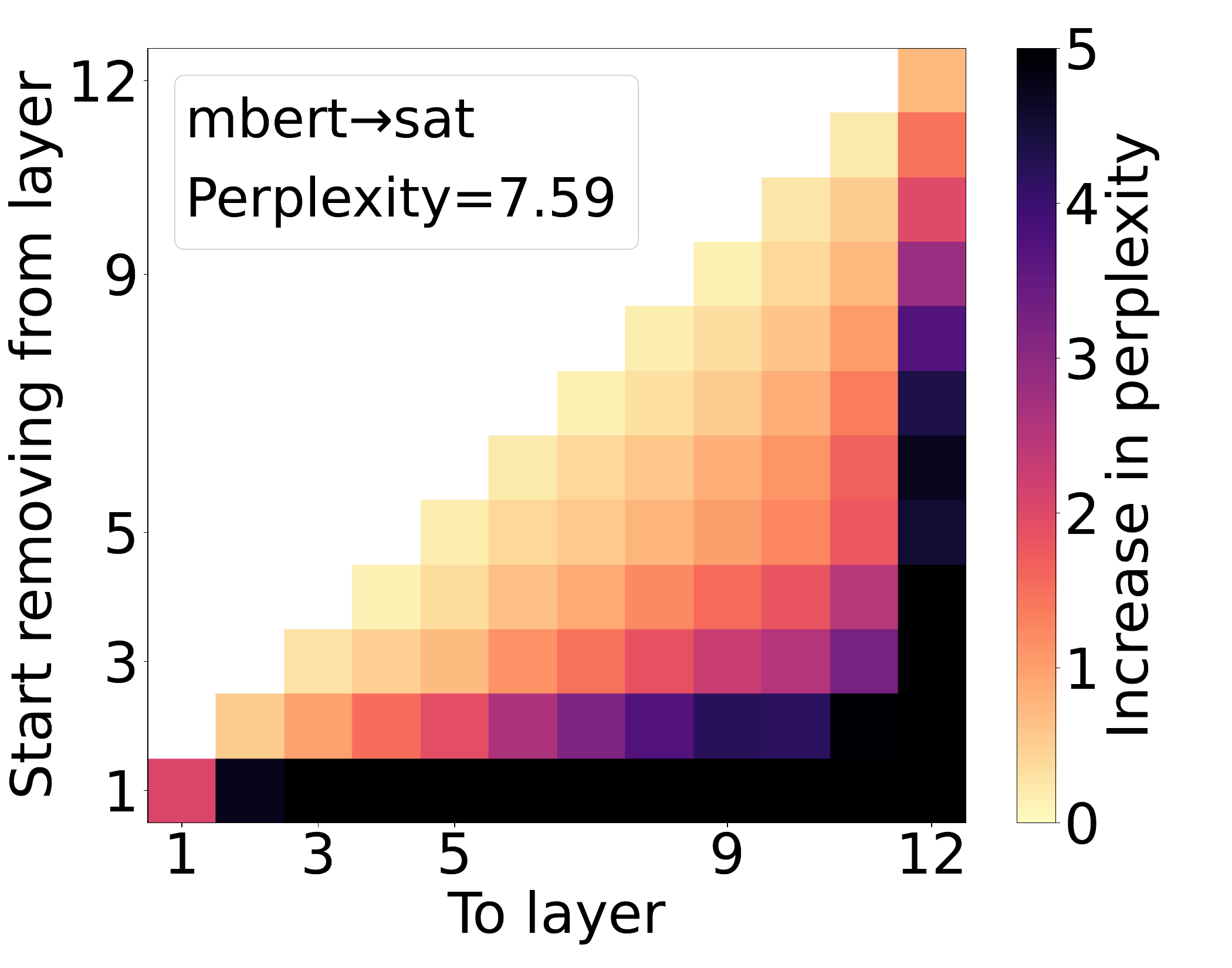}
        \caption{mBERT $\rightarrow$ sat}
        \label{fig:mbert_sat}
    \end{subfigure}
    ~
    \begin{subfigure}{0.235\textwidth} 
        \includegraphics[width=\textwidth]{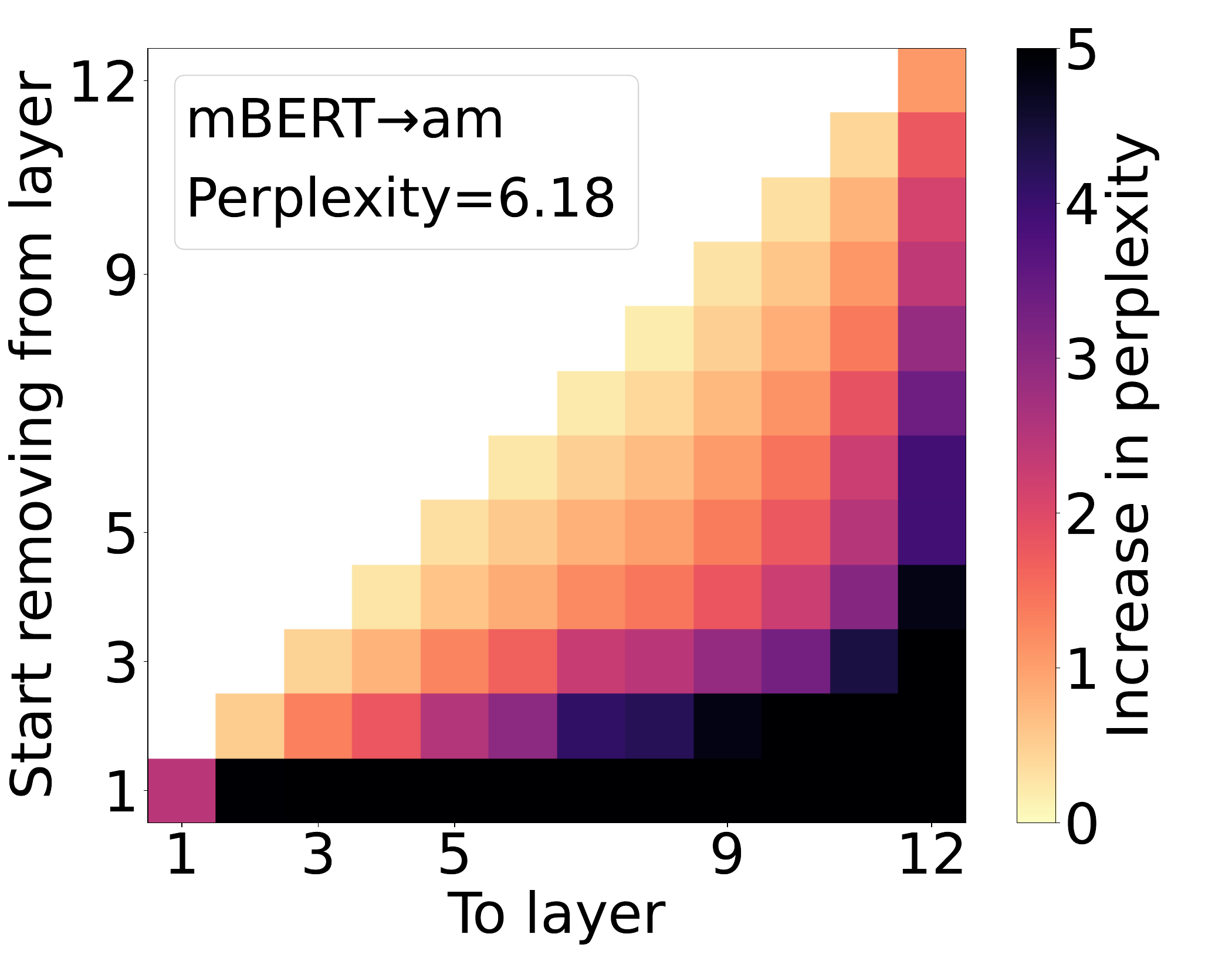}
        \caption{mBERT $\rightarrow$ am}
        \label{fig:mbert_am}
    \end{subfigure}
    ~
    \begin{subfigure}{0.235\textwidth}
        \includegraphics[width=\textwidth]{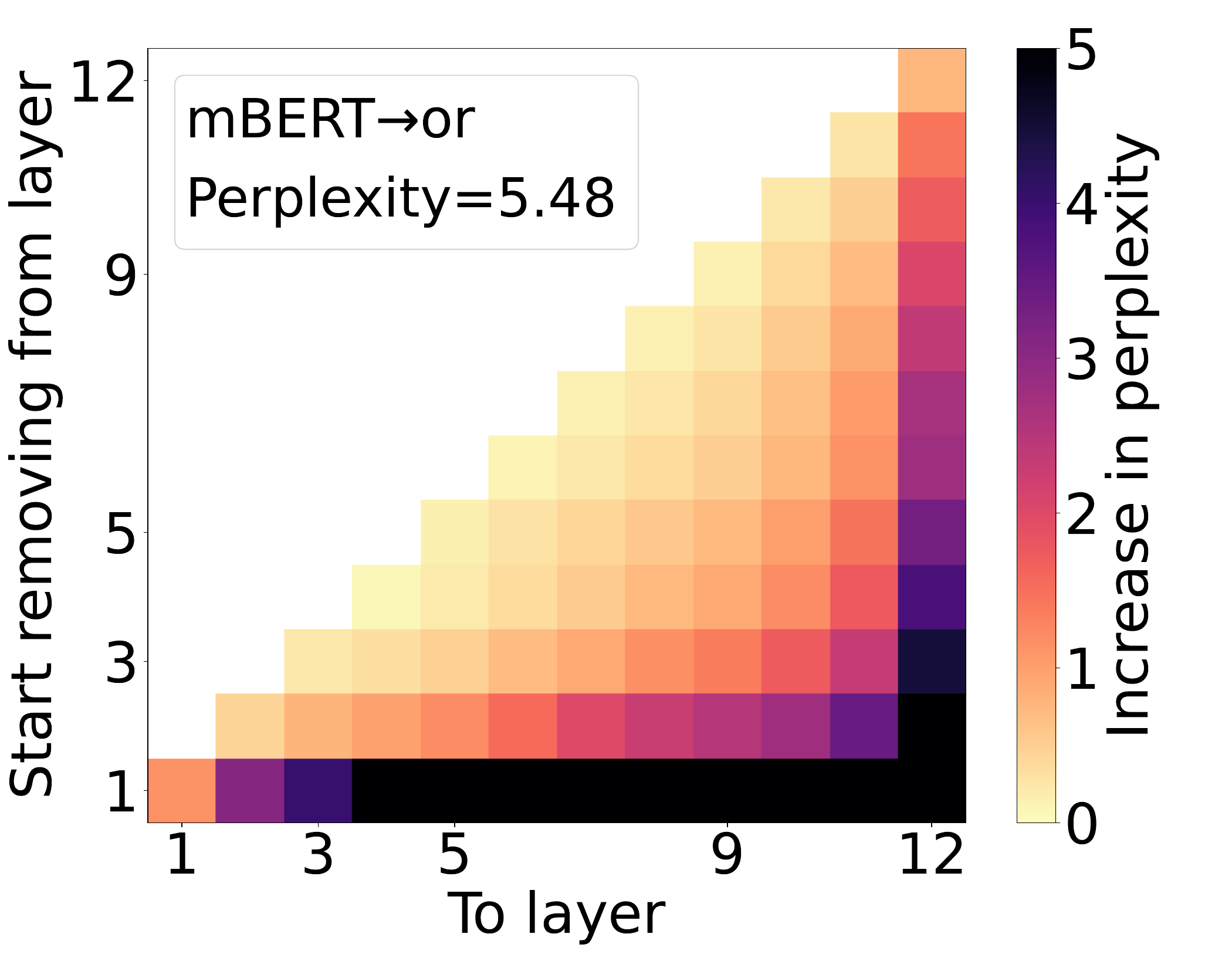}
        \caption{mBERT $\rightarrow$ or}
        \label{fig:mbert_or}
    \end{subfigure}
    ~
    \begin{subfigure}{0.235\textwidth} 
        \includegraphics[width=\textwidth]{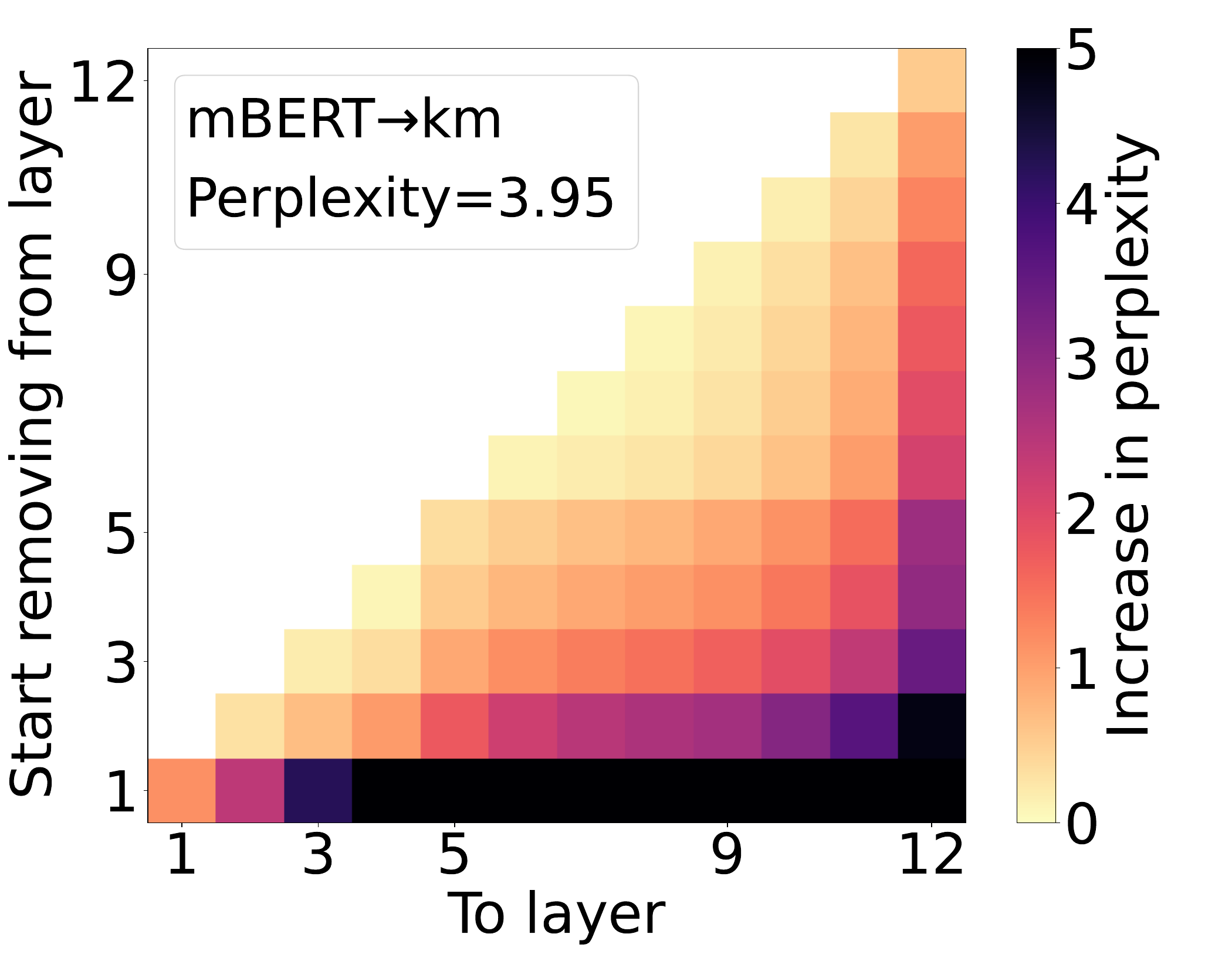}
        \caption{mBERT $\rightarrow$ km}
        \label{fig:mbert_km}
    \end{subfigure}
\caption{Difference in perplexity after removing some adapters
}
\label{fig:appendix:perp_diff_mbert}
\end{figure*}

\begin{figure*}[t]
  \centering
    \begin{subfigure}{0.235\textwidth}
        \includegraphics[width=\textwidth]{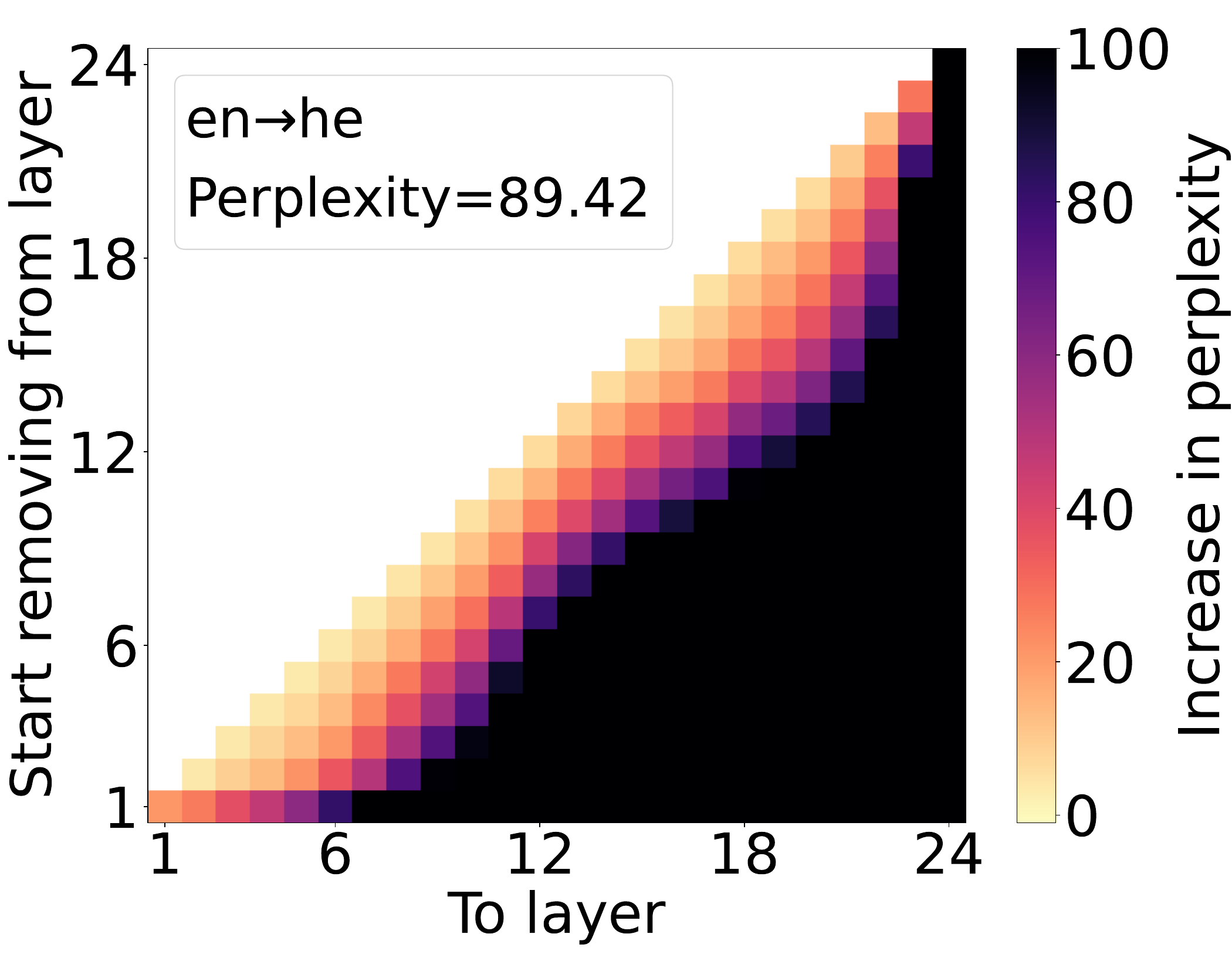}
        \caption{en $\rightarrow$ he (LoRA)}
        \label{fig:lora_he}
    \end{subfigure}
    ~
    \begin{subfigure}{0.235\textwidth} 
        \includegraphics[width=\textwidth]{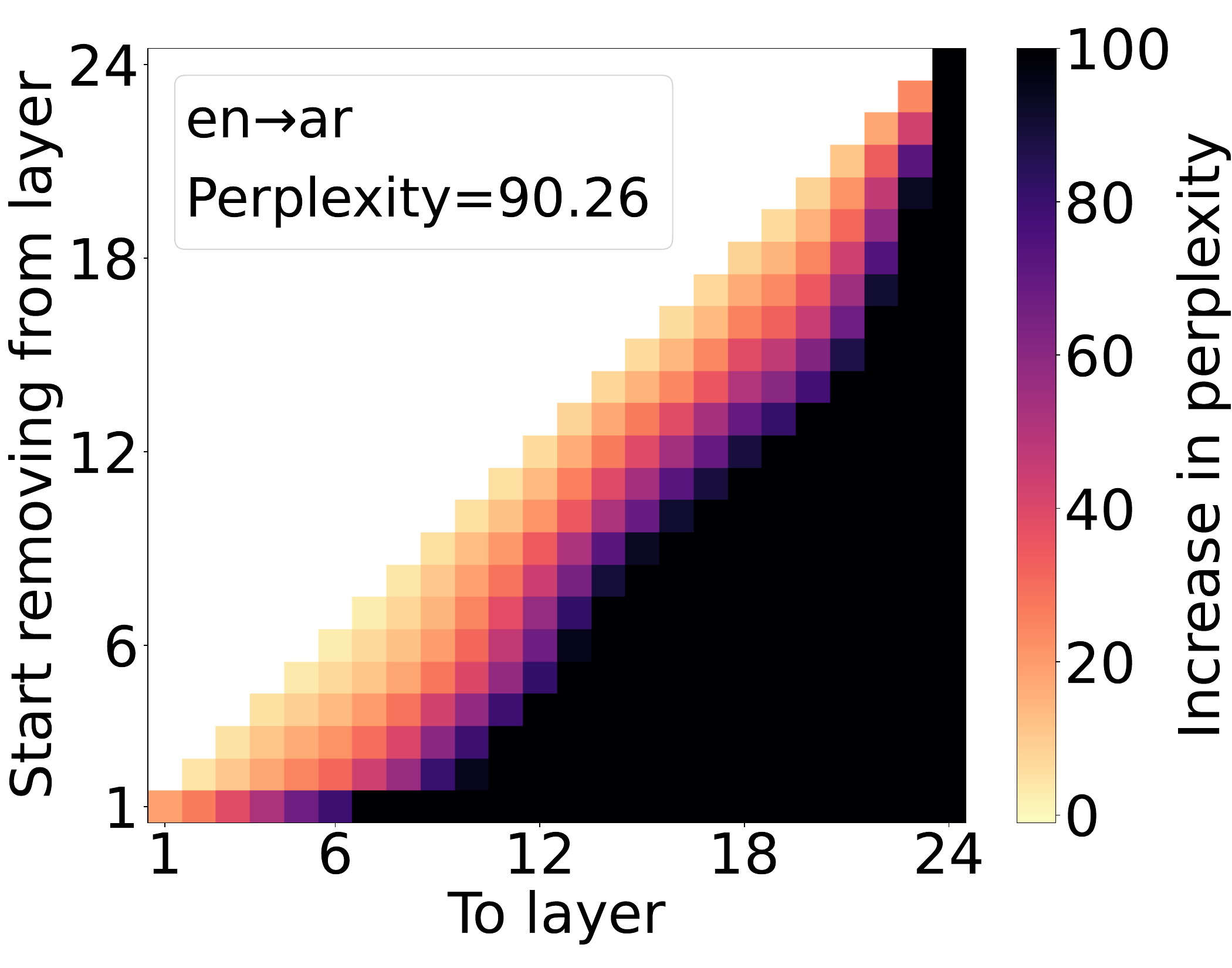}
        \caption{en $\rightarrow$ ar (LoRA)}
        \label{fig:lora_ar}
    \end{subfigure}
    ~
    \begin{subfigure}{0.235\textwidth}
        \includegraphics[width=\textwidth]{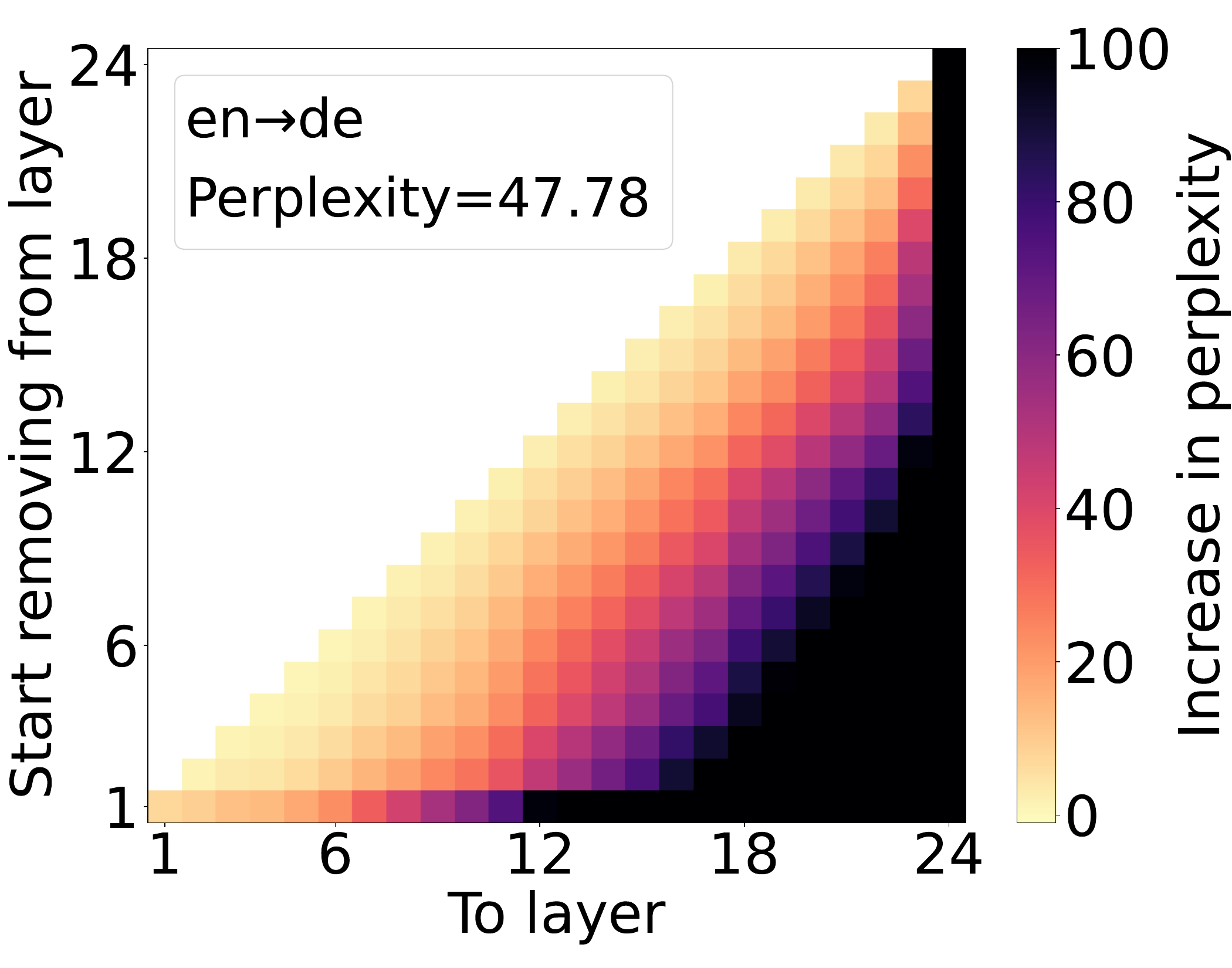}
        \caption{en $\rightarrow$ (LoRA)}
        \label{fig:lora_de}
    \end{subfigure}
    ~
    \begin{subfigure}{0.235\textwidth} 
        \includegraphics[width=\textwidth]{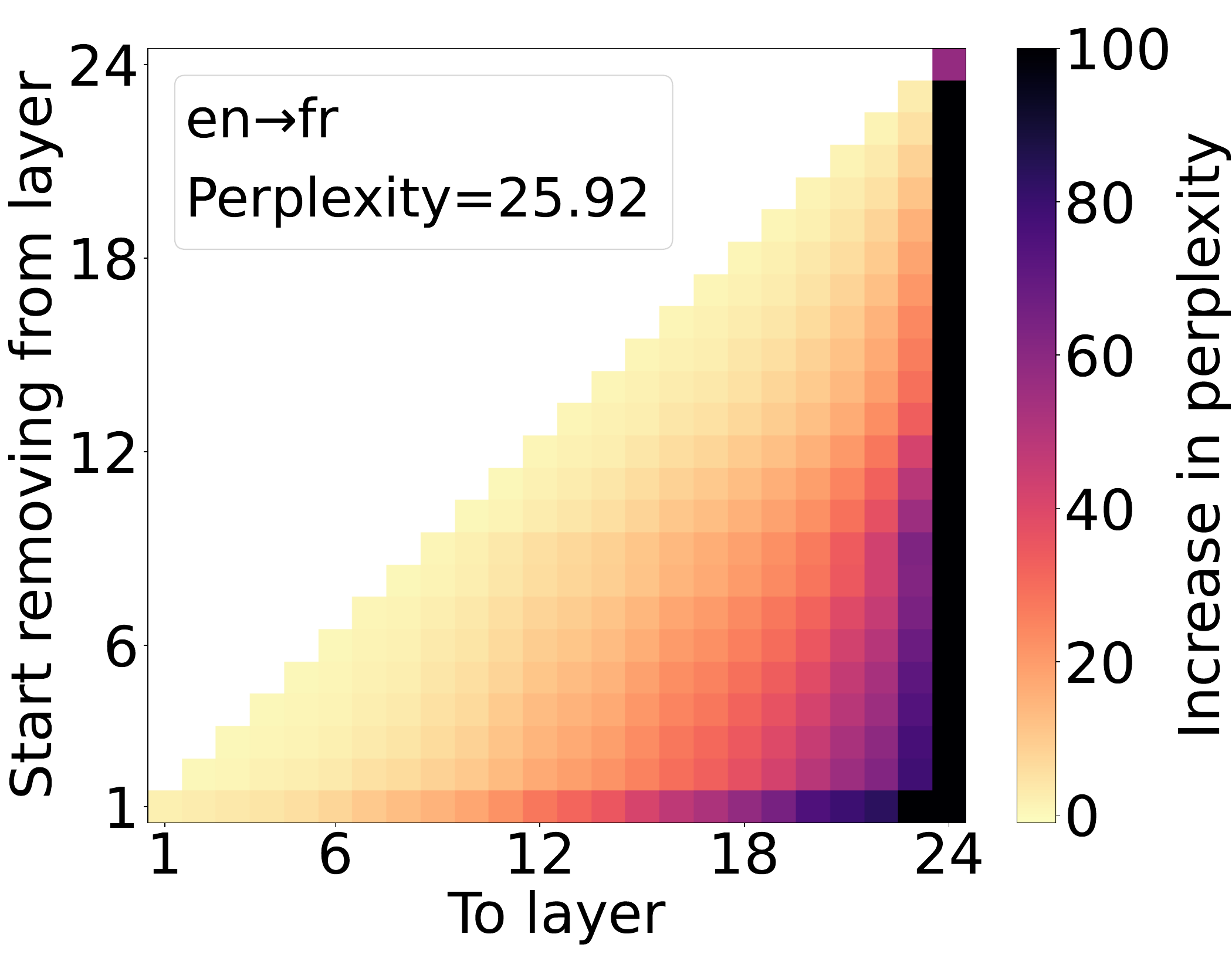}
        \caption{en $\rightarrow$ fr (LoRA)}
        \label{fig:lora_fr}
    \end{subfigure}
\caption{Difference in perplexity after removing some adapters with LoRA
}
\label{fig:appendix:perp_diff_lora}
\end{figure*}

\subsection{Effect of removing adapters on perplexity}
\label{sec:appendix:adapter_drop_mbert}

\Cref{fig:appendix:perp_diff_mbert} present results for the increase in perplexity when dropping individual or chunks of adapters of mBERT during inference. Similar to the results in \Cref{fig:ppl_results}, we observe little impact on perplexity when removing individual adapters. However, when removing larger consecutive blocks of adapters, perplexity increases for all languages. Compared to \Cref{fig:ppl_results}, the increase in perplexity is generally less dramatic, which we attribute to the fact that mBERT is a highly optimized multilingual model as well as the fact that masked language modeling is generally an easier task that causal language modeling. 

\Cref{fig:appendix:perp_diff_lora} shows result for the same analysis using LoRA. The result are highly consistent with the Pfeiffer adapter setup. Removing individual adapters has little impact on perplexity. However, removing larger consecutive chunks of adapters affects perplexity considerably.

\begin{figure}[t]
\setlength{\belowcaptionskip}{-2pt}
  \centering
    \begin{subfigure}{0.45\columnwidth}
        \includegraphics[width=\textwidth]{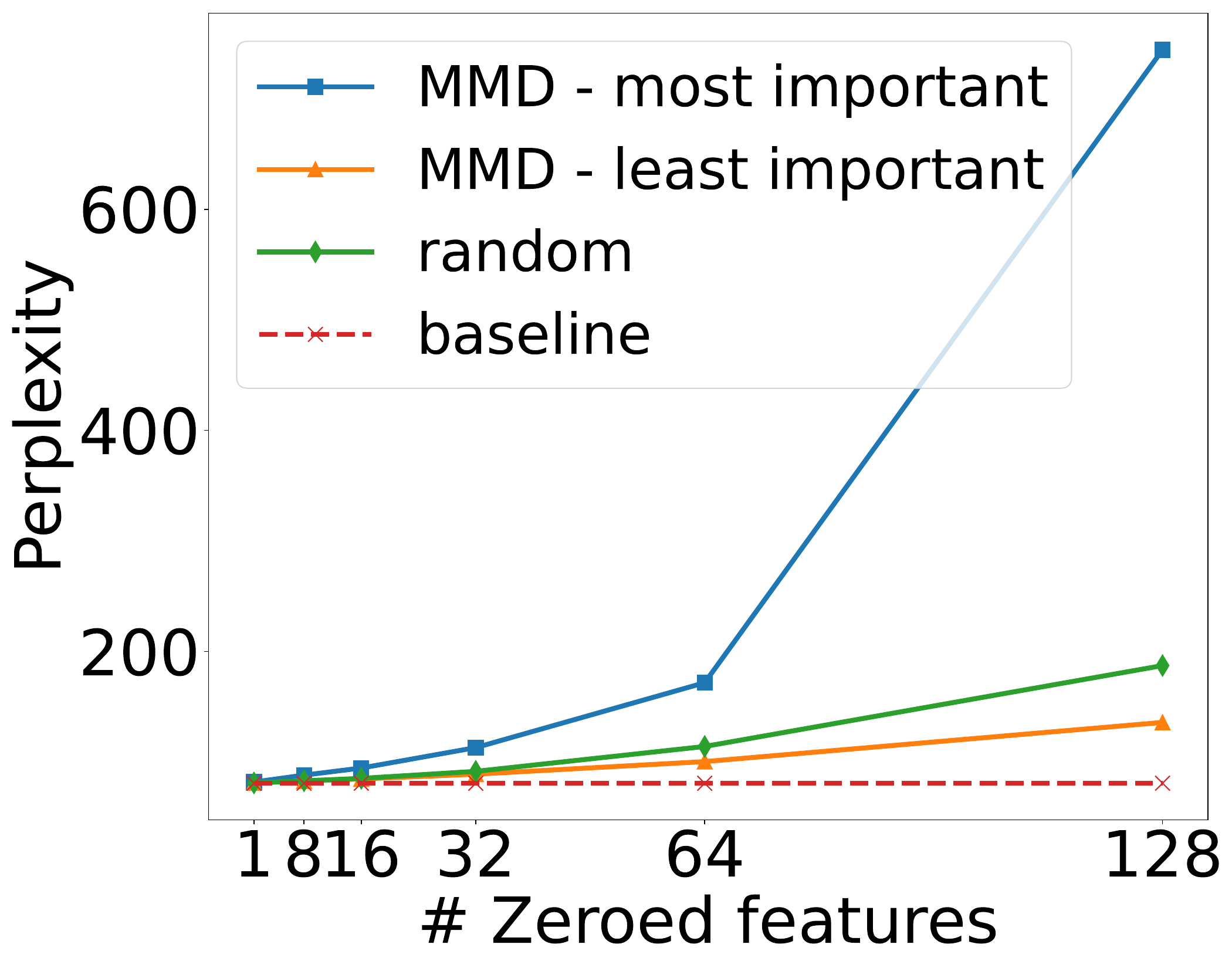}
        \caption{\texttt{en} $\rightarrow$ \texttt{he}}
        \label{fig:zero-intervention-he2}
    \end{subfigure}
    ~
    \begin{subfigure}{0.45\columnwidth} 
        \includegraphics[width=\textwidth]{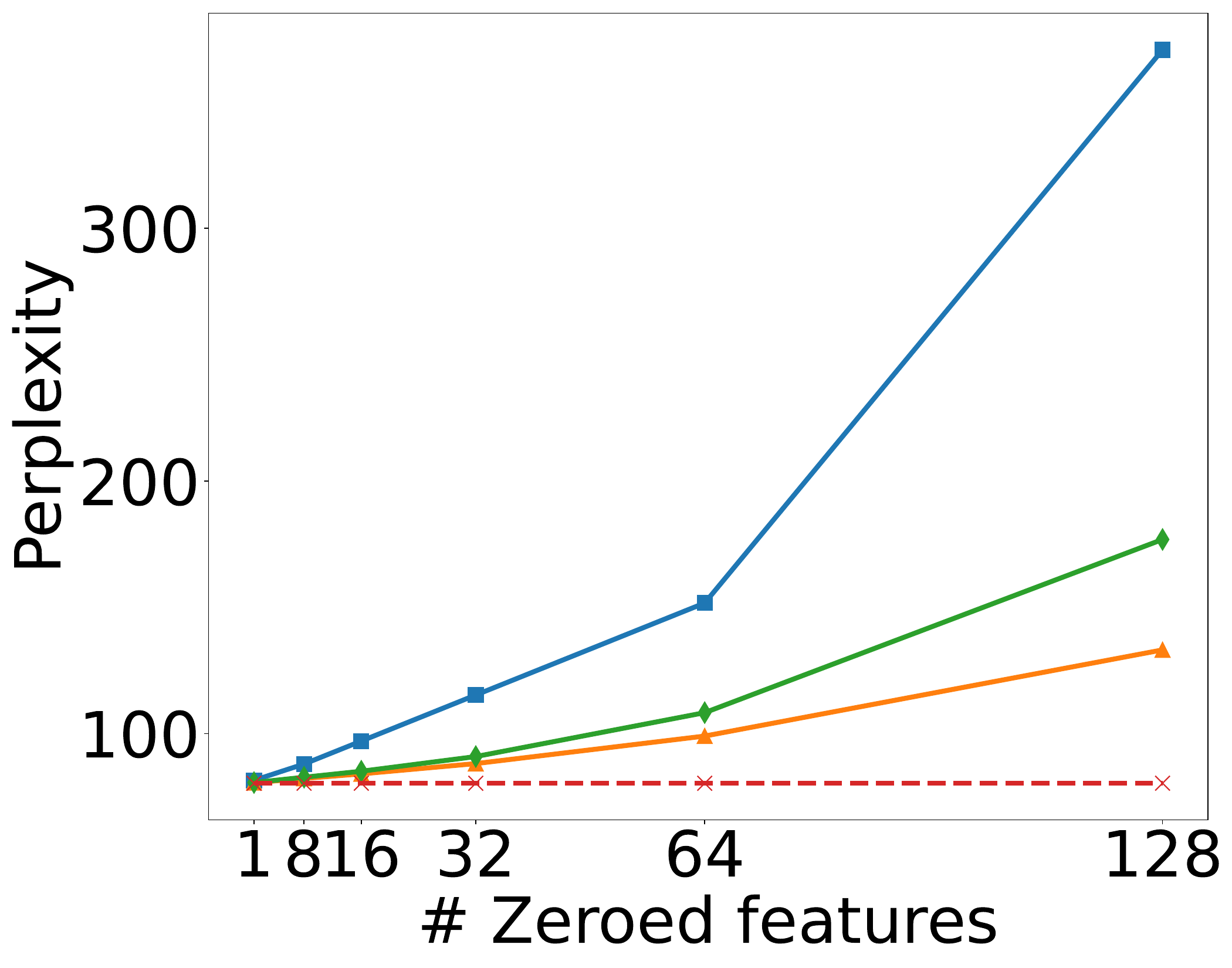}
        \caption{\texttt{en} $\rightarrow$ \texttt{ar}}
        \label{fig:zero-intervention-ar2}
    \end{subfigure}
    \\
    \begin{subfigure}{0.45\columnwidth}
        \includegraphics[width=\textwidth]{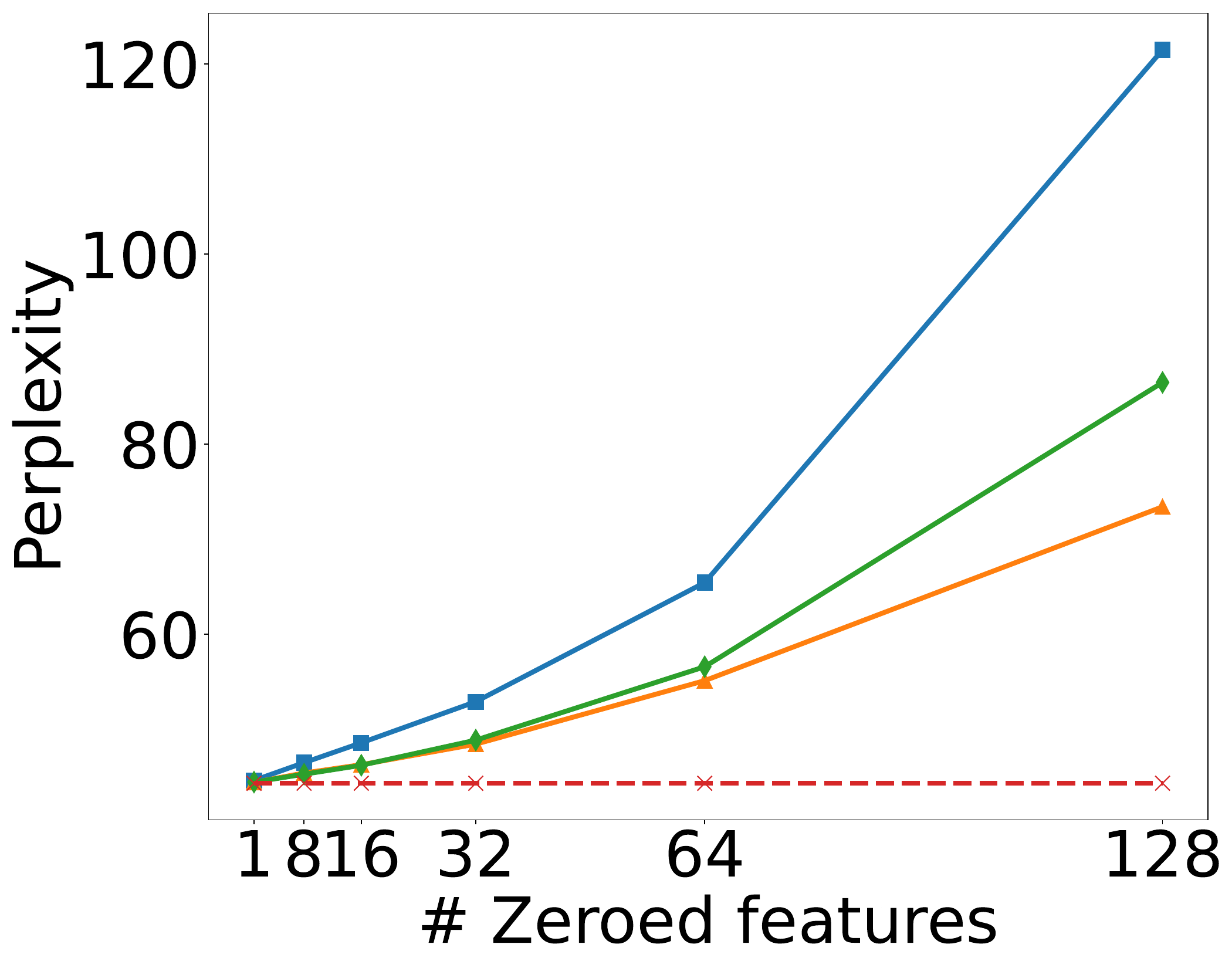}
        \caption{\texttt{en} $\rightarrow$ \texttt{de}}
       \label{fig:zero-intervention-de2}
    \end{subfigure}
    ~
    \begin{subfigure}{0.45\columnwidth} 
        \includegraphics[width=\textwidth]{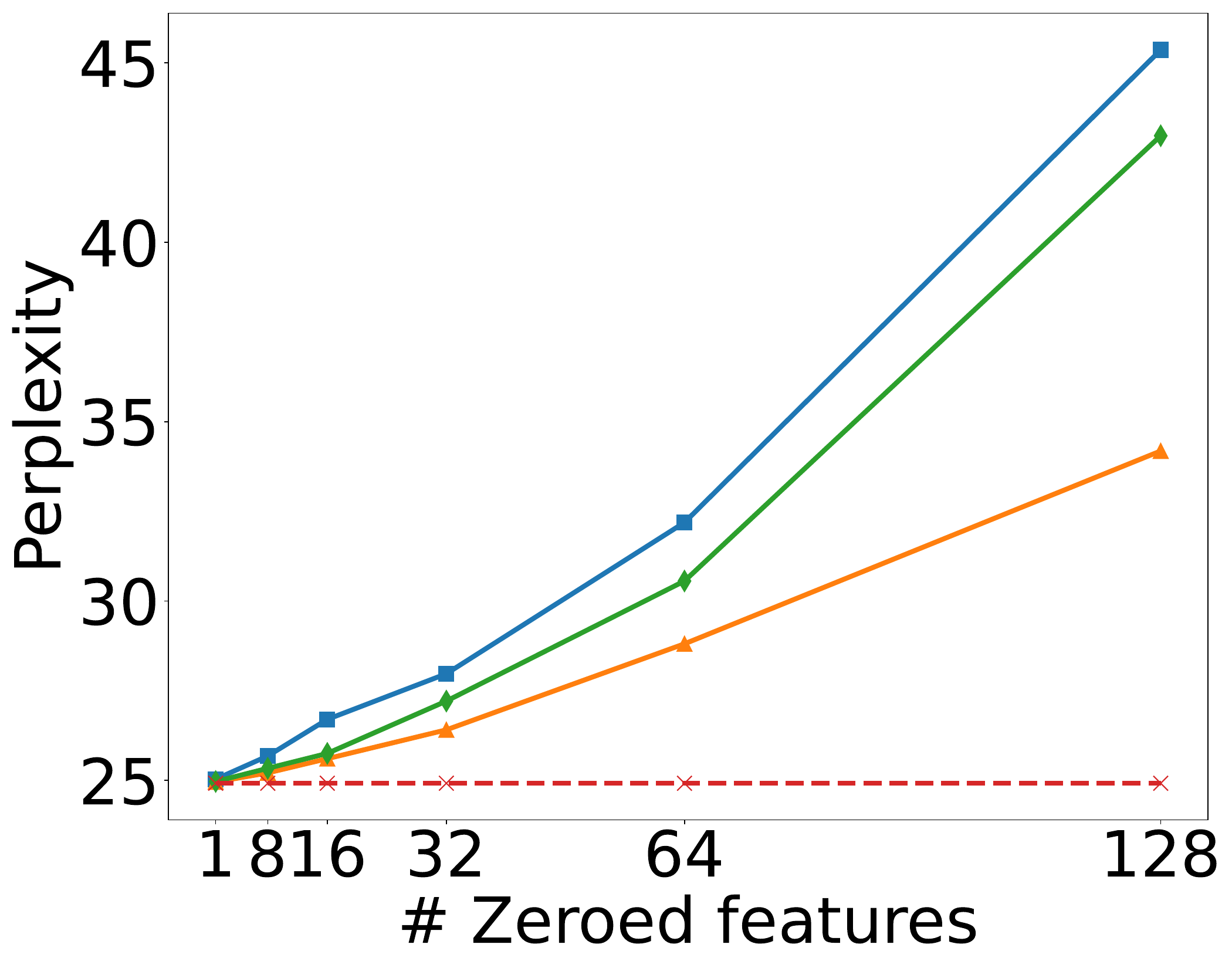}
        \caption{\texttt{en} $\rightarrow$ \texttt{fr}}
        \label{fig:zero-intervention-fr2}
    \end{subfigure}
\caption{Adapted model perplexity after intervention (\textbf{case 2}).
}
\label{fig:interventions2}
\end{figure}

\begin{figure}[t]
\setlength{\belowcaptionskip}{-4pt}
  \centering
    \begin{subfigure}{0.48\columnwidth}
        \includegraphics[width=\textwidth]{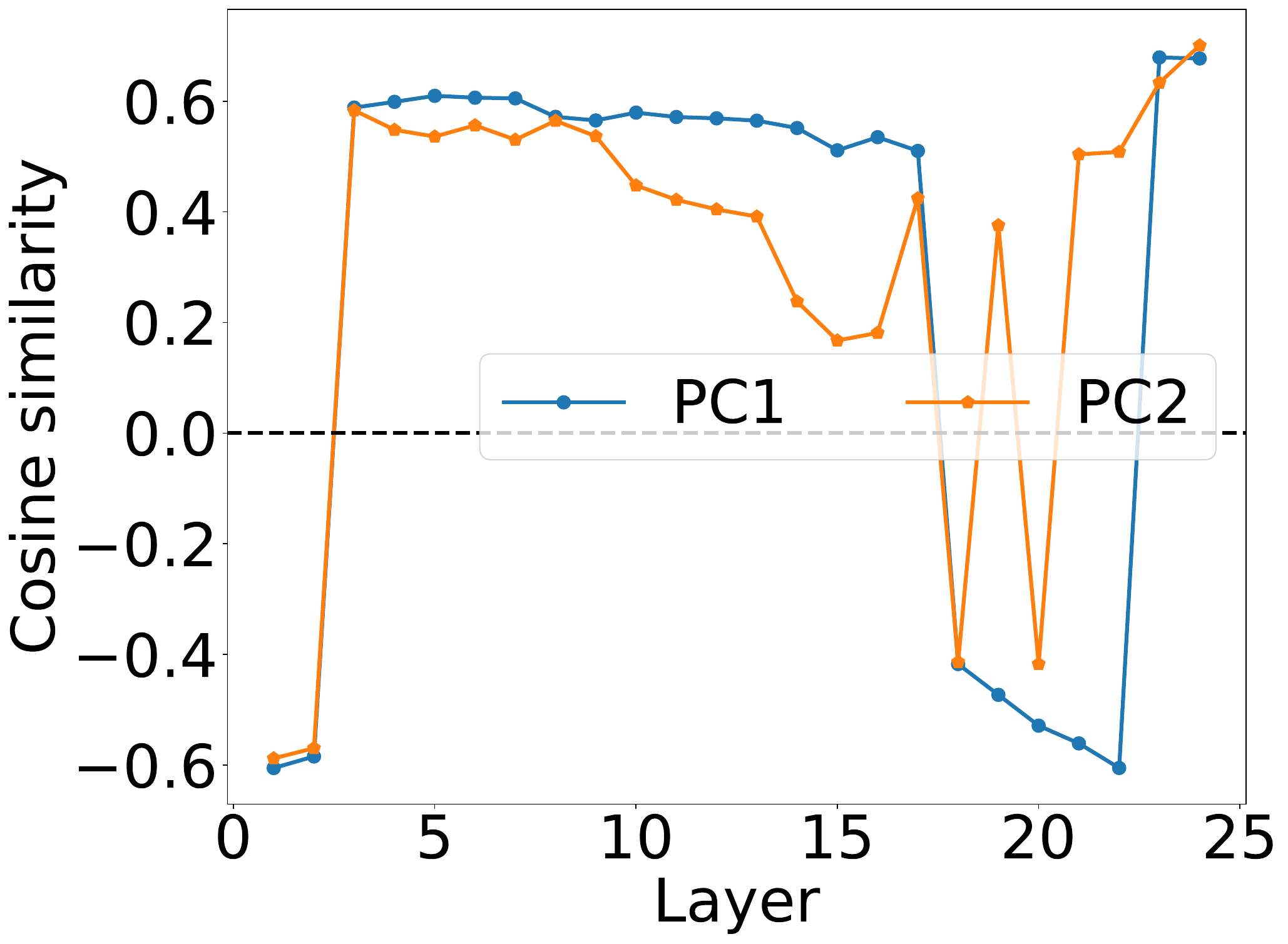}
        \caption{\texttt{en} \& \texttt{de} (POS)}
        \label{fig:cosine_en_de_lr}
    \end{subfigure}
    ~
    \begin{subfigure}{0.48\columnwidth}
        \includegraphics[width=\textwidth]{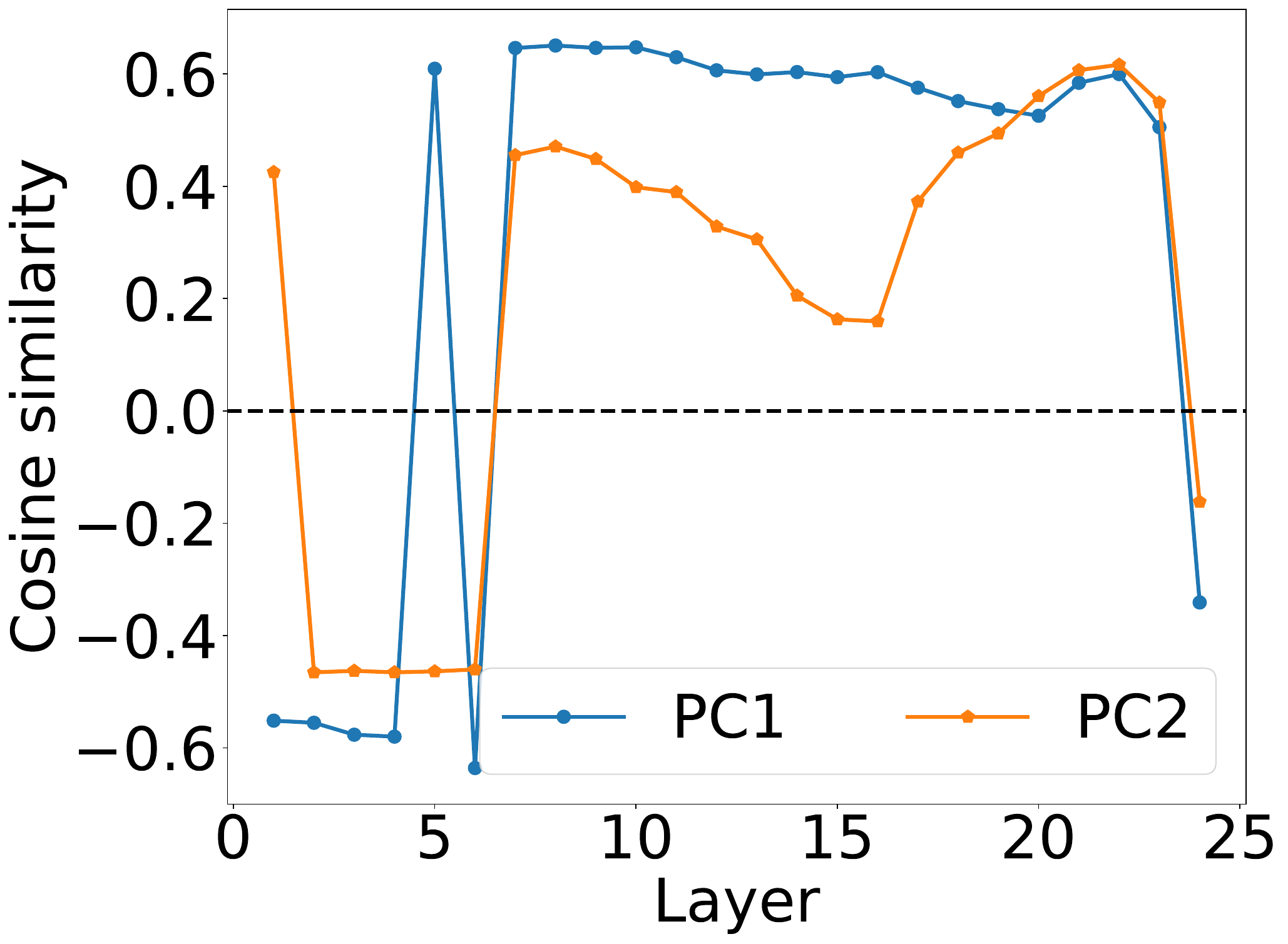}
        \caption{\texttt{en} \& \texttt{fr} (POS)}
        \label{fig:cosine_en_fr_lr}
    \end{subfigure}
\caption{Cosine similarity between the first two principal components of residual stream. PCA is computed on token representations with different POS on English and the target languages separately using LoRA adapters.
 }
\label{fig:cosine_lr}
\end{figure} 

\subsection{Details on identifying features critical for adaptation}
\label{sec:subspace_id}

By design, adapters modify outputs at each layer by adding to the residual stream. Building on our previous observations, we hypothesize that adapters function within a specific subspace of the residual stream. In the following, our objective is to identify the subspace used by adapters at each layer. 

We approach this via a sparse probing experiment, where we train probing classifiers to predict whether a given layer output has been adapted or not. We collect positive and negative examples by feeding sequences of each target language to the adapted models and select $6,000$ tokens randomly per layer. 
We considered two cases, and for both cases, for the positive examples, we kept adapters in place at every layer. For the first case (case 1), for the negative examples at layer $l$, we remove adapters at all layers, including $l$ itself. And secondly (case 2), for the negative examples at layer $l$, we keep adapters at all previous layers but remove the adapter at layer $l$ itself.

Given these representations, we follow \citet{gurnee2023finding} and use the scoring based maximum mean difference (MMD) algorithm to score each of the $1024$ features of the residual stream based on the absolute mean difference between the examples from the positive and negative class

\vspace{-0.25cm}
\begin{align}
    s_j^l = \frac{1}{\left|P \right|}\sum_{i=1}^{P}X_{ij}^l - \frac{1}{\left|N \right|}\sum_{i=1}^{N}X_{ij}^l~,
\end{align}

where $l$ is the layer index, $j$ is a single feature dimension, and $P$ and $N$ are the number of positive and negative examples, respectively.

After scoring, we select the top-k ranked features for $k\in \{1, 8, 16, 32, 64, 128, 256, 512\}$ and train a logistic regression probe to discriminate between the positive and negative examples.

\Cref{fig:probe_acc,fig:probe_acc2} show the probing accuracies from the two cases described above respectively. The results show that sparse probing classifiers can detection adaptation with high accuracy.

\subsection{More results on intervening on critical features increases perplexity}
\label{sec:appendix:intervention}

\begin{figure}[t]
\setlength{\belowcaptionskip}{-4pt}
  \centering
    \begin{subfigure}{0.45\columnwidth}
        \includegraphics[width=\textwidth]{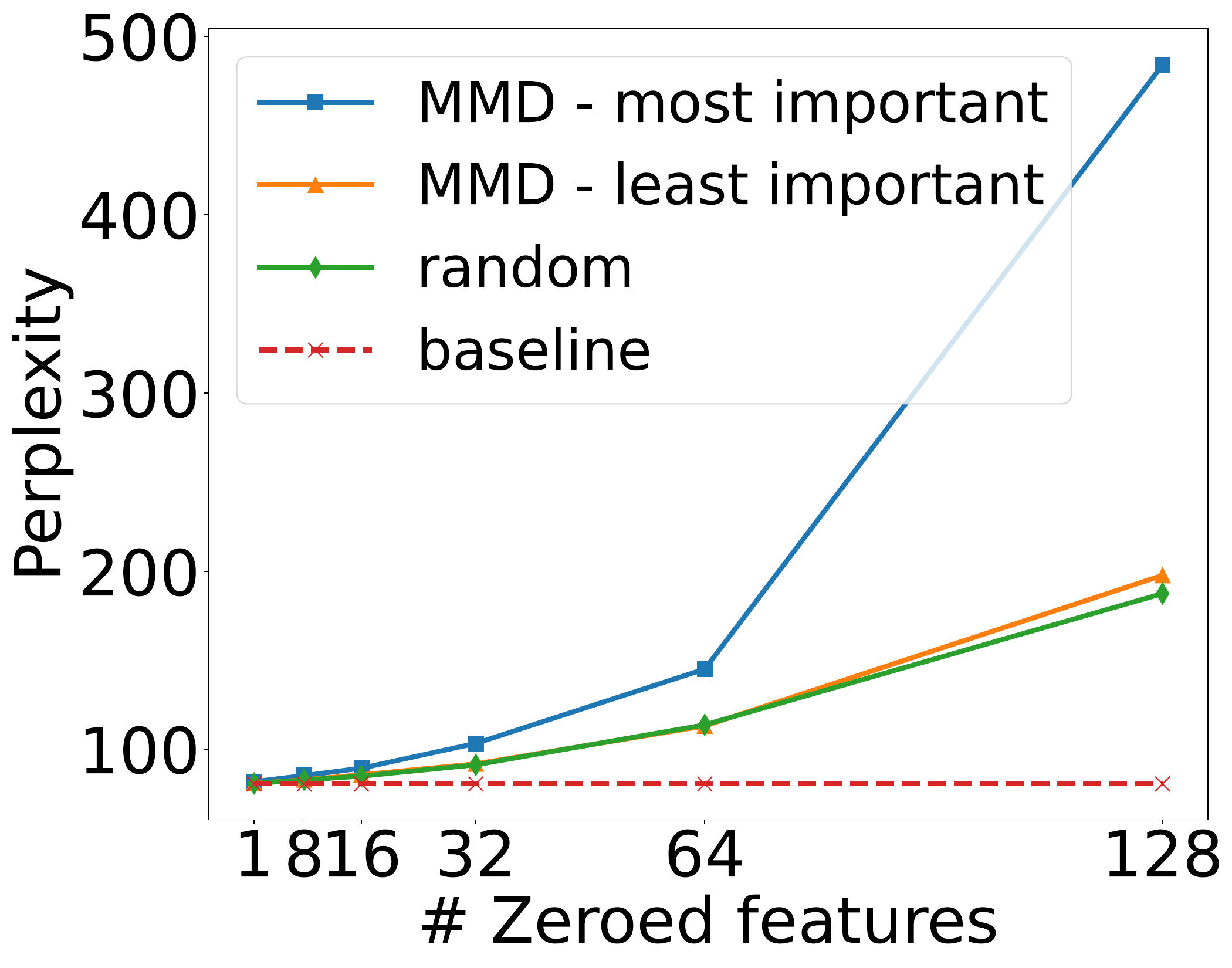}
        \caption{\texttt{en} $\rightarrow$ \texttt{he}}
        \label{fig:avg-intervention-he}
    \end{subfigure}
    ~
    \begin{subfigure}{0.45\columnwidth} 
        \includegraphics[width=\textwidth]{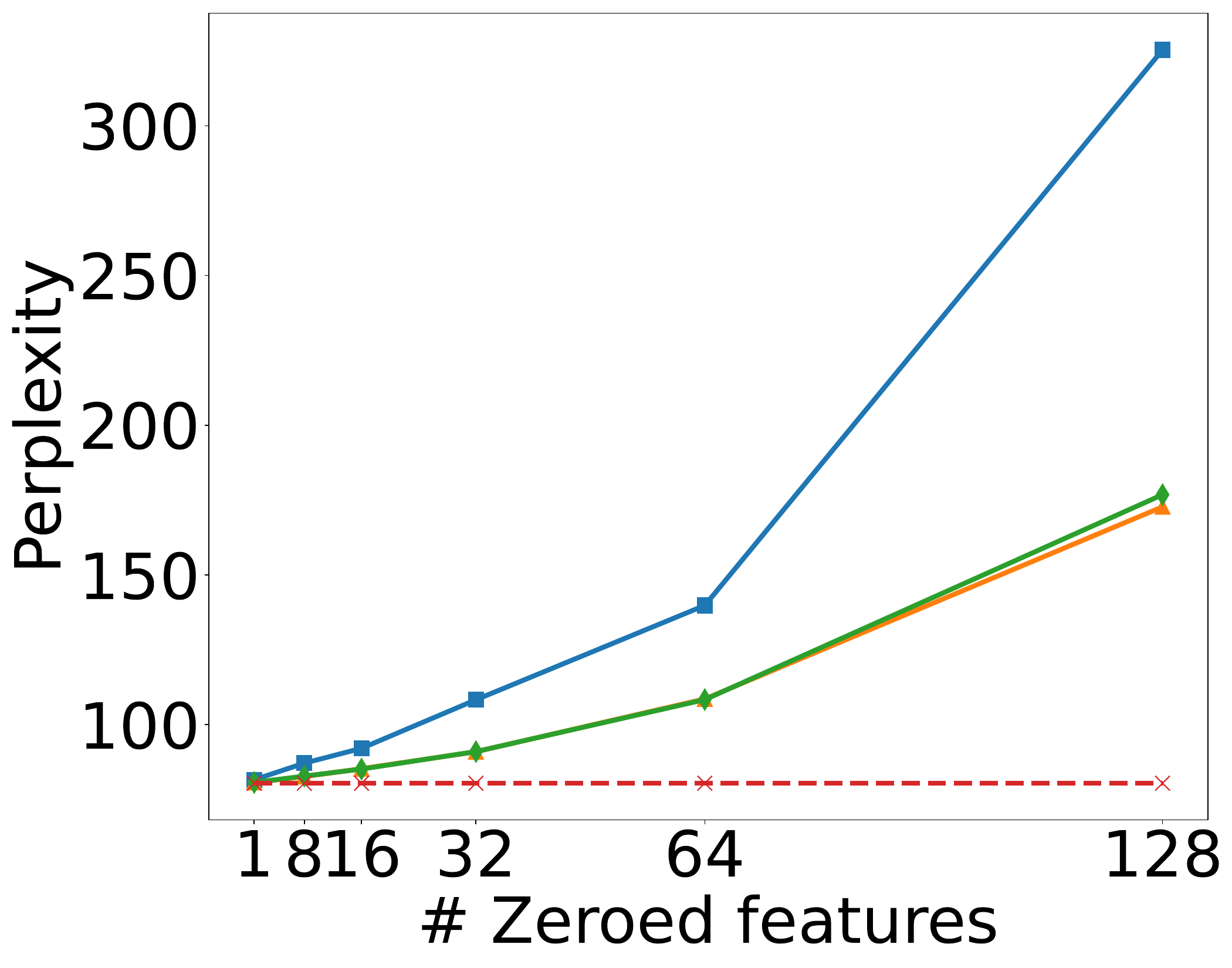}
        \caption{\texttt{en} $\rightarrow$ \texttt{ar}}
        \label{fig:avg-intervention-ar}
    \end{subfigure}
    \\
    \begin{subfigure}{0.45\columnwidth}
        \includegraphics[width=\textwidth]{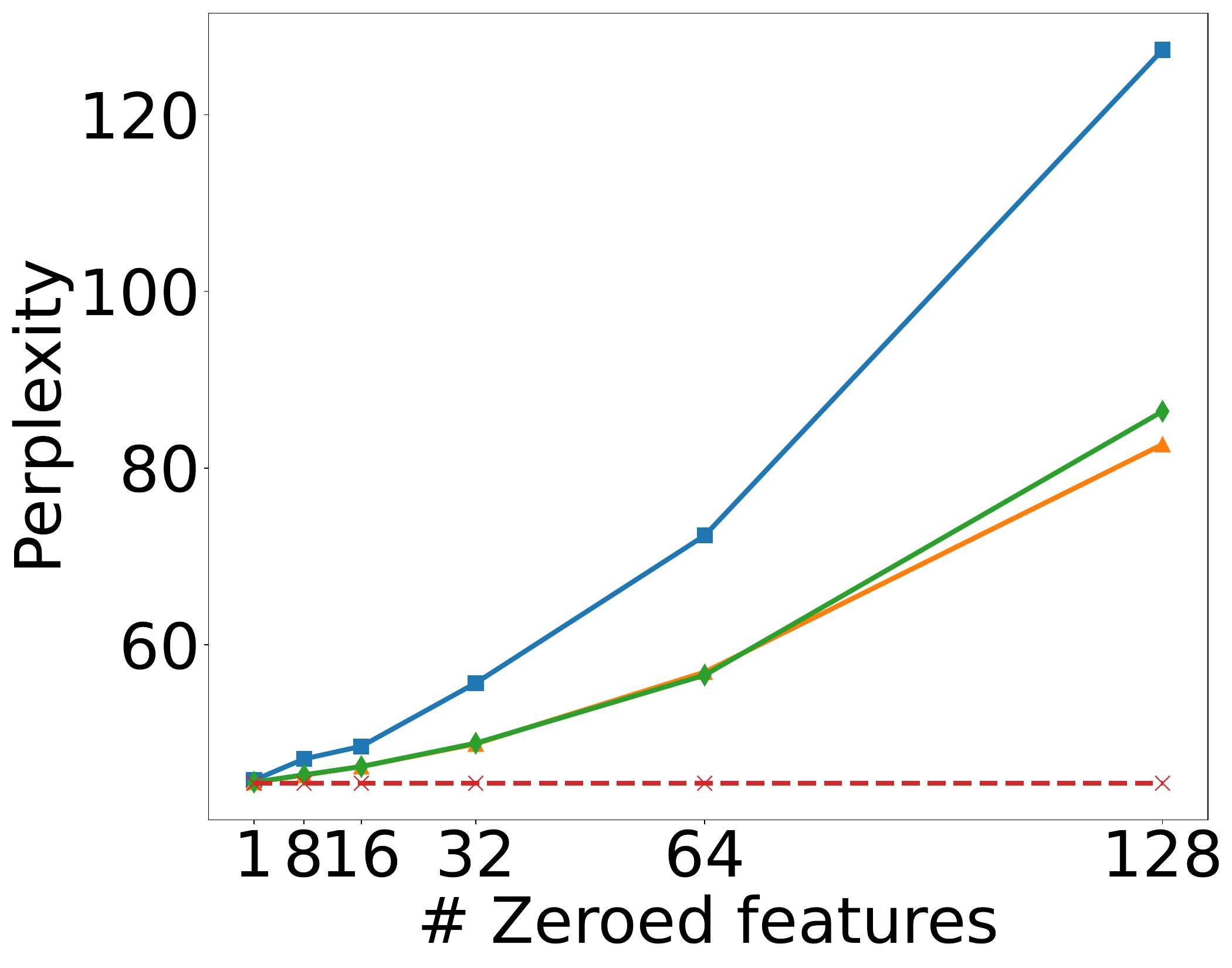}
        \caption{\texttt{en} $\rightarrow$ \texttt{de}}
        \label{fig:avg-intervention-de}
    \end{subfigure}
    ~
    \begin{subfigure}{0.45\columnwidth} 
        \includegraphics[width=\textwidth]{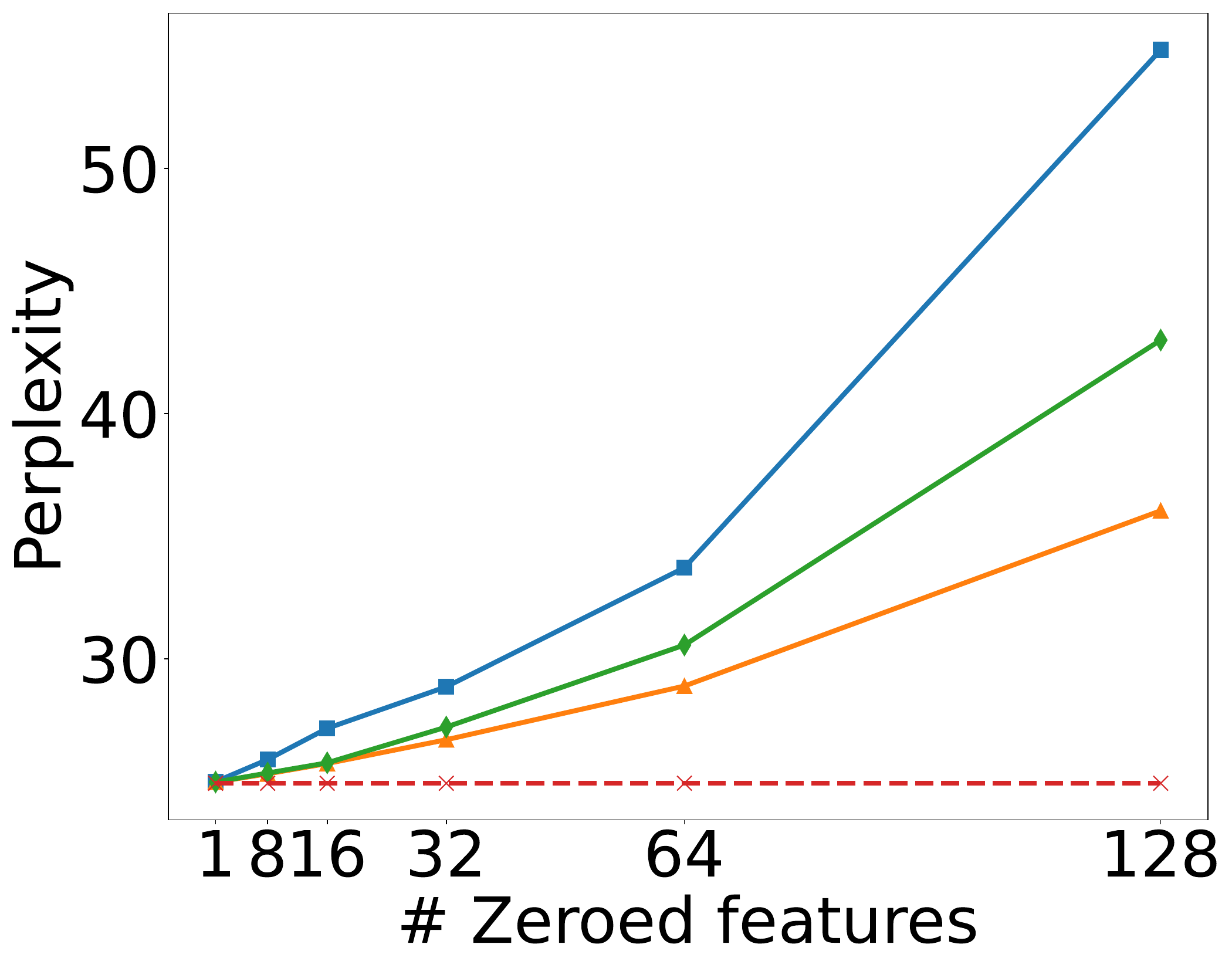}
        \caption{\texttt{en} $\rightarrow$ \texttt{fr}}
        \label{fig:avg-intervention-fr}
    \end{subfigure}
\caption{Adapted model perplexity after using the average of non-zero features for intervention.
}
\label{apx:avg_interventions}
\end{figure}

Instead of zeroing out the adapter representation before updating the residual stream we experiment with an alternative intervention which replaces the features to be intervened on by the average values of the remaining features. We compare the target language validation perplexity after the intervention in \Cref{apx:avg_interventions}. Our results show that while intervening on the most important features leads to the largest increase in perplexity, intervening on the least important or randomly selection similarly leads to a considerable increase in perplexity.

\subsection{More results on adapters largely preserve the structure of the underlying model}
\label{sec:appendix:overlap}

\begin{figure}[t]
  \centering
    \begin{subfigure}{0.48\columnwidth}
        \includegraphics[width=\textwidth]{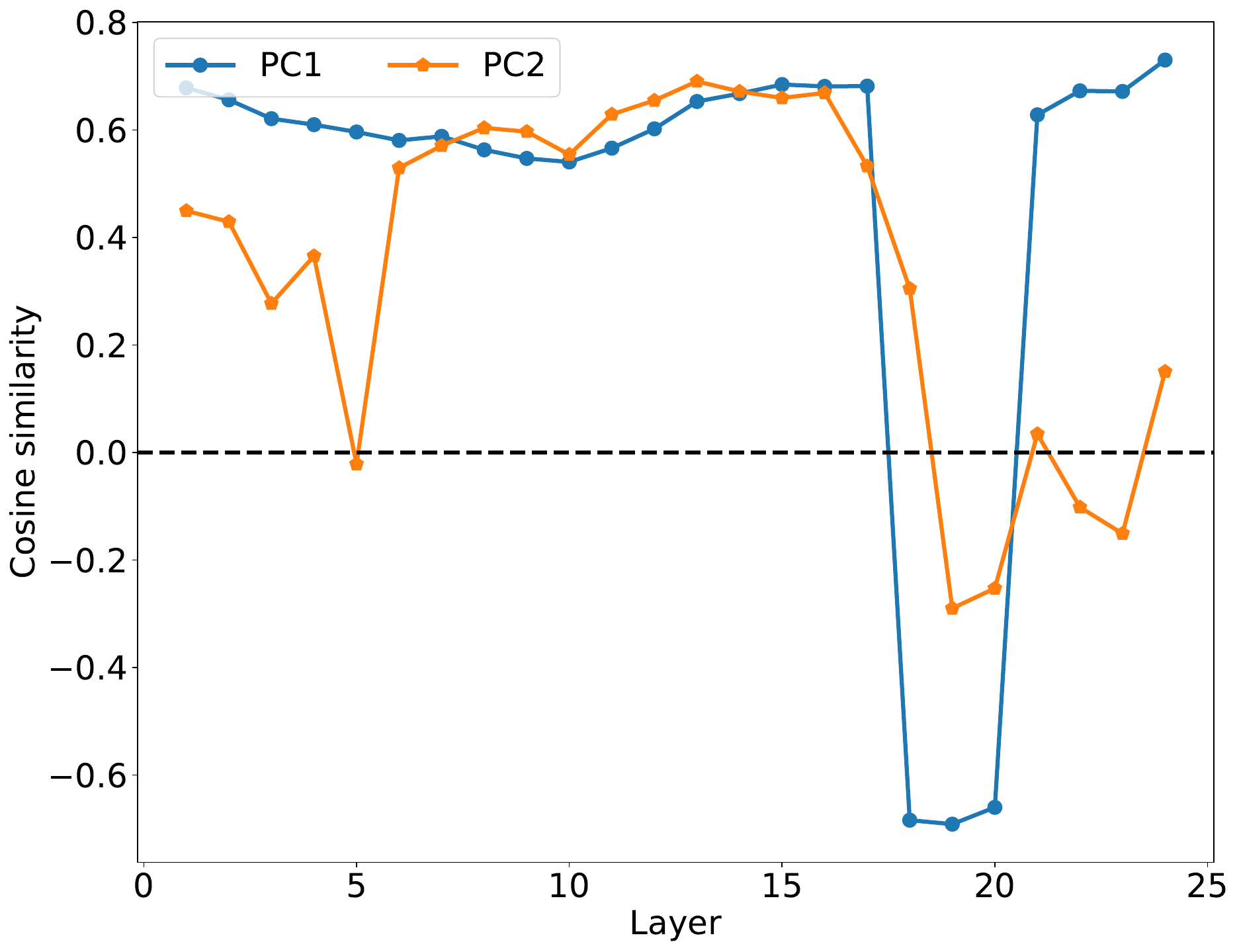}
        \caption{\texttt{en} \& \texttt{de} (number)}
        \label{fig:appendix:cosine_en_de_nn}
    \end{subfigure}
    ~
    \begin{subfigure}{0.48\columnwidth}
        \includegraphics[width=\textwidth]{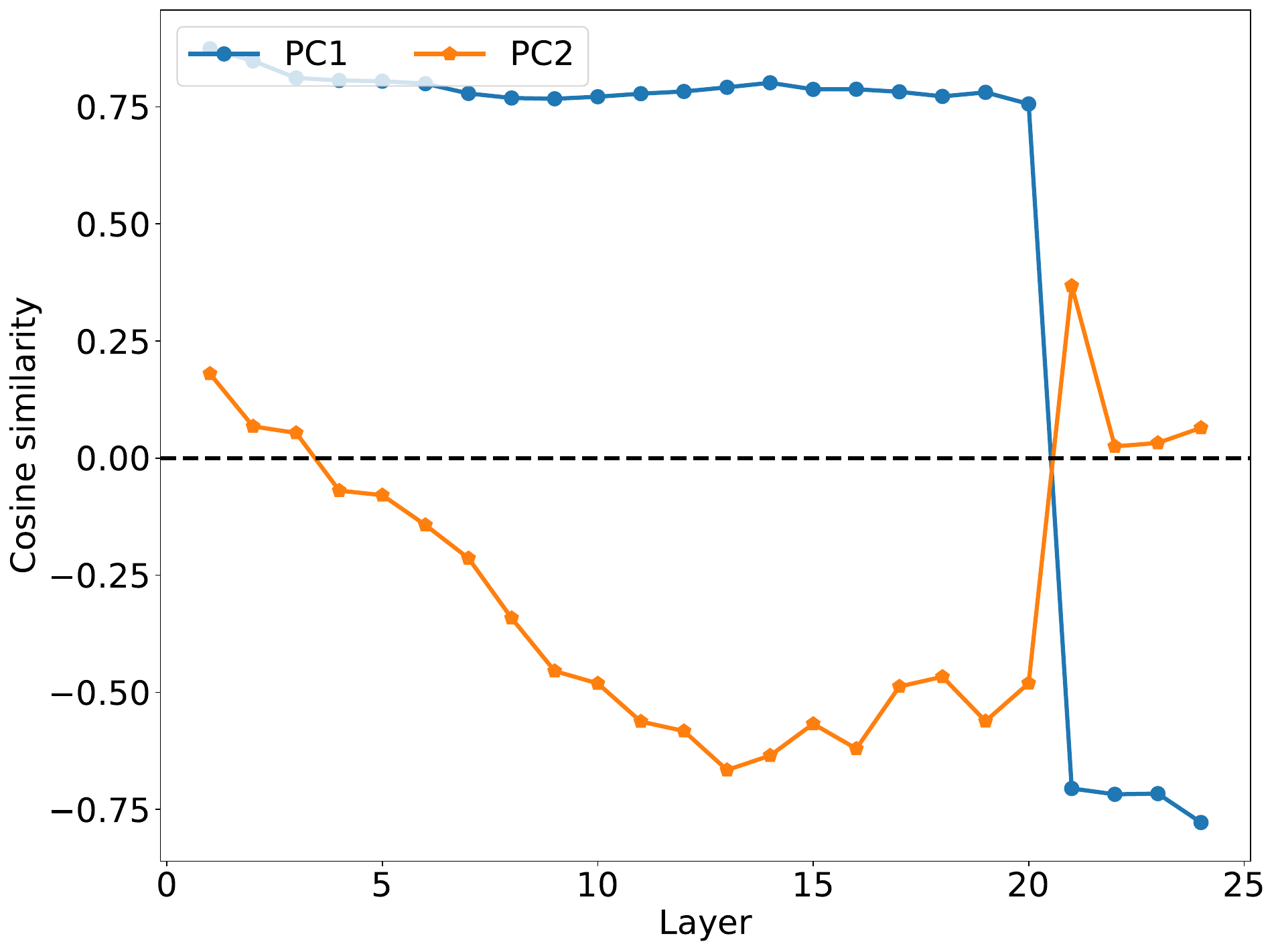}
        \caption{\texttt{en} \& \texttt{fr} (number)}
        \label{fig:appendix:cosine_en_fr_nn}
    \end{subfigure}
    \\
    \begin{subfigure}{0.48\columnwidth}
        \includegraphics[width=\textwidth]{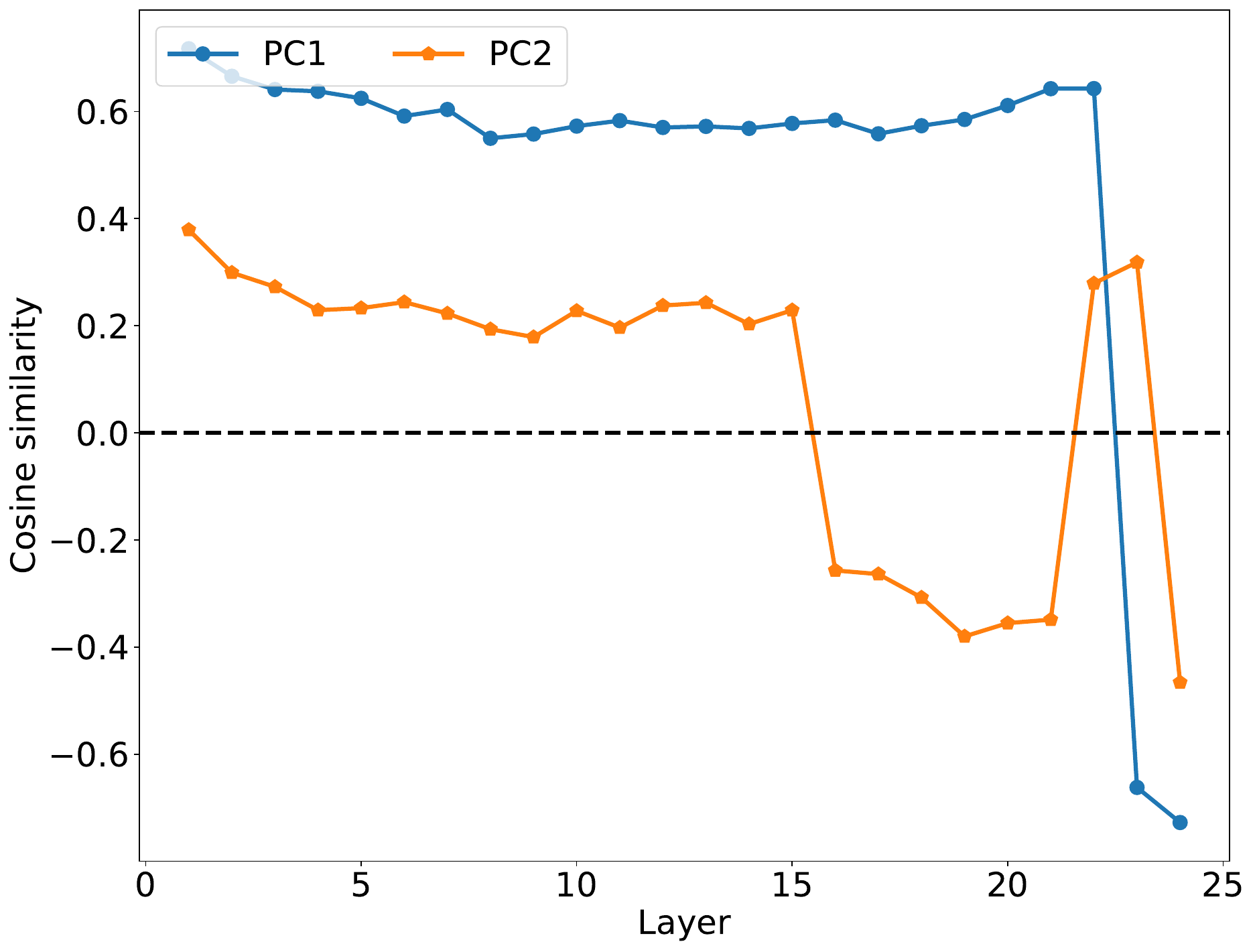}
        \caption{\texttt{en} \& \texttt{de} (tense)}
        \label{fig:appendix:cosine_en_de}
    \end{subfigure}
    ~
    \begin{subfigure}{0.48\columnwidth}
        \includegraphics[width=\textwidth]{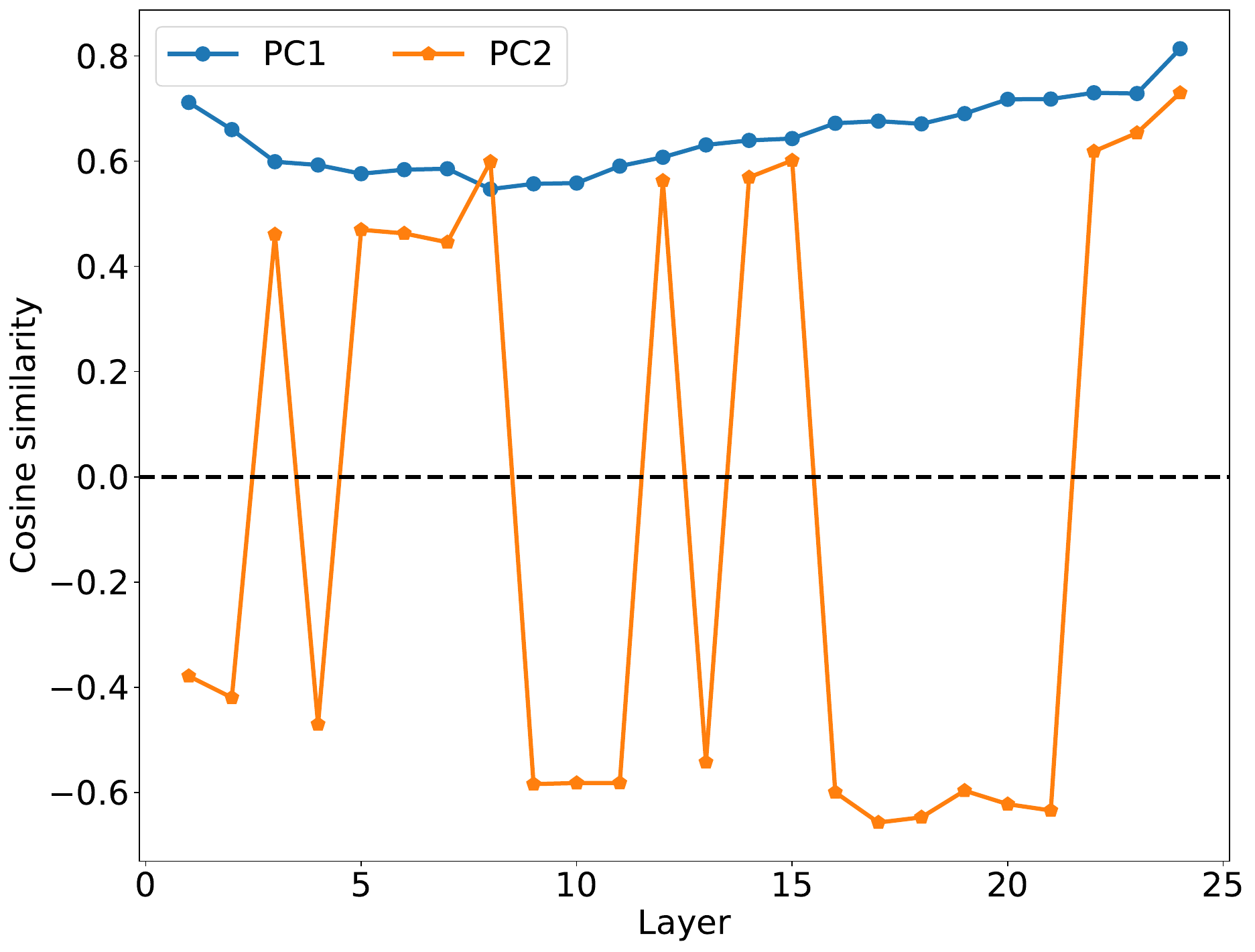}
        \caption{\texttt{en} \& \texttt{fr} (tense)}
        \label{fig:appendix:cosine_en_fr}
    \end{subfigure}
\caption{Cosine similarity between the first two principal components of residual stream. PCA is computed on token representations with different number or tense on English and the target languages separately.
}
\label{fig:appendix:cosine_number_tense}
\end{figure} 

\paragraph{Part of speech}

\begin{figure*}[p]
    \centering    
    \begin{subfigure}{0.25\textwidth}
        \includegraphics[width=\textwidth]{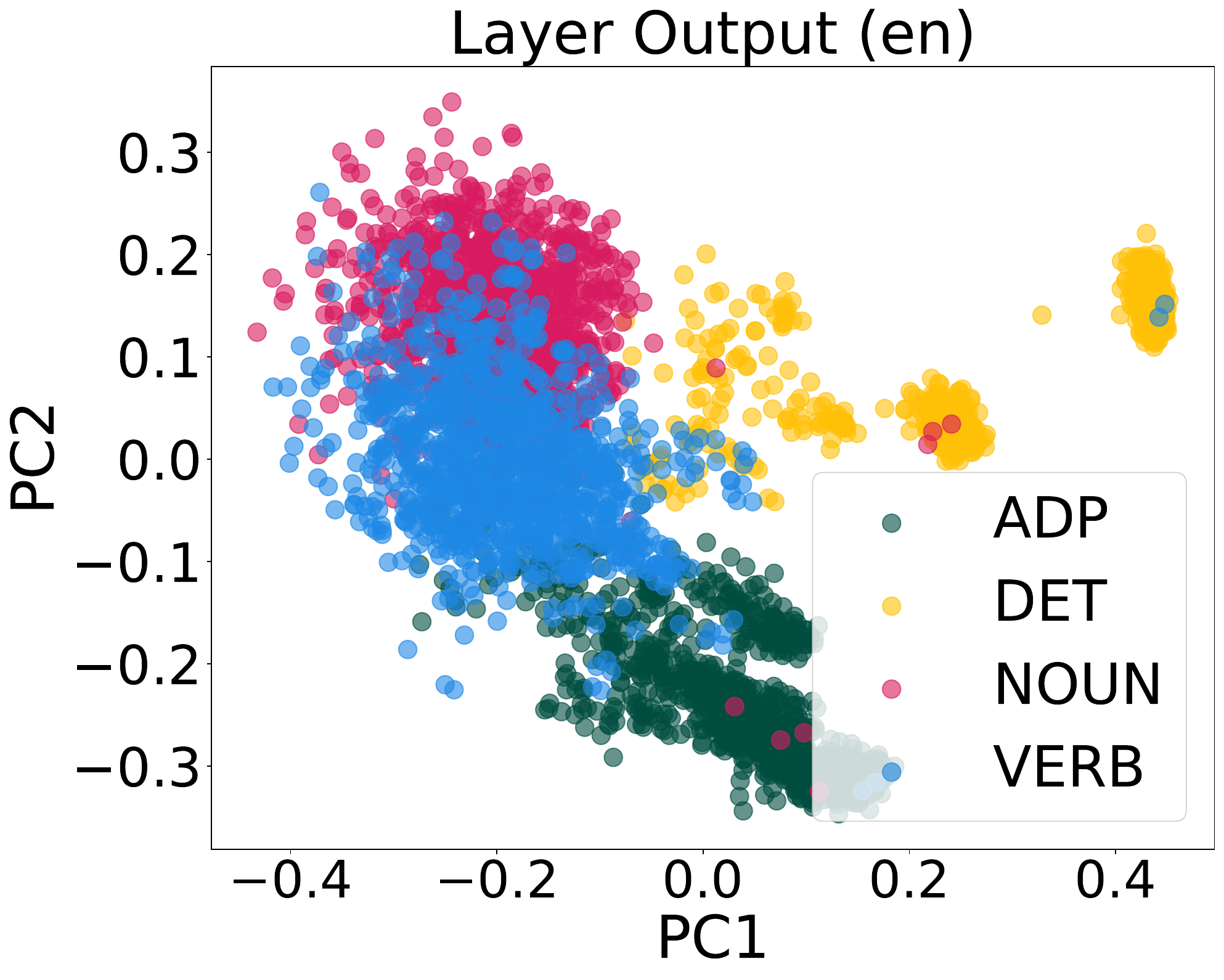}
        \caption{Layer 1}
    \end{subfigure}
    ~
    \begin{subfigure}{0.25\textwidth}
        \includegraphics[width=\textwidth]{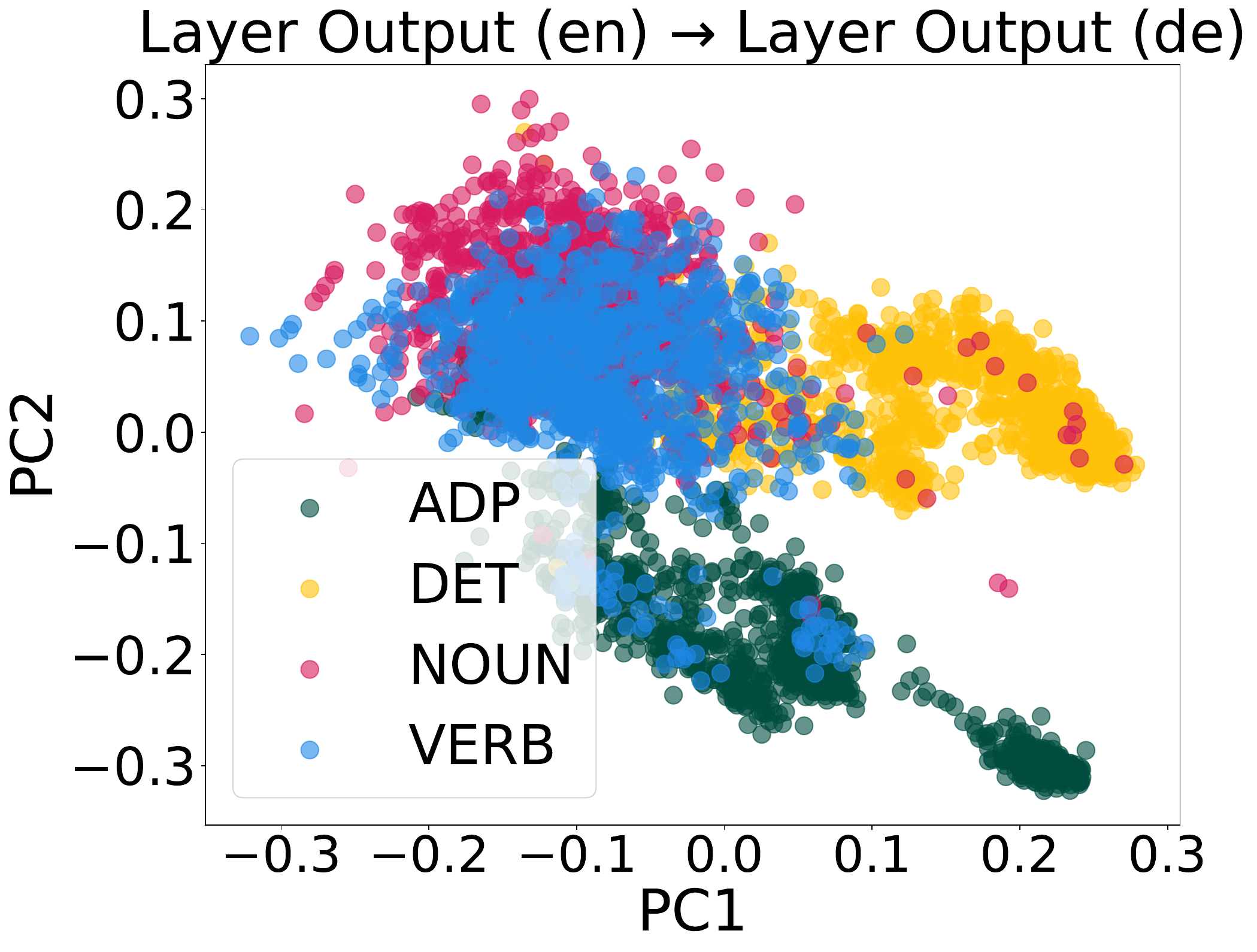}
        \caption{Layer 1}
    \end{subfigure}
    ~
    \begin{subfigure}{0.25\textwidth}
        \includegraphics[width=\textwidth]{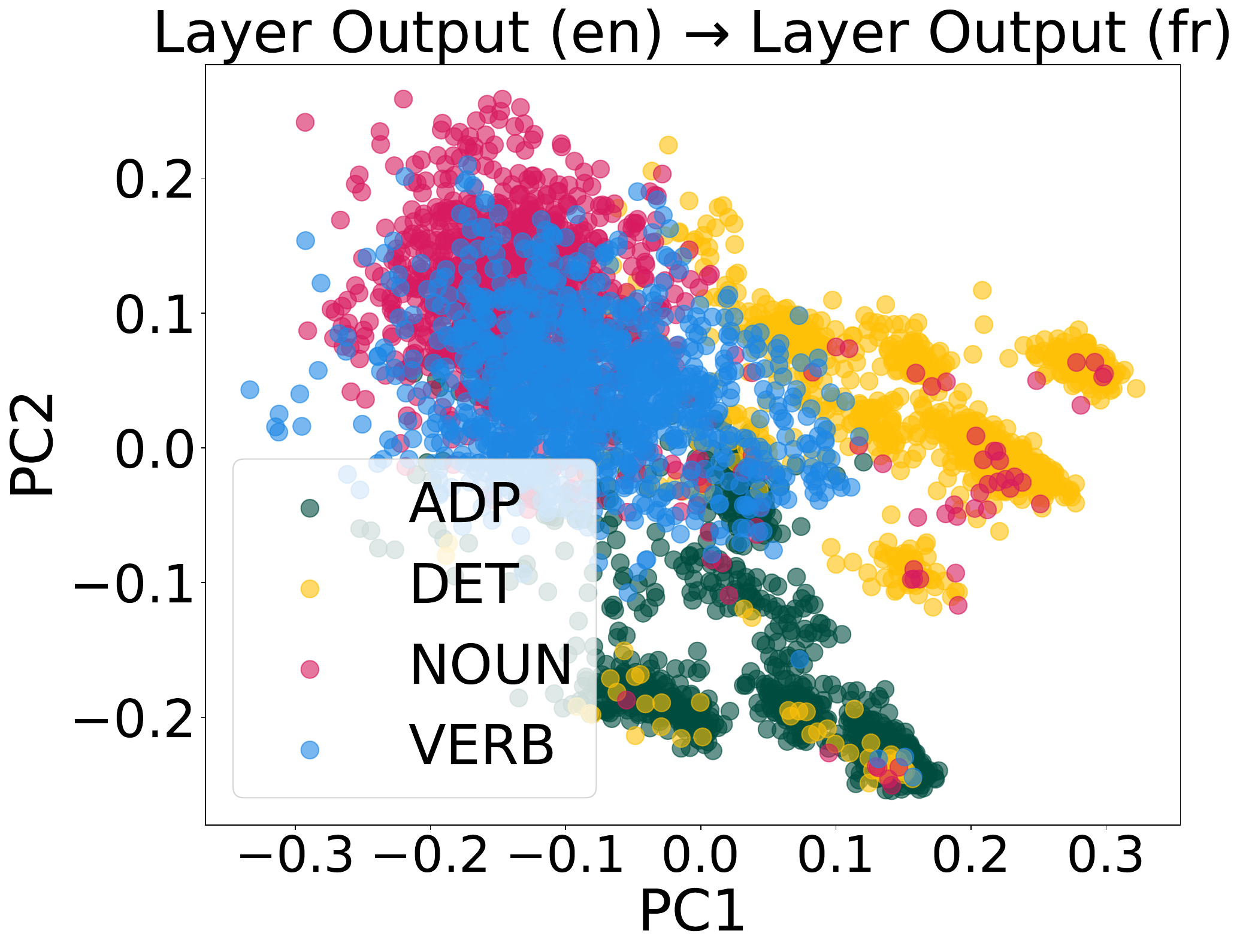}
        \caption{Layer 1}
    \end{subfigure}
    \\
    \begin{subfigure}{0.25\textwidth}
        \includegraphics[width=\textwidth]{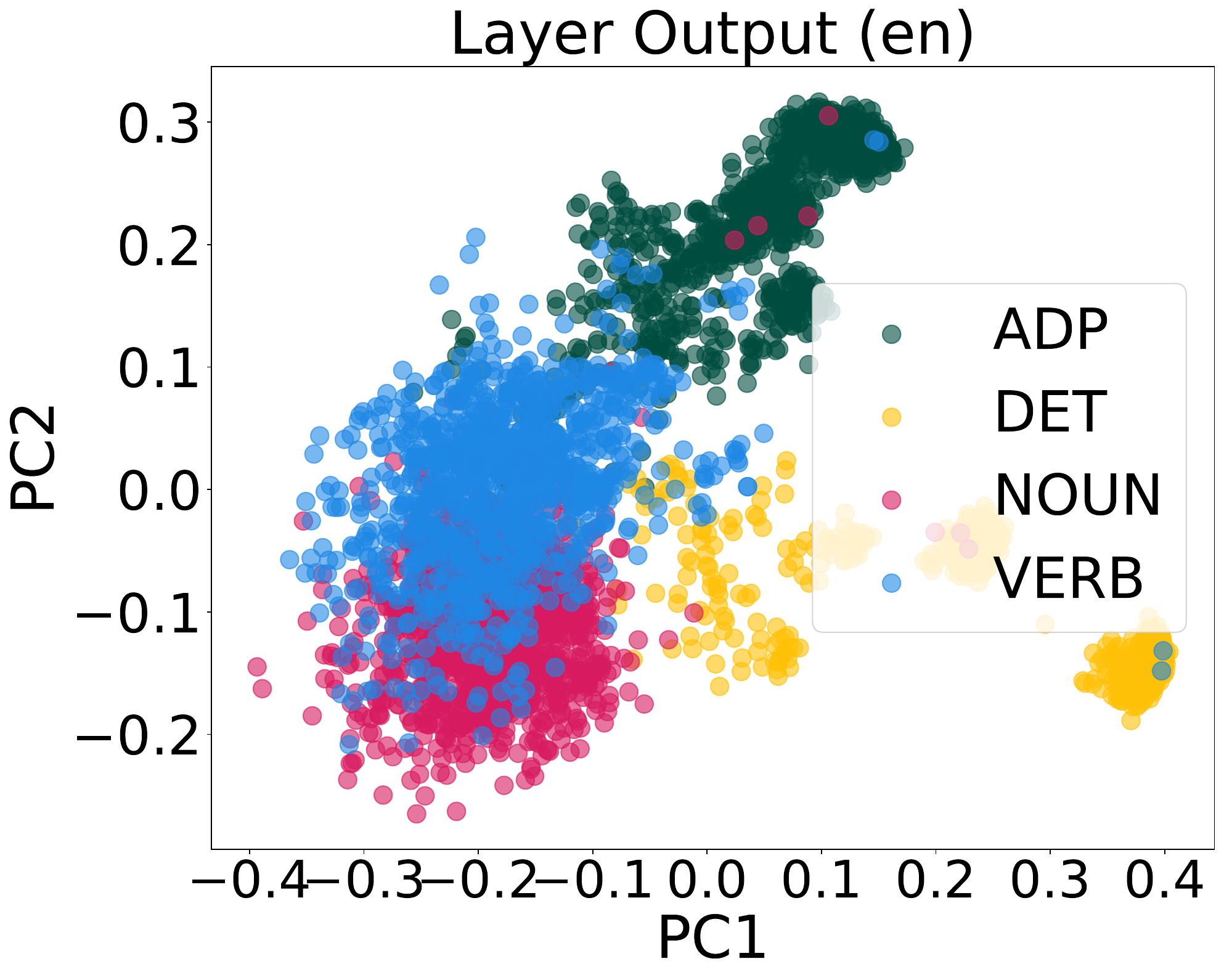}
        \caption{Layer 2}
    \end{subfigure}
    ~
    \begin{subfigure}{0.25\textwidth}
        \includegraphics[width=\textwidth]{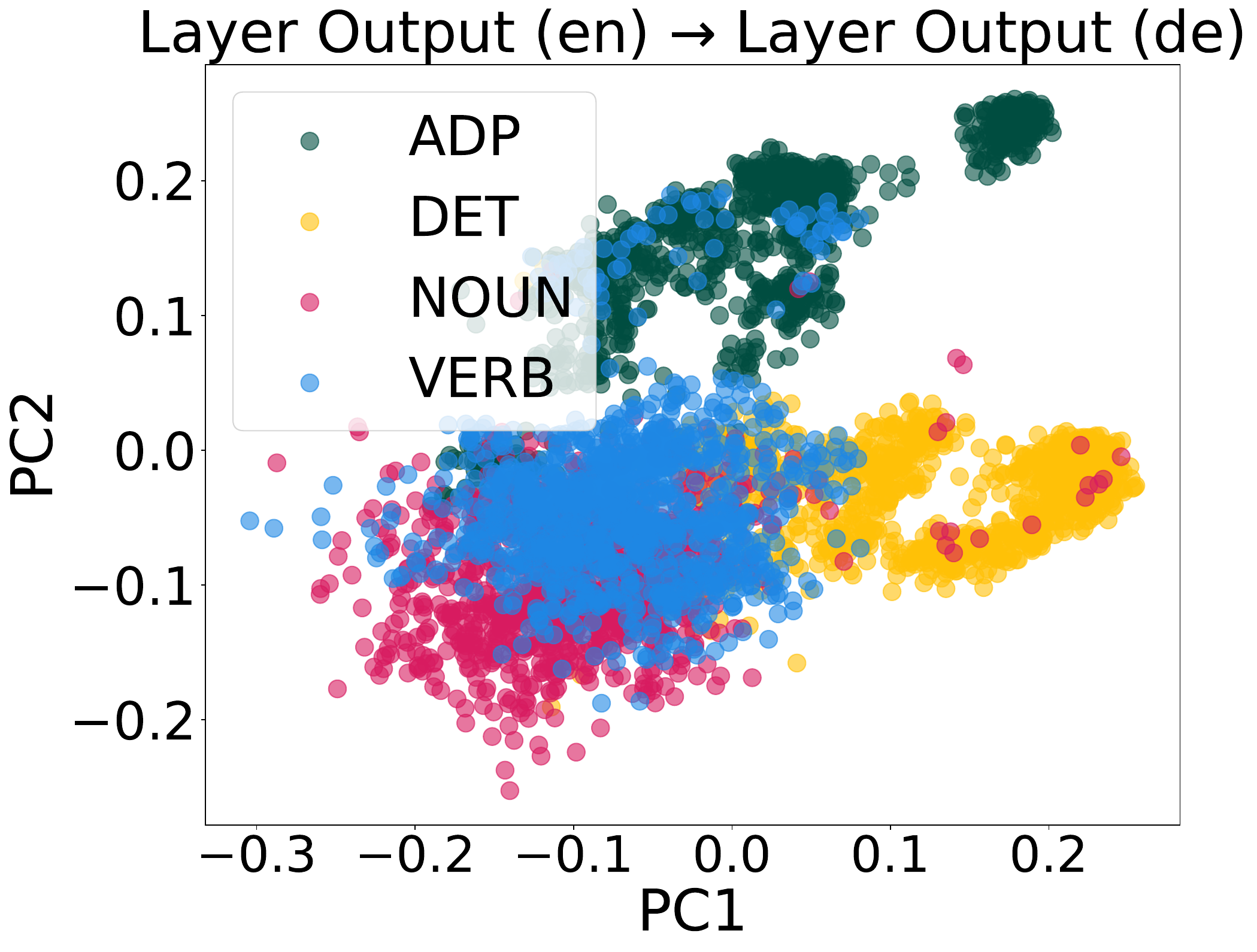}
        \caption{Layer 2}
    \end{subfigure}
    ~
    \begin{subfigure}{0.25\textwidth}
        \includegraphics[width=\textwidth]{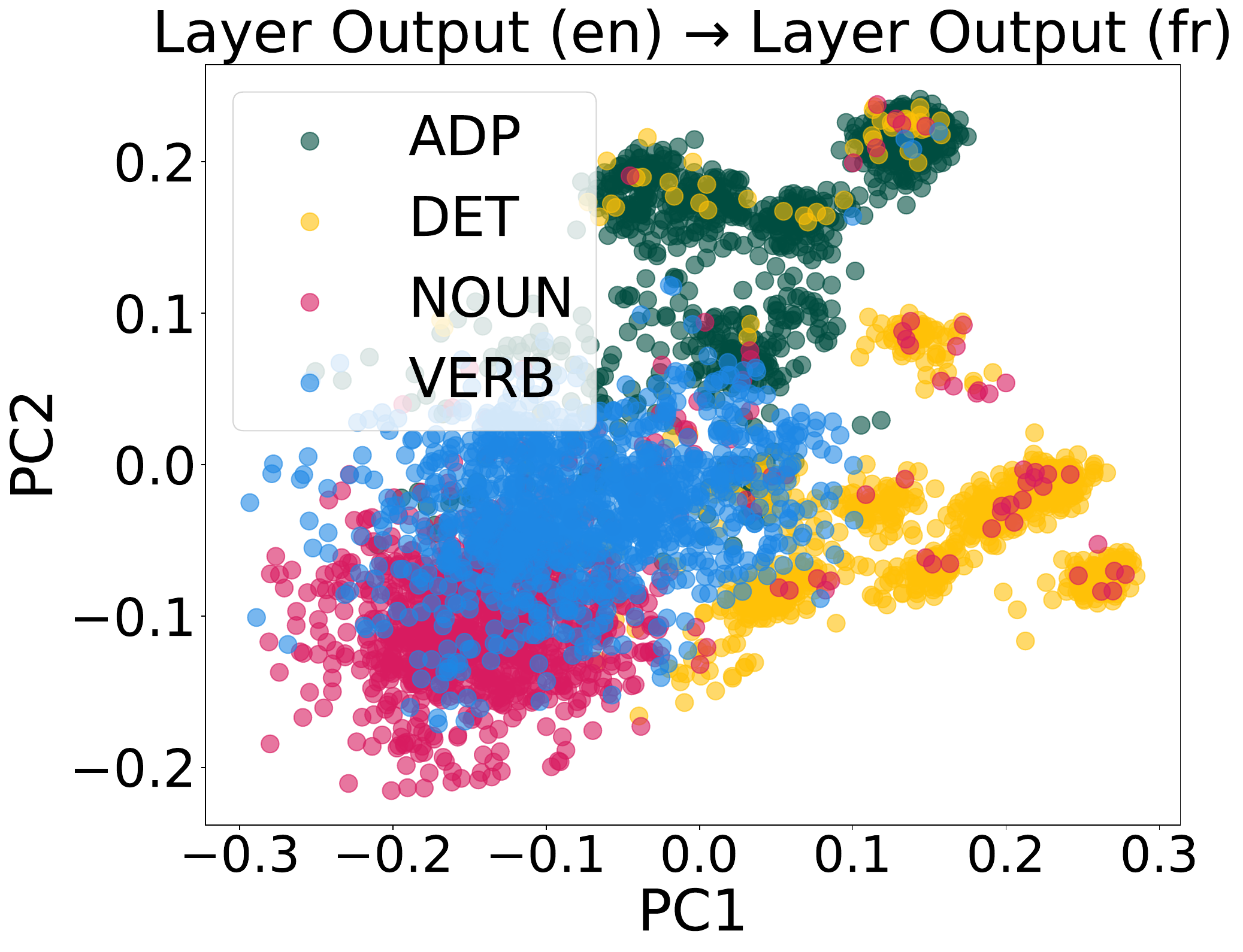}
        \caption{Layer 2}
    \end{subfigure}
    \\
    \begin{subfigure}{0.25\textwidth}
        \includegraphics[width=\textwidth]{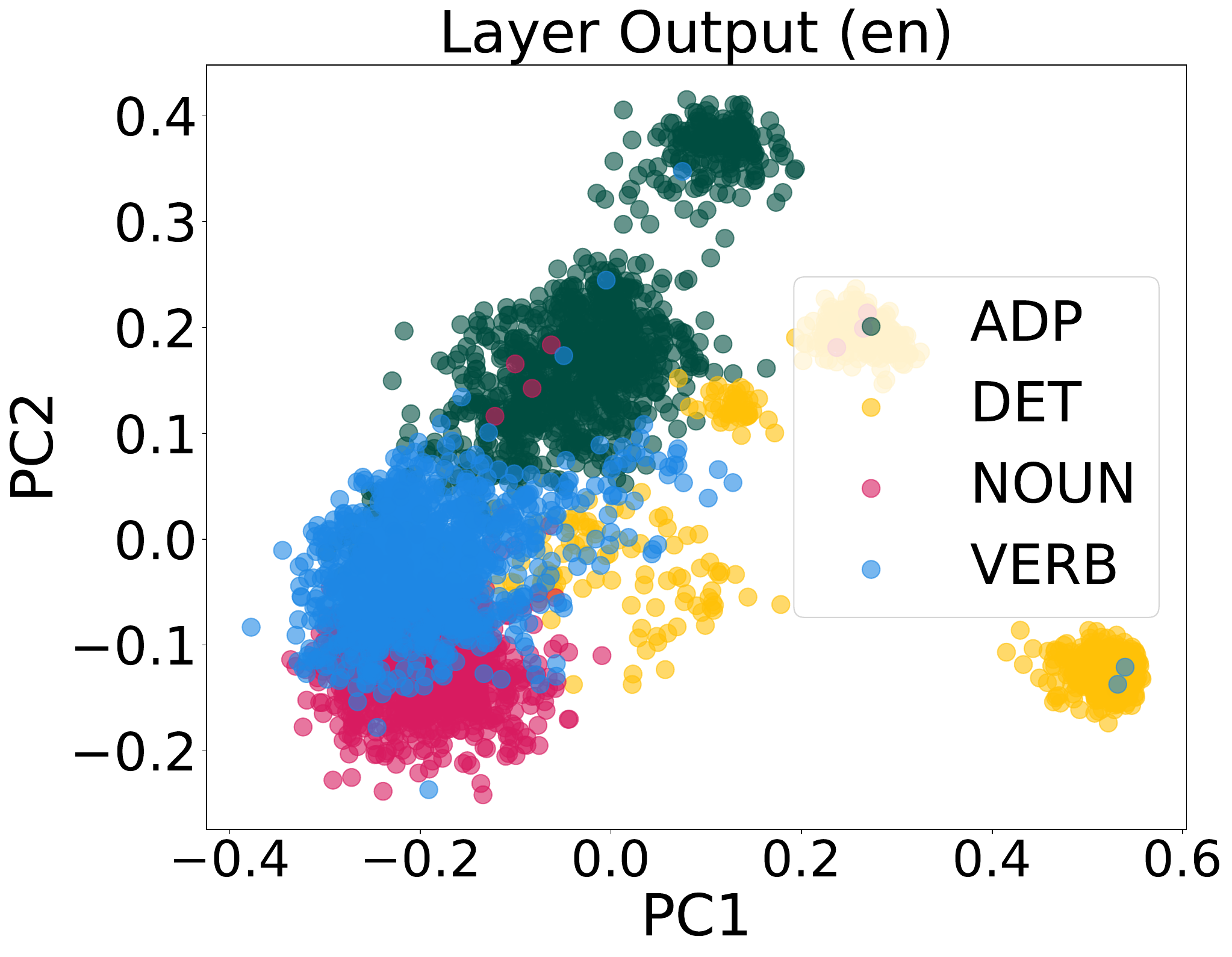}
        \caption{Layer 8}
    \end{subfigure}
    ~
    \begin{subfigure}{0.25\textwidth}
        \includegraphics[width=\textwidth]{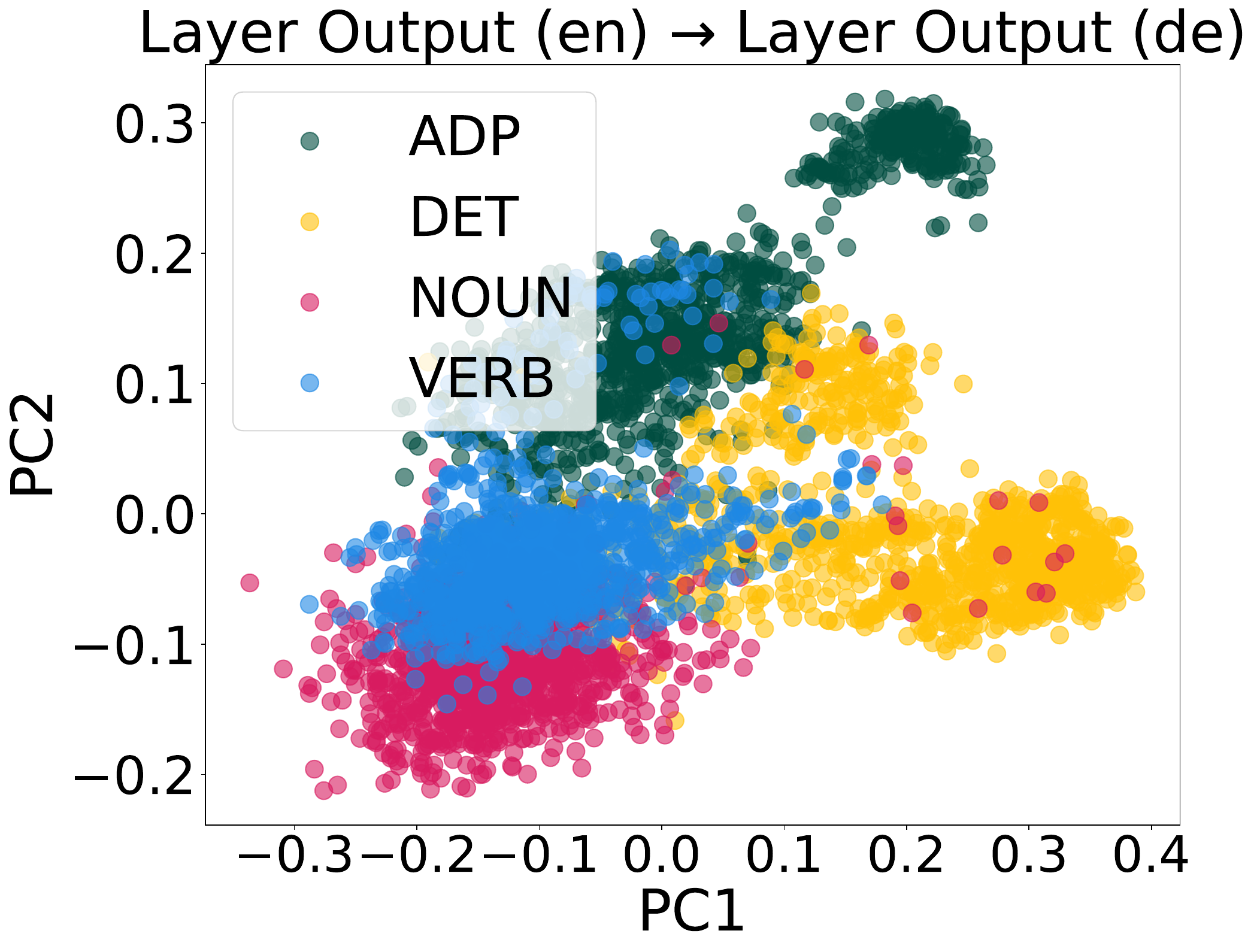}
        \caption{Layer 8}
    \end{subfigure}
    ~
    \begin{subfigure}{0.25\textwidth}
        \includegraphics[width=\textwidth]{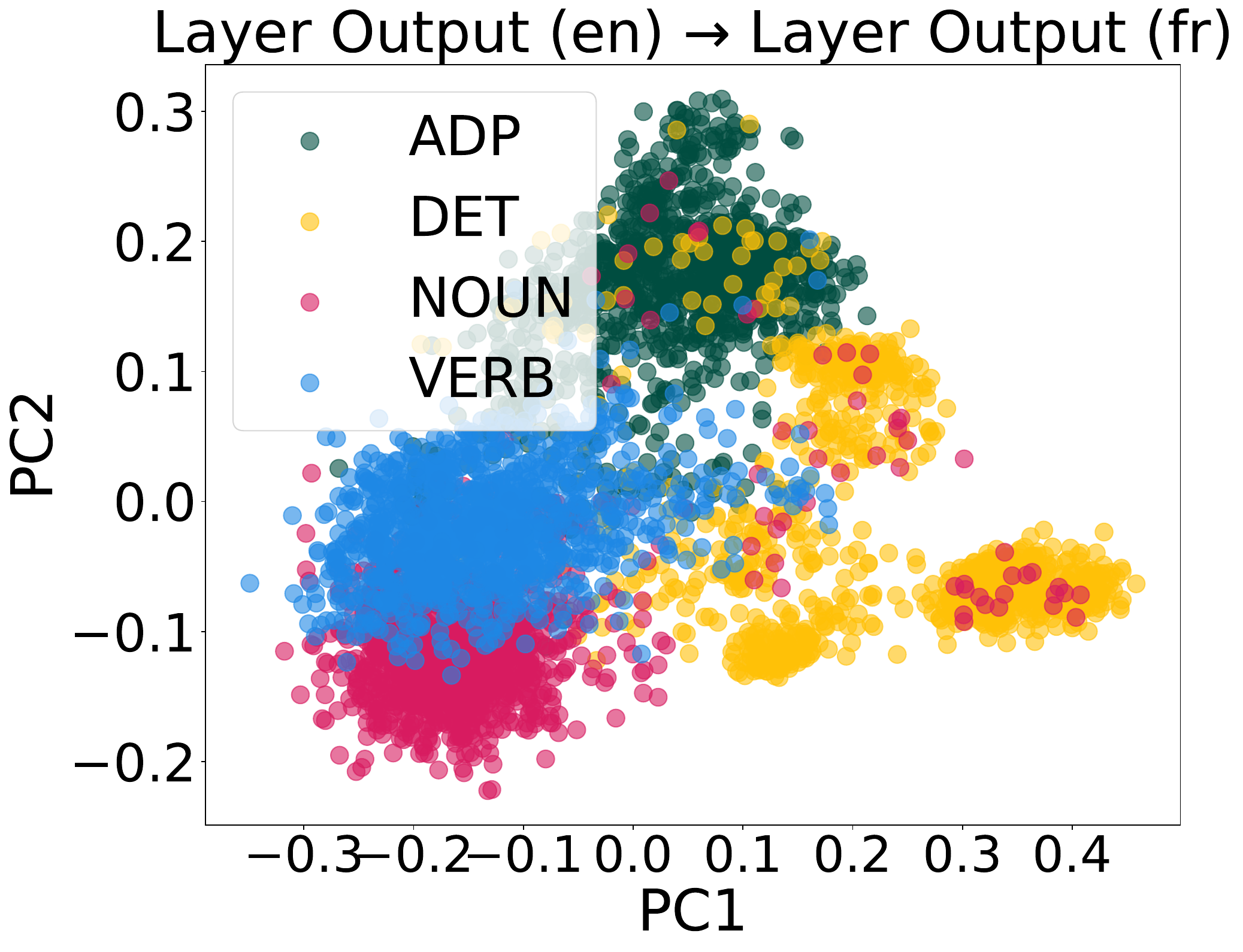}
        \caption{Layer 8}
    \end{subfigure}
    \\
    \begin{subfigure}{0.25\textwidth}
        \includegraphics[width=\textwidth]{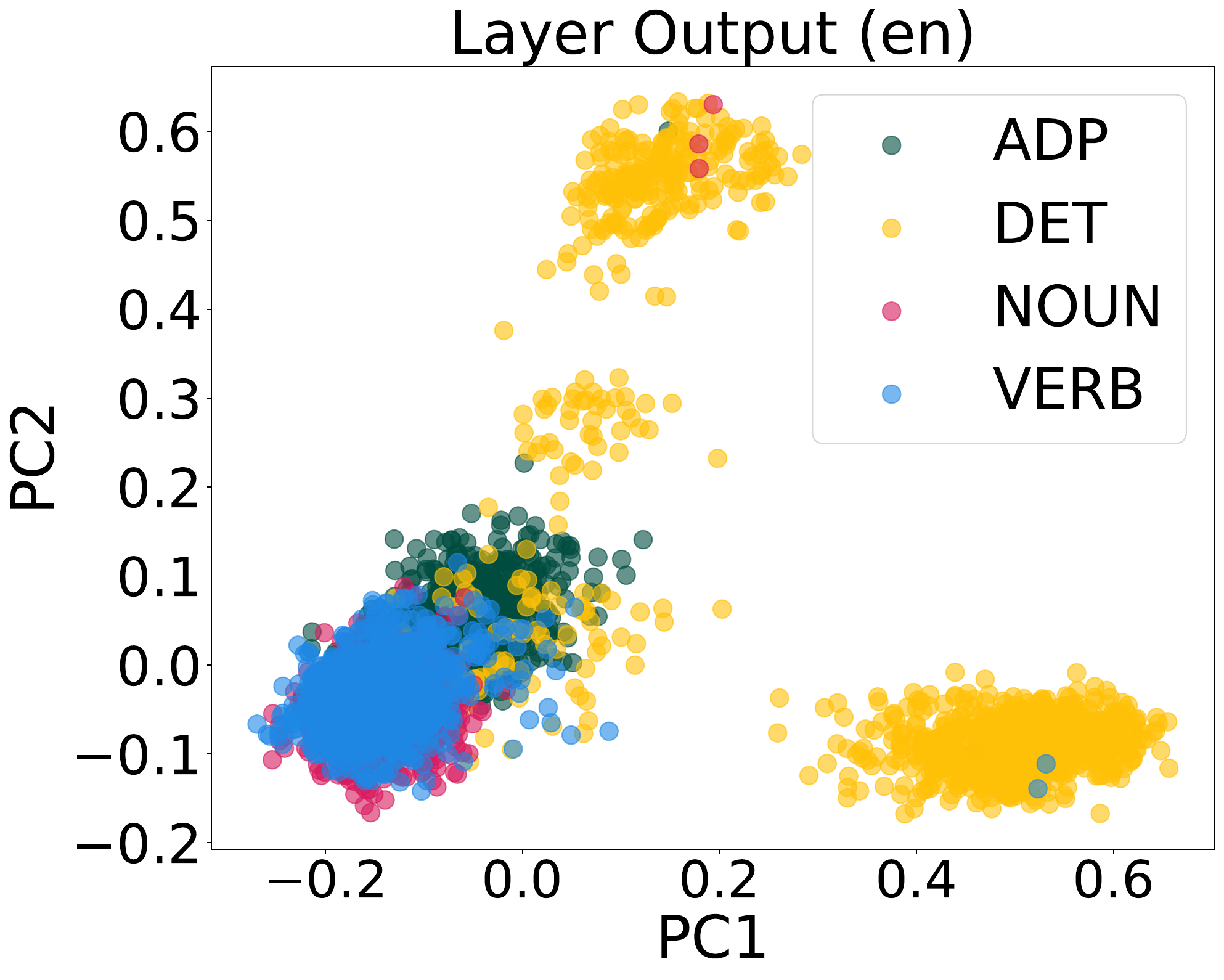}
        \caption{Layer 16}
    \end{subfigure}
    ~
    \begin{subfigure}{0.25\textwidth}
        \includegraphics[width=\textwidth]{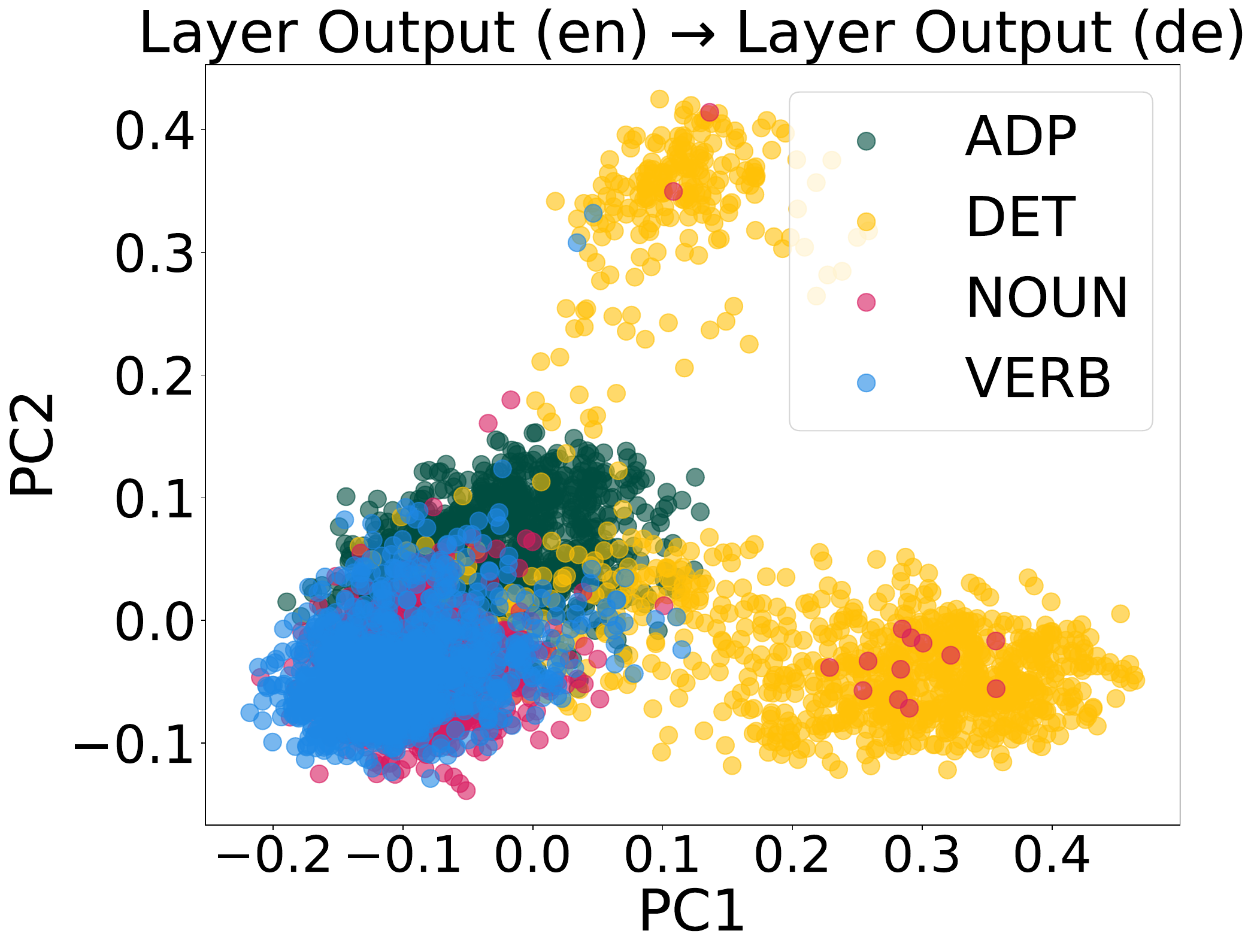}
        \caption{Layer 16}
    \end{subfigure}
    ~
    \begin{subfigure}{0.25\textwidth}
        \includegraphics[width=\textwidth]{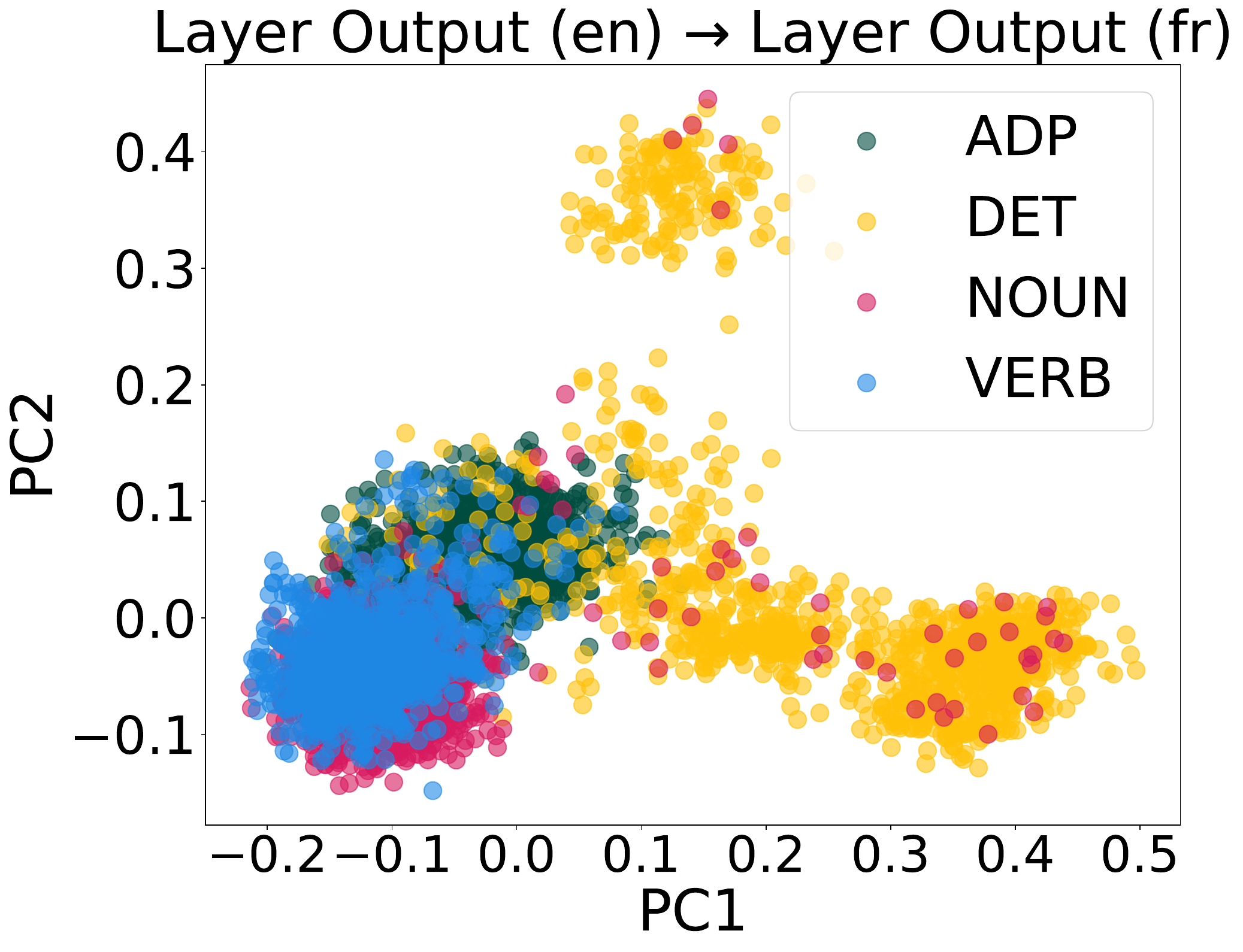}
        \caption{Layer 16}
    \end{subfigure}
    \\
    \begin{subfigure}{0.25\textwidth}
        \includegraphics[width=\textwidth]{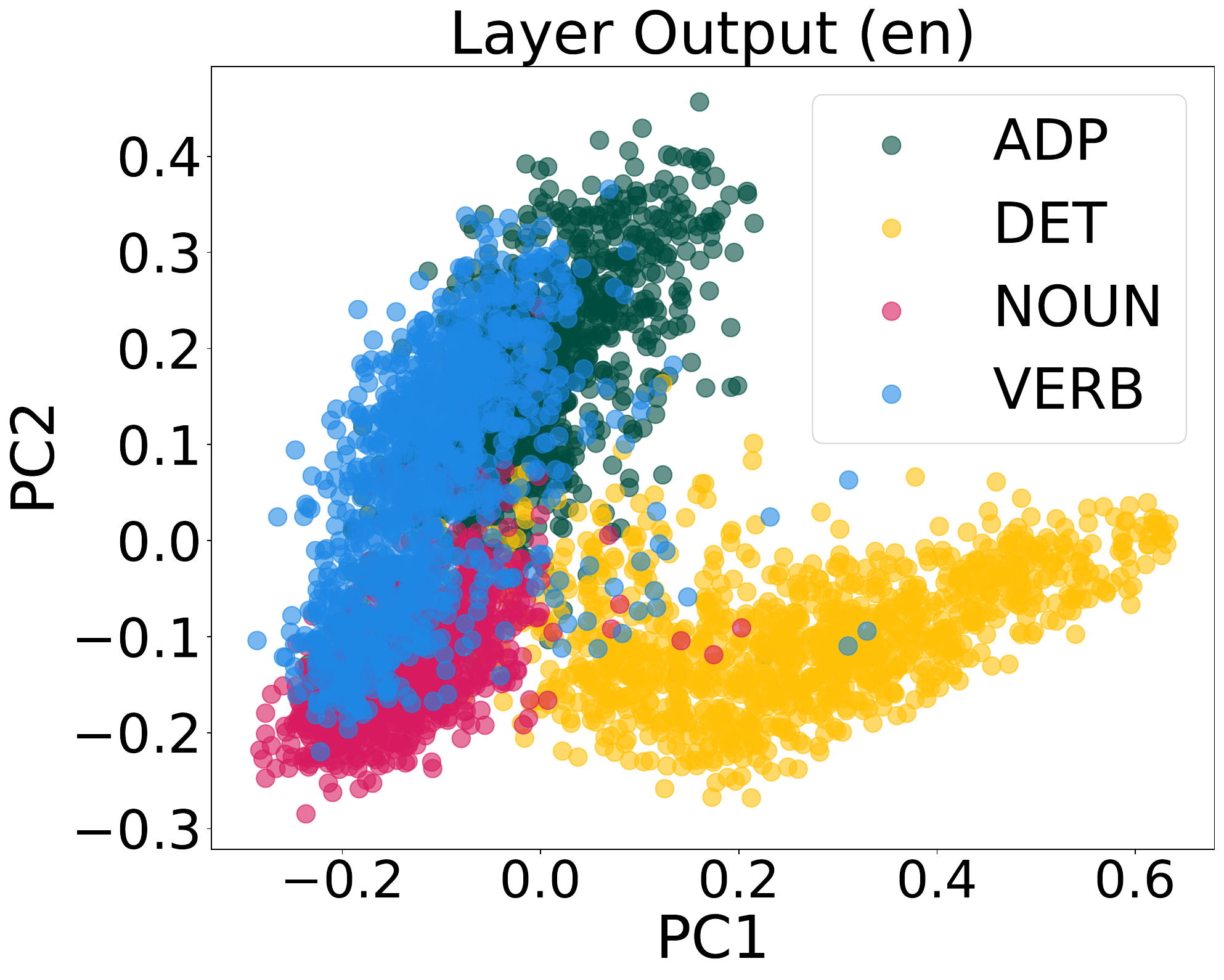}
        \caption{Layer 22}
    \end{subfigure}
    ~
    \begin{subfigure}{0.25\textwidth}
        \includegraphics[width=\textwidth]{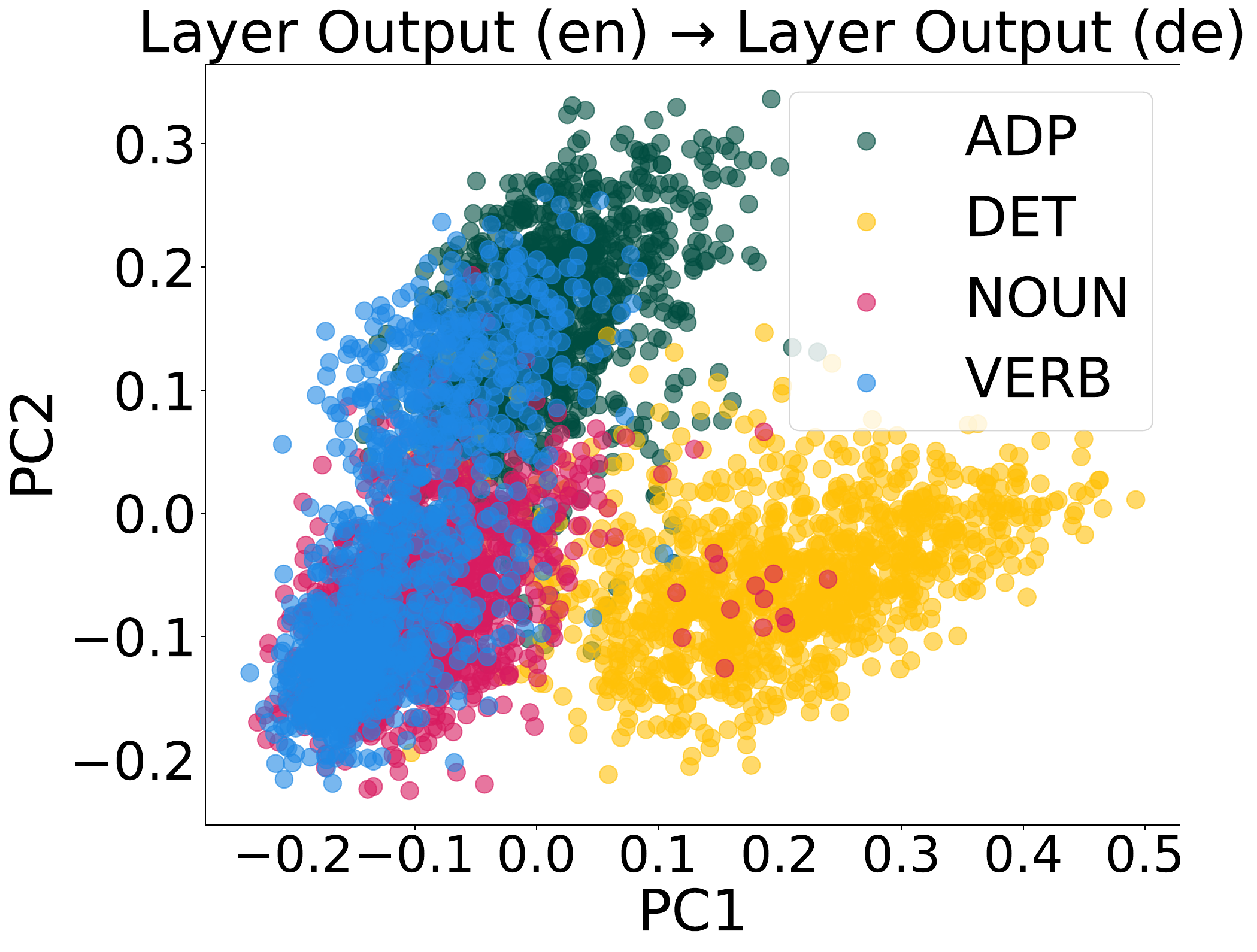}
        \caption{Layer 22}
    \end{subfigure}
    ~
    \begin{subfigure}{0.25\textwidth}
        \includegraphics[width=\textwidth]{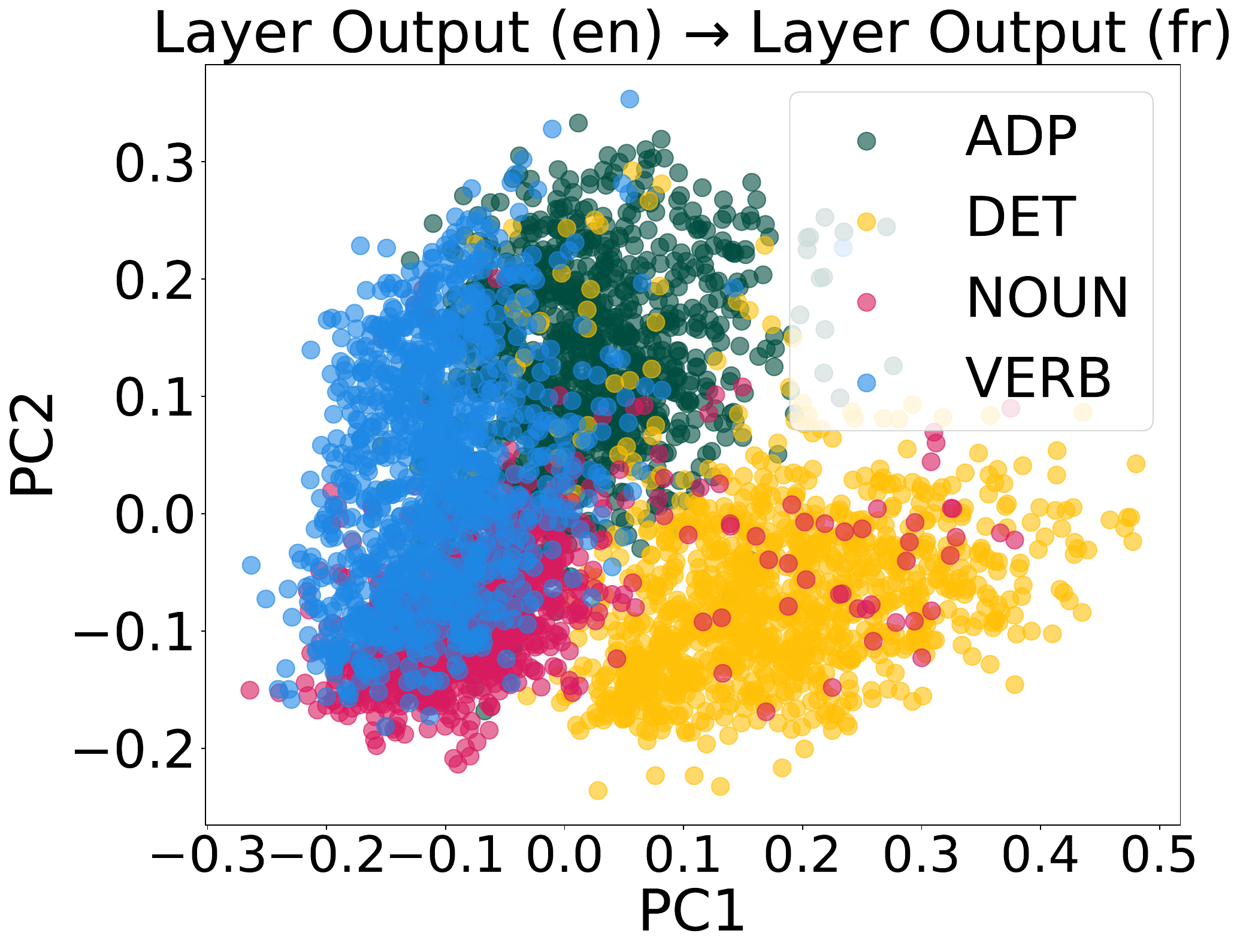}
        \caption{Layer 22}
    \end{subfigure}
    \caption{2D projections for tokens with different POS of the model pre-trained on English (first column) and the adapted models trained on German (second column) and French (third column) at various layers. In all three cases, the projection matrix is computed via PCA on the English representations only.}
    \label{fig:appendix:pos_pca_de}
\end{figure*}

\begin{figure*}[p]
    \centering    
    \begin{subfigure}{0.25\textwidth}
        \includegraphics[width=\textwidth]{emnlp2023-latex/images/pos_alignment/pc_plots/en/pos_pc_b_lay_en_1.pdf}
        \caption{Layer 1}
    \end{subfigure}
    ~
    \begin{subfigure}{0.25\textwidth}
        \includegraphics[width=\textwidth]{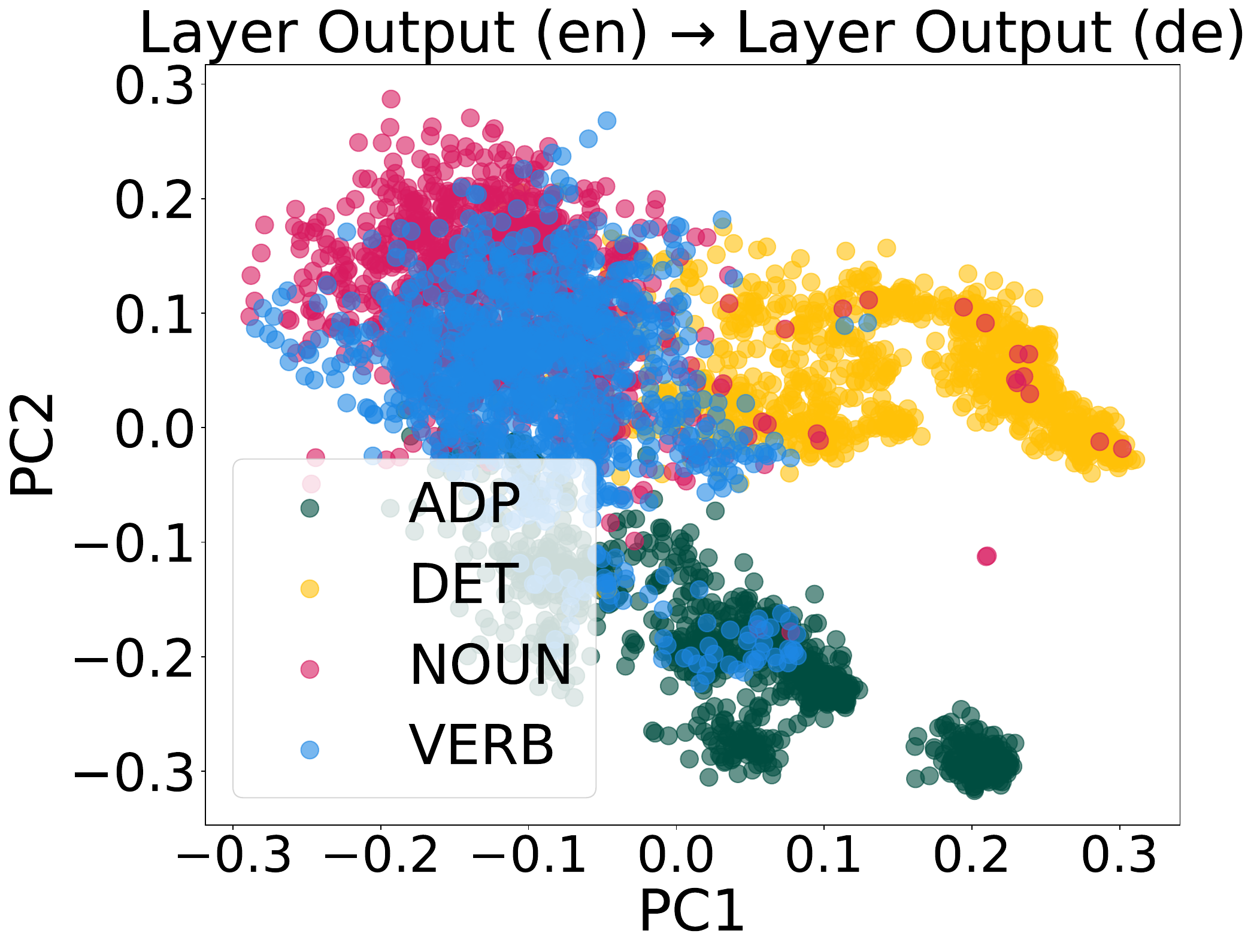}
        \caption{Layer 1}
    \end{subfigure}
    ~
    \begin{subfigure}{0.25\textwidth}
        \includegraphics[width=\textwidth]{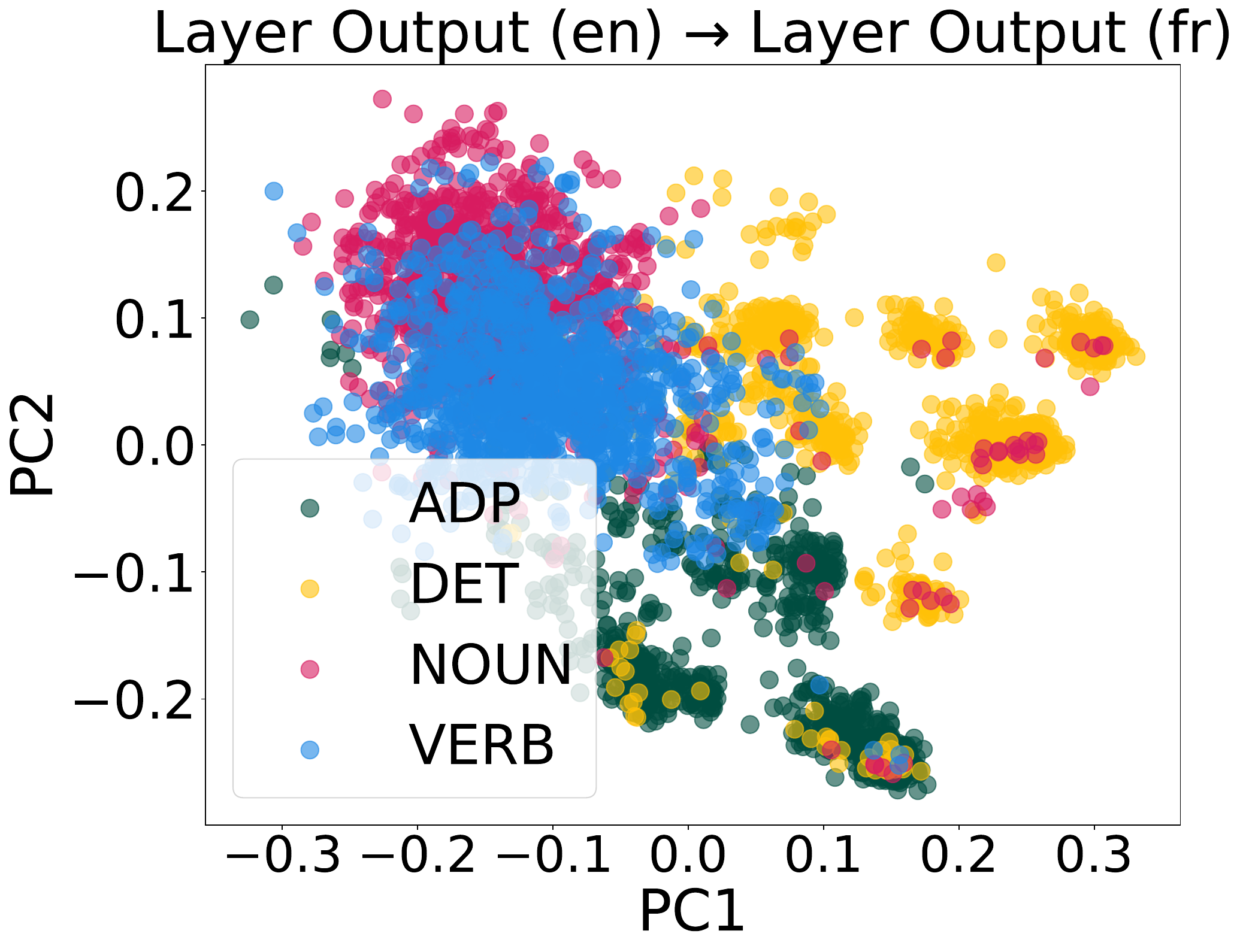}
        \caption{Layer 1}
    \end{subfigure}
    \\
    \begin{subfigure}{0.25\textwidth}
        \includegraphics[width=\textwidth]{emnlp2023-latex/images/pos_alignment/pc_plots/en/pos_pc_b_lay_en_2.pdf}
        \caption{Layer 2}
    \end{subfigure}
    ~
    \begin{subfigure}{0.25\textwidth}
        \includegraphics[width=\textwidth]{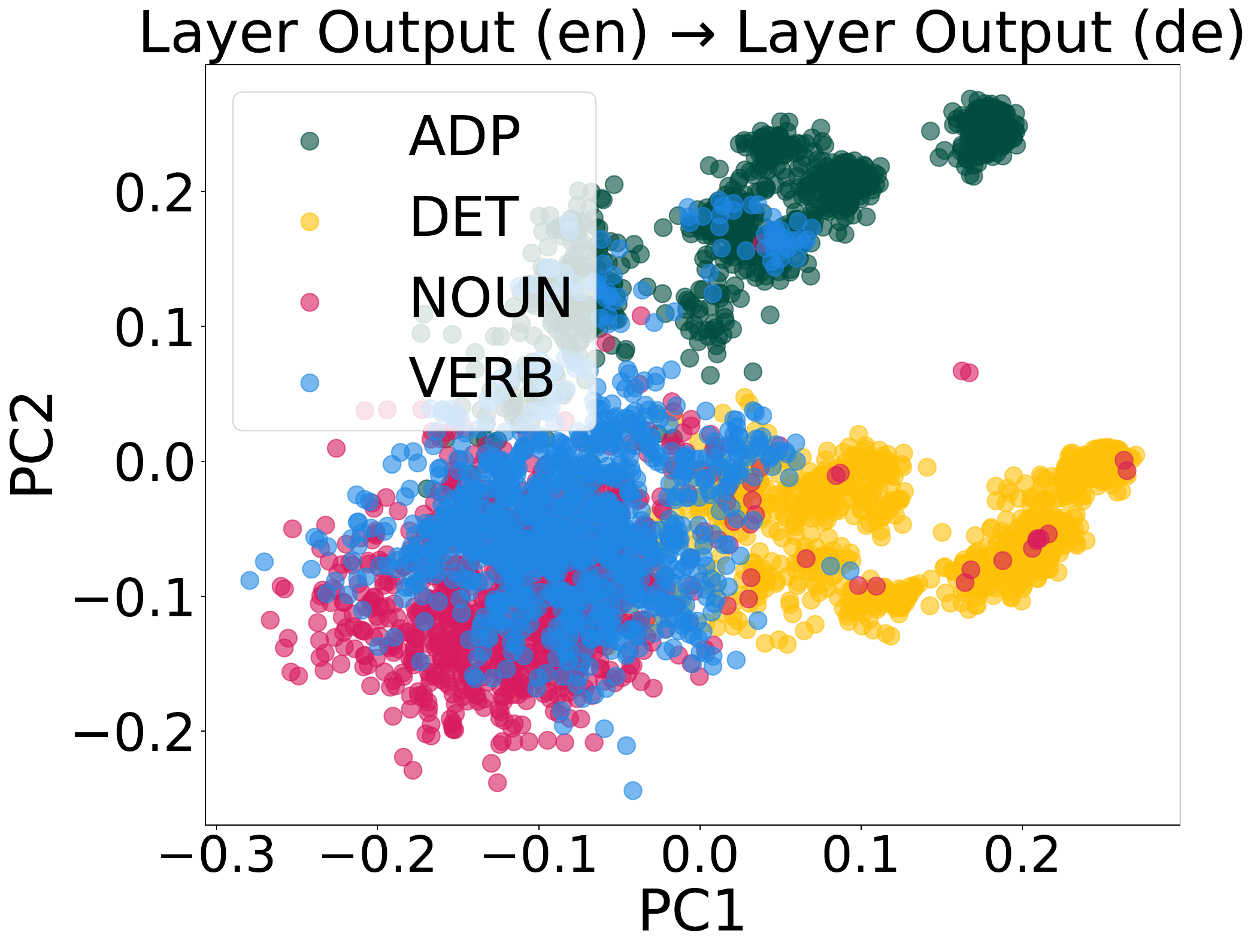}
        \caption{Layer 2}
    \end{subfigure}
    ~
    \begin{subfigure}{0.25\textwidth}
        \includegraphics[width=\textwidth]{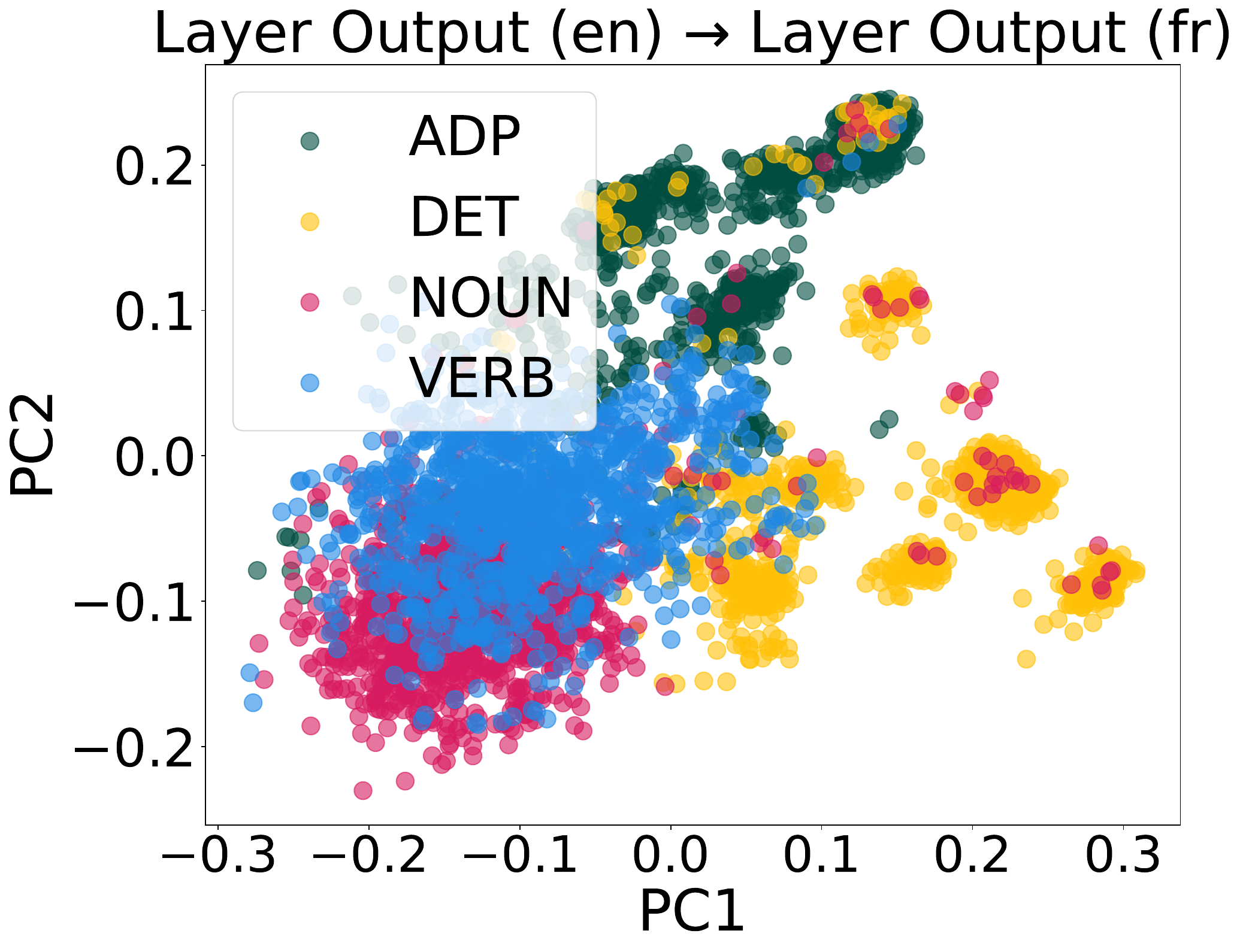}
        \caption{Layer 2}
    \end{subfigure}
    \\
    \begin{subfigure}{0.25\textwidth}
        \includegraphics[width=\textwidth]{emnlp2023-latex/images/pos_alignment/pc_plots/en/pos_pc_b_lay_en_8.pdf}
        \caption{Layer 8}
    \end{subfigure}
    ~
    \begin{subfigure}{0.25\textwidth}
        \includegraphics[width=\textwidth]{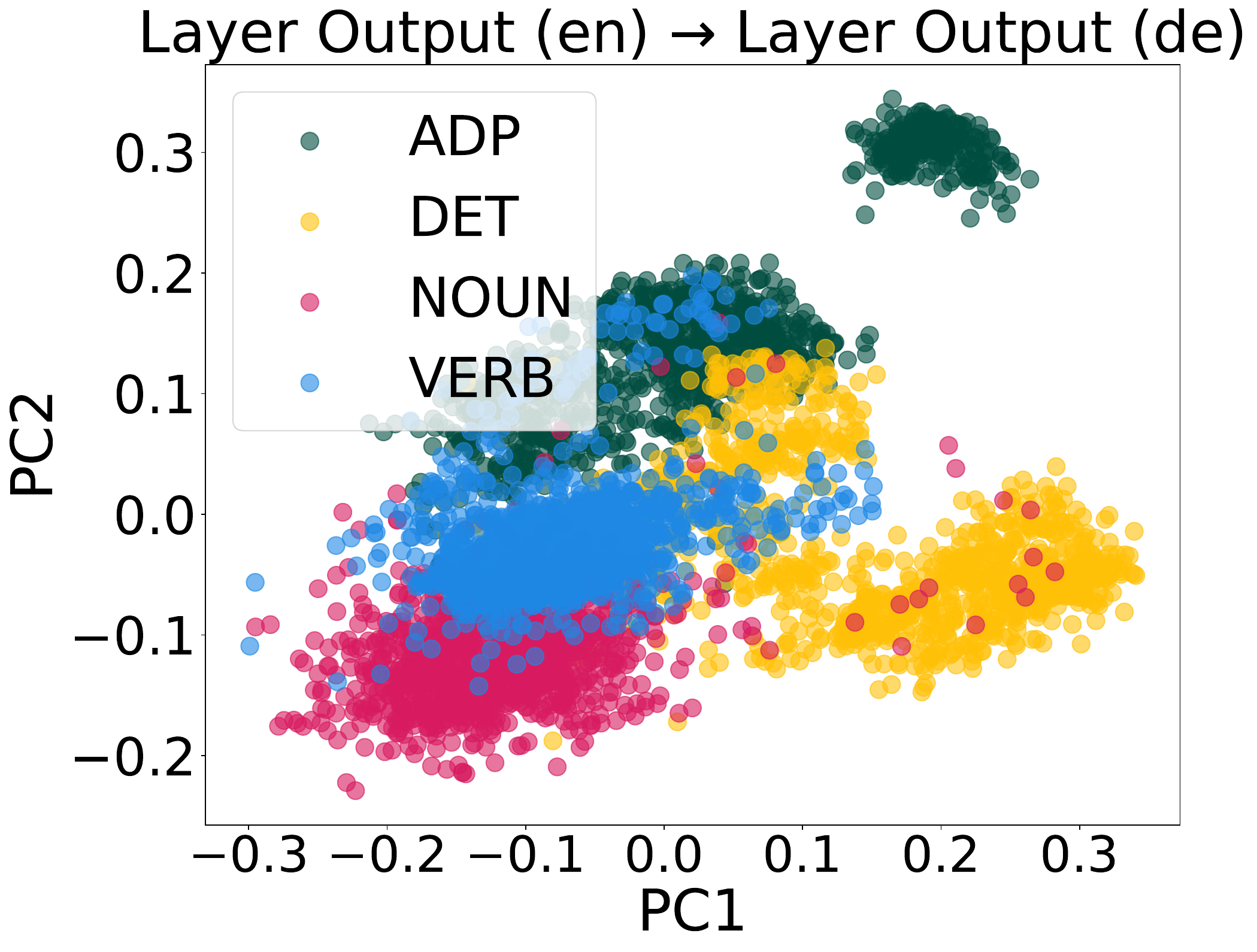}
        \caption{Layer 8}
    \end{subfigure}
    ~
    \begin{subfigure}{0.25\textwidth}
        \includegraphics[width=\textwidth]{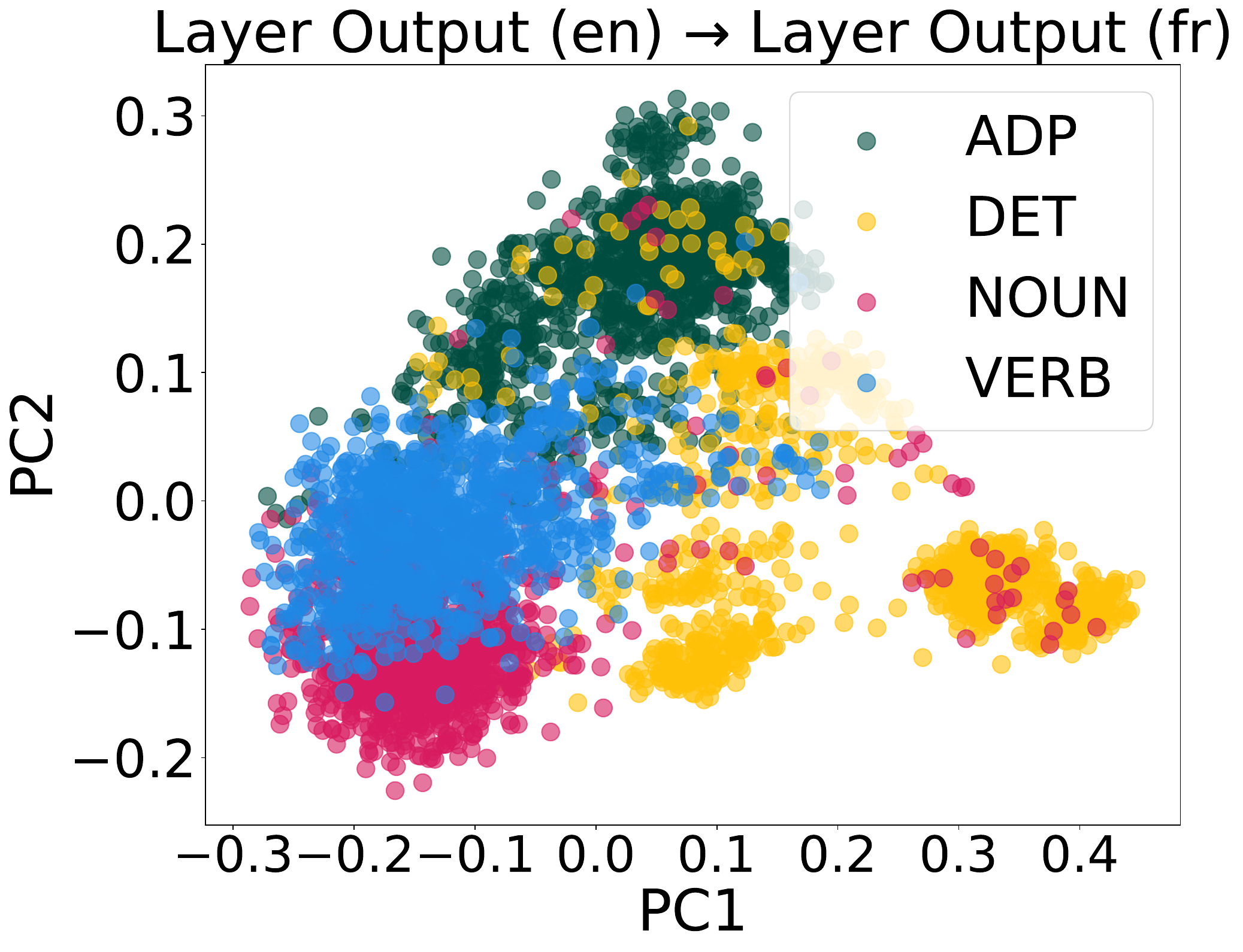}
        \caption{Layer 8}
    \end{subfigure}
    \\
    \begin{subfigure}{0.25\textwidth}
        \includegraphics[width=\textwidth]{emnlp2023-latex/images/pos_alignment/pc_plots/en/pos_pc_b_lay_en_16.pdf}
        \caption{Layer 16}
    \end{subfigure}
    ~
    \begin{subfigure}{0.25\textwidth}
        \includegraphics[width=\textwidth]{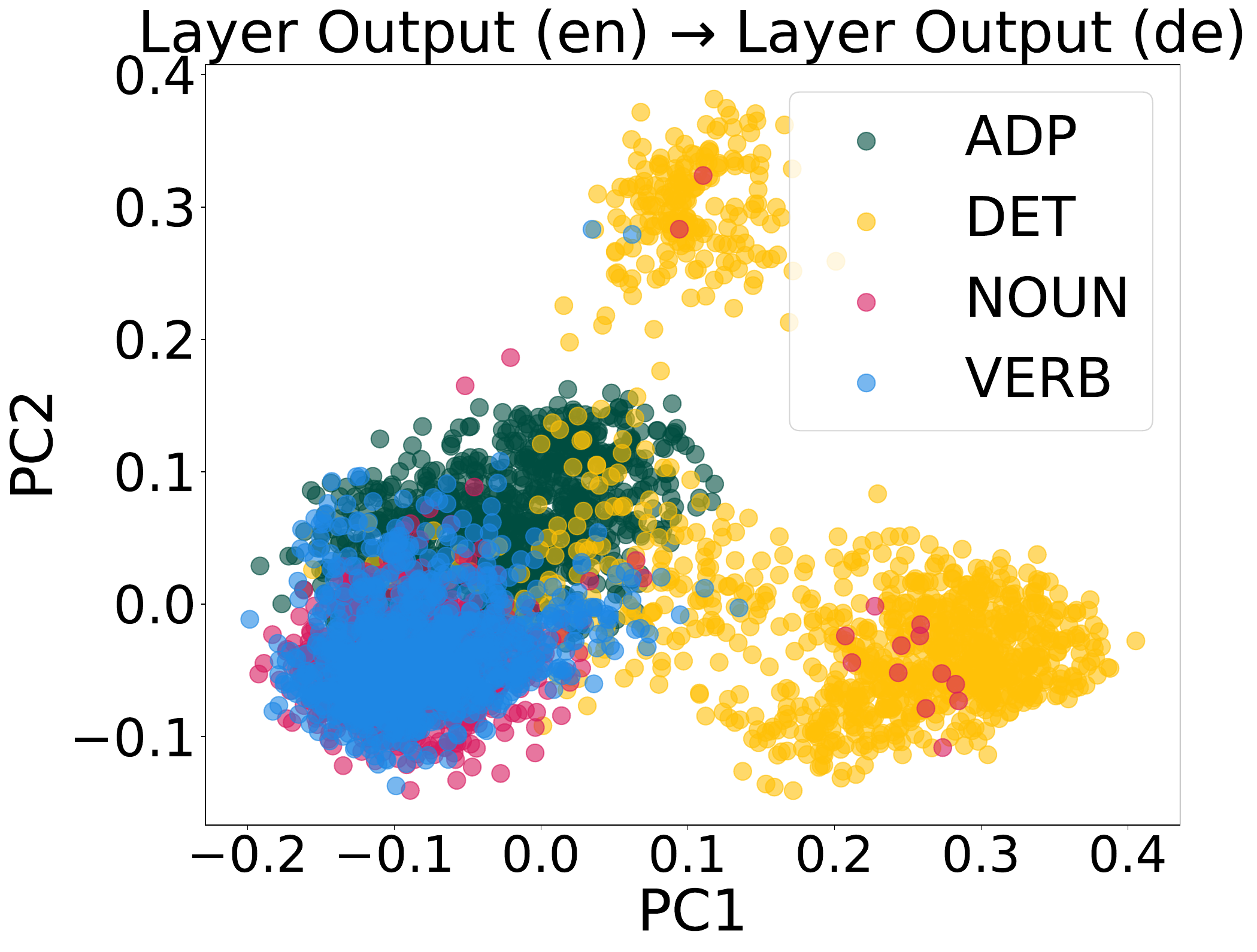}
        \caption{Layer 16}
    \end{subfigure}
    ~
    \begin{subfigure}{0.25\textwidth}
        \includegraphics[width=\textwidth]{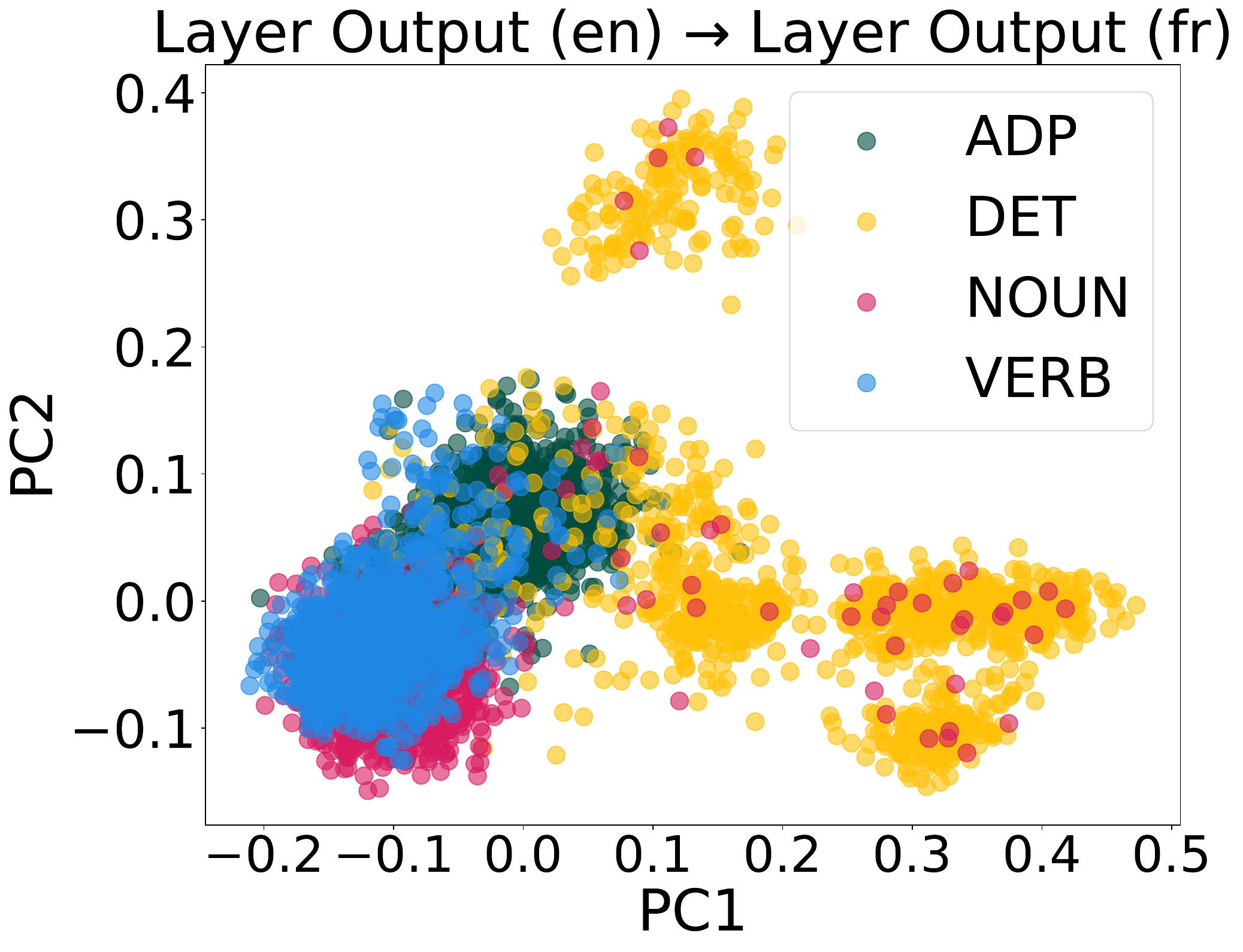}
        \caption{Layer 16}
    \end{subfigure}
    \\
    \begin{subfigure}{0.25\textwidth}
        \includegraphics[width=\textwidth]{emnlp2023-latex/images/pos_alignment/pc_plots/en/pos_pc_b_lay_en_22.pdf}
        \caption{Layer 22}
    \end{subfigure}
    ~
    \begin{subfigure}{0.25\textwidth}
        \includegraphics[width=\textwidth]{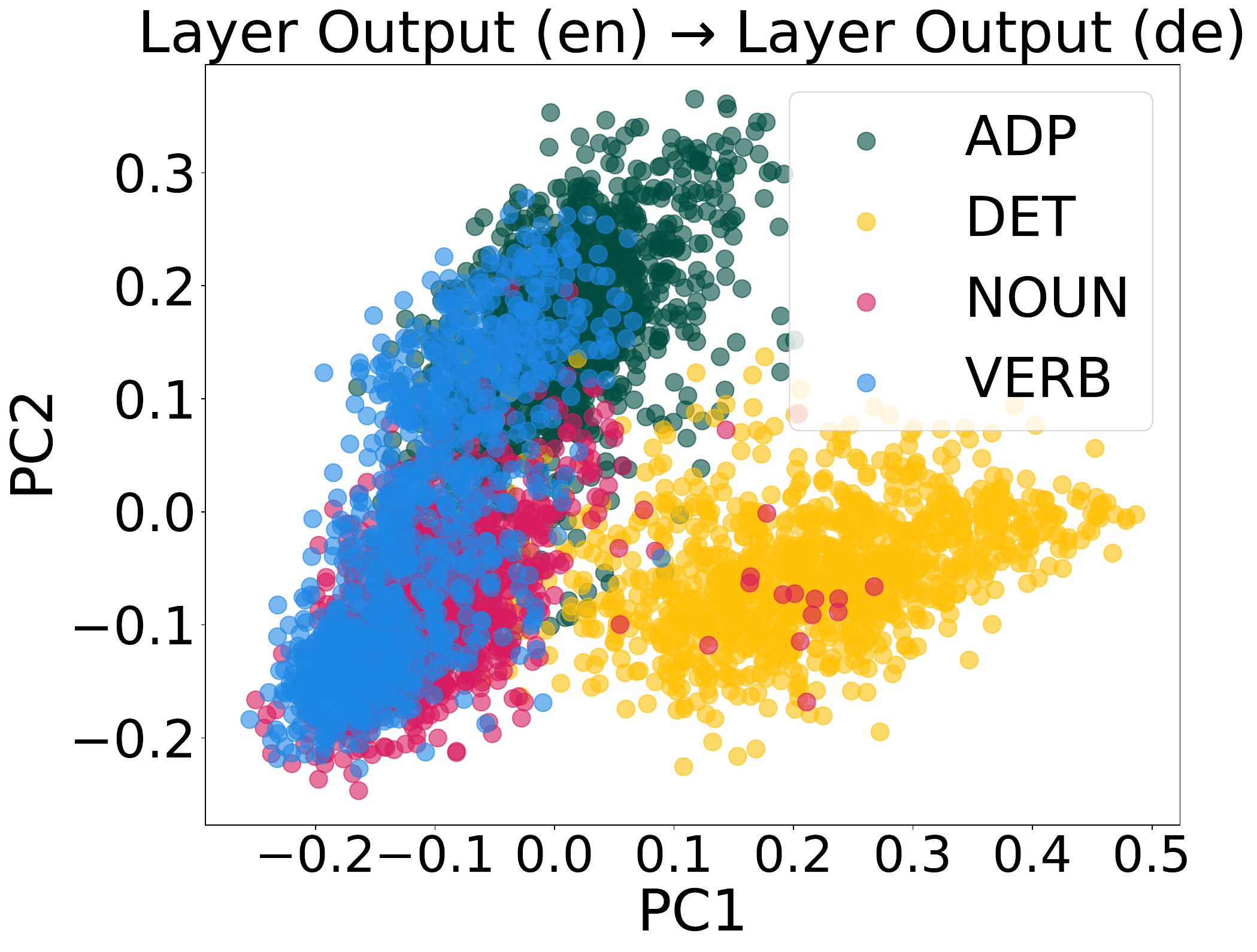}
        \caption{Layer 22}
    \end{subfigure}
    ~
    \begin{subfigure}{0.25\textwidth}
        \includegraphics[width=\textwidth]{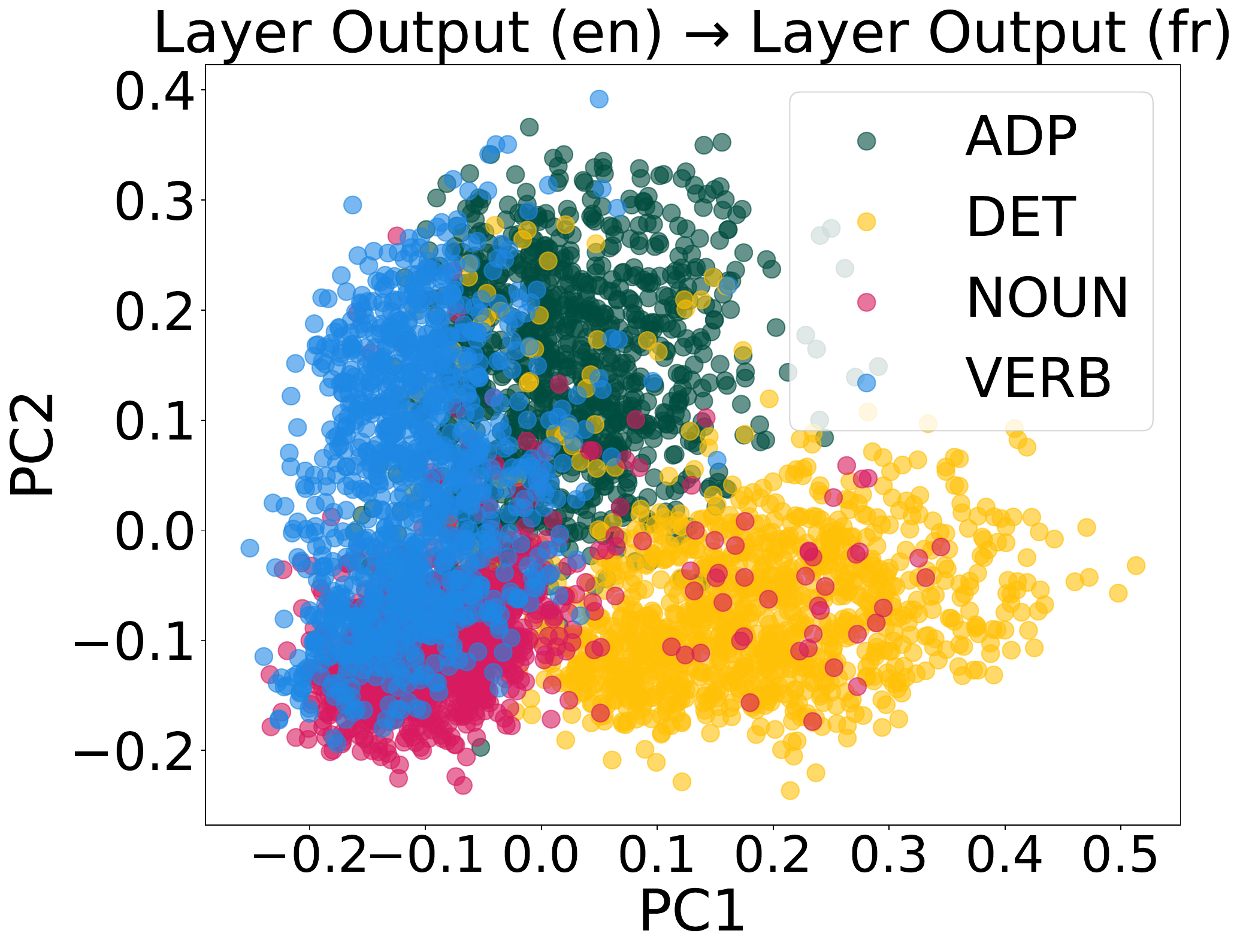}
        \caption{Layer 22}
    \end{subfigure}
    \caption{2D projections for tokens with different POS of the model pre-trained on English (first column) and the adapted models with LoRA trained on German (second column) and French (third column) at various layers. In all three cases, the projection matrix is computed via PCA on the English representations only.}
    \label{fig:appendix:pos_pca_de_lora}
\end{figure*}

To evaluate whether adapters operate on top of the already existing structure in the representation space, we analyze the structure corresponding to POS (ADP/DET/NOUN/VERB) for the adapted and non-adapted models. For every layer, we run PCA on the hidden representations for English tokens and create a projection matrix consisting of the first two principal components. We apply the projection matrix obtained from the English representations to layer output representations of German and French inputs with different POS tags. \Cref{fig:appendix:pos_pca_de} shows the result of this projection for several layers. \Cref{fig:appendix:pos_pca_de_lora} shows highly similar results when using LoRA. Furthermore, \Cref{fig:cosine_lr} shows the cosine similarity between the first two principal components computed on the English and target language representations from the LM adapted with LoRA. Similar to the result from the classical adapters presented in \Cref{fig:cosine}, we observe a very high alignment (absolute cosine similarity $\approx 0.6$) between the principal components.

\paragraph{Tense and number}

\begin{figure*}[p]
    \centering    
    \begin{subfigure}{0.25\textwidth}
        \includegraphics[width=\textwidth]{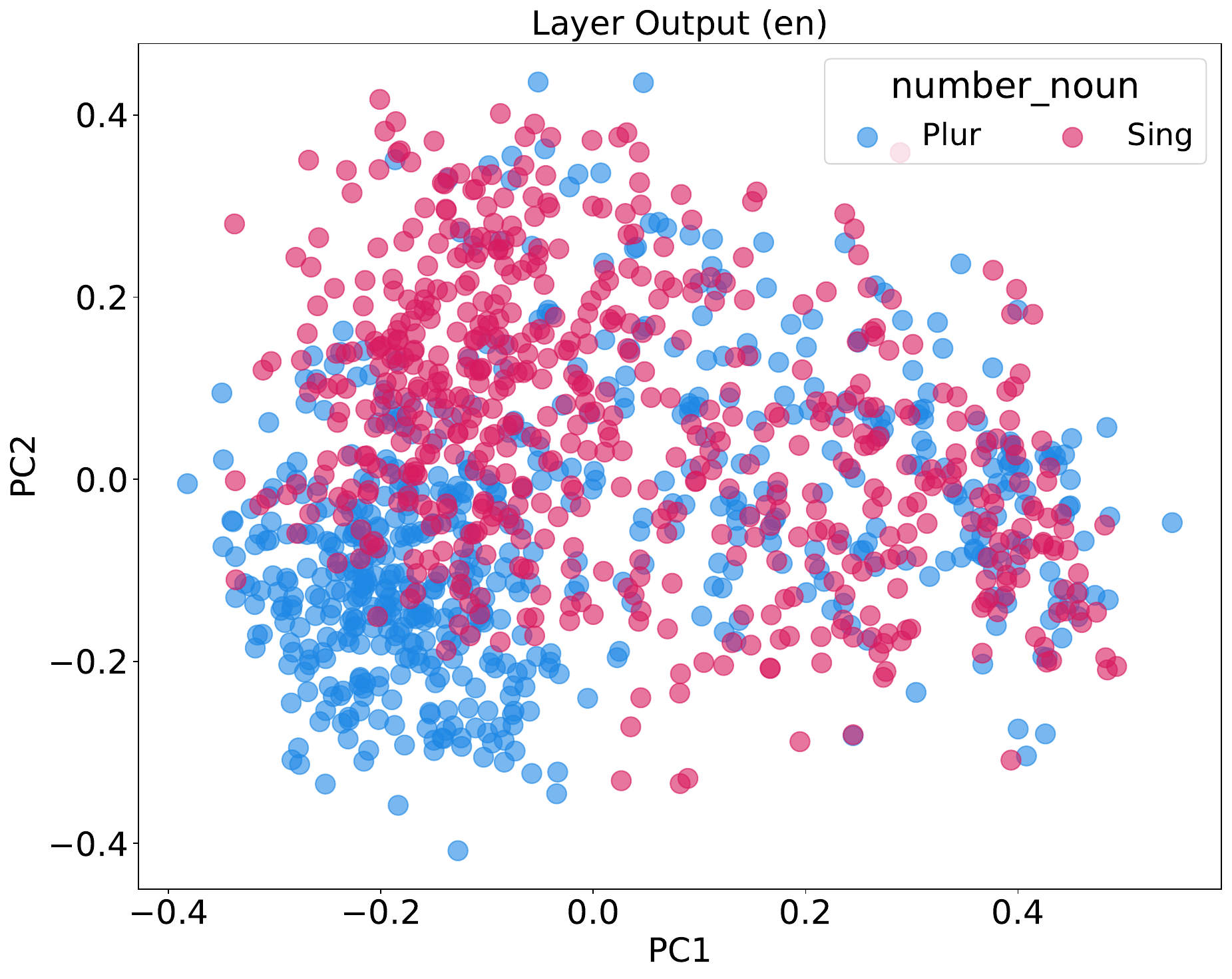}
        \caption{Layer 1}
    \end{subfigure}
    ~
    \begin{subfigure}{0.25\textwidth}
        \includegraphics[width=\textwidth]{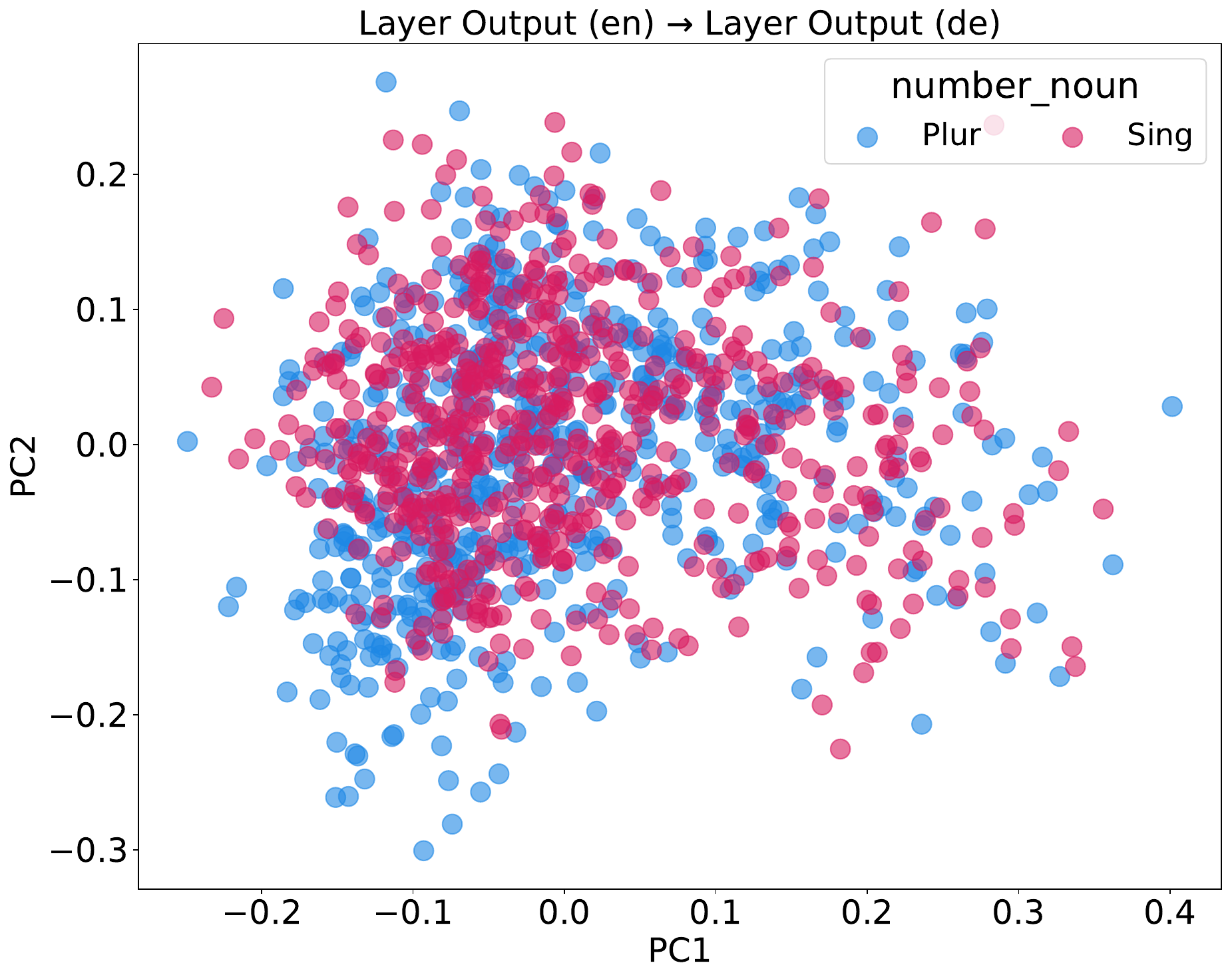}
        \caption{Layer 1}
    \end{subfigure}
    ~
    \begin{subfigure}{0.25\textwidth}
        \includegraphics[width=\textwidth]{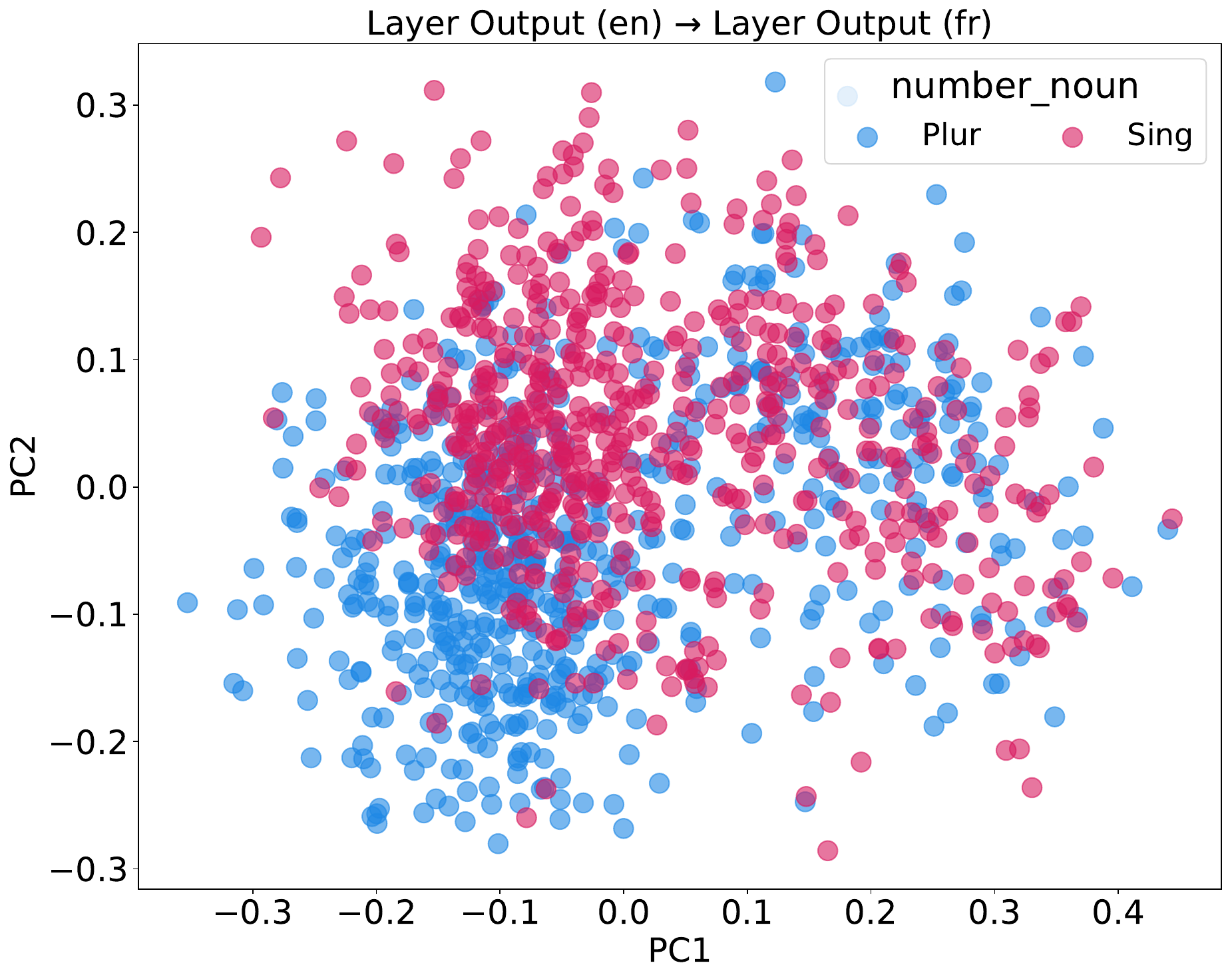}
        \caption{Layer 1}
    \end{subfigure}
    \\
    \begin{subfigure}{0.25\textwidth}
        \includegraphics[width=\textwidth]{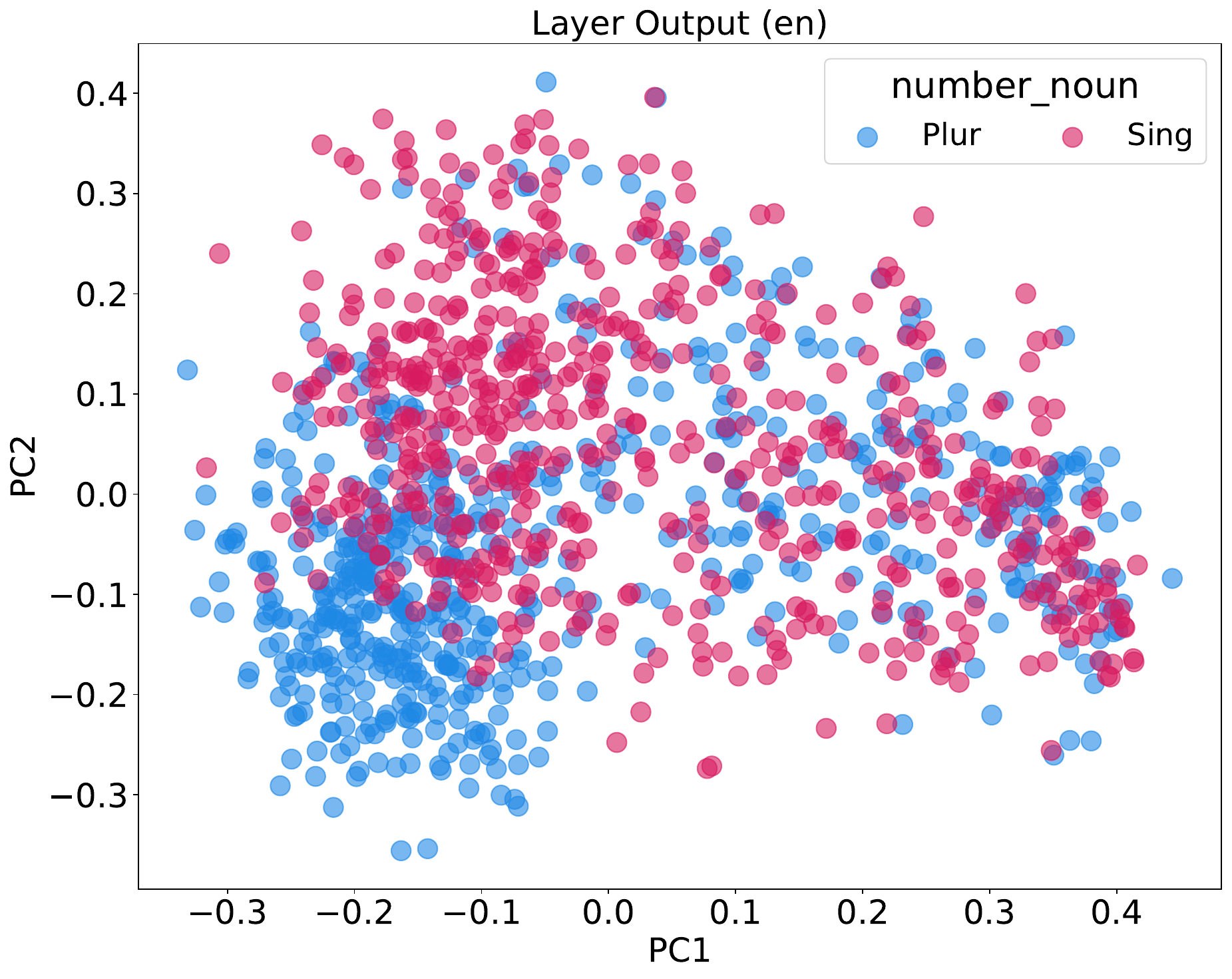}
        \caption{Layer 2}
    \end{subfigure}
    ~
    \begin{subfigure}{0.25\textwidth}
        \includegraphics[width=\textwidth]{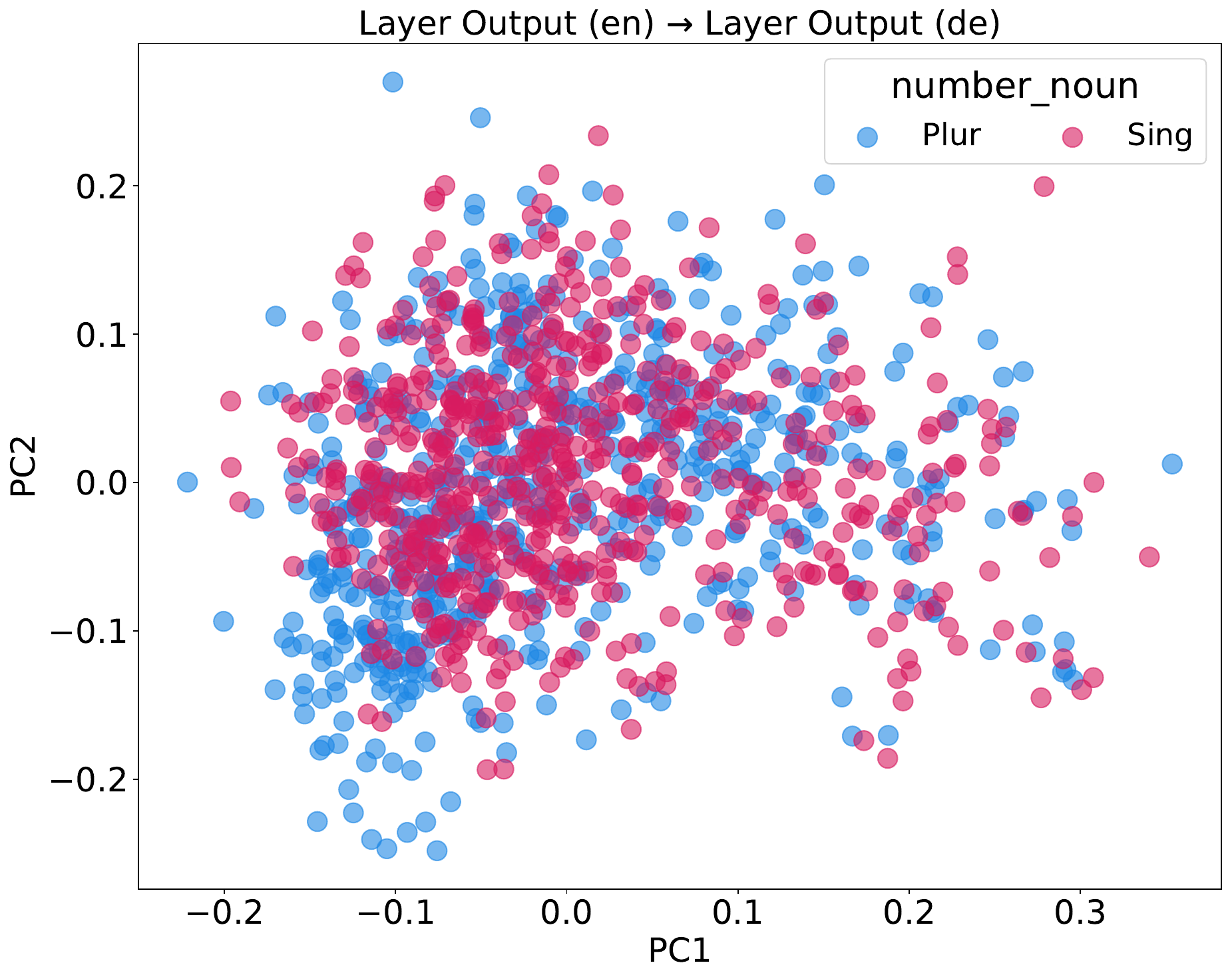}
        \caption{Layer 2}
    \end{subfigure}
    ~
    \begin{subfigure}{0.25\textwidth}
        \includegraphics[width=\textwidth]{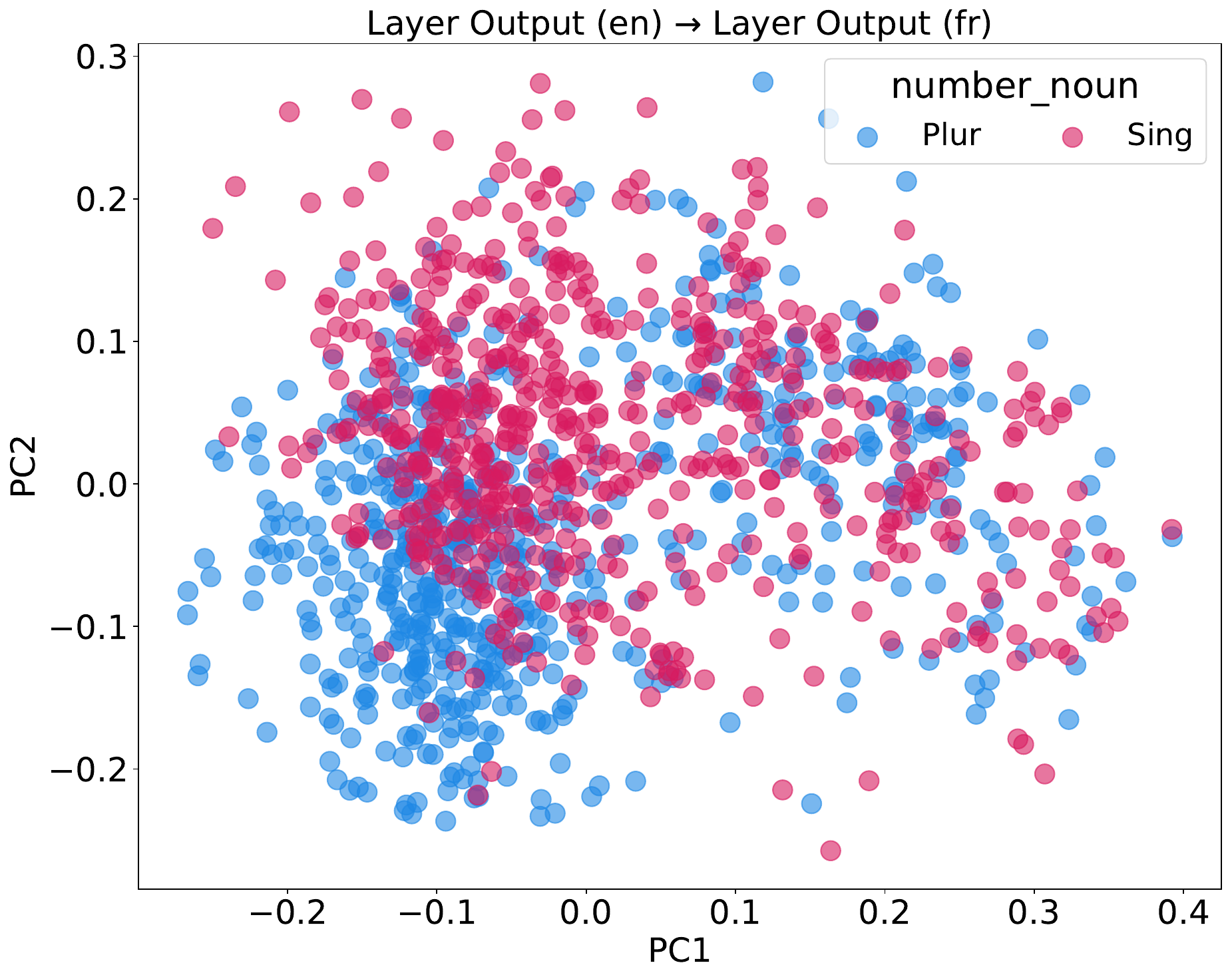}
        \caption{Layer 2}
    \end{subfigure}
    \\
    \begin{subfigure}{0.25\textwidth}
        \includegraphics[width=\textwidth]{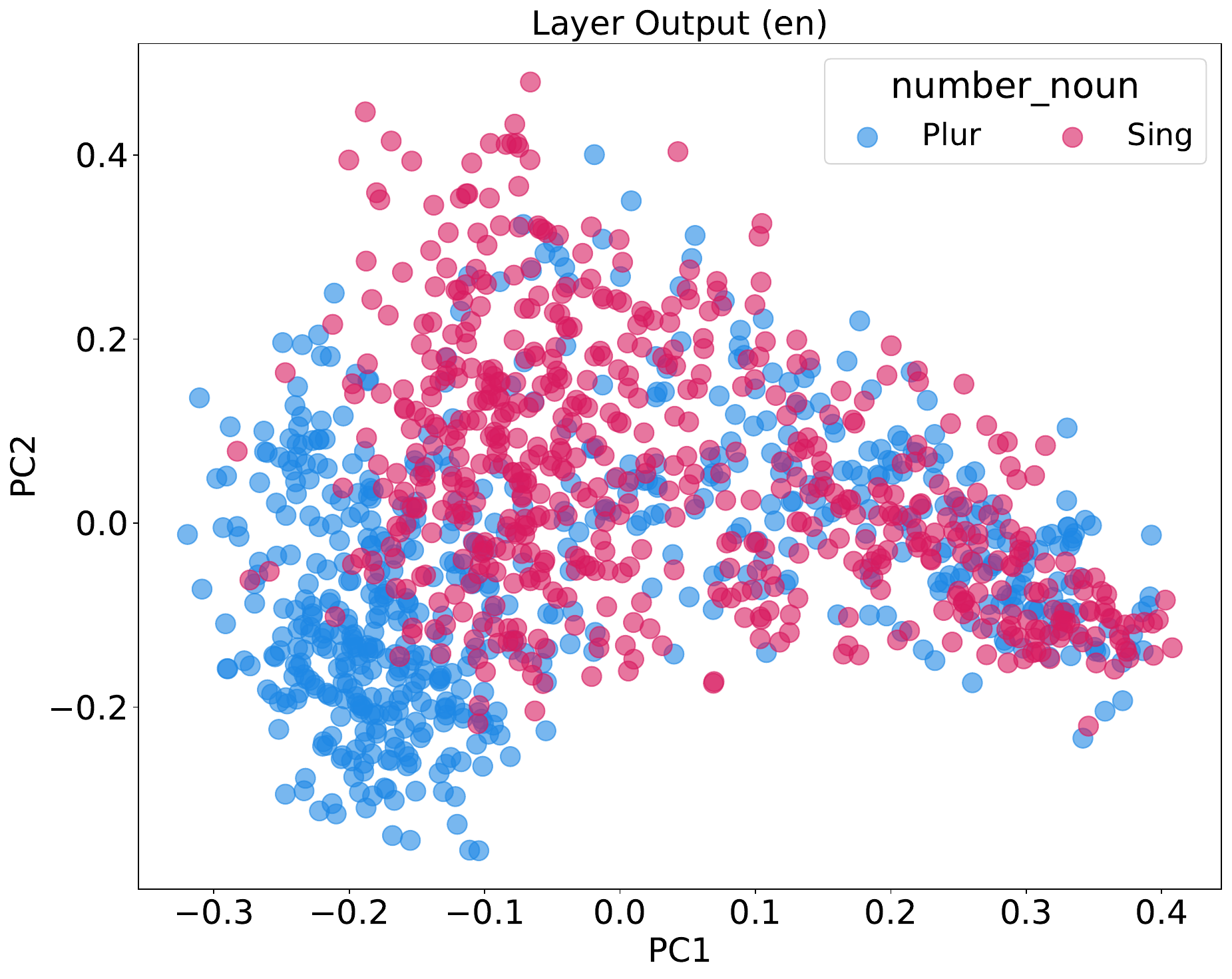}
        \caption{Layer 6}
    \end{subfigure}
    ~
    \begin{subfigure}{0.25\textwidth}
        \includegraphics[width=\textwidth]{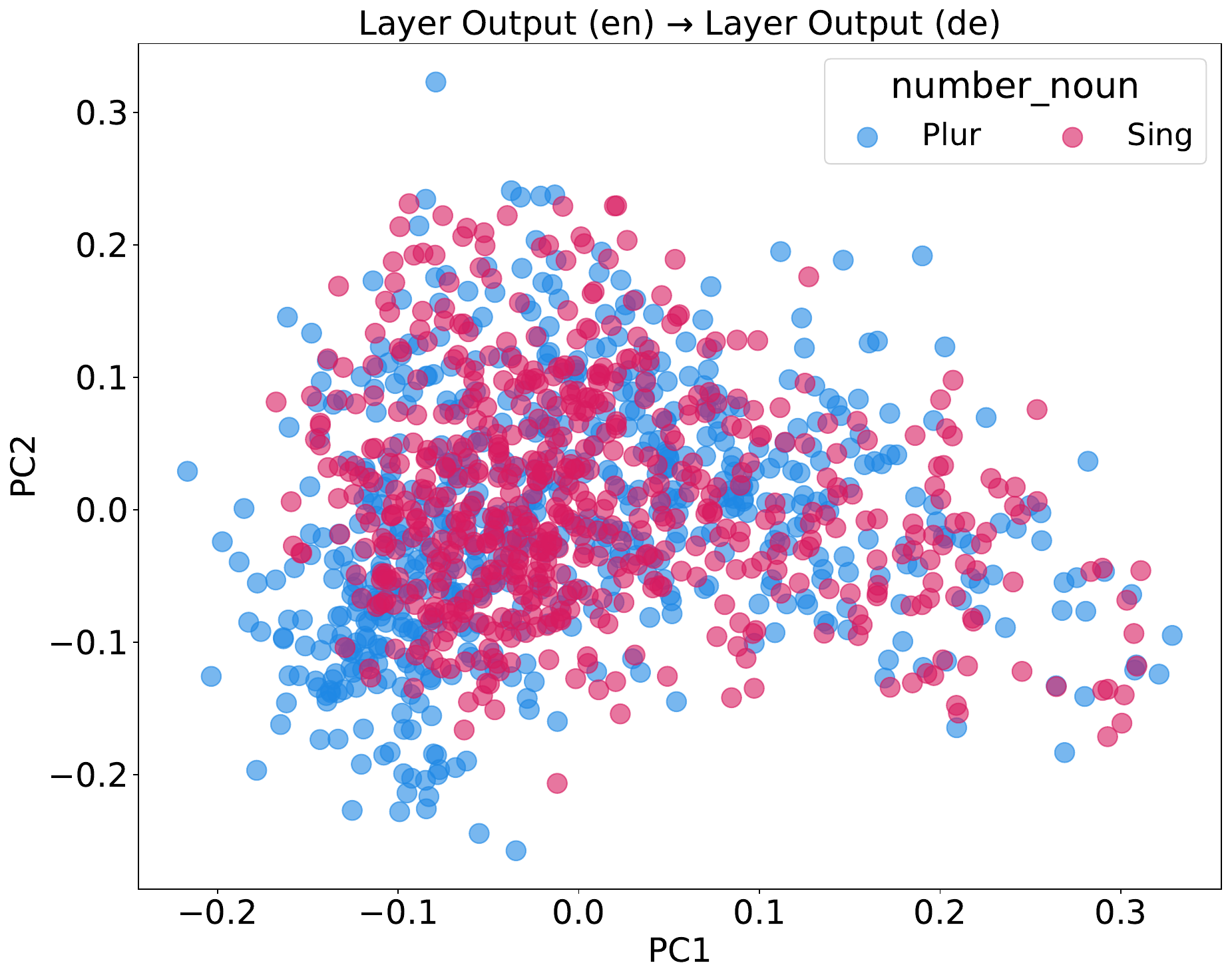}
        \caption{Layer 6}
    \end{subfigure}
    ~
    \begin{subfigure}{0.25\textwidth}
         \includegraphics[width=\textwidth]{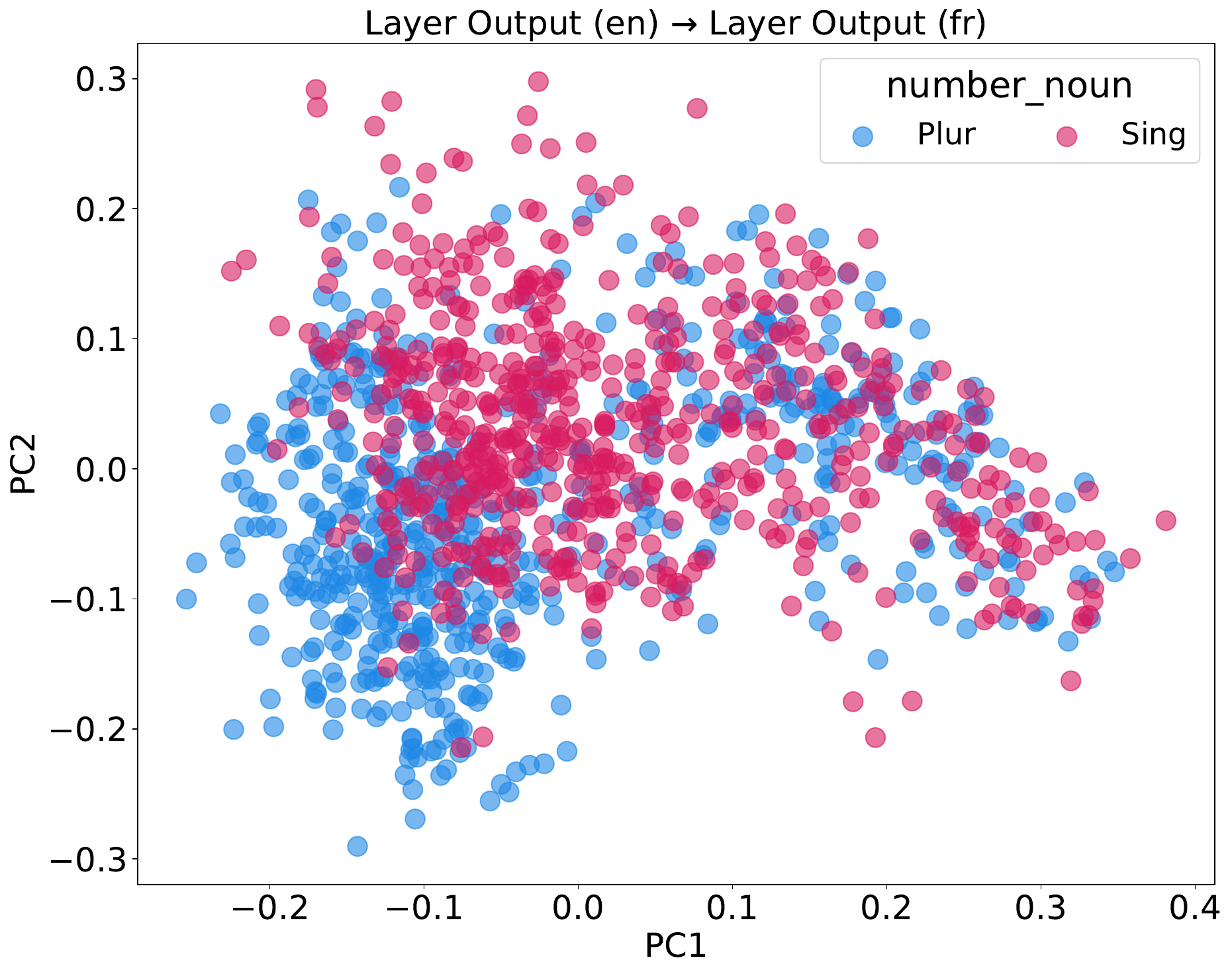}
        \caption{Layer 6}
    \end{subfigure}
    \\
    \begin{subfigure}{0.25\textwidth}
        \includegraphics[width=\textwidth]{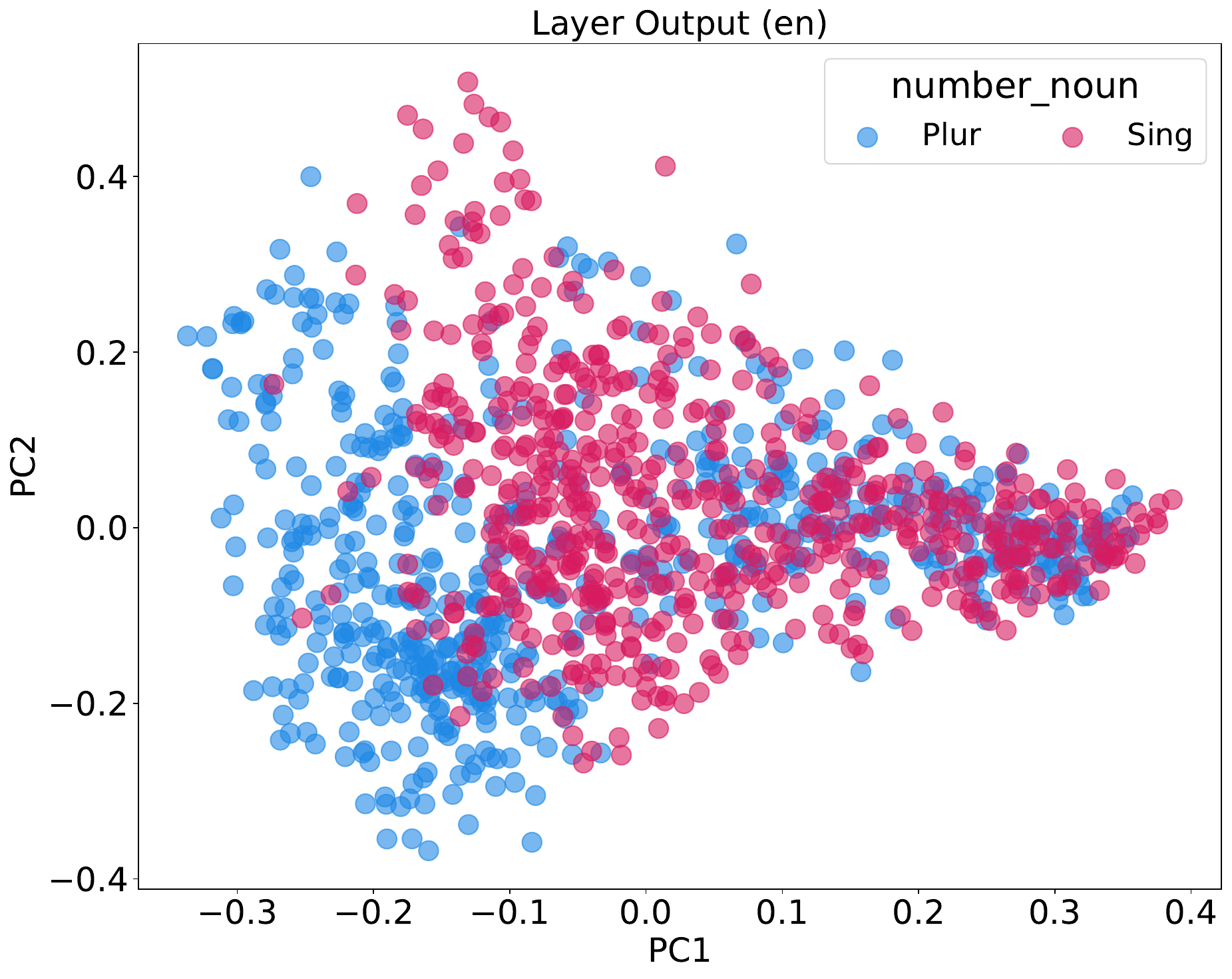}
        \caption{Layer 12}
    \end{subfigure}
    ~
    \begin{subfigure}{0.25\textwidth}
        \includegraphics[width=\textwidth]{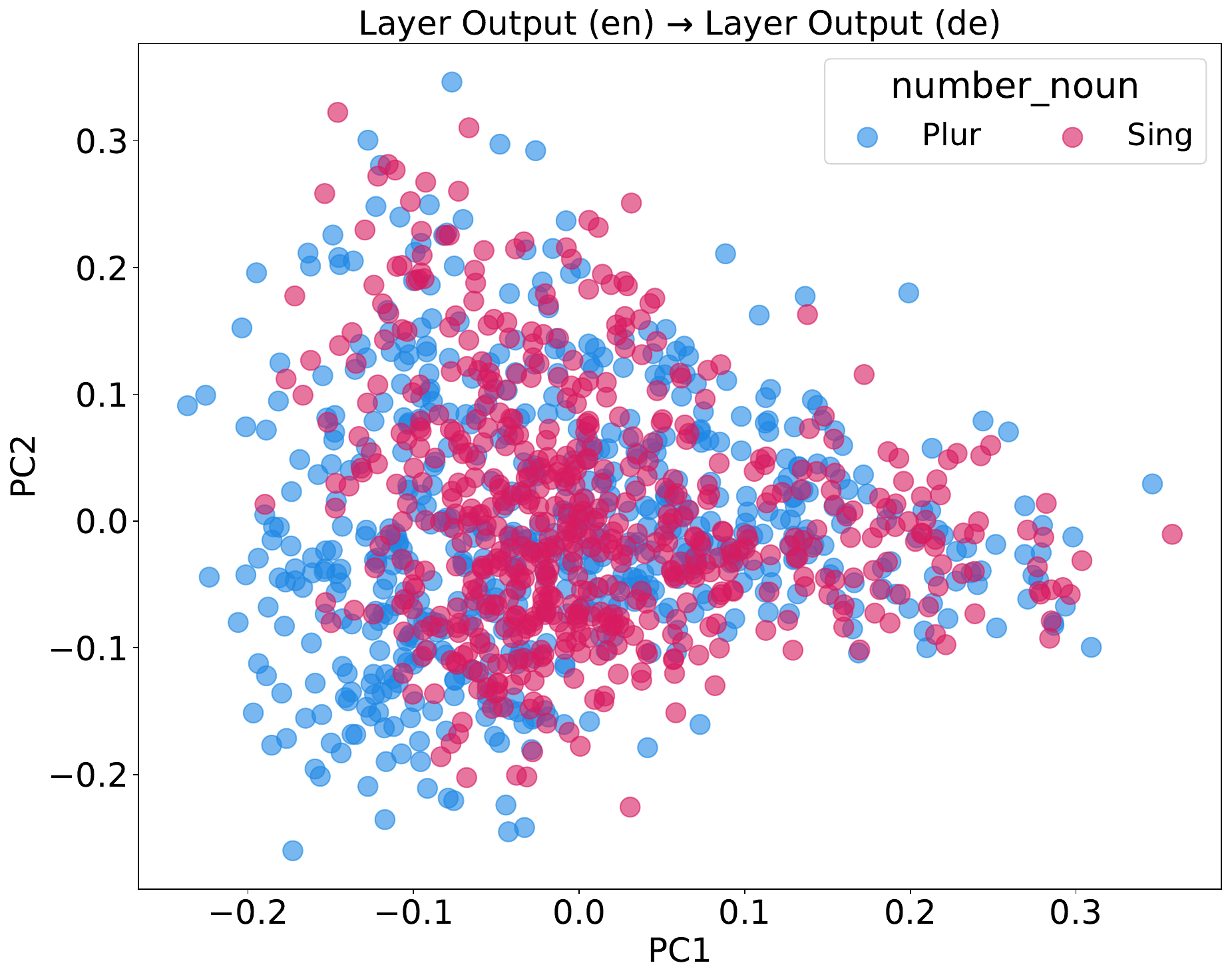}
        \caption{Layer 12}
    \end{subfigure}
    ~
    \begin{subfigure}{0.25\textwidth}
         \includegraphics[width=\textwidth]{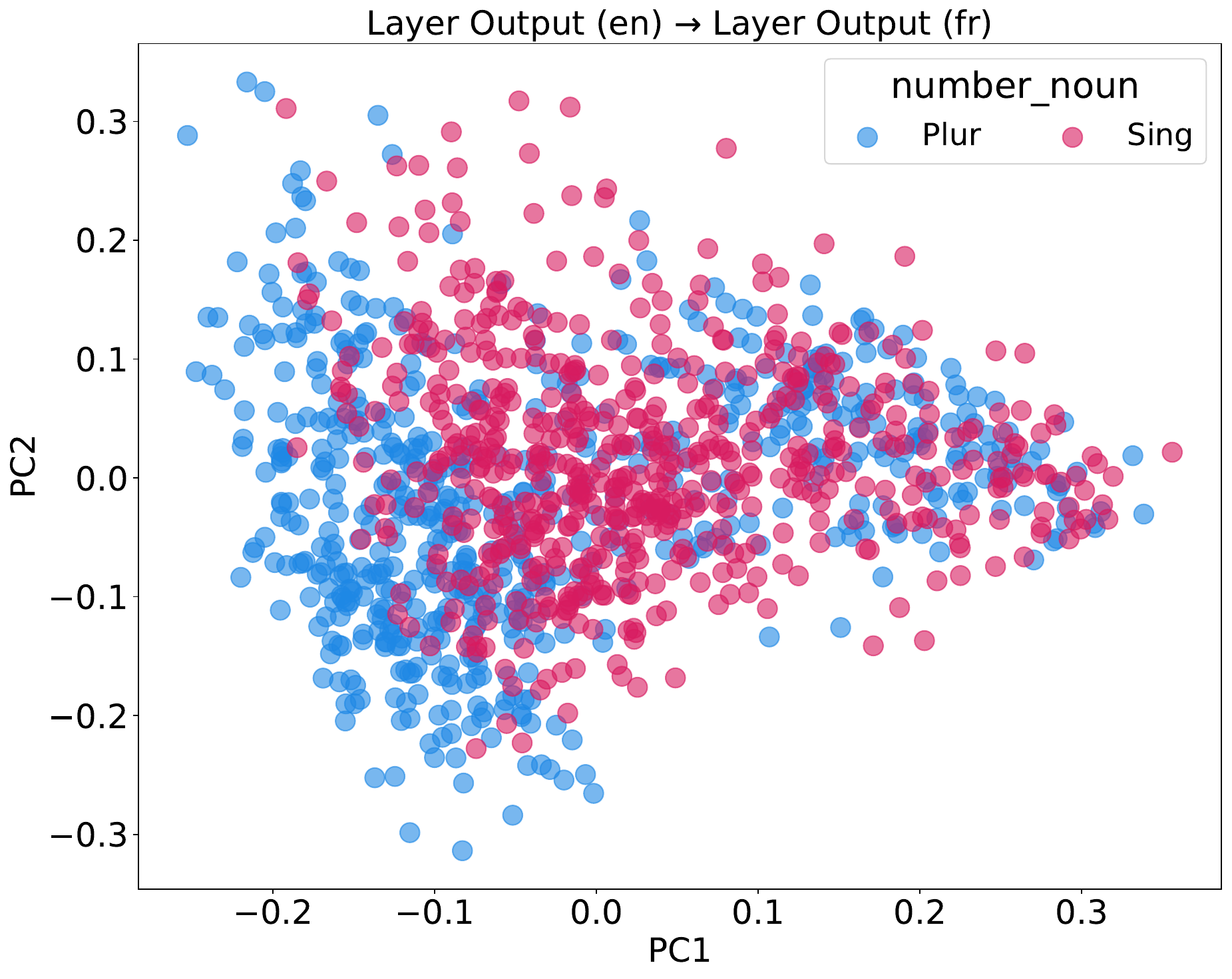}
        \caption{Layer 12}
    \end{subfigure}
    \\
    \begin{subfigure}{0.25\textwidth}
        \includegraphics[width=\textwidth]{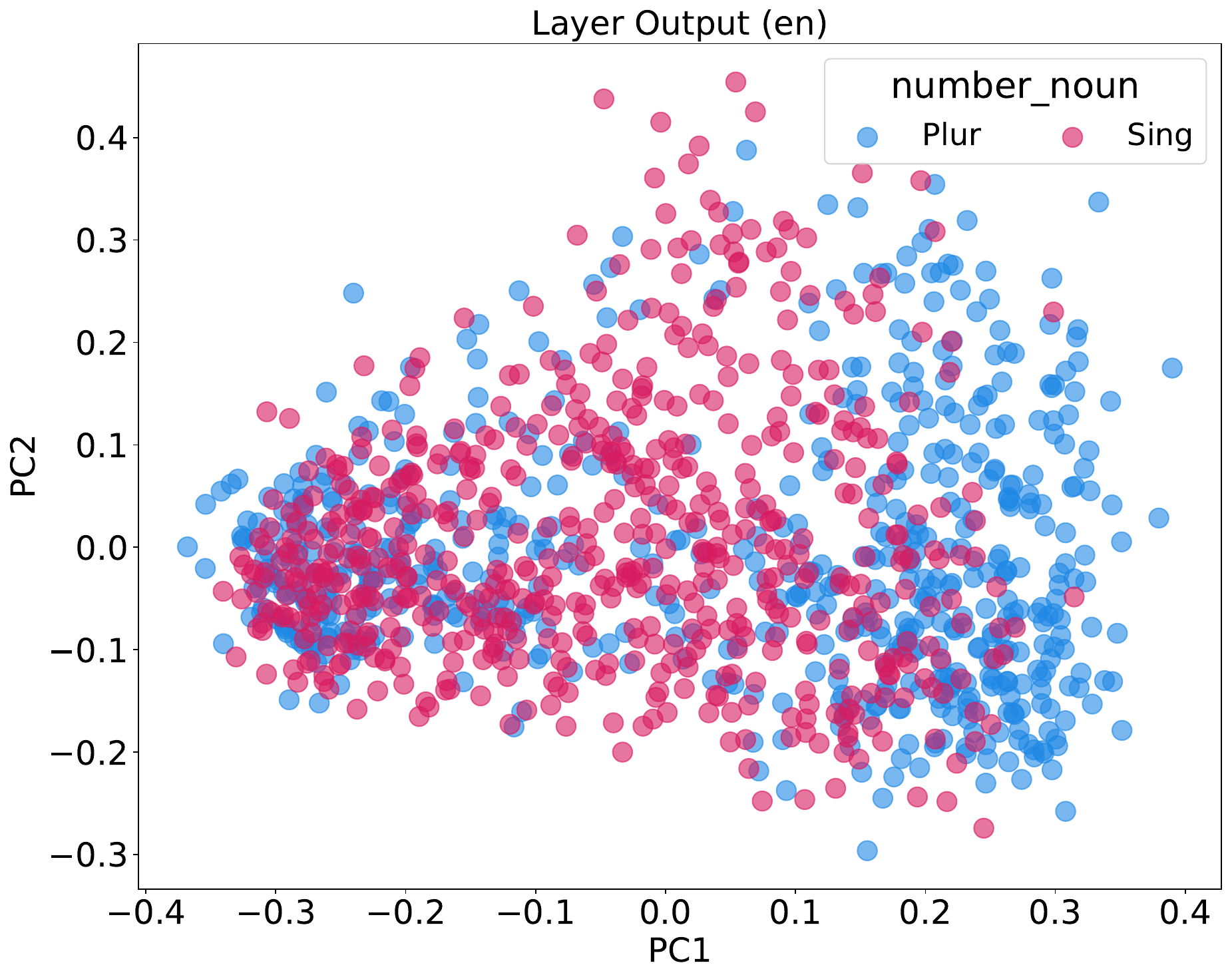}
        \caption{Layer 22}
    \end{subfigure}
    ~
    \begin{subfigure}{0.25\textwidth}
        \includegraphics[width=\textwidth]{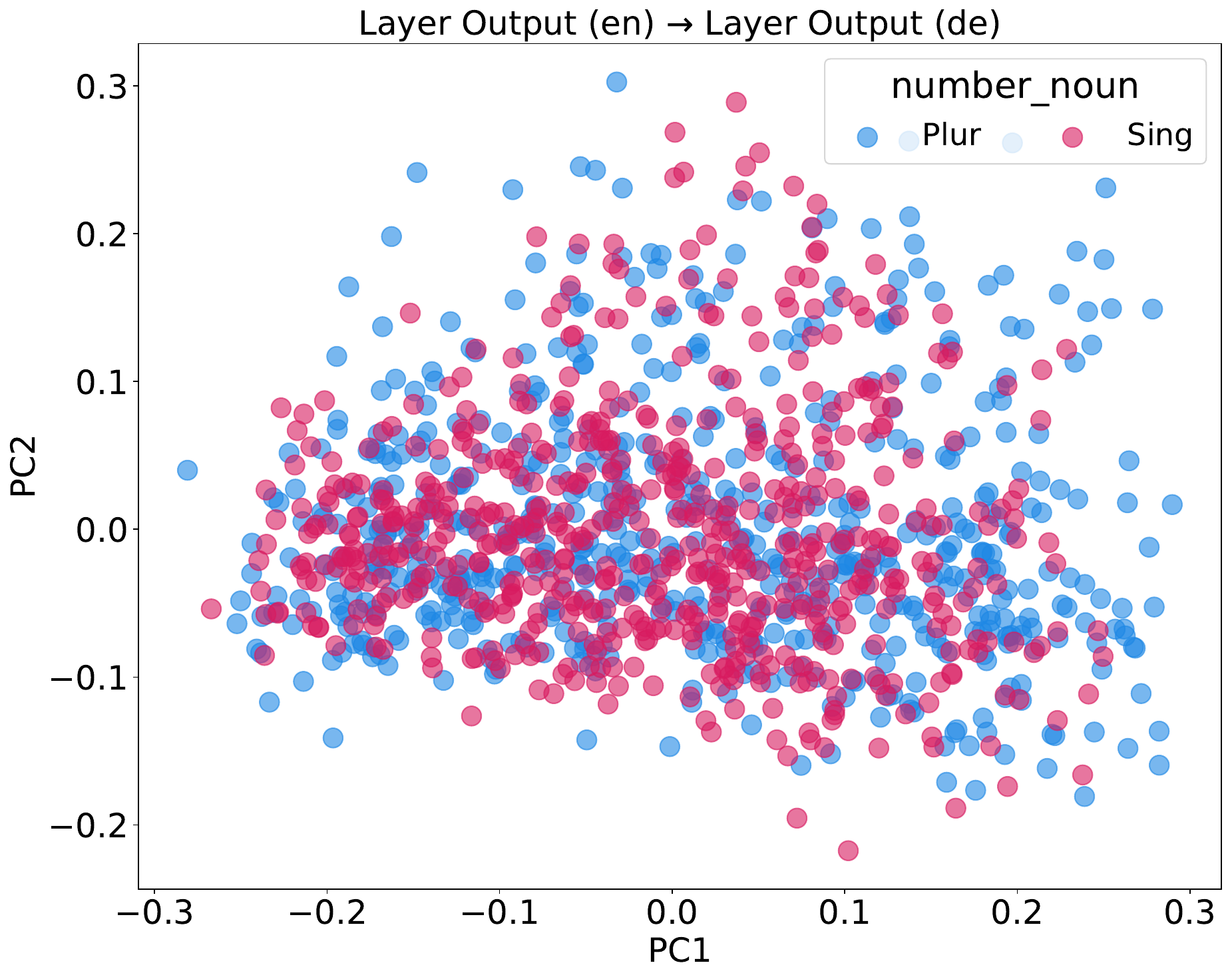}
        \caption{Layer 22}
    \end{subfigure}
    ~
    \begin{subfigure}{0.25\textwidth}
         \includegraphics[width=\textwidth]{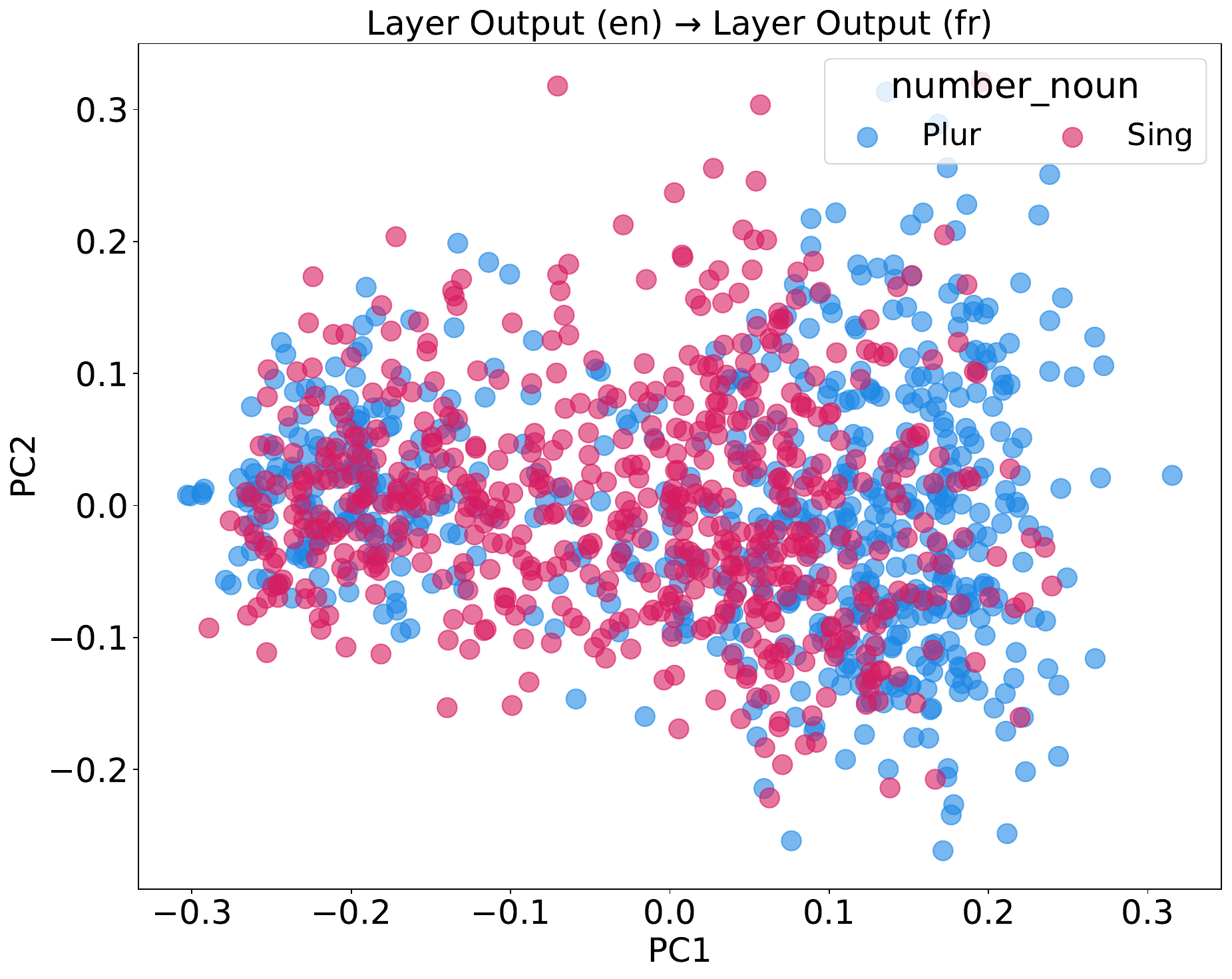}
        \caption{Layer 22}
    \end{subfigure}
    \\
    \begin{subfigure}{0.25\textwidth}
        \includegraphics[width=\textwidth]{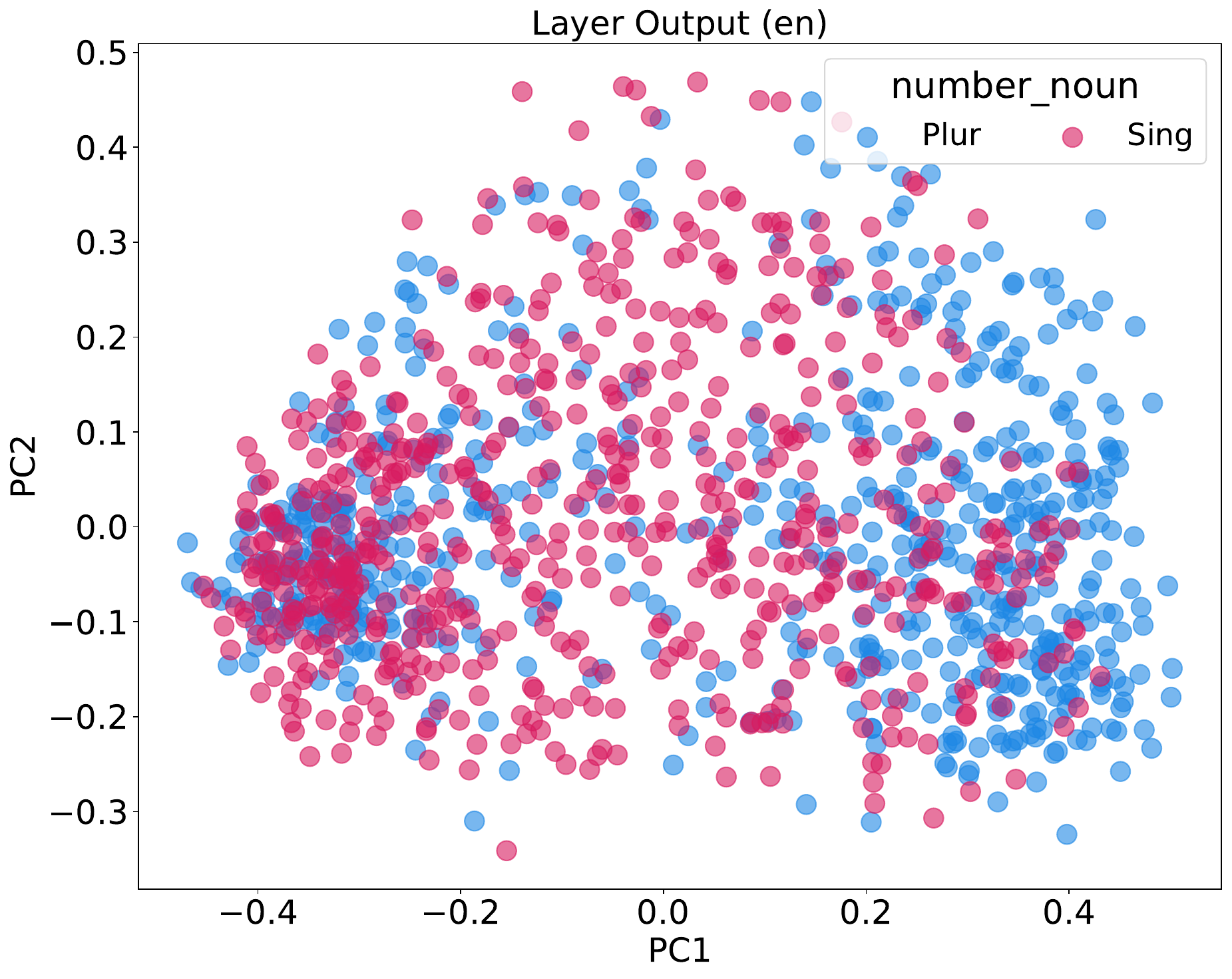}
        \caption{Layer 24}
    \end{subfigure}
    ~
    \begin{subfigure}{0.25\textwidth}
        \includegraphics[width=\textwidth]{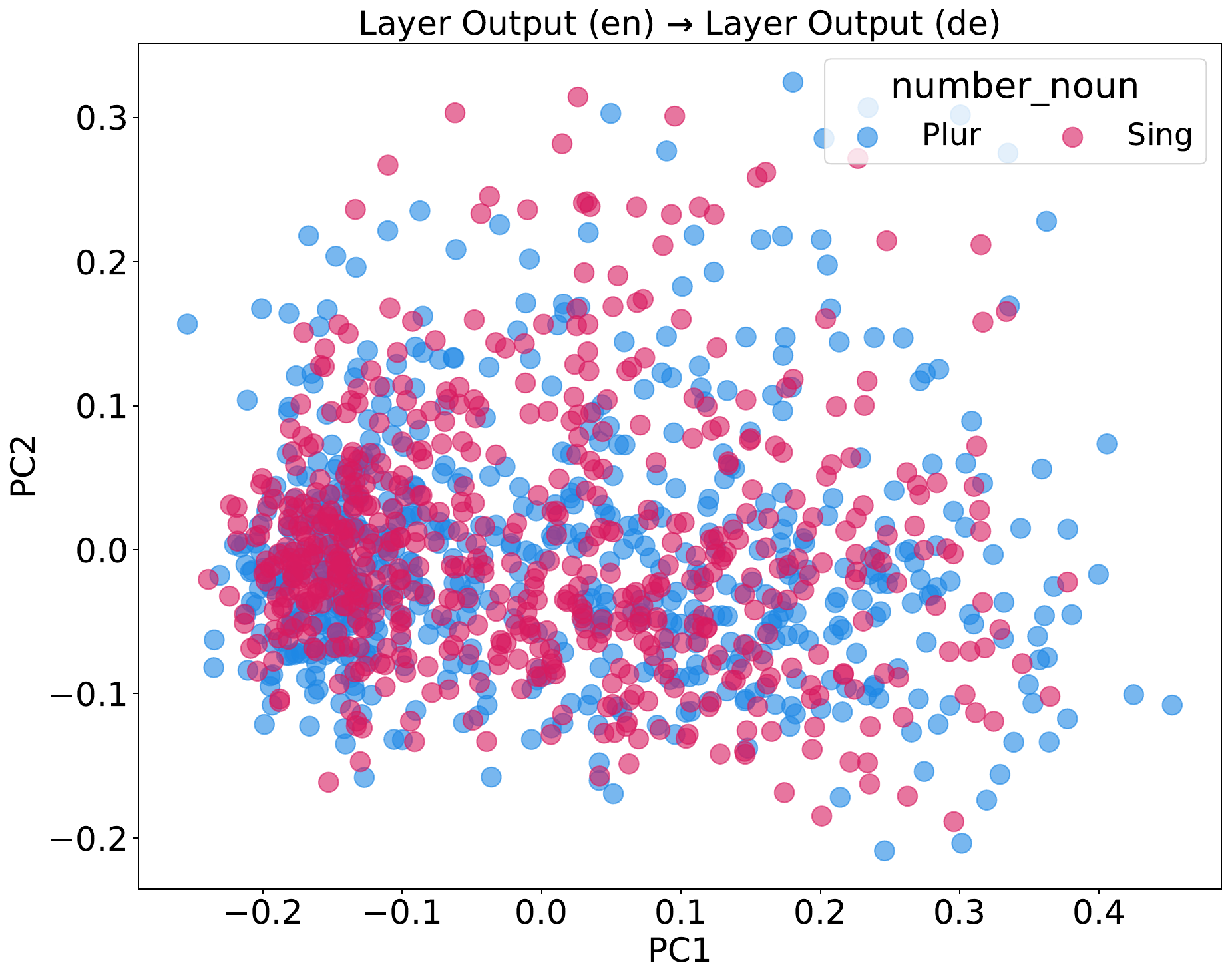}
        \caption{Layer 24}
    \end{subfigure}
    ~
    \begin{subfigure}{0.25\textwidth}
         \includegraphics[width=\textwidth]{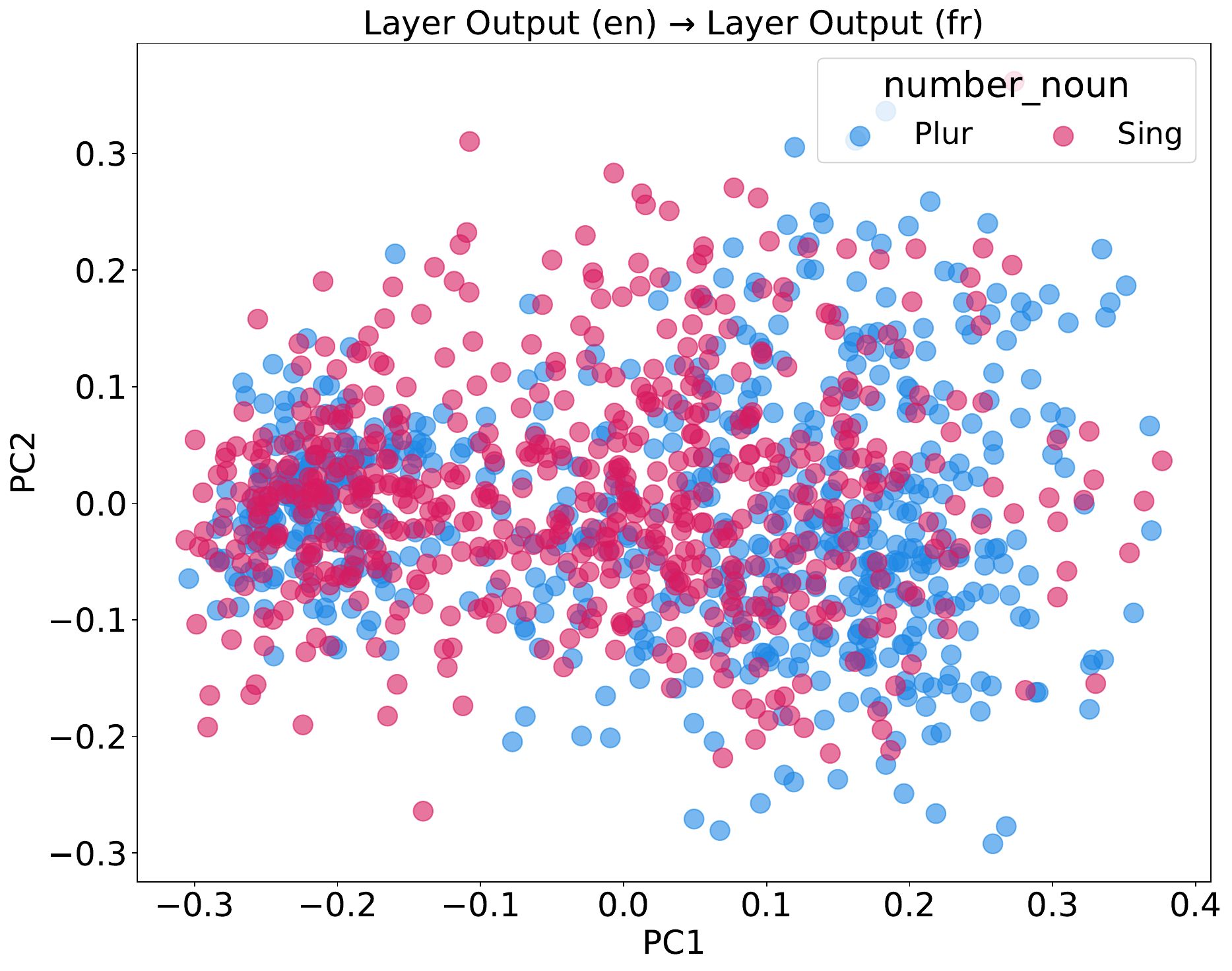}
        \caption{Layer 24}
    \end{subfigure}
    \caption{2D projections for tokens with different number (plural vs. singular) of the model pre-trained on English (first column) and the adapted models trained on German (second column) and French (third column) at various layers. In all three cases, the projection matrix is computed via PCA on the English representations only.}
    \label{fig:appendix:pos_number}
\end{figure*}

\begin{figure*}[t]
    \centering    
    \begin{subfigure}{0.25\textwidth}
        \includegraphics[width=\textwidth]{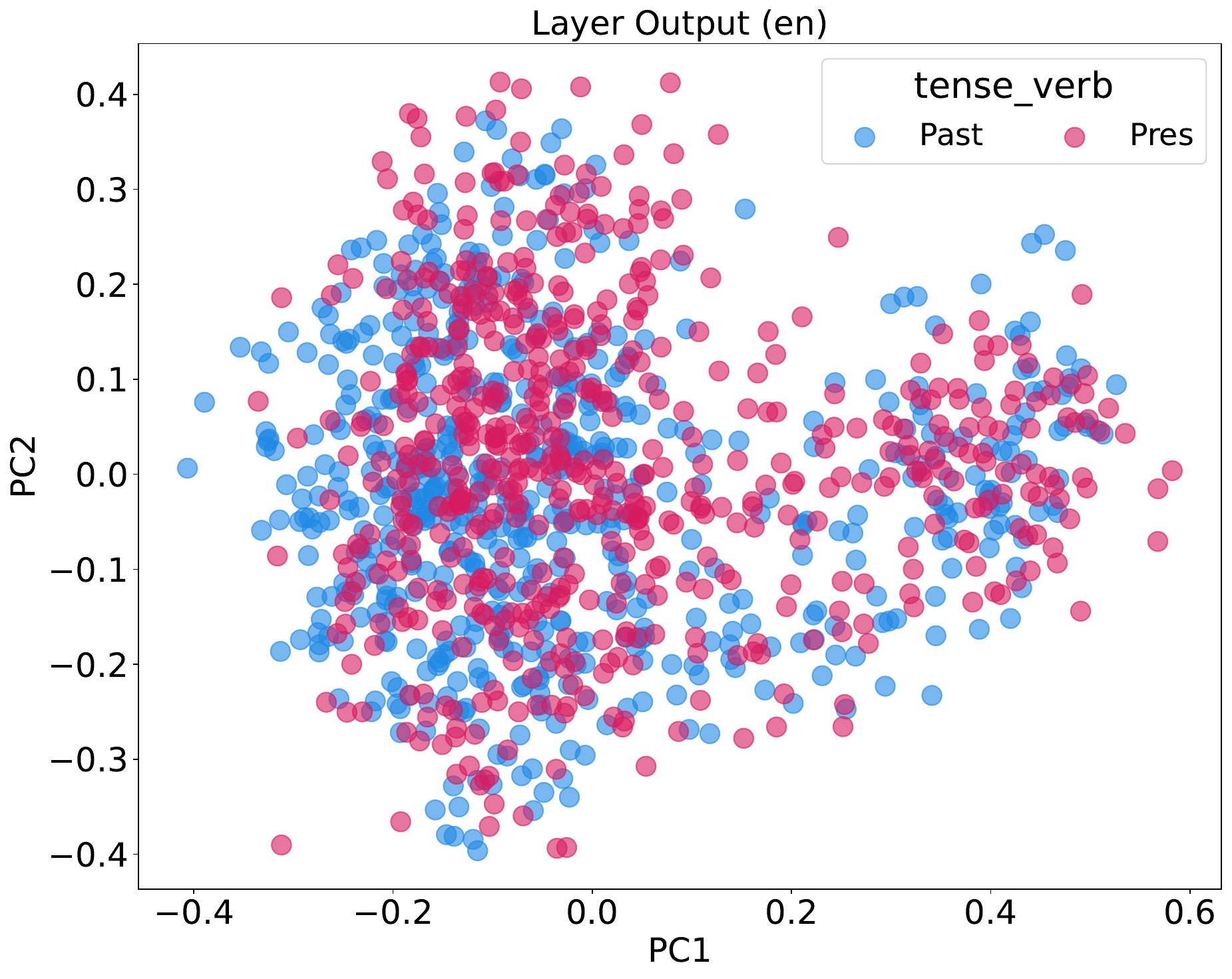}
        \caption{Layer 1}
    \end{subfigure}
    ~
    \begin{subfigure}{0.25\textwidth}
        \includegraphics[width=\textwidth]{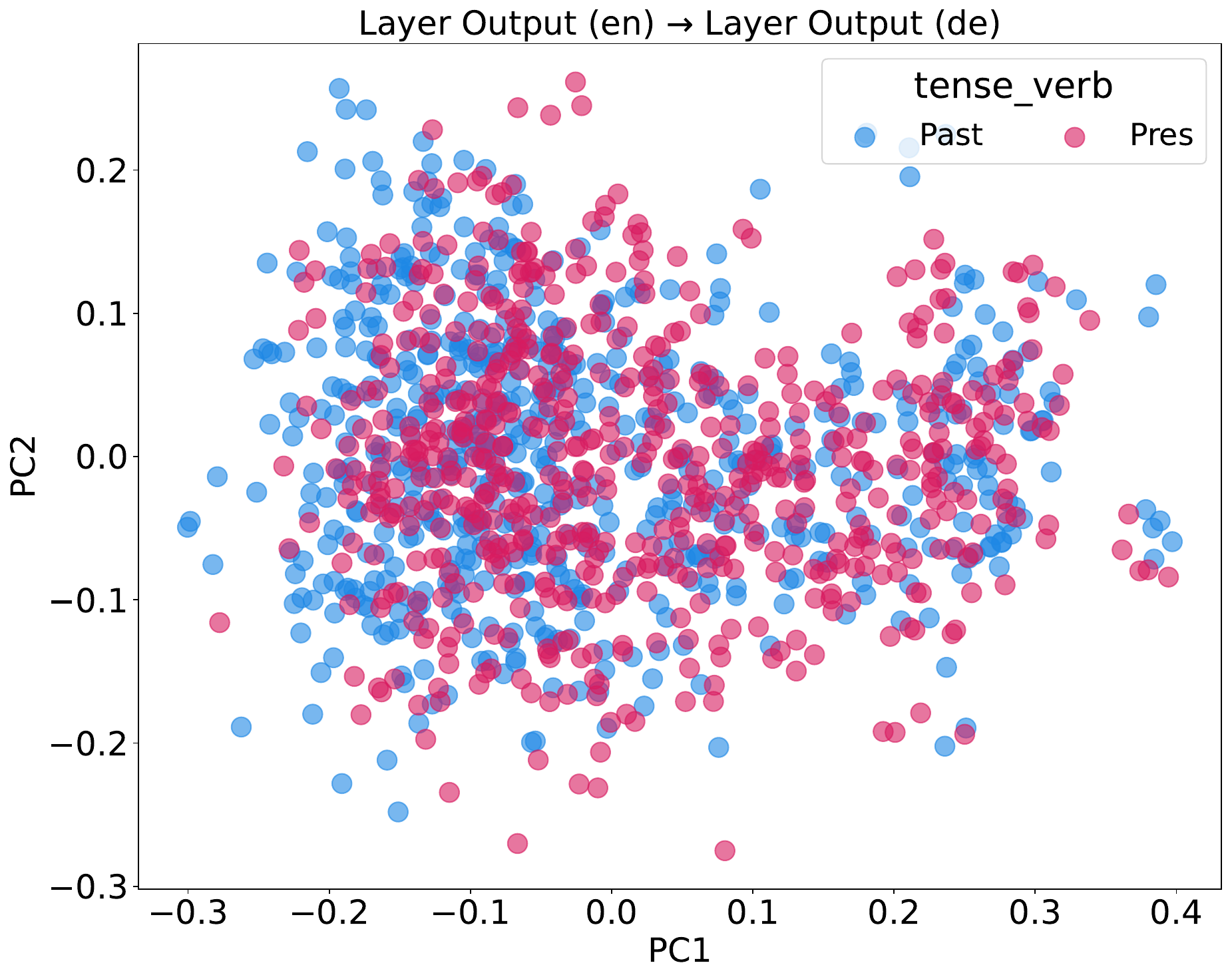}
        \caption{Layer 1}
    \end{subfigure}
    ~
    \begin{subfigure}{0.25\textwidth}
         \includegraphics[width=\textwidth]{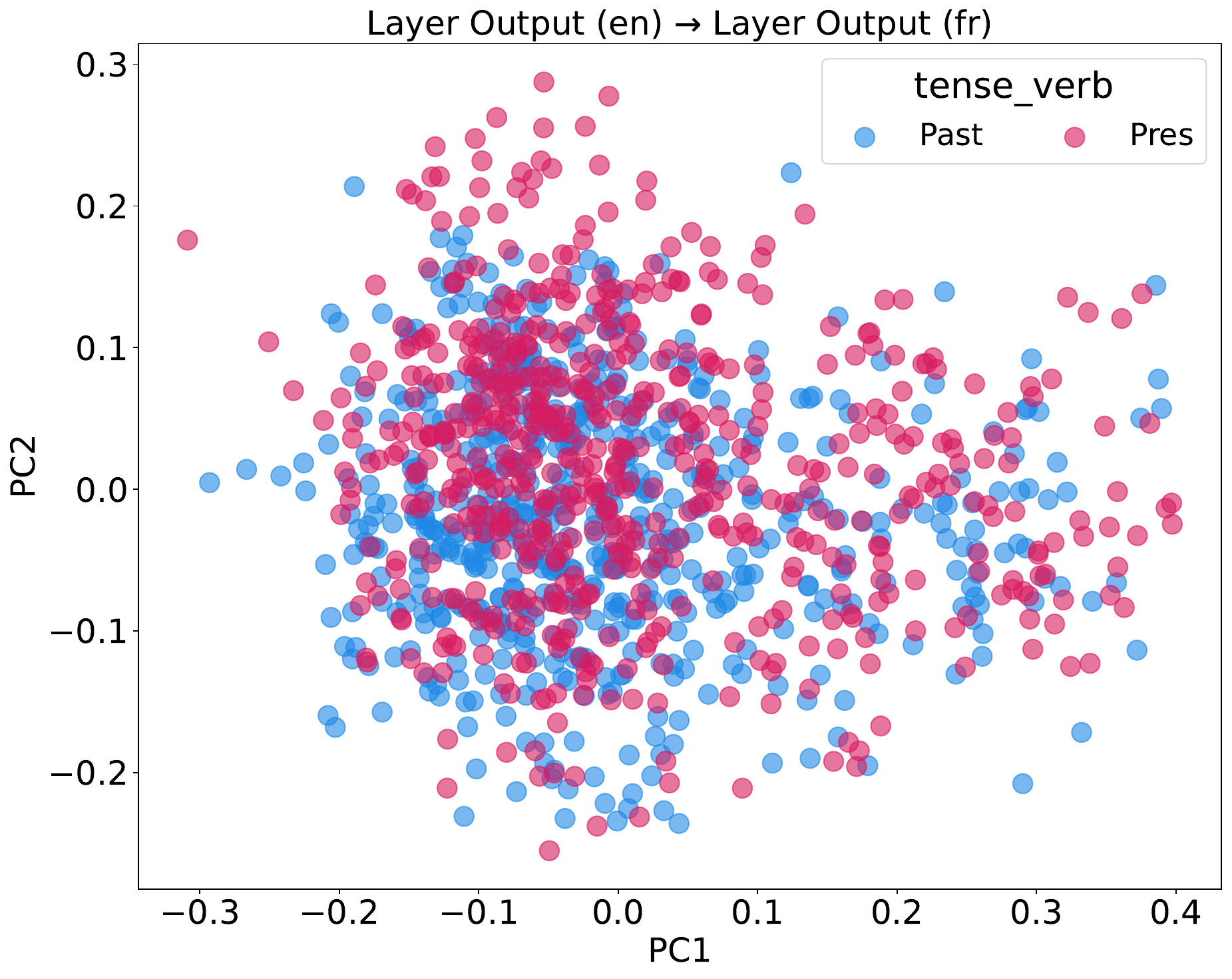}
        \caption{Layer 1}
    \end{subfigure}
    \\
    \begin{subfigure}{0.25\textwidth}
        \includegraphics[width=\textwidth]{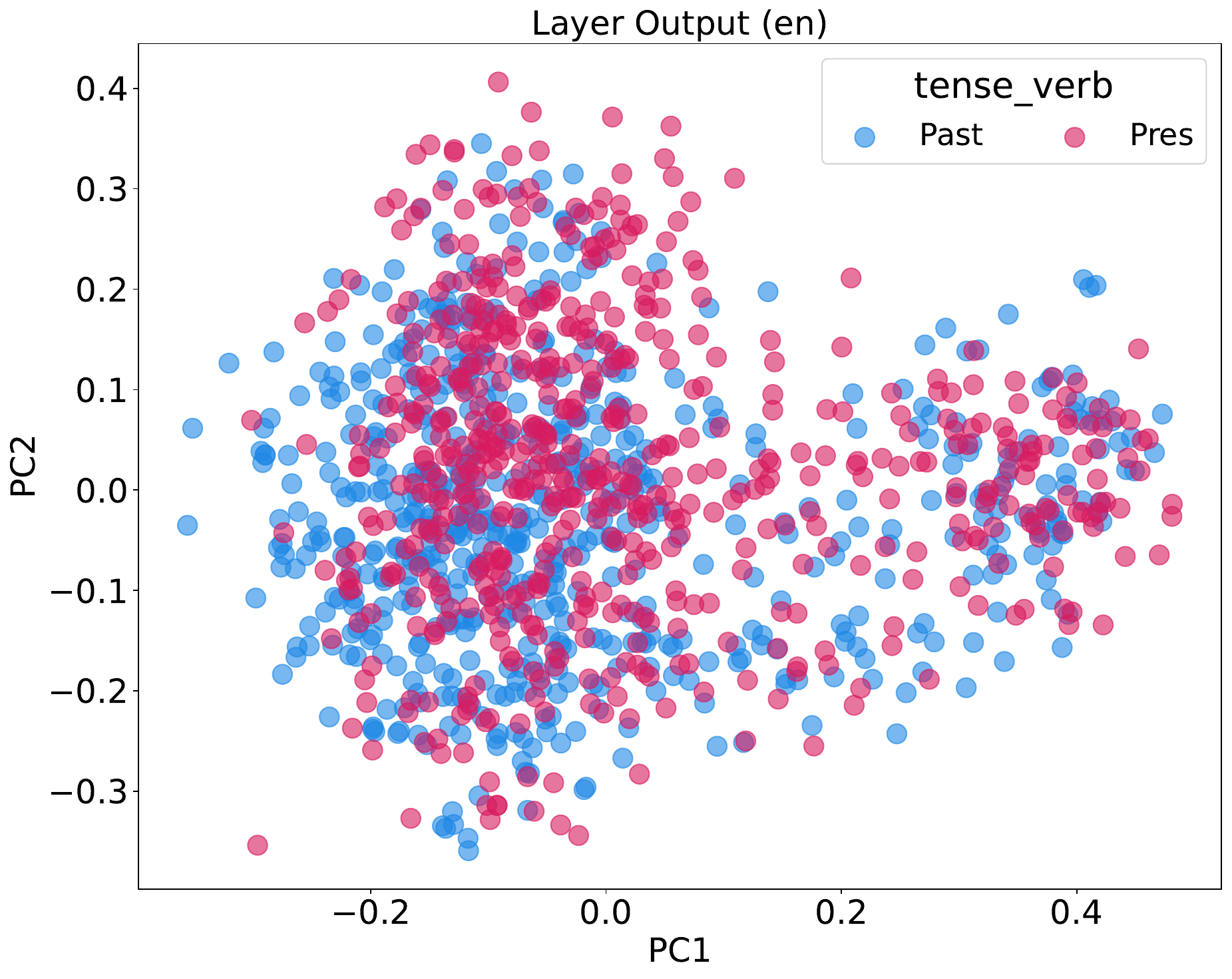}
        \caption{Layer 2}
    \end{subfigure}
    ~
    \begin{subfigure}{0.25\textwidth}
        \includegraphics[width=\textwidth]{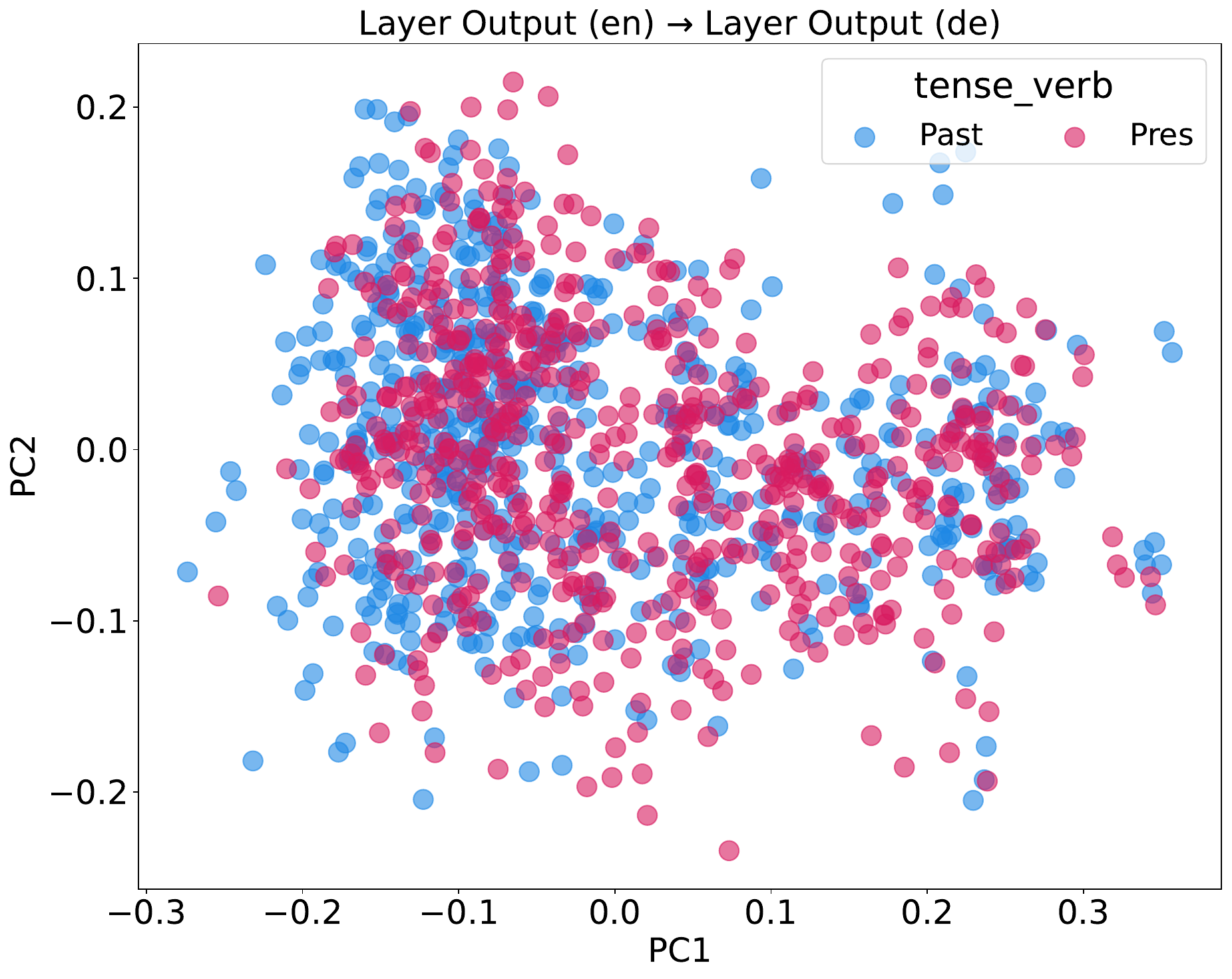}
        \caption{Layer 2}
    \end{subfigure}
    ~
    \begin{subfigure}{0.25\textwidth}
         \includegraphics[width=\textwidth]{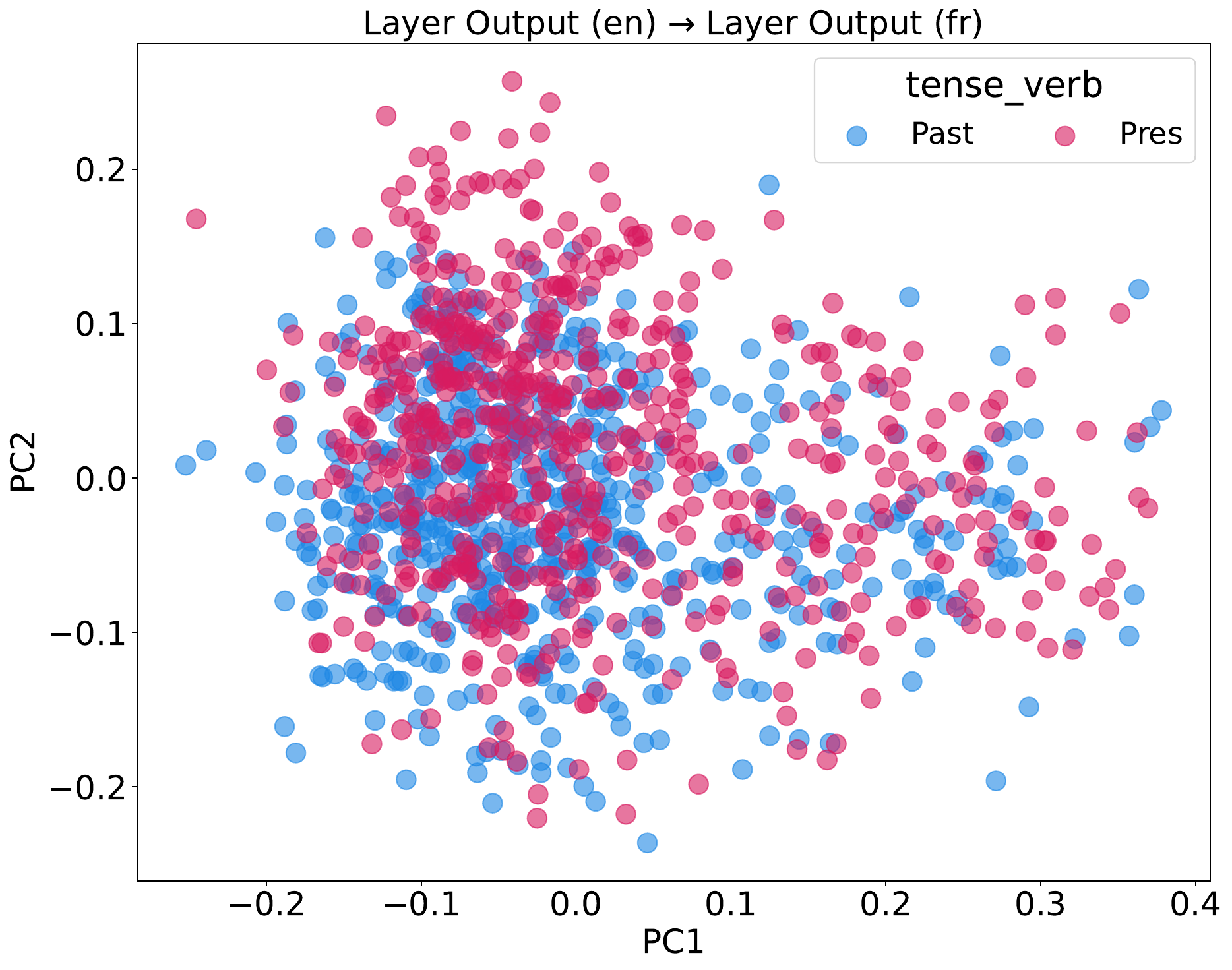}
        \caption{Layer 2}
    \end{subfigure}
    \\
    \begin{subfigure}{0.25\textwidth}
        \includegraphics[width=\textwidth]{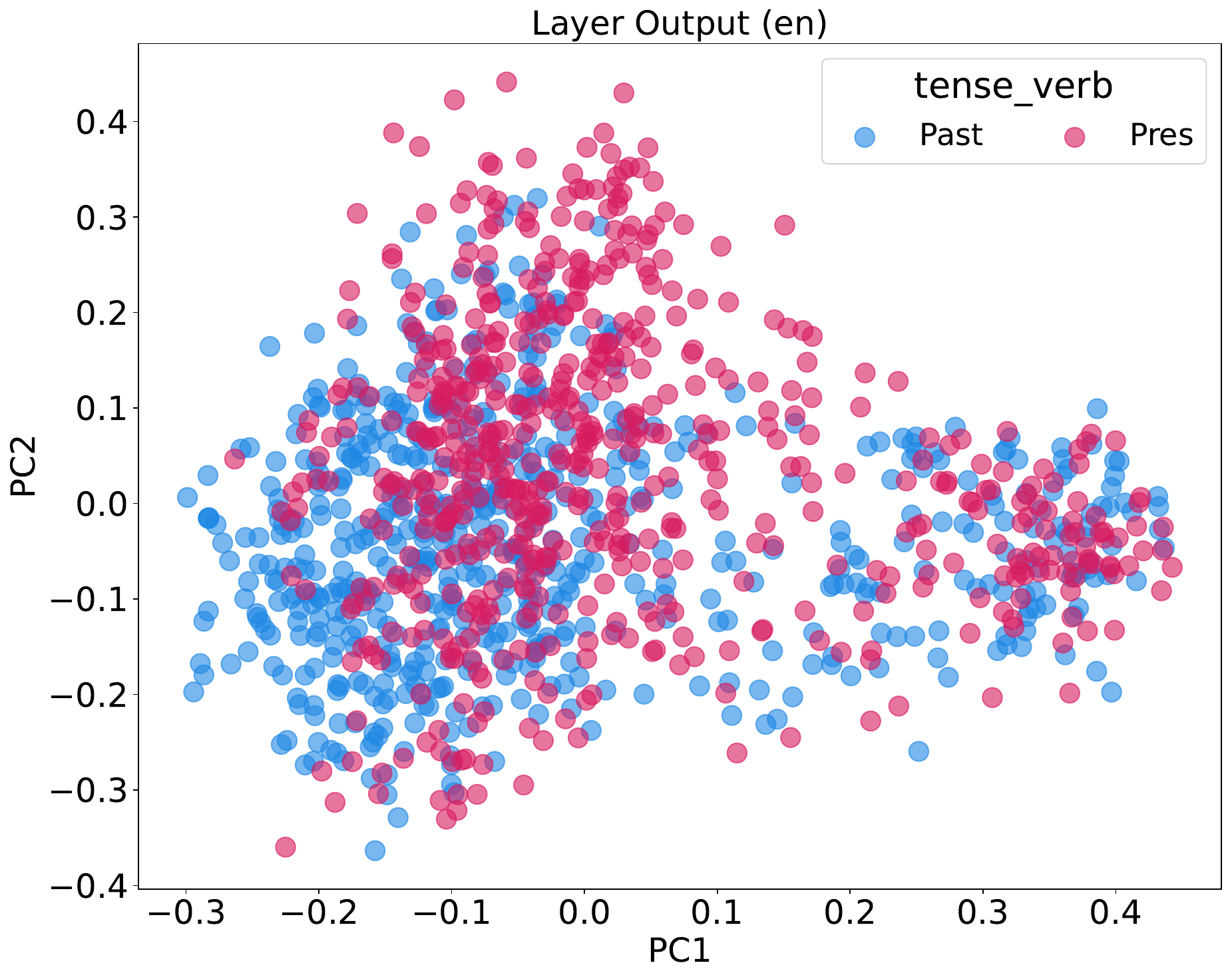}
        \caption{Layer 6}
    \end{subfigure}
    ~
    \begin{subfigure}{0.25\textwidth}
        \includegraphics[width=\textwidth]{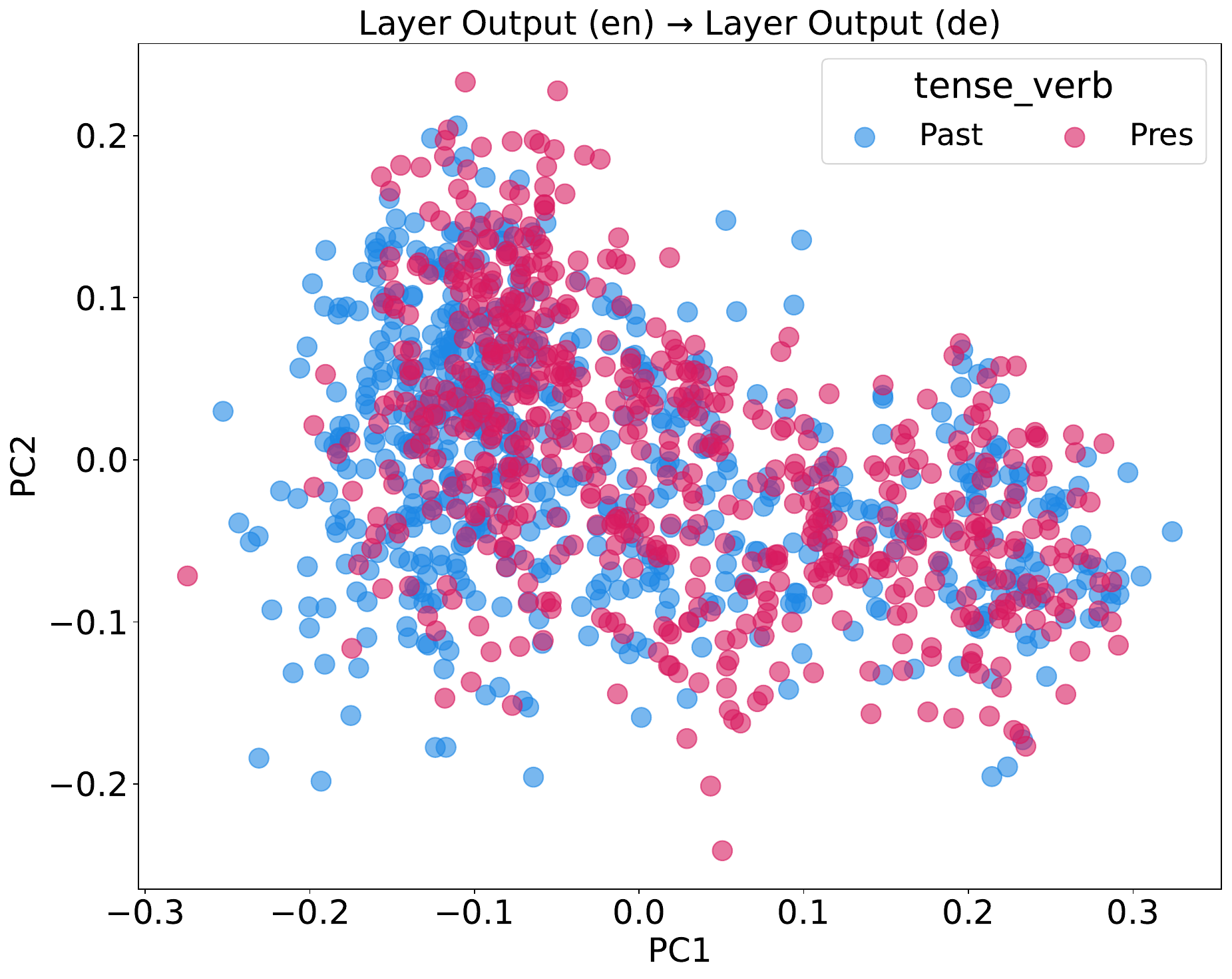}
        \caption{Layer 6}
    \end{subfigure}
    ~
    \begin{subfigure}{0.25\textwidth}
         \includegraphics[width=\textwidth]{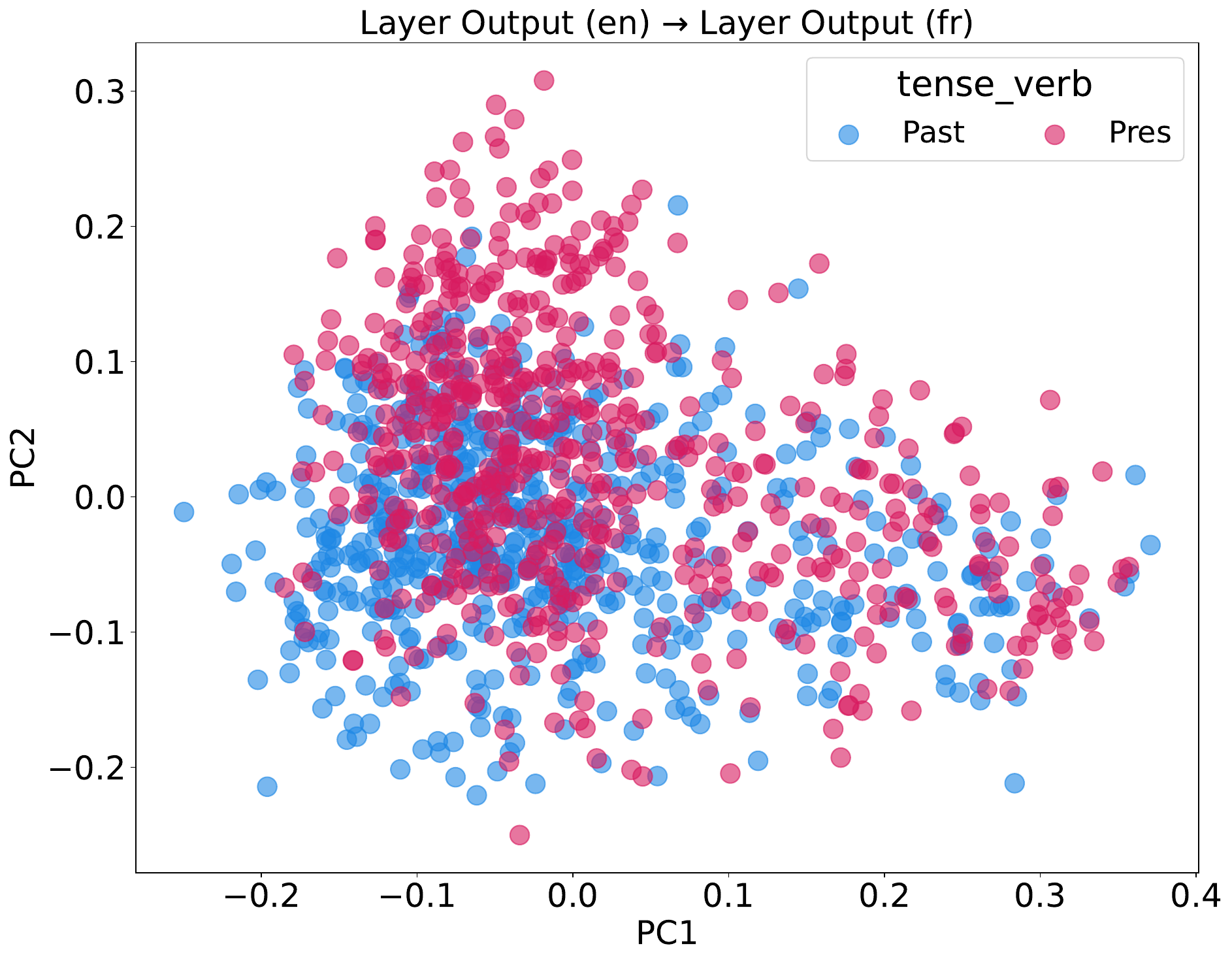}
        \caption{Layer 6}
    \end{subfigure}
    \\
    \begin{subfigure}{0.25\textwidth}
        \includegraphics[width=\textwidth]{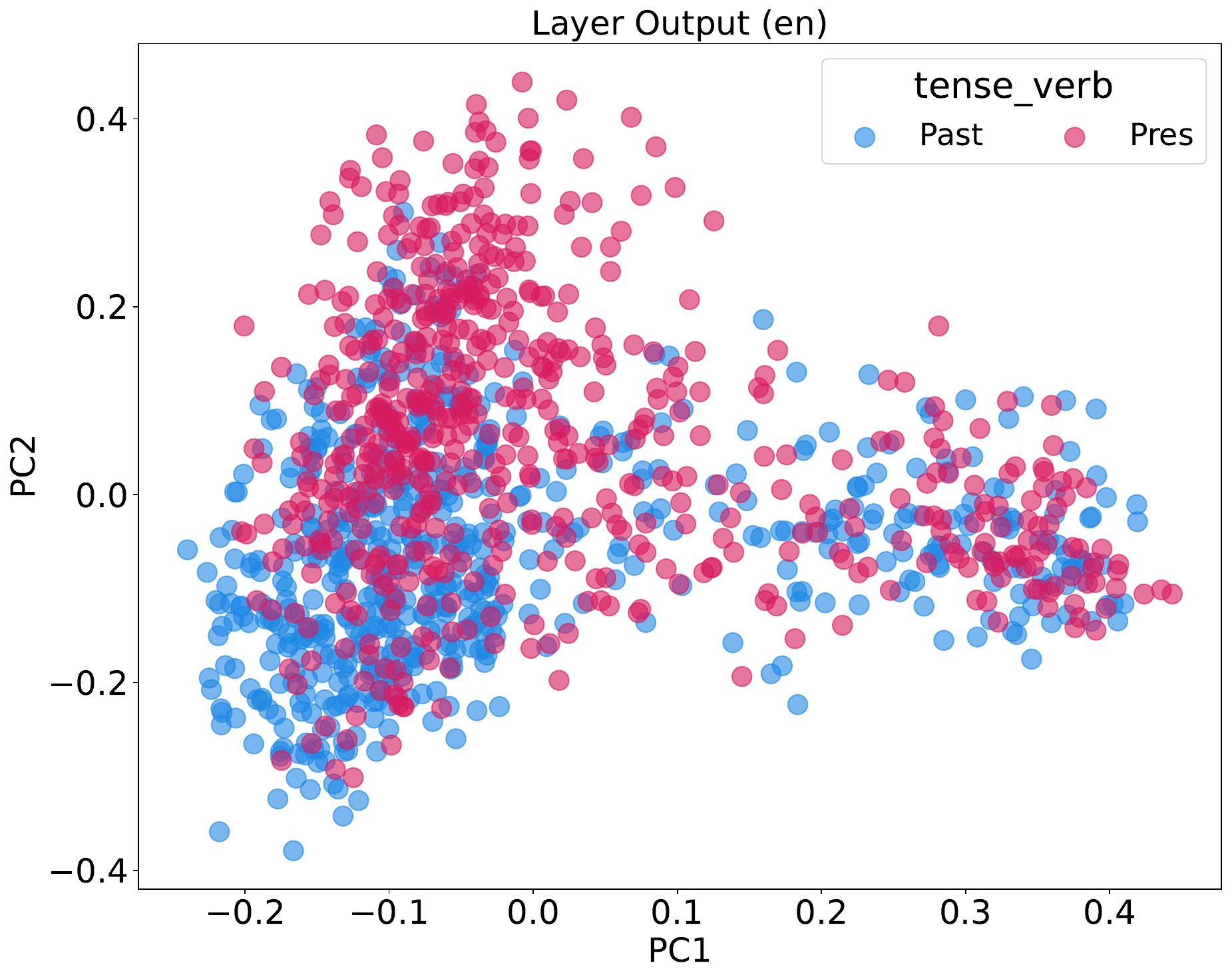}
        \caption{Layer 12}
    \end{subfigure}
    ~
    \begin{subfigure}{0.25\textwidth}
        \includegraphics[width=\textwidth]{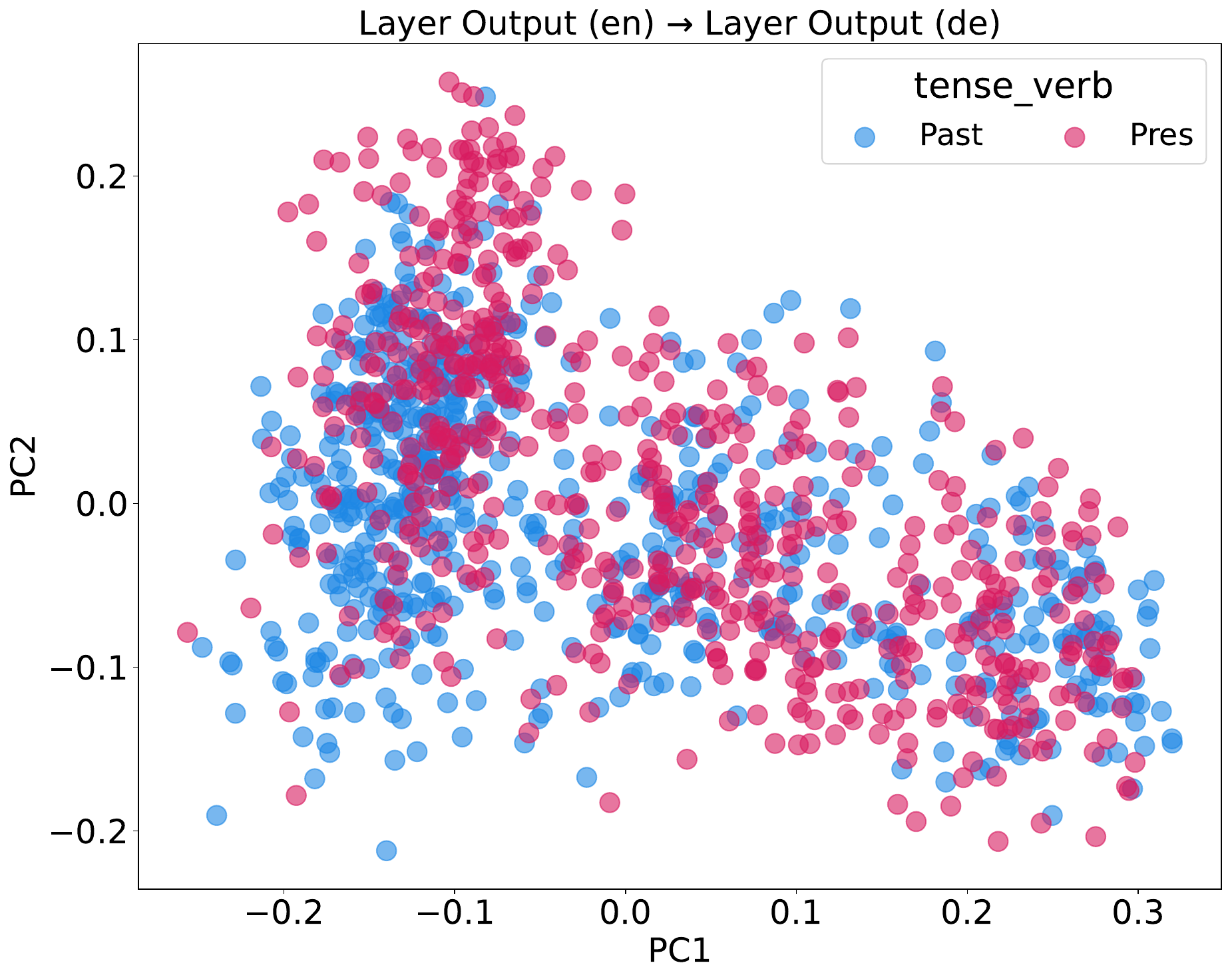}
        \caption{Layer 12}
    \end{subfigure}
    ~
    \begin{subfigure}{0.25\textwidth}
         \includegraphics[width=\textwidth]{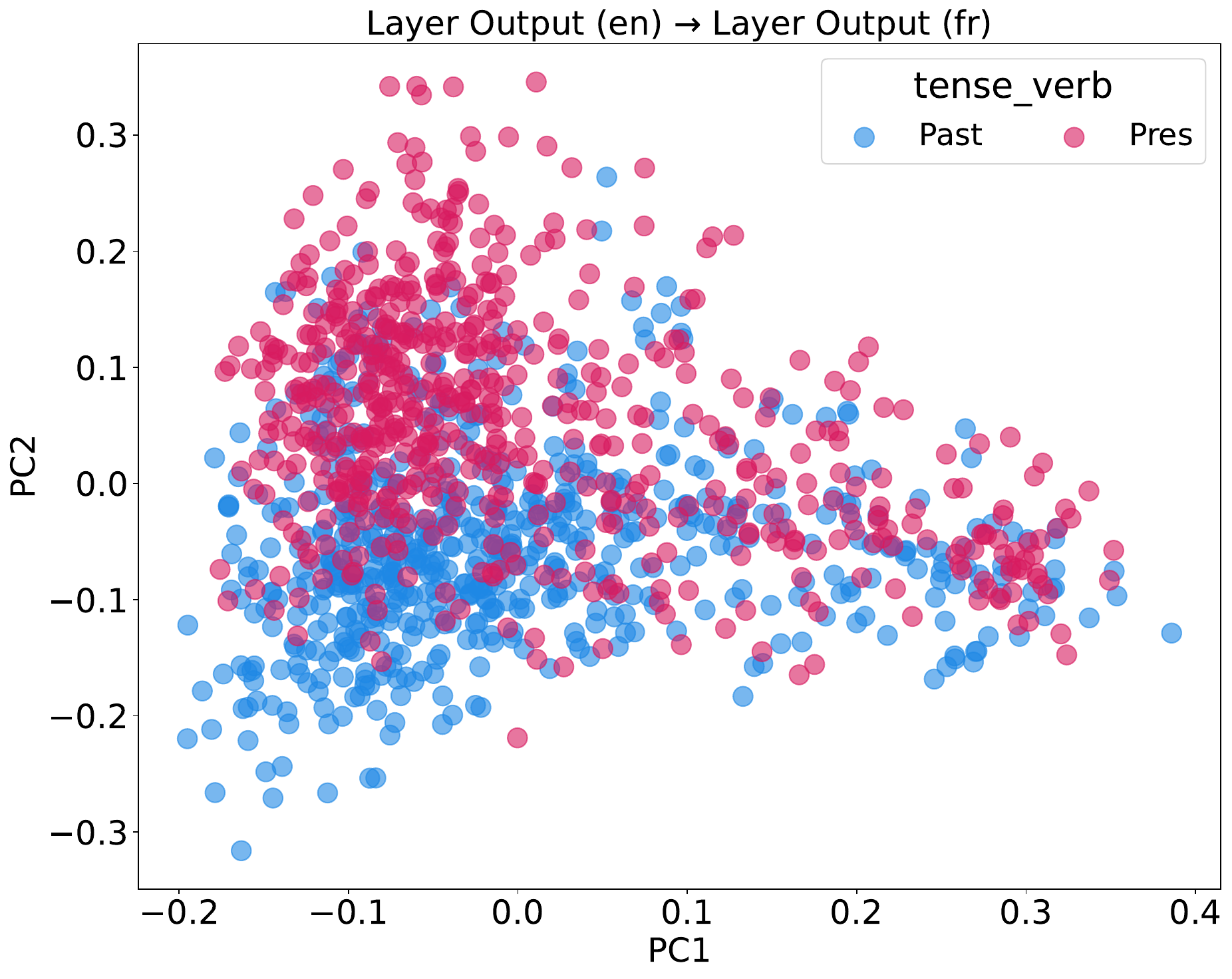}
        \caption{Layer 12}
    \end{subfigure}
    \\
    \begin{subfigure}{0.25\textwidth}
        \includegraphics[width=\textwidth]{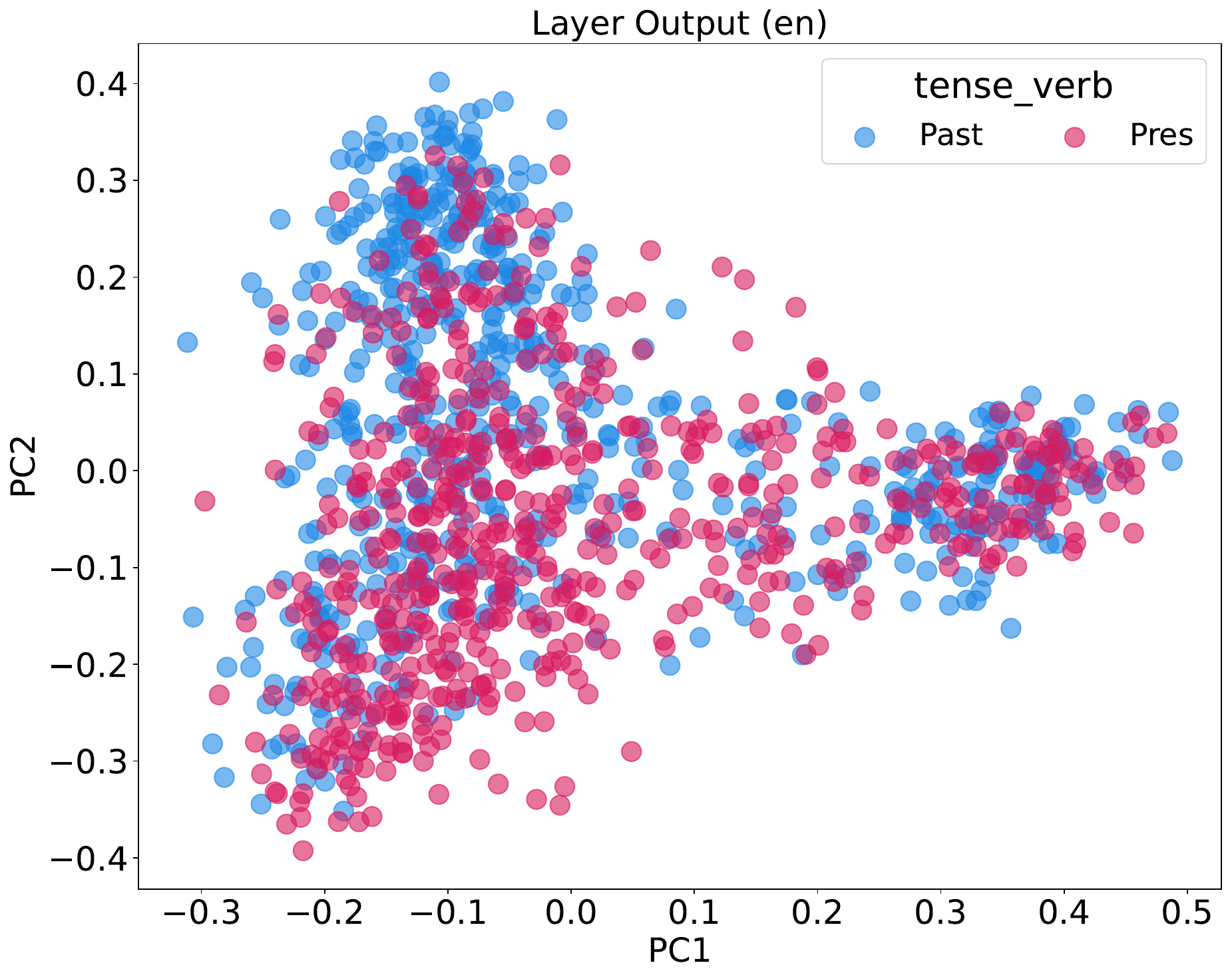}
        \caption{Layer 22}
    \end{subfigure}
    ~
    \begin{subfigure}{0.25\textwidth}
        \includegraphics[width=\textwidth]{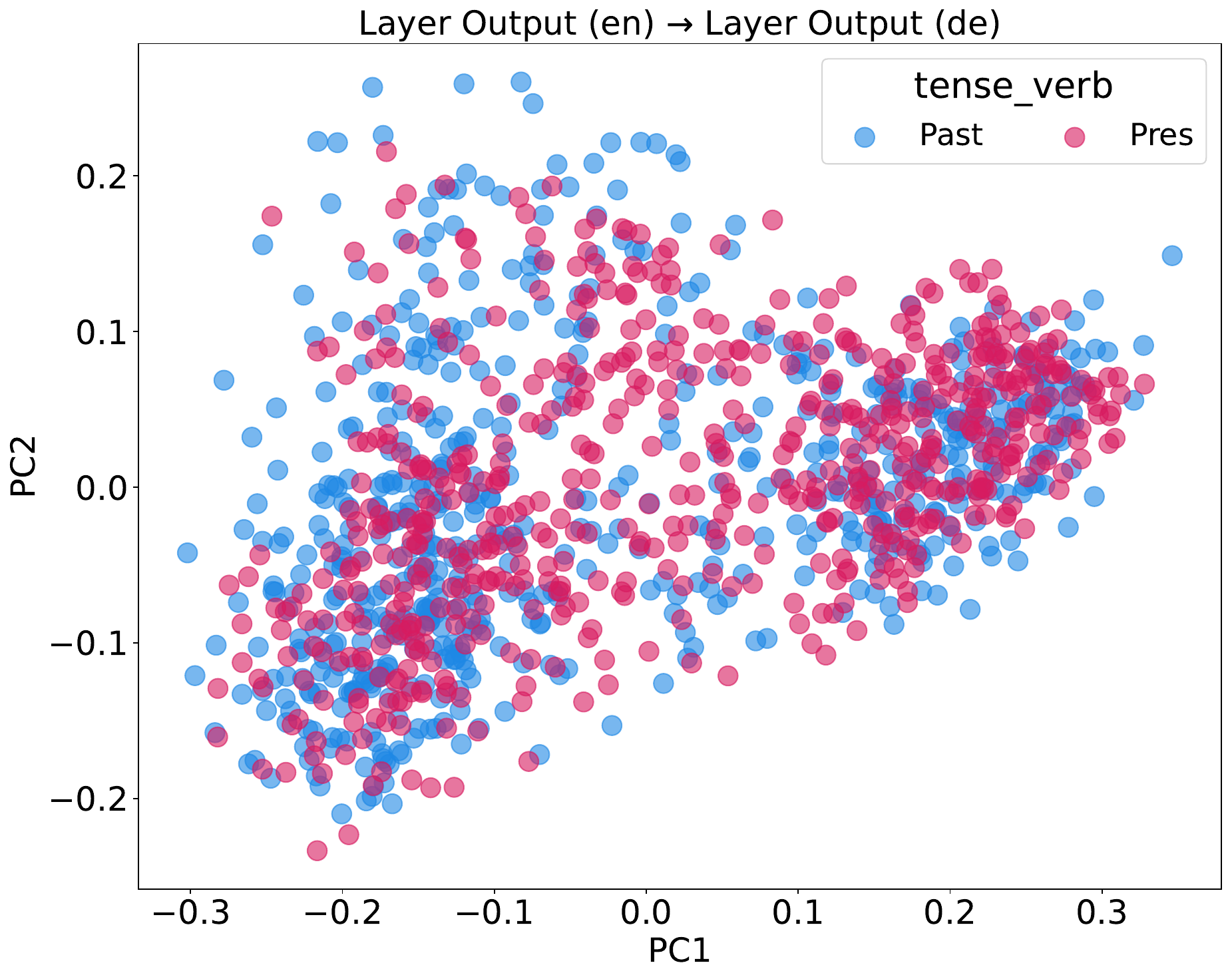}
        \caption{Layer 22}
    \end{subfigure}
    ~
    \begin{subfigure}{0.25\textwidth}
         \includegraphics[width=\textwidth]{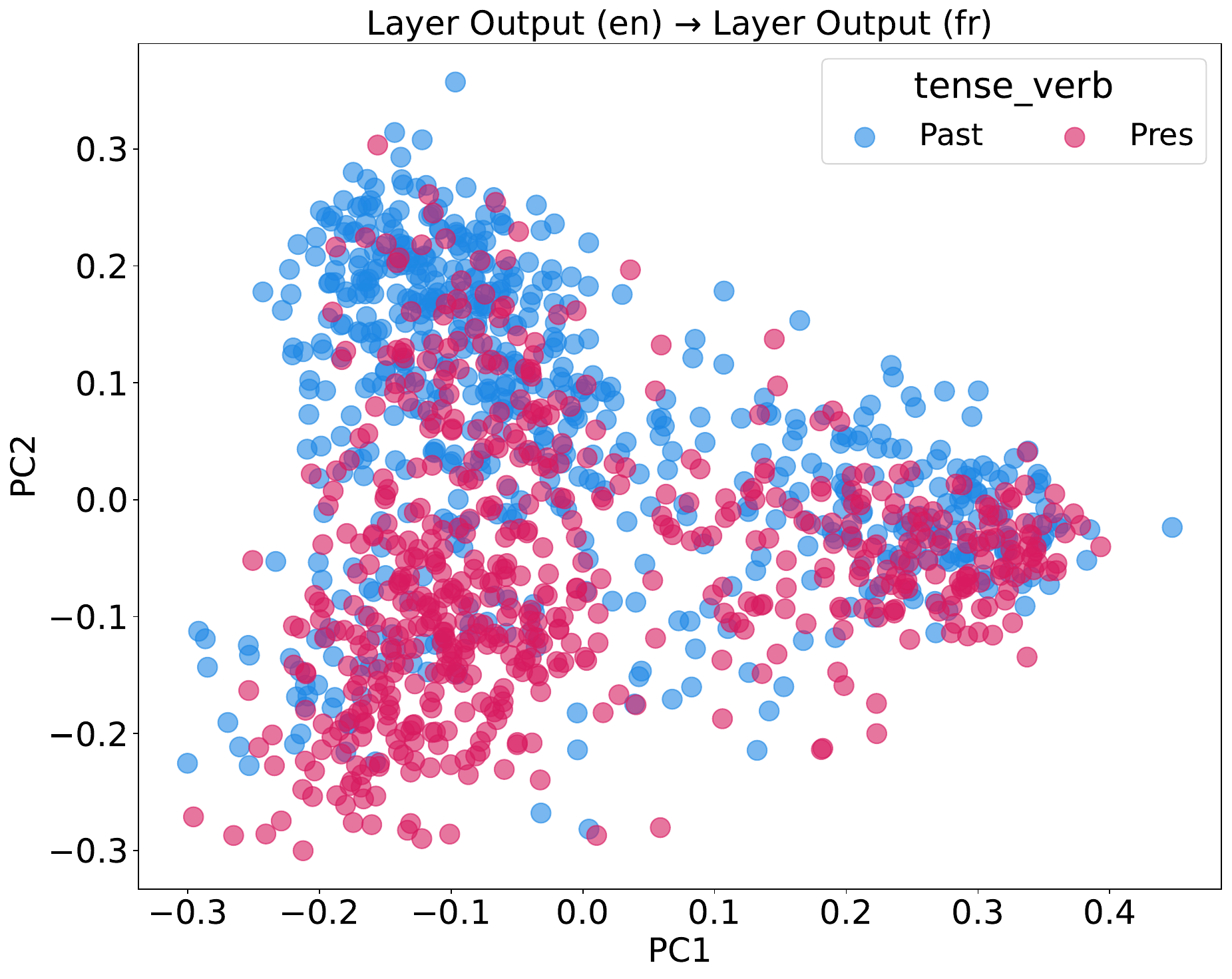}
        \caption{Layer 22}
    \end{subfigure}
    \\
    \begin{subfigure}{0.25\textwidth}
        \includegraphics[width=\textwidth]{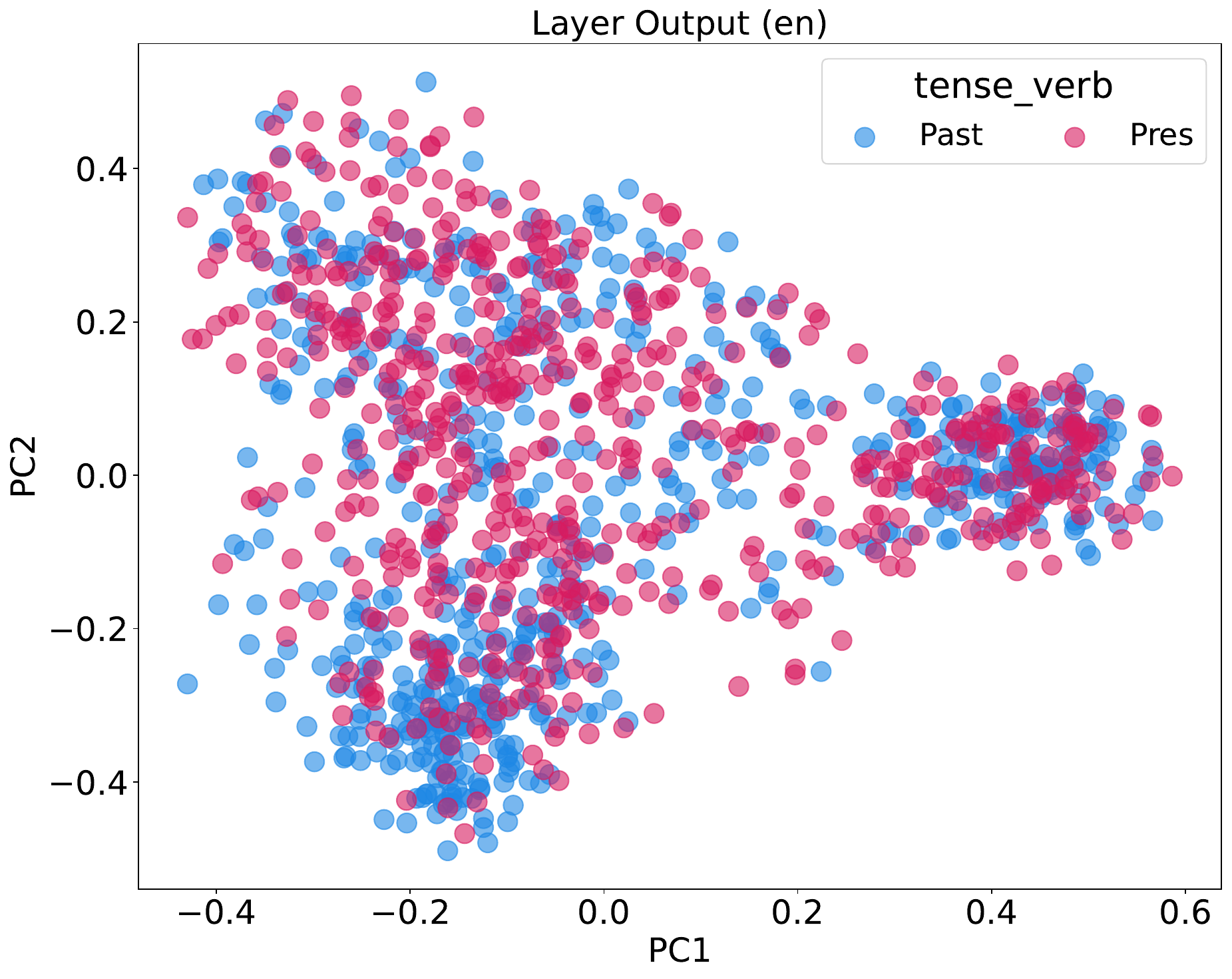}
        \caption{Layer 24}
    \end{subfigure}
    ~
    \begin{subfigure}{0.25\textwidth}
        \includegraphics[width=\textwidth]{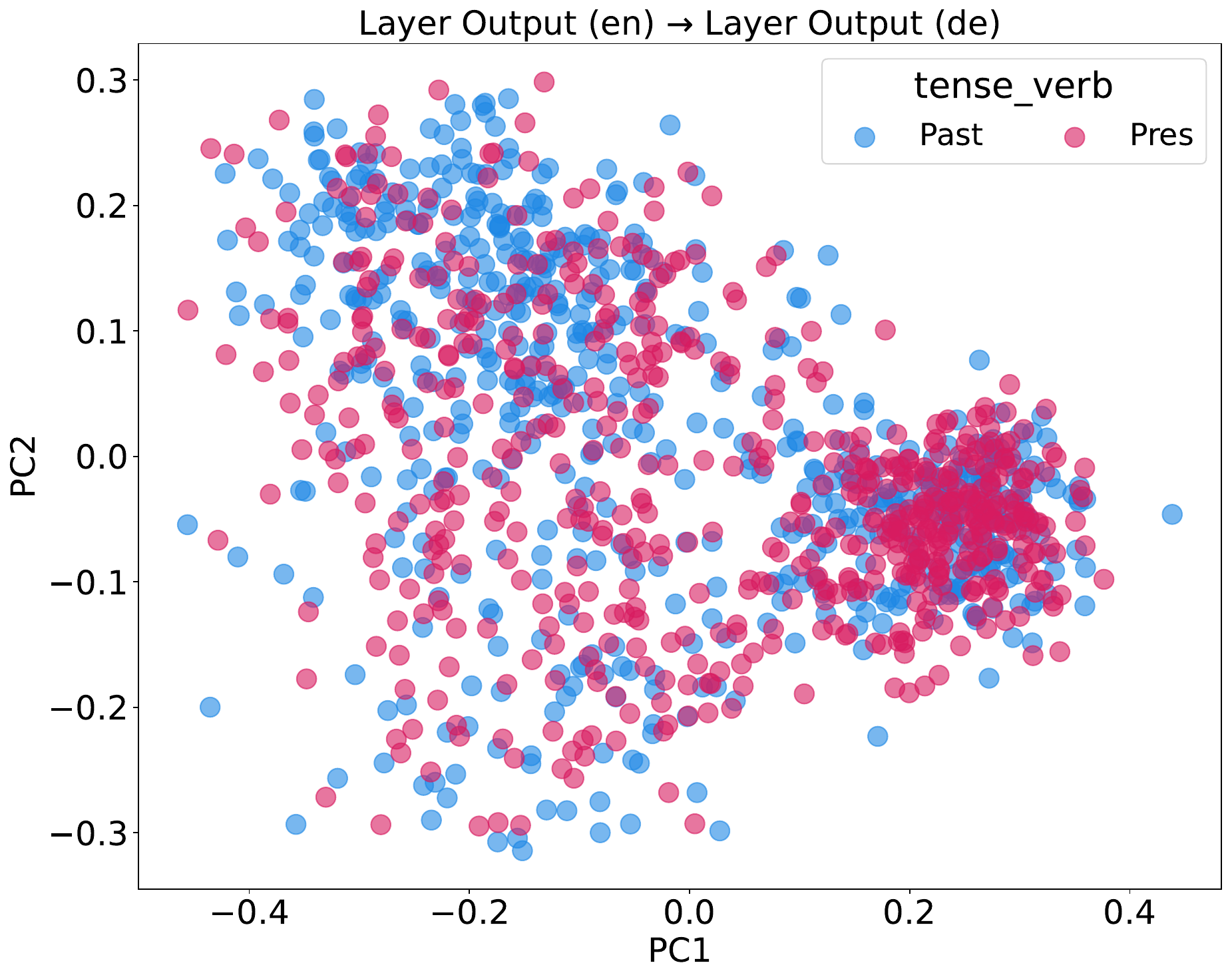}
        \caption{Layer 24}
    \end{subfigure}
    ~
    \begin{subfigure}{0.25\textwidth}
        \includegraphics[width=\textwidth]{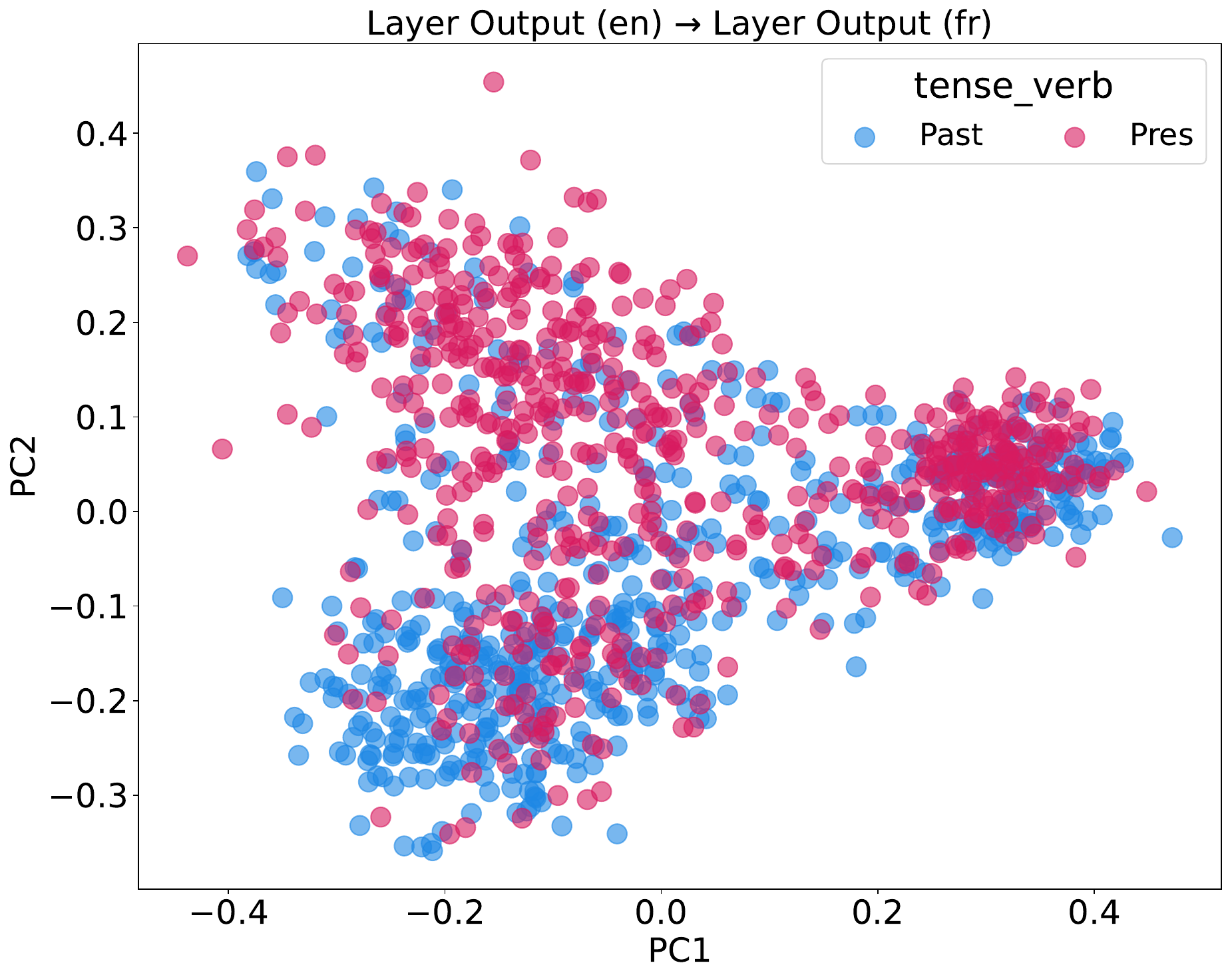}
        \caption{Layer 24}
    \end{subfigure}
    \caption{2D projections for tokens with different tense (past vs. present) of the model pre-trained on English (first column) and the adapted models trained on German (second column) and French (third column) at various layers. In all three cases, the projection matrix is computed via PCA on the English representations only.}
    \label{fig:appendix:pos_tense}
\end{figure*}

In addition analyzing the structure corresponding to POS, we also examine more nuanced linguistic features, such as verb tense (present and past) and noun number (singular and plural). Following a similar methodology to the POS experiment, we conduct a principal component analysis (PCA) on the hidden representations of English tokens and applied the resulting projection matrix to the output representations of German and French inputs. \Cref{fig:appendix:pos_number,fig:appendix:pos_tense} shows the results from this experiment.

In addition, \Cref{fig:appendix:cosine_number_tense} shows the cosine similarity between the first two principal components computed on the English and target language representations separately. As with POS, we observe a large alignment between the principal components, providing further evidence for our hypothesis that adapters mostly operate on top of the existing structure of the pre-trained model.

\end{document}